\documentclass[10pt,journal,compsoc]{IEEEtran}
\ifCLASSOPTIONcompsoc
  \usepackage[nocompress]{cite}
\else
  \usepackage{cite}
\fi
\ifCLASSINFOpdf
\else
\fi

\usepackage{times}
\usepackage{epsfig}
\usepackage{graphicx}
\usepackage{amsmath}
\usepackage{amssymb}
\usepackage{subfigure}
\usepackage{multirow}
\usepackage{booktabs}
\usepackage{color}
\usepackage[colorinlistoftodos]{todonotes}
\usepackage{url}

\begin{document}
%
\title{Refine-Net: Normal Refinement Neural Network for Noisy Point Clouds}
%
%
%
%

\author{\renewcommand{\thefootnote}{\fnsymbol{footnote}}Haoran Zhou, Honghua Chen, Yingkui Zhang, Mingqiang Wei, \textit{Senior Member}, \textit{IEEE}, Haoran Xie, \textit{Senior Member}, \textit{IEEE},
Jun Wang, Tong Lu, Jing Qin, and Xiao-Ping Zhang, \textit{Fellow}, \textit{IEEE}

   \thanks{H. Zhou and T. Lu are with the State Key Laboratory for Novel Software Technology, Nanjing University, Nanjing, China (hrzhou98@gmail.com; lutong@nju.edu.cn).}
   \thanks{H. Chen, J. Wang and M. Wei are with the School of Computer Science and Technology, Nanjing University of Aeronautics and Astronautics, Nanjing, China, and also with the MIIT Key Laboratory of Pattern Analysis and Machine Intelligence, Nanjing, China (chenhonghuacn@gmail.com; davis.wjun@gmail.com; mingqiang.wei@gmail.com).}
   \thanks{H. Xie is with the Department of Computing and Decision Sciences, Lingnan University, Hong Kong SAR, China (hrxie2@gmail.com).}    
   \thanks{Y. Zhang and J. Qin are with the School of Nursing, The Hong Kong Polytechnic University, Hong Kong SAR, China (yingkui.zhang@polyu.edu.hk; harry.qin@polyu.edu.hk).}
   \thanks{X.-P. Zhang is with the Department of Electrical, Computer and Biomedical Engineering, Ryerson University, Toronto, Canada (xzhang@ee.ryerson.ca).} 
    \thanks{Corresponding authors: M. Wei and T. Lu.} 

}

%
%

\markboth{}%
{Shell \MakeLowercase{\textit{et al.}}: Bare Demo of IEEEtran.cls for Computer Society Journals}
%



\IEEEtitleabstractindextext{%
\begin{abstract}

 Point normal, as an intrinsic geometric property of 3D objects, not only serves conventional geometric tasks such as surface consolidation and reconstruction, but also facilitates cutting-edge learning-based techniques for shape analysis and generation. In this paper, we propose a normal refinement network, called Refine-Net, to predict accurate normals for noisy point clouds. Traditional normal estimation wisdom heavily depends on priors such as surface shapes or noise distributions, while learning-based solutions settle for single types of hand-crafted features. Differently, our network is designed to refine the initial normal of each point by extracting additional information from multiple feature representations. To this end, several feature modules are developed and incorporated into Refine-Net by a novel connection module. Besides the overall network architecture of Refine-Net, we propose a new multi-scale fitting patch selection scheme for the initial normal estimation, by absorbing geometry domain knowledge. 
 Also, Refine-Net is a generic normal estimation framework: 1) point normals obtained from other methods can be further refined, and 2) any feature module related to the surface geometric structures can be potentially integrated into the framework. Qualitative and quantitative evaluations demonstrate the clear superiority of Refine-Net over the state-of-the-arts on both synthetic and real-scanned datasets. Our code is available at \url{https://github.com/hrzhou2/refinenet}.
\end{abstract}

}

\maketitle

\IEEEdisplaynontitleabstractindextext

%
\IEEEpeerreviewmaketitle

\IEEEraisesectionheading{\section{Introduction}\label{sec:introduction}}
\IEEEPARstart{P}oint cloud is a set of unstructured points to efficiently represent 3D surfaces/scenes. Each point in the point cloud possesses a spatial position $(x, y, z)$ and potentially a vector of attributes, such as the normal, color or material reflection.  
Currently, point clouds are routinely captured by laser/optical scanners and depth cameras, e.g., Velodyne LiDAR Velabit, Intel RealSense, LiDAR scanner of Apple iPad Pro, or Microsoft Kinect. As standard outputs of these 3D sensors, point clouds have been flexibly used in various applications, ranging from 6-degree virtual reality \cite{he2020pvn3d,peng2019pvnet}, robotics \cite{rusu2008towards} to autonomous driving \cite{liang2018deep,wang2018deep}. As known, even the state-of-the-art 3D sensors all inevitably arise noise, due to both the measurement and reconstruction errors. Noise hinders existing normal estimation methods heavily, since surface geometric properties are sensitive to such local perturbations. Besides noise, the objects in real scans are even more complicated, which introduce additional challenges in normal estimation. 

For any noisy point cloud as input, normal estimation aims to accurately predict the normal of each point; it is an indispensable step for an input point cloud with different noise distributions, geometric characteristics and sampling rates. Accurate normals will contribute to either fundamental geometric tasks, such as point cloud consolidation \cite{chen2019multi,dinesh2020point,hu2020feature,hermosilla2019total}, surface reconstruction \cite{FleishmanCS05,xie2019surface,chen2021multiscale} and segmentation \cite{che2018multi,wang2019dynamic}, or high-level shape analysis neural networks \cite{li2018so,te2018rgcnn,qian2020pugeo}. Ideally, the normal field of a noisy point cloud, which is formed by the calculated point normals, should approach the normal field of its clean counterpart (i.e., the underlying surface).

\begin{figure}[t]
	\includegraphics[width=0.99\linewidth]{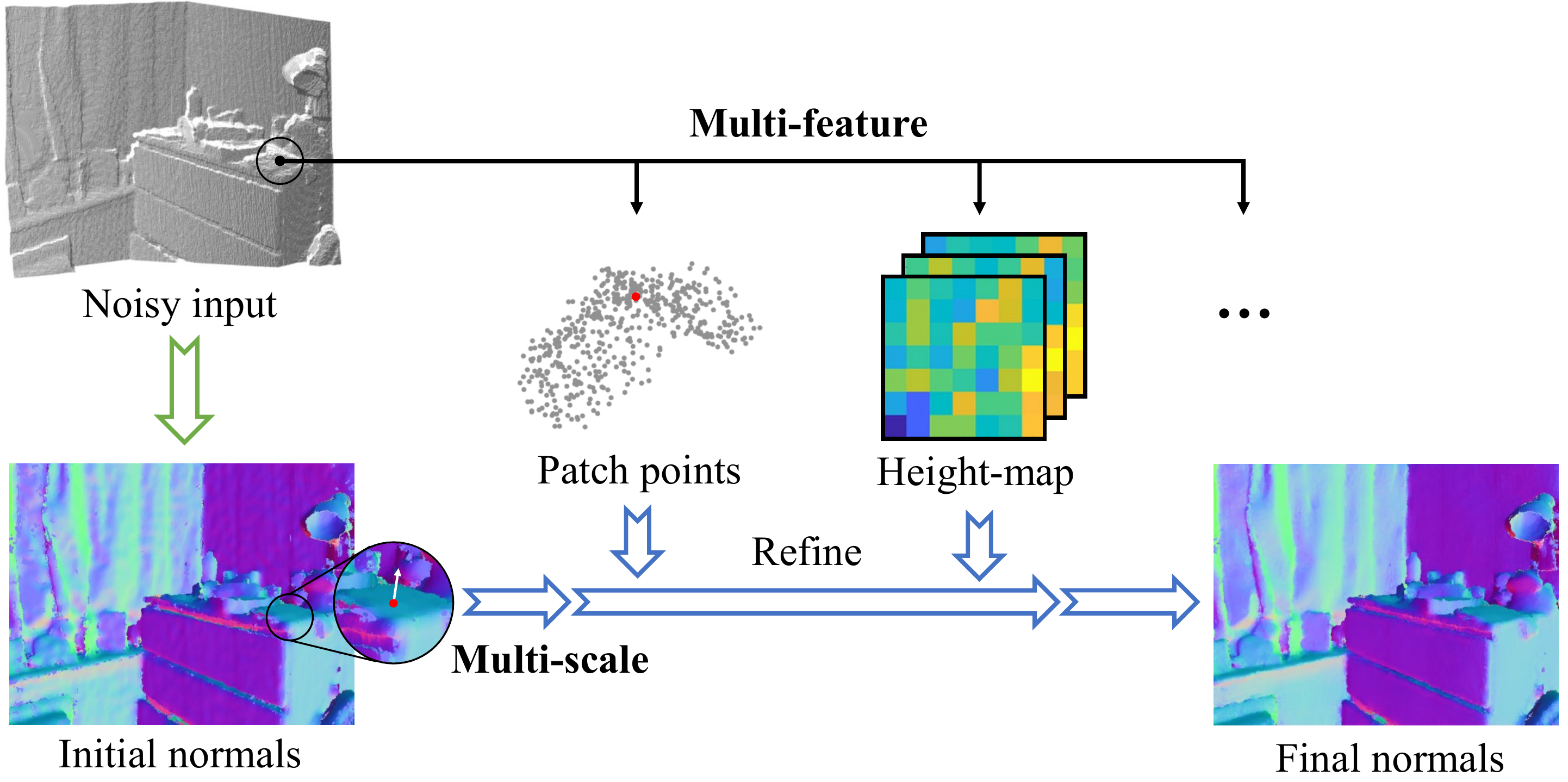}
	\caption{An effective multi-scale and -feature scheme is leveraged in our normal refinement paradigm. We explore different types of feature representations from an input point cloud (black arrows), and incorporate them into Refine-Net (blue arrows). The green arrow denotes the normal estimation effort to obtain the initial normals which are further processed by a multi-scale scheme in the refinement.}
	\label{fig:intro1}
\end{figure}

A good normal estimation solution will possess two merits of 1) maintaining data fidelity by preserving geometric features, and 2) adapting to noise with unknown distributions. However, the fundamental difficulty of normal estimation exists in how to differentiate surface features and noise, which are both of high frequency and sometimes small in scale. Traditional estimation methods often utilize several elaborately-designed regularities (e.g., $l_{0}$ minimization, low rank) or noise distributions (e.g., Gaussian/impulsive noise) to preserve geometric features. These prior-based approaches, though achieving notable successes in certain surface types or noise models, may not generalize well to resolve diverse inputs. In addition, they rely on parameters tuning heavily, like the neighborhood scale for plane fitting \cite{zhang2018multi,zhang2013point}, which tends to over-smooth or over-sharpen normal estimation results.

\begin{figure}[t]
	\newlength{\unitT}
	\setlength{\unitT}{0.33\linewidth}
	\centering
	
	\subfigure[Depth image]{\includegraphics[width=\unitT]{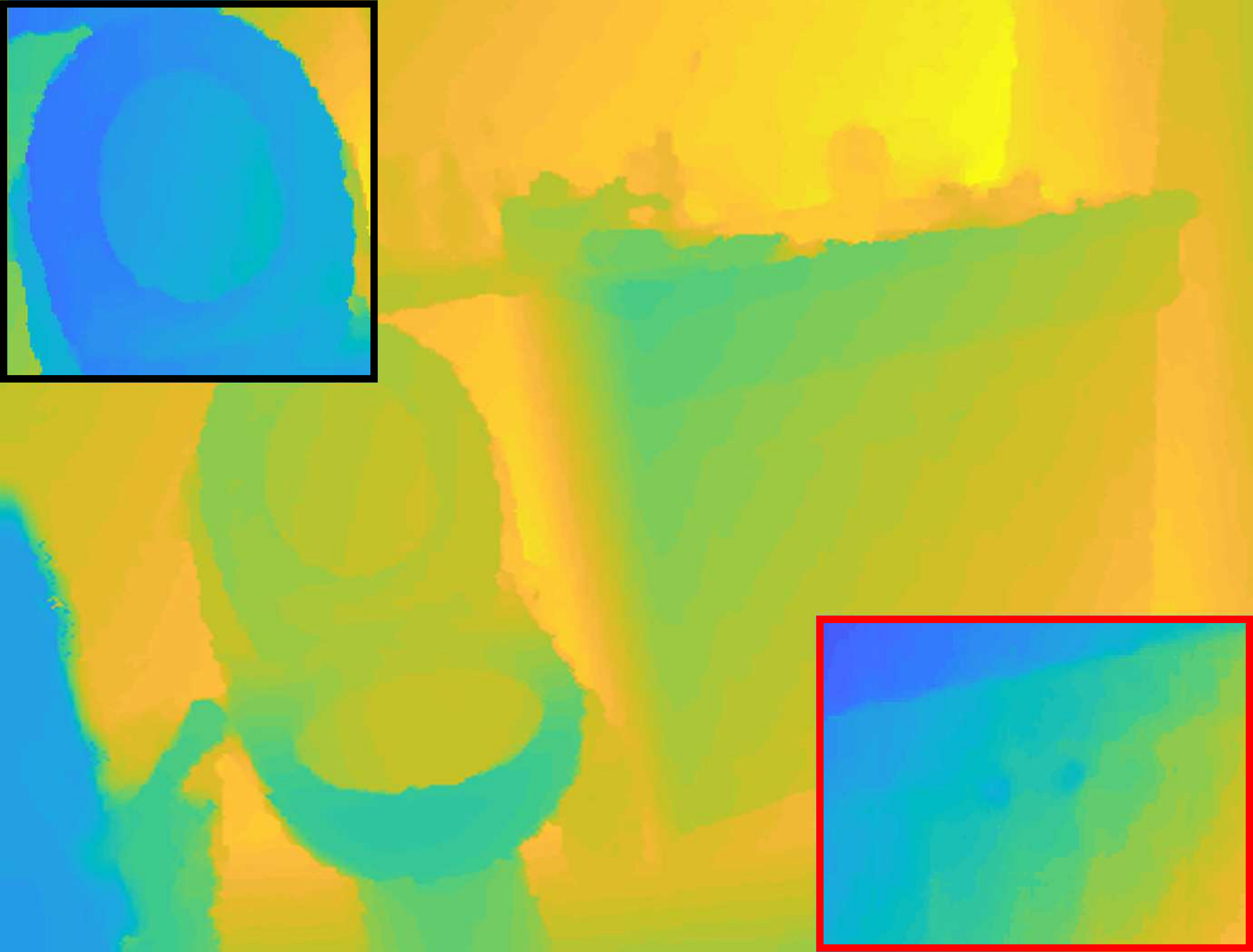}}%
	\subfigure[Nesti-Net \cite{Ben-ShabatLF19} \label{fig:intro2:Nesti}]{\includegraphics[width=\unitT]{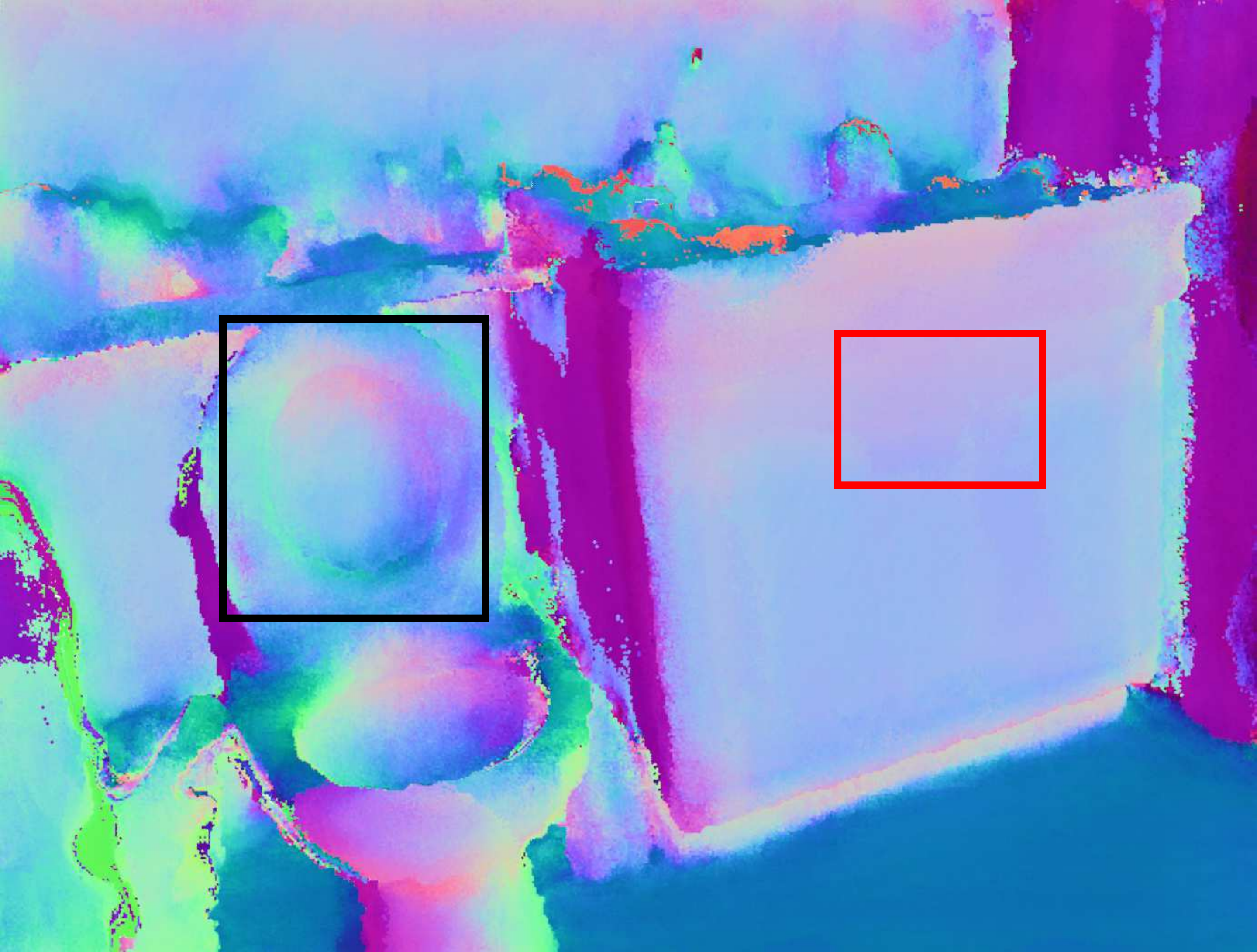}}%
	\subfigure[Initial normal (MFPS)]{\includegraphics[width=\unitT]{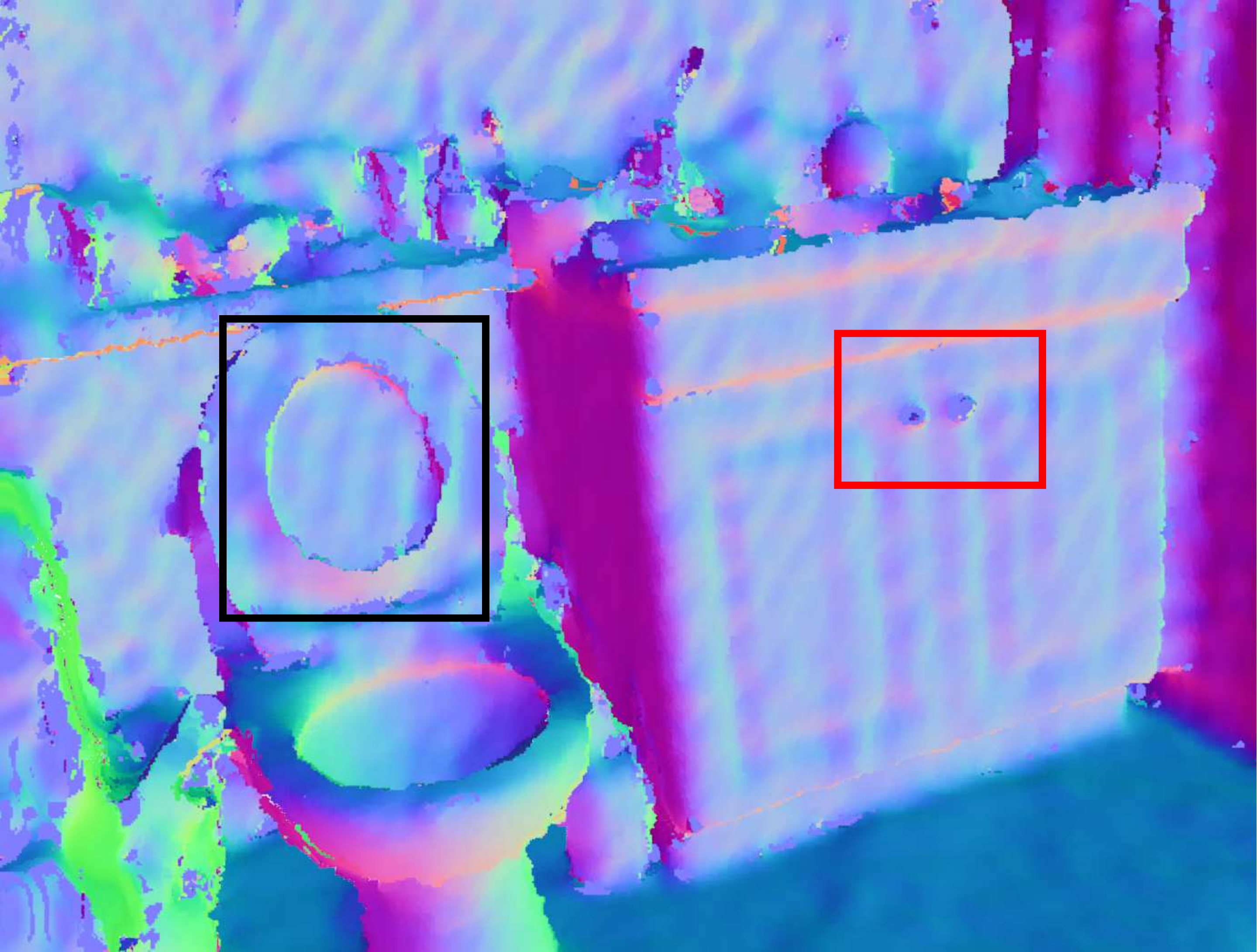}}%
	
	\vspace{-5pt}
	
	\subfigure[RGB image]{\includegraphics[width=\unitT]{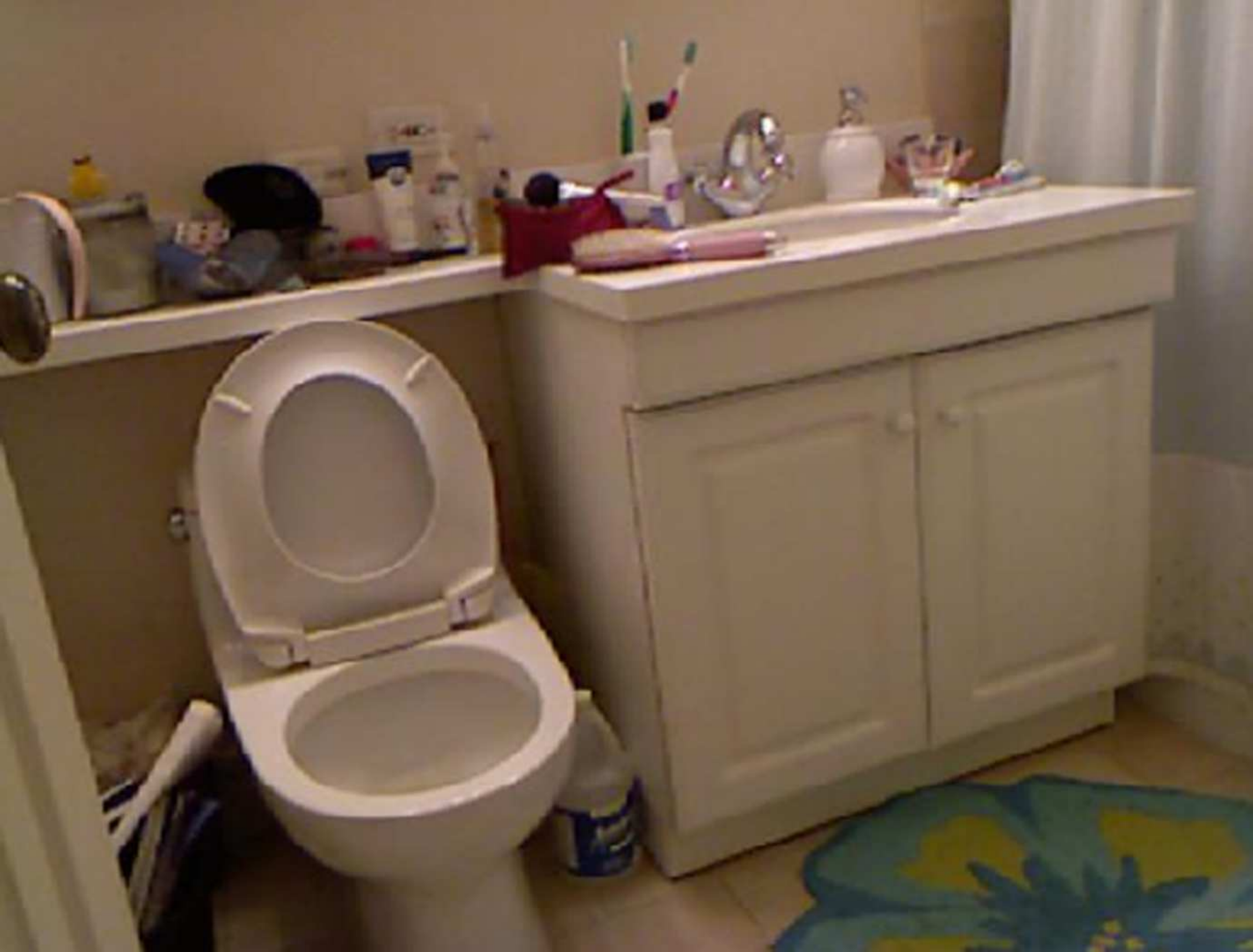}}%
	\subfigure[Nesti-Net \cite{Ben-ShabatLF19} + Ours]{\includegraphics[width=\unitT]{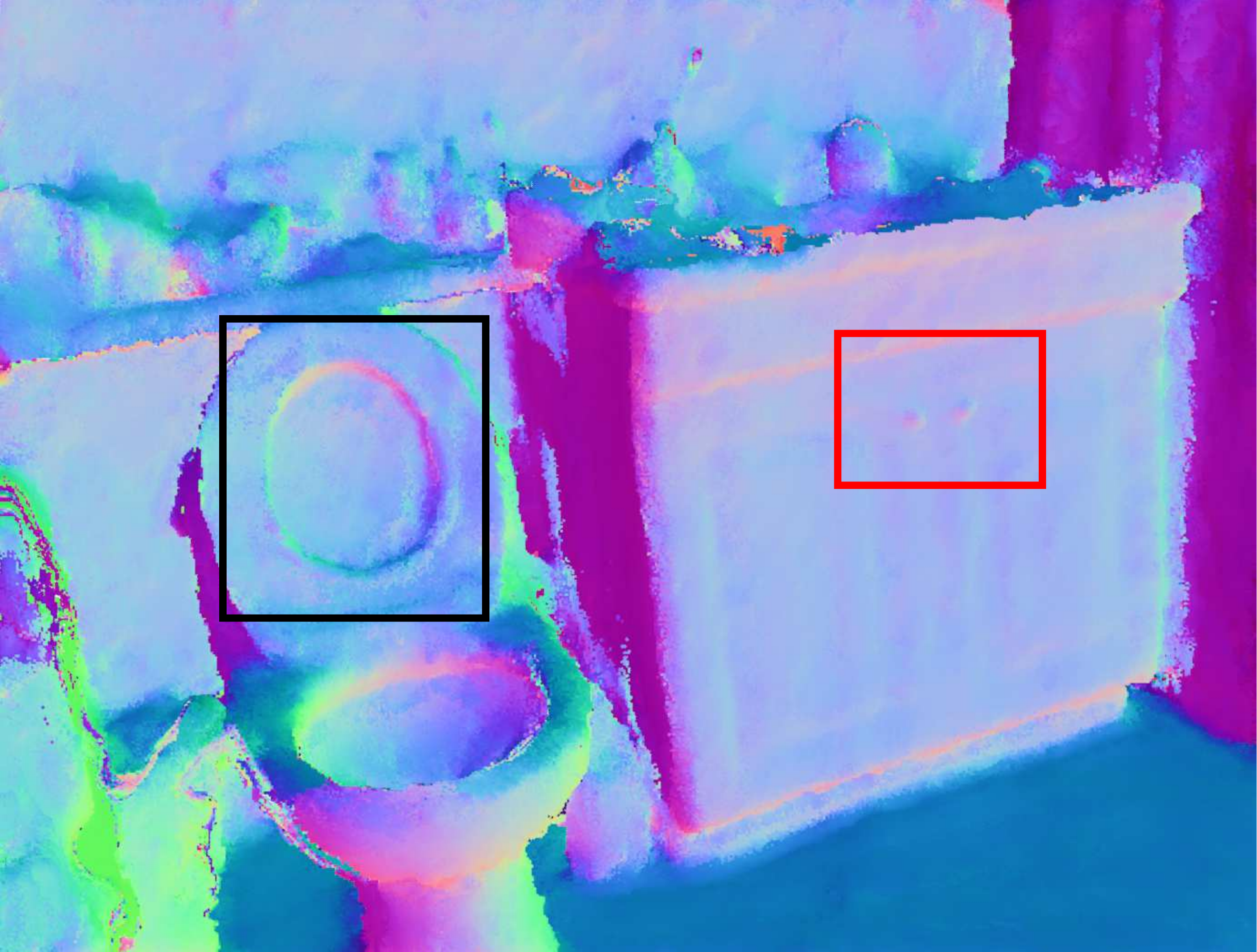}}%
	\subfigure[Ours]{\includegraphics[width=\unitT]{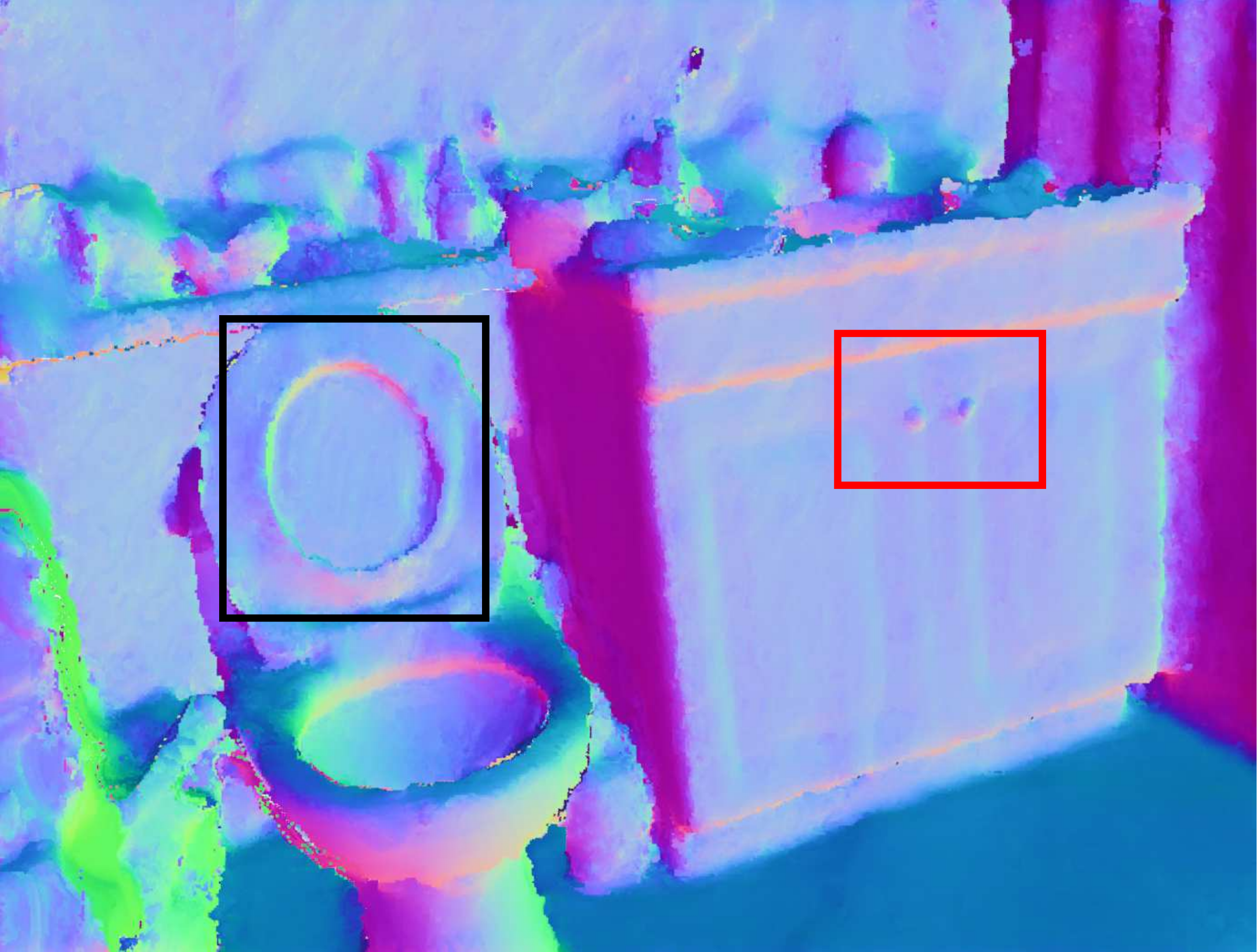}}
	\caption{Visual comparison of estimated normal results, in which normals are rendered to RGB colors. The proposed Refine-Net is better at recovering details and sharp edges. (a) is the depth image (noisy input) with high resolution zoom-ins. (d) is the original RGB image. (b) and (e) show the normal estimation result of Nesti-Net \cite{Ben-ShabatLF19} and the corresponding refined result by our network. (c) is our initial normal result, and (f) is the result of our full pipeline. Clear improvements can be observed from tiny objects (in the red bounding box) and sharp feature regions (in the black bounding box).}
	\label{fig:intro2}
\end{figure}

Recent years have witnessed considerable research efforts of learning-based normal estimation. Existing wisdom, either using the convolutional neural network (CNN) architecture \cite{boulch2016deep,Ben-ShabatLF19}, or the PointNet architecture \cite{GuerreroEtAl:PCPNet:EG:2018}, attempts to design a complete form of feature representation to predict point normals. 
However, two core problems are adverse to accurate normal estimation. 
First, while many methods \cite{GuerreroEtAl:PCPNet:EG:2018, wang2016mesh,Ben-ShabatLF19,boulch2016deep} exploit the multi-scale neighborhood size, they may ignore the limited ability of networks that are built on a single type of feature inputs. We exhibit in Fig.~\ref{fig:intro1} that several kinds of features are explored from a noisy input, such as the point coordinates of a local patch and the grid-like structures describing height distances, etc. 
Previous methods \cite{chen2019multi,GuerreroEtAl:PCPNet:EG:2018} have proved that each of the extracted feature representations can be independently modelled for the point geometric properties. We validate that these features can mutually support each other, from different perspectives, and can be jointly incorporated into the estimation of point normals. 
This incorporation of multiple features is an effort in this work. Besides, we still apply the multi-scale scheme following \cite{wang2016mesh} to promote learning robustness.
These are the two main schemes adopted in this work (see Fig.~\ref{fig:intro1}). 
Second, for severely degraded data, it is open to recover sharp features and fine details by a general mapping learned from deep neural networks without any support of geometric information. We refer the readers to Fig.~\ref{fig:intro2:Nesti} that small handles are blurred in the red bounding box where the input domain is similar to planar areas. Over-smoothed results that ignore sharp edges (in the black bounding box of Fig.~\ref{fig:intro2}) are not allowed in the accuracy-required tasks like robotic grasp and surface reconstruction.


Motivated by the aforementioned challenges, we propose a normal refinement network, namely Refine-Net, which fuses important geometric information based on a multi-feature scheme. Our main contributions are summarized as follows:

\begin{itemize}
\item The proposed Refine-Net is a normal-to-normal system, designed to refine the initial normal of each point. We introduce two additional feature representations into the refinement steps in the context of neural networks. 


\item We present a novel connection module, which applies a flexible transformation to integrate useful information learned from additional features with the initial normal.

\item We further extend Refine-Net into a normal refinement framework, with the generalization ability of both feature and normal modules.

\item We propose a geometric method, namely multi-scale fitting patch selection (MFPS), for the initial normal estimation. It serves as our own initial normal estimator, which is better at capturing geometric supports by using an elaborately-designed patch selection scheme.

\item A new point update algorithm is designed for the downstream construction tasks.

\end{itemize}



\vspace{5pt}\noindent\textbf{Difference from our conference paper.}
This work covers and extends our conference version \cite{zhou2020geometry} as follows: (i) we exploit more feature modules in the system, making a breakthrough from single-feature to a multi-feature refinement framework. Thus, our previous work can be seen as part of the network (normals\&HMPs version in Sec.~\ref{sec:eval:ablation}); (ii) we extend the connection module to three alternative choices including the newly proposed one which is more flexible in the high-dimensional space; (iii) the original geometric method is further improved, by using a novel selection solution, to determine normal directions in challenging regions of sharp edges; (iv) we demonstrate that Refine-Net can also improve previous normal estimation networks \cite{GuerreroEtAl:PCPNet:EG:2018,Ben-ShabatLF19}, by directly taking their results as input; (v) we conduct more experiments on the outdoor scene dataset \cite{serna:hal-00963812}; (vi) our results show clear improvements over Zhou et al. \cite{zhou2020geometry} in terms of normal estimation accuracy and noise-robustness.

\label{SECintro}

\section{Related work}
Normal estimation for point clouds is a long-standing problem in academic. We will review previous researches from traditional normal estimators to recent prevalent learning-based techniques.

\subsection{Traditional normal estimators}
Traditional approaches  mainly focus on exploiting low-level geometric priors.
The most well-known one is the Principal Component Analysis (PCA) \cite{hoppe1992surface}, by analyzing covariance in a local structure around a point and defining its normal as the eigenvector corresponding to the smallest eigenvalue. Following this work, a lot of improvements have been proposed \cite{MitraN03, cazals2005estimating, GuennebaudG07} to increase estimation precision. In particular, Mitra et al. \cite{MitraN03} analyze the effects of neighborhood size, curvature, sampling density, and noise level for estimating normals. By carefully analyzing these local information, this method can find a suitable neighborhood size for each point. Another kind of normal estimation methods is based on Voronoi cells \cite{amenta1999surface, dey2006provable, AlliezCTD07, merigot2010voronoi}. However, this kind of methods cannot well estimate the normals of points near/on sharp features. Some other works are based on the improvement of a preliminary normal result. Typical approaches use adaptive moving least squares  \cite{alexa2001point} and robust local kernel regression \cite{oztireli2009feature} to estimate normals as the gradient of an implicit surface which fits the local points and their preliminary normals.
Later, based on the observation that neighbors belonging to different surface patches often result in large fitting errors, recent works are dedicated to selecting a plane approximating points from the same surface patch to estimate normals \cite{li2010robust, wang2013adaptive, zhang2018multi}. Under the assumptions that: i) surfaces are commonly composed of piecewise flat patches, and ii) geometric features are sparsely distributed over the entire shape, sparsity-based methods \cite{avron2010L1, sun2015denoising, chen2019structure} show impressive results for those CAD-like models, especially in sharp feature preservation. Low-rank based methods \cite{chen2019structure,wei2018mesh,lu2020low,cad/LiZFXWWH20} explore the non-local geometric similarity to generate better normal fields on 3D surfaces. Some other methods, like Hough Transform \cite{boulch2012fast}, also yield pleasing results.

Overall, the geometric methods often resolve the normal estimation problem under certain assumptions or specific observations. Besides, these approaches contain a number of parameters, which require careful trial-and-error tuning to obtain decent results, especially for complex models.

\subsection{Learning-based normal estimators}
Recently, learning-based methods have gradually shown their power in the task of normal estimation. We will review the related prior works in two main types: CNN-based and PointNet-based networks.

\subsubsection{CNN-based methods} Boulch et al. \cite{boulch2016deep} propose to project a discrete Hough space representing normal directions onto a structure amenable to CNN-based deep networks. Similarly, Roveri et al. \cite{RoveriOPG18} define a grid-like regular input to learn ideal normal results. Ben-Shabat et al. \cite{Ben-ShabatLF19} propose to parameterize the local field using a multi-scale 3D modified Fisher vector which serves as the input to a deep 3D CNN architecture. In addition, a scale manager module is designed for selecting a suitable neighborhood size in a data-driven manner. As we investigate above, how to map the unstructured point data into a regular domain, is the key to these CNN-based methods.

\subsubsection{PointNet-based methods} Another point cloud learning framework, namely PointNet \cite{QiSMG17}, becomes increasingly popular in 3D domains, thanks to its ability to directly manipulate raw point cloud data. Inspired by this, Guerrero et al. \cite{GuerreroEtAl:PCPNet:EG:2018} 
apply the PointNet architecture in a patch-based multi-scale form to predict normals. With a data-driven manner, this method allows to replace the difficult and error-prone manual tuning of parameters presented in the majority of existing traditional techniques. More recently, Zhou et al. \cite{zhou2020normal} introduce an extra feature constraint mechanism and a multi-scale neighborhood selection strategy to estimate normals for 3D point clouds. Considering the typical problem that PointNet is not suitable for encoding local structures of a point cloud, Hashimoto et al. \cite{HashimotoS19} propose a two-branch network that extracts both local and spatial features required for learning. 
The aforementioned learning-based deep networks try to directly regress point normals from a large dataset. However, they have downsides in terms of robustness to small input changes and interpretability. Contrary to these methods, Lenssen et al. \cite{lenssen2020deep} propose to embed deep learning in least-squares plane fitting, which leads to a fast and accurate algorithm. Specifically, a graph neural network is utilized to iteratively parameterize an adaptive anisotropic kernel that produces point weights for least-squares plane fitting.

In general, the aforementioned learning-based methods have two main drawbacks: 1) the weak ability of detail perceiving in/near challenging feature regions; 2) an existing learning-based method often exploits one single type of feature representation, neglecting the useful information from multiple types of features that can be combined together to co-support the normal estimation task. In this work, we focus on deploying a multi-feature scheme to explore different geometric supports in our normal refinement system.

\label{SECrelated}

\begin{figure*}
	\begin{center}
		\includegraphics[width=0.95\linewidth]{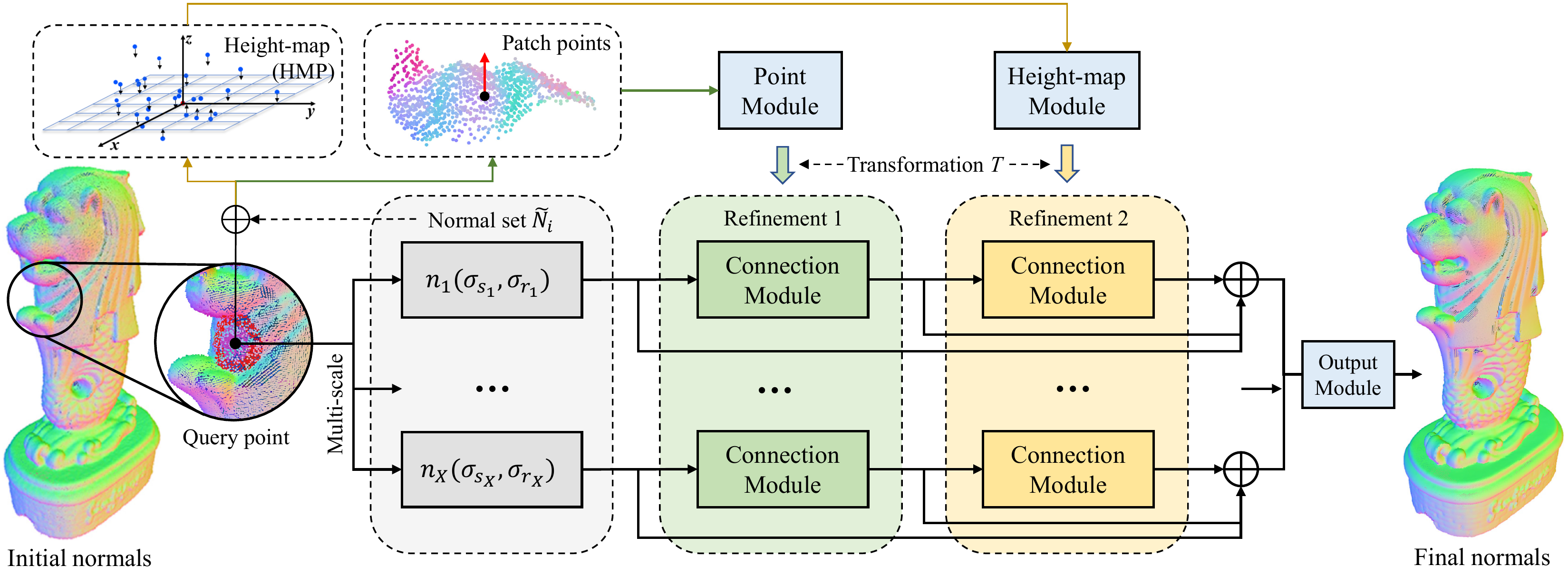} 	
	\end{center}
	\caption{The overall architecture of Refine-Net. For a target point, its initial normal is first extended to a multi-scale version using bilateral filtering. 
	Then, in each branch, the normal feature is extracted and refined using a multi-feature scheme: 1) \textbf{Point Module} introduces the additional point feature from a local patch to produce a new feature in the first refinement; 2) \textbf{Height-map Module} uses the constructed height-maps to further improve the normal; 3) \textbf{Connection Module} is designed to combine the normal feature and the learned transformation $T$ in each step of the refinement. Finally, the outputs from all branches are collected to predict the final normal.}
	\label{fig:architecture}
\end{figure*}

\section{Refine-Net}
\label{sec:refnet}
In this section, we first describe the overall architecture of Refine-Net (Sec.~\ref{sec:refnet:overview}), followed by detailed elaboration of the component modules (Sec.~\ref{sec:refnet:filter}-Sec.~\ref{sec:refnet:connect}). Furthermore, we show that Refine-Net can be well extended to a generic normal refinement framework in Sec.~\ref{sec:refnet:framework}.

\subsection{Architecture overview}
\label{sec:refnet:overview}
The overall architecture of Refine-Net is shown in Fig.~\ref{fig:architecture}, which is designed to refine the initial normals by introducing multiple types of feature representations from an input point cloud. 
The initial normal of each point (Sec.~\ref{sec:geometric}) serves as the main feature in the network, and is processed by a two-step refinement for the final normal. First, we use bilateral filtering to extend the initial normal into a multi-scale version of normal vectors in order to expand the receptive field of local points (Sec.~\ref{sec:refnet:filter}). Then, the filtered normals are fed into individual branches for a multi-feature refinement. 
Based on the extracted information, an additional point feature is introduced, 
which fuses with the normal in the connection module, in order to apply a rough refinement (Sec.~\ref{sec:refnet:point}). 
Similarly, in the second step, we use the height-map feature, which is generated according to point positions, to further fine-tune the intermediate result (Sec.~\ref{sec:refnet:hmp}). 
The connection modules which incorporate different features into the refinement are discussed in Sec.~\ref{sec:refnet:connect}. 
Finally, Refine-Net collects outputs from all branches and obtains the final normal. We show in Sec.~\ref{sec:refnet:framework} that our network is a generic normal refinement framework, which can be potentially combined with any other initial normal and feature module.

\subsection{Multi-scale normal filtering}
\label{sec:refnet:filter}
The network starts from the initial normals $\mathbf{N} = \{\hat{n}_i\}_{i=1}^N$ of noisy points $\mathbf{P} = \{p_i\}_{i=1}^N$, where $N$ is the number of points in the input point cloud. We will discuss this preliminary normal estimation later in Sec.~\ref{sec:geometric}. Previous attempts \cite{wang2016mesh,cad/WangHWWXQ19,cgf/WeiGHXZKWQ19,li2020dnf,junjiecao2021} that apply deep neural networks in the normal field have shown impressive results of recovering ground-truth normals from the corresponding noisy inputs. We consider the initial normal of each point as the main feature in the recovery, and extend it to multiple scales using bilateral normal filtering.

\vspace{5pt}\noindent\textbf{Multi-scale bilateral filters.}
We first give a brief introduction to the commonly-used bilateral filter \cite{jones2004normal}. Given the centered point $p_i$ and its neighborhood $\mathcal{N}(i)$, we can calculate the filtered normal as:
\begin{equation}
n_i' = \Lambda(\sum_{{p}_{j}\in \mathcal{N}(i)} W_s(||p_i-p_j||) W_r(||\hat{n}_i-\hat{n}_j||) \hat{n}_j),
\end{equation}
where $\Lambda(*)$ is the vector normalization function, and $W_s$ and $W_r$ are Gaussian weight functions, i.e., $W_{\sigma}(x) = \exp(-x^2/(2\sigma^2))$, representing the spatial similarity and normal similarity respectively between a pair of points. Here, standard deviations $\sigma_s$ and $\sigma_r$ are used. Instead of carefully tuning these parameters, we apply a multi-scale filtering \cite{wang2016mesh} and obtain filtered normals from two designed parameter sets $P_s = \{\sigma_{s_1}, \sigma_{s_2}, ...\}$ and $P_r = \{\sigma_{r_1}, \sigma_{r_2}, ...\}$ by using every combination of the two parameters. Thus, together with the initial one, we get a total of $X$ normals in the normal set of $p_i$, denoted as $\widetilde{N}_i = \{ n_1, n_2, ... , n_X \}$. Each filtered normal will be fed into an individual branch for the following refinement operations.

\vspace{5pt}\noindent\textbf{Point-wise normal reorientation.}
Before diving into these branches, we consider the point-wise normal reorientation in order to lower the learning difficulty. To this end, we apply a global rotation to the filtered normals to make them invariant to rigid transformation. 
%
Specifically, we first compute the normal tensor of point $p_i$:
\begin{equation}
    T_i = \sum_{j=1}^X n_j \times n_j^T, n_j \in \widetilde{N}_i.
\end{equation}
$T_i$ fuses the normal characteristics of $p_i$'s local neighbors and can be decomposed as:
\begin{equation}
    T_i = \lambda_1 {e}_1 {e}_1^T + \lambda_2 {e}_2 {e}_2^T + \lambda_3 {e}_3 {e}_3^T,
\end{equation}
where $\lambda_1 \leq \lambda_2 \leq \lambda_3$ are the eigenvalues and ${e}_1, {e}_2, {e}_3$ are the corresponding eigenvectors. Then, a rotation matrix $R_i$ can be constructed as $[{e}_1, {e}_2, {e}_3]$. $R_i$ rotates all normals in $\widetilde{N}_i$ to a local frame near the Z-axis of the global frame. 
Furthermore, if they are in the direction of negative Z-axis, we reverse them to the positive one. The ground-truth normal is also applied by the same rotation matrix to match the input. Such a pre-processing operation explicitly reorients the normals of each point to the same direction in order to tackle the inconsistency of global directions between input samples.

\vspace{5pt}\noindent\textbf{Normal-based clustering.}
Before training, we partition the training samples into $K_c$ clusters via the k-means algorithm \cite{wang2016mesh}. It is performed on the filtered normals $\widetilde{N}_i$ to gather similar inputs together in one group. For each cluster, we train a separate Refine-Net to recover the ground-truth normals. In this way, the network can focus on a specific kind of geometric features and intensively refine this type of normals. In the runtime test stage, a single point sample is distributed into one of these clusters in terms of $||\widetilde{N}_i - C_l|| \leq ||\widetilde{N}_i - C_k||, \forall k$. Here, $C_k$ is the center of the k-th cluster. We evaluate the performance of this cluster scheme in Sec.~\ref{sec:eval:ablation}.

\subsection{Point module}
\label{sec:refnet:point}
We introduce point features to prepare for the refinement. For a target point $p_i$, a local patch $\mathbb{P}_i$ is defined as the neighboring point set $\{p_1, p_2, ..., p_n\}$ which is centered at $p_i$. We propose to extract useful information from the patch points to refine the normal. To simplify the input space, the patch $\mathbb{P}_i$ is translated to origin and normalized by the patch radius. In order to match the filtered normals obtained above, $\mathbb{P}_i$ is also applied by the corresponding matrix $R_i$, rotating the patch to the same direction. The rotation constraint is nontrivial \cite{GuerreroEtAl:PCPNet:EG:2018}, and in this case we can use the global directions extracted from the initial normals instead of computing directly from raw points. 

To process the point set, we use the PointNet architecture \cite{QiSMG17} which is denoted as Point Module in Fig.~\ref{fig:architecture}. PointNet suggests to apply a set of shared functions independently on each point and collect the global feature using a symmetric function:
\begin{equation}
f(\mathbb{P}_i) = g(h(p_1), ..., h(p_n)),
\end{equation}
where $g(*)$ is a symmetric function, i.e., max-pooling or sum, and $h(*)$ denotes a multi-layer perceptron network (MLP). Here, the input $p_j \in \mathbb{R}^3$ only contains the $(x, y, z)$ coordinates. Then, our point module takes the global feature and processes it using several fully-connected layers which output a $d_1$ dimensional feature vector. The output dimension $d_1$ is determined according to the design of connection module, and we will discuss several choices later in Sec.~\ref{sec:refnet:connect}.


\subsection{Height-map module}
\label{sec:refnet:hmp}
The height-map (HMP) feature is constructed related to both the point normals and coordinates which maps the local points onto a grid-like structure.
The point $p_i$ itself as well as one of its filtered normals $n_t \in \widetilde{N}_i$ defines a tangent plane, and the associated HMP is built on this plane. Suppose that we build a matrix with $m\times m$ bins that describes the local point positions, and the center of the matrix is located at $p_i$. Similar as \cite{chen2019multi}, we fill each bin by weightedly averaging the height distances of points in a ball neighborhood $\mathcal{N}_{ball}(b_j)$ of bin center $b_j$:
\begin{equation}
v_{b_j} = \frac{\sum_{p_k \in \mathcal{N}_{ball}(b_j)} w(b_j, p_k) H(T(p_i, n_t), p_k)}{\sum_{p_k}w(b_j, p_k)},
\end{equation}
where $T(p_i, n_t)$ is the tangent plane defined by $p_i$ and $n_t$, $H(T(p_i, n_t), p_k)$ returns the signed distance from $p_k$ to the plane and $w(b_j, p_k) = \exp(- \frac{||b_j-p_k||^2}{\sigma_d^2})$ is a spatial Gaussian weight function in which $\sigma_d$ is the residual bandwidth. Assembling all HMPs yields each point's height-map features $\{\textbf{HMP}_1, \textbf{HMP}_2, ..., \textbf{HMP}_X \}$. Note that, for the defined planes associated with HMPs, their Z-axes of local coordinate systems are also rotated by the matrix $R_i$. Thus, the directions of X-axes and Y-axes are consistent among all samples.

The computed representation is fed into the height-map module (refinement 2 in Fig.~\ref{fig:architecture}). The $X$ HMPs are concatenated and processed using several convolutional layers and max-pooling layers, followed by fully-connected layers (see Tab.~\ref{table:network_architecture}). Similarly, the output of this module is a $d_2$ dimensional vector prepared for refinement.

\subsection{Connection module}
\label{sec:refnet:connect}
In the aforementioned cases, we introduce additional features to refine the normal in each branch. The main discussion behind is how to establish a connection, e.g., between the normal vector $V \in \mathbb{R}^3$ and the point feature $F$ (connection module of refinement 1 in Fig.~\ref{fig:architecture}), since they belong to distinct data types. The key idea is to use the point feature $F$ to learn an intermediate transformation matrix $T$, which should match the size of $V$, and apply matrix multiplication to obtain the higher-level feature $Y = T \cdot V$ (a new normal feature). Several choices to construct the transformation target $T$ are discussed below.

\vspace{5pt}\noindent\textbf{Rotation matrix.}
The simplest solution is to apply 3D rotation to normal vectors for the refinement.
In this way, we can learn a rotation matrix from the input feature $F$ which parameterizes the local points related to $V$. Thus, the vector $V$ can be rotated in the guidance of the point feature. To be more specific, we use the point module to learn a quaternion, which means $d_1=4$ is set in the output dimension, and construct a $3 \times 3$ rotation matrix $T$ from the unit quaternion.

\vspace{5pt}\noindent\textbf{Transformation matrix.}
Another option that generates a transformation matrix does not require a quaternion for rotating. Intuitively, we set the output dimension as $d_1=9$ and simply reshape it to a $3 \times 3$ matrix, which is a more direct solution. Note that, this does not act the same as a rotation operator. This construction can be seen as a common 3D transformation applied to the input normal vector $V$ which is not designed strictly for rotating.

\vspace{5pt}\noindent\textbf{Weight matrix.}
More generally, we construct a weight matrix $T=[t_1, t_2, ..., t_p]^T\in \mathbb{R}^{p\times3}$, in which each $t_i = [w_{x,i}, w_{y,i}, w_{z,i}]^T$ contains learned weights corresponding to the 3D coordinate of $V$. Thus, the transformation can be more flexible, producing output features with arbitrary dimensions. Through matrix multiplication, a weight matrix can extract more information from the input feature compared with a single $3 \times 3$ matrix that is applied within the limits of 3D space. We set $d_1 = p \times 3$ in the output dimension, and reshape it to a $(p, 3)$ matrix $T$. This is the option that we adopt in this work.

\begin{figure}[t]
	\begin{center}
		\includegraphics[width=0.99\linewidth]{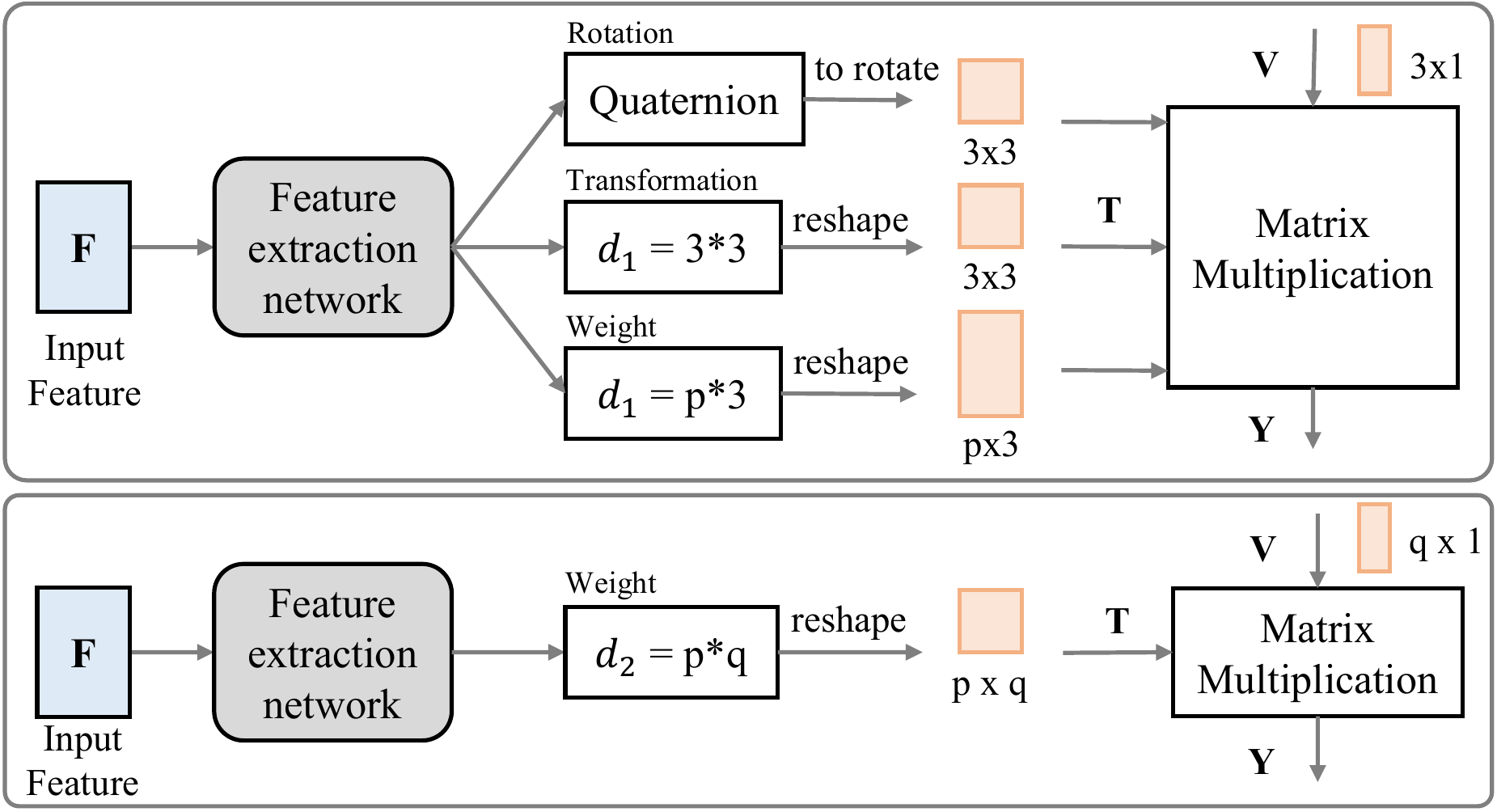} 	
	\end{center}
	\caption{Structure of our connection module. Three alternative choices (top) for constructing a transformation matrix $T$ are provided when inputting a normal vector $V \in \mathbb{R}^3$ (refinement 1). If the input is a normal feature $V \in \mathbb{R}^q$ (refinement 2), we propose to leverage the weight matrix option (bottom).}
	\label{fig:connect}
\end{figure}

To be more specific, in refinement 1 of Fig.~\ref{fig:architecture}, we set $d_1 = 64 \times 3$ which converts the filtered normal into a new normal feature of dimension 64 (see the top part of Fig.~\ref{fig:connect}). Similarly, a second refinement is applied by the height-map module with $d_2 = 64 \times 64$ (see the bottom part of Fig.~\ref{fig:connect}). The two features together with their filtered normal are stacked for the output.

By inference, the choices designed above can also provide a solid connection in other tasks with similar inputs. The weight matrix option receives two inputs with different sizes and can produce a high-dimensional output. We find in our experiments that the way of simply concatenating two distinct feature vectors leads to an incompatible connection which may oppositely limit the learning ability. The attempt to use matrix multiplication in the connection module shows its advantages compared with other learning schemes (see Sec.~\ref{sec:eval:ablation}).

\vspace{5pt}\noindent\textbf{Output module and loss function.}
The refined features from all branches are collected for the output. Refine-Net applies several fully-connected layers and outputs the final predicted normal $N^* = (N_x, N_y, N_z)$. Batch normalization and ReLU are included in each layer of this module.

We train the network by minimizing the MSE loss between the output normal $N^*$ and its ground truth $\hat{N}$:
\begin{equation}
Loss = || \Lambda(N^*) - \hat{N} ||^2 + \lambda E_{reg},
\end{equation}
where $\Lambda(*)$ is the vector normalization function, and $E_{reg}$ is the commonly-used $L_2$ regularization to avoid overfitting, where $\lambda = 0.02$ is used.

\begin{figure}[t]
	\begin{center}
		\includegraphics[width=0.32\linewidth]{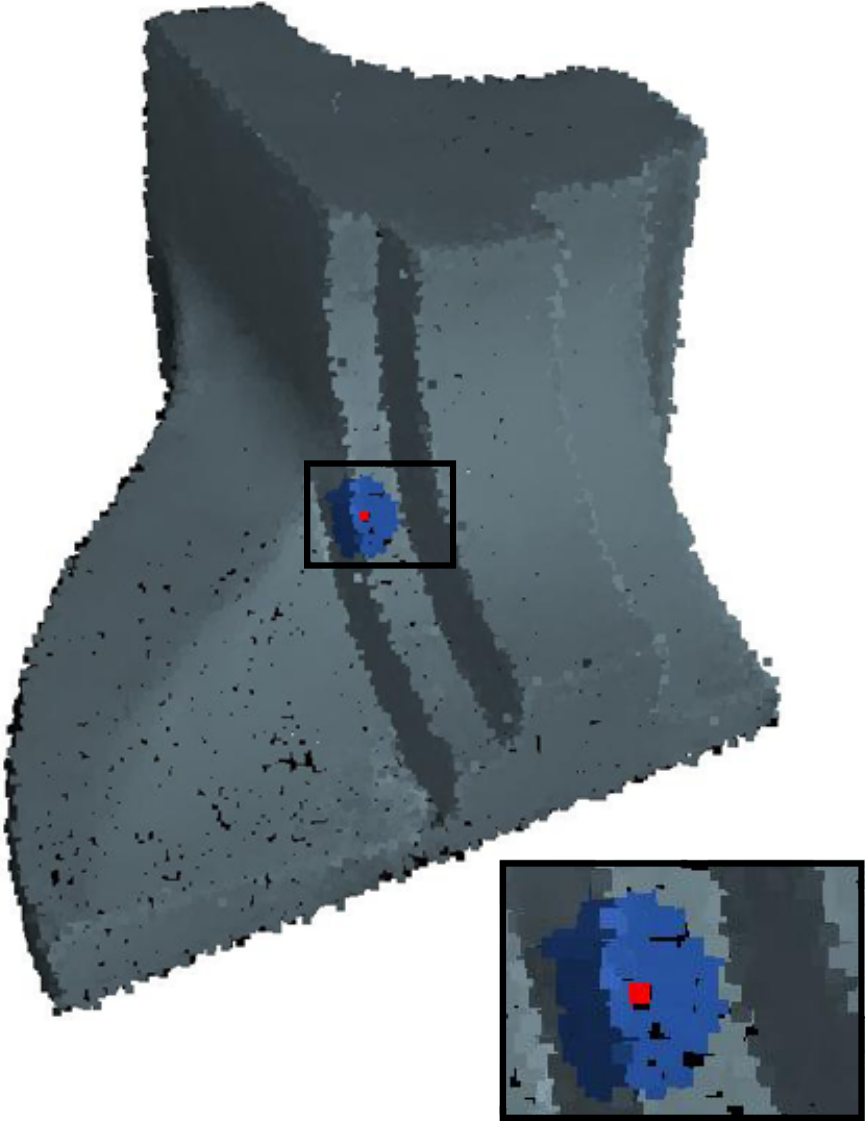}%
		\hspace{3pt}\includegraphics[width=0.32\linewidth]{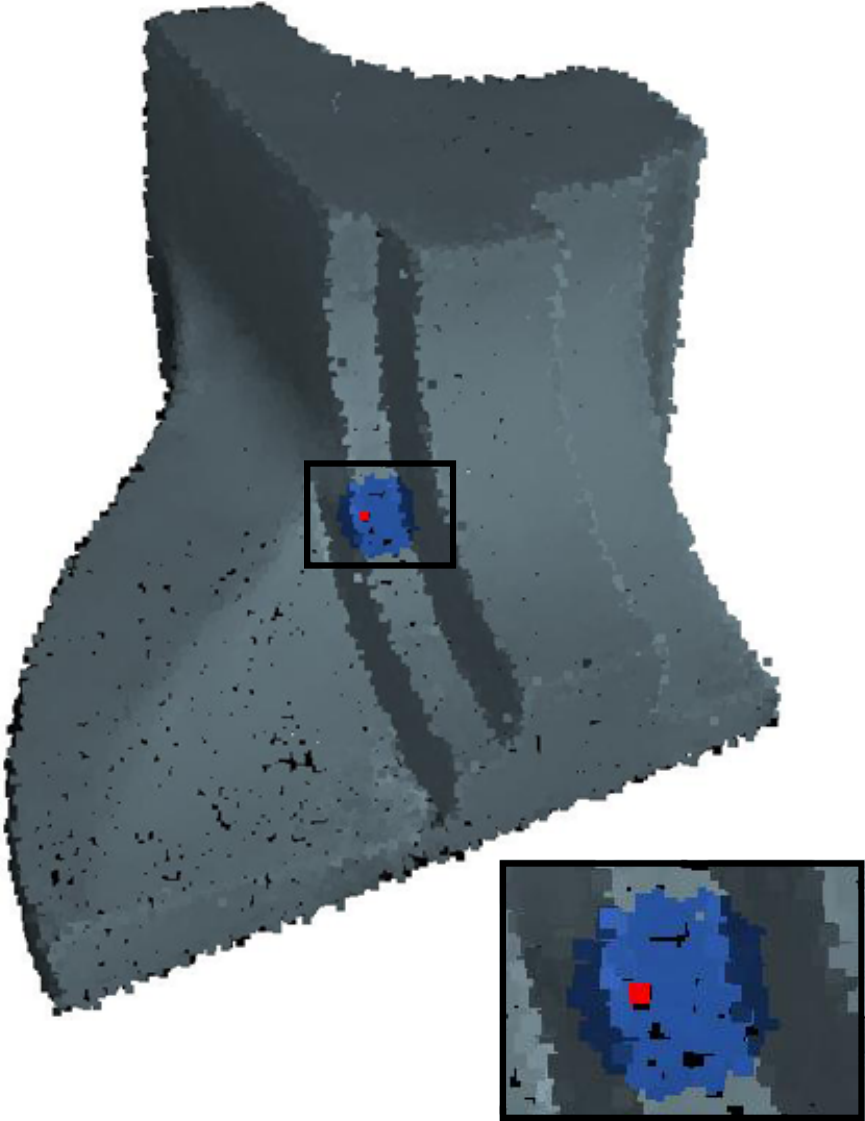}%
		\hspace{3pt}\includegraphics[width=0.32\linewidth]{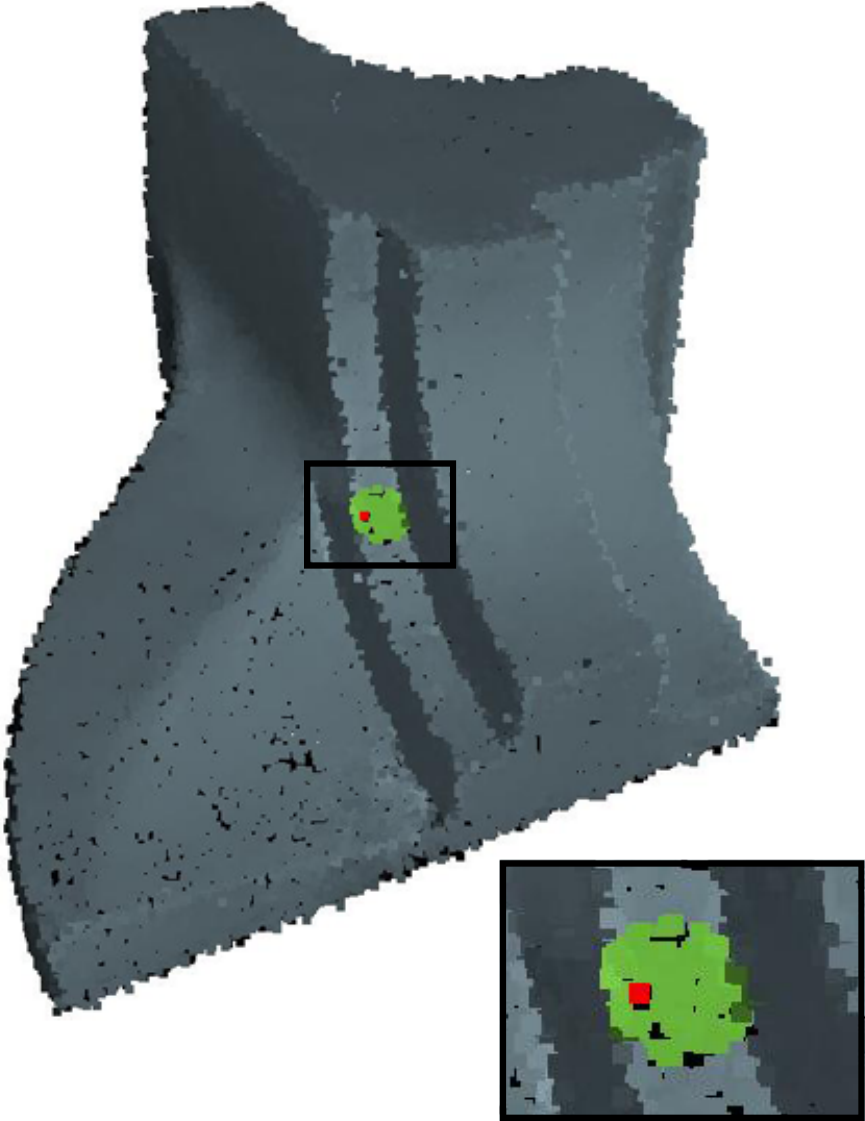}
		
	\end{center}
	\caption{Three candidate patches around the target point (red point) are shown. The patch centered at the target point (left) is unsuitable for the estimation since it contains undesired points from other sides of the edge. Instead, we propose to select a better neighboring patch (right) which is more consistent for detecting the underlying surface. The multi-scale scheme makes more choices available to further promote estimation robustness.}
	\label{fig:MFPS}
\end{figure}

\subsection{Normal refinement framework}
\label{sec:refnet:framework}
Refine-Net uses two feature modules to refine the initial normal (normal module). When extended to a general normal refinement framework, these modules can be potentially replaced. That is, the normals from other methods can be further refined, and other feature modules can be developed and incorporated in the network. They are included by the connection modules in order to obtain the predicted results.

In this work, we introduce both point and height-map features in the network. We show that a single feature module is already enough for a promising result in Sec.~\ref{sec:eval:ablation}. Further, more information related to the point geometric details can be absorbed by appending other feature modules in the branches. 

The initial normal choice has a significant influence on the recovery results. We choose to use a traditional estimator, namely MFPS in Sec.~\ref{sec:geometric}, which can focus more on the geometric information compared with learning-based techniques. It is designed to preserve sharp features and fine details which are hard to recover by learning a straightforward mapping. Refine-Net that is trained on other typical initial normals can also produce a well-improved normal result.
\label{SECrefnet}


\section{Multi-scale fitting patch selection}
\label{sec:geometric}
As aforementioned, we choose to design a traditional geometric estimator for the initial normal estimation.
Before computing the initial normal for each point, we first apply the local covariance analysis (Sec. 4.2 in \cite{zhang2013point}), to classify all points into candidate points (near sharp features) and smooth points (far from sharp features). The normals of smooth points can be simply computed by PCA.

To estimate the normals of candidate points, one common strategy \cite{li2010robust,zhang2018multi} is to randomly select three non-collinear points to construct a set of candidate planes and pick the one that best describes the underlying surface patch. Typically, let's define the local patch $Q_j^k$ (the $k$-nearest points of the point $p_j$) as the neighborhood of $p_j$. Then, the selected plane on this patch can be determined by the following objective function:
\begin{equation}
E_{Q_j^k}(\theta) =  \frac{1}{|Q_j^k|} \sum_{p_k \in Q_j^k} W_{\sigma_{j}}(p_k, \theta), \label{equ:E_plane}
\end{equation}
\begin{equation}
\theta_{Q_j^k}^{*} = \mathop{\arg\max}_{\theta} E_{Q_j^k}(\theta), \label{equ:patch_plane}
\end{equation}
where $W_{\sigma_{j}}(p_k, \theta) = \exp(- r_{k,\theta}^2 / \sigma_{j}^2)$ is the Gaussian function in which $r_{k,\theta}$ denotes the distance from the point $p_k$ to the plane $\theta$, and $\sigma_{j}$ is the residual bandwidth of $p_j$. Thus, $\theta_{Q_j^k}^{*}$ is the selected plane, and the normal of this plane can be seen as the patch normal of $Q_j^k$.


\begin{figure}[t]
	\vspace{-5pt}
	\begin{center}
		\subfigure[Ground truth]{\label{fig:fps:gt}\includegraphics[width=0.33\linewidth]{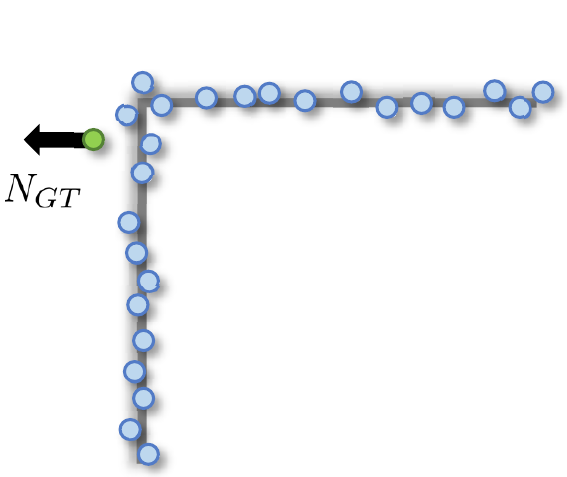}}%
		\subfigure[Anisotropic patch 1]{\label{fig:fps:ani1}\includegraphics[width=0.33\linewidth]{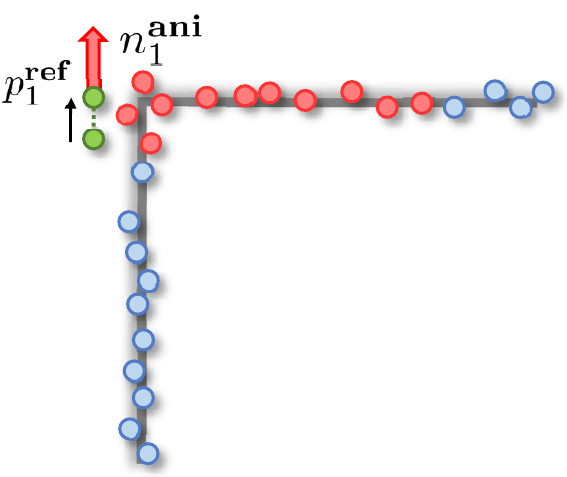}}%
		\subfigure[Anisotropic patch 2]{\label{fig:fps:ani2}\includegraphics[width=0.33\linewidth]{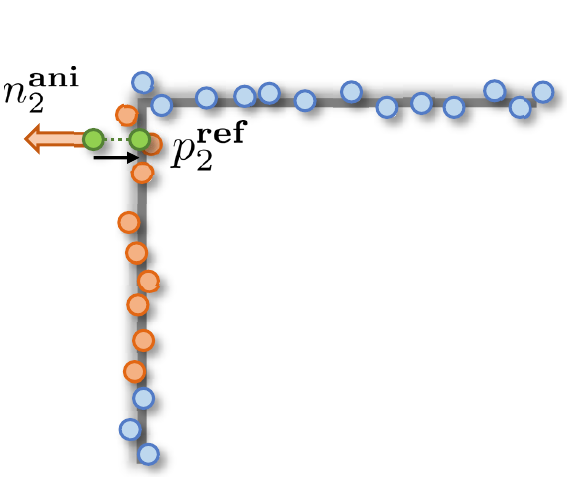}}
		
	\end{center}
	\caption{Selection for the best-fitting normal of the green point. (a) represents the input points near a sharp edge and the ground truth normal $N_{GT}$. (b) and (c) describe two selected anisotropic normals and the corresponding patch points. The black arrow denotes the signed distance from $p_i$ to $p^{\textbf{ref}}_j$, which determines the better one (patch 2).}
	\label{fig:fps}
\end{figure}

\subsection{Multi-scale scheme}
If $p_j$ lies in the neighborhood of an edge/intersection, the corresponding patch $Q_j^k$ might be corrupted by points from other sides, over-smoothing the resulting estimated normal of $p_j$ (see Fig.~\ref{fig:MFPS} left).
Considering this, we propose to select a more consistent and suitable neighborhood which can make it easier to determine the normal. For each candidate point $p_i$, our method tries to find the best fitting patch that contains $p_i$.
Please note that normals of smooth points are already computed by PCA.

First, the patch size is extended to a multi-scale parameter set $K = \{k_1, k_2, ..., k_{\alpha}\}$. For each point $p_j$ in a point cloud, every scale size $k_t \in K$ defines a patch $Q_j^{t}$ similarly as above, and we compute the plane of this patch by Eq.~\ref{equ:patch_plane}.
Then, for a candidate point $p_i$, we denote $S_{i} = \{Q_j^t | p_i \in Q_j^t \} $ as all patches containing $p_i$. We aim to search for, among all patches in $S_i$, the most isotropic one and to use the normal of the plane of this patch as the estimated initial normal of $p_i$. 
Fig.~\ref{fig:MFPS} (right) shows a better neighboring patch to determine the normal on sharp edges, rather than the one centered at the target point (Fig.~\ref{fig:MFPS} left). Moreover, the multi-scale scheme provides more choices for complicated regions where a single-scale size cannot fit in structures with varying point densities (Fig.~\ref{fig:MFPS} middle). The scale parameter is set as $K = 50, 100, 150$ by default in our experiments.  
The consistency of any patch $Q_j^t$ can be measured as:
\begin{equation}
\mathcal D(Q_j^t) = E_{Q_j^t}(\theta_{Q_j^t}^{*}) \eta(k_t), \label{equ:point_plane}
\end{equation}
\begin{equation}
\eta(k_t) = \beta + (1-\beta) \frac{k_t-k_{min}}{k_{max}-k_{min}},
\end{equation}
where $E_{Q_j^t}(\theta_{Q_j^t}^{*})$ comes from Eq.~\ref{equ:E_plane} and $k_t \in K$. The larger $ E_{Q_j^t}(\theta_{Q_j^t}^{*})$ is, the smaller the plane fitting error is, and $\eta(k_t)$ is a tradeoff parameter used to punish relatively small patches because larger ones are preferred in noisy areas. We set $\beta=0.9$ empirically. The measurement values are sorted, preparing for the following patch selection. 

\begin{table*}
	\centering
	\footnotesize
	\setlength{\tabcolsep}{3.5mm}
	\caption{Normal estimation results on the synthetic PCPNet dataset \cite{GuerreroEtAl:PCPNet:EG:2018}, evaluated as angular RMS errors.}
	\begin{tabular}{c |cccc|cc|c} 
		\toprule[1pt]
		Category & \multicolumn{4}{|c|}{Noise $\delta$} & \multicolumn{2}{|c|}{Density} & \multicolumn{1}{|c}{Average}  \\
		& None & 0.00125 & 0.006 & 0.012 & Gradient & Stripes & \\ 
		\midrule[0.3pt]
		\midrule[0.3pt]
		PCA \cite{hoppe1992surface} 					& 12.29  & 12.87 & 18.38  & 27.5 & 12.81  & 13.66  & 16.25  \\
		Jet \cite{cazals2005estimating} & 12.23 & 12.84 & 18.33 & 27.68 & 13.39 & 13.13 & 16.29 \\
		HoughCNN-ss \cite{boulch2016deep} & 10.23 & 11.62 & 22.66 & 33.39 & 12.47 & 11.02 & 16.9 \\
		HoughCNN-ms \cite{boulch2016deep} & 10.02 & 11.51 & 23.36 & 36.7 & 10.67 & 11.95 & 17.37 \\
		PCPNet-ss \cite{GuerreroEtAl:PCPNet:EG:2018} & 9.68 & 11.46 & 18.26 & 22.8 & 11.74 & 13.42 & 14.56 \\
		PCPNet-ms \cite{GuerreroEtAl:PCPNet:EG:2018} & 9.62 & 11.37 & 18.87 & 23.28 & 11.7 & 11.16 & 14.34 \\
		Nesti-Net \cite{Ben-ShabatLF19} & 6.99 & 10.11 & 17.63 & 22.28 & 8.47 & 9.00 & 12.41 \\
		Lenssen et al. \cite{lenssen2020deep} & 6.72 & 9.95 & 17.18 & \textbf{21.96} & 7.73 & 7.51 & 11.84 \\
		DeepFit \cite{ben2020deepfit} & 6.51 & 9.21 & 16.72 & 23.12 & 7.31 & 7.92 & 11.8 \\
		\midrule[0.3pt]
		MFPS & 7.22 & 11.19 & 17.91 & 24.07 & 7.27 & 7.87 & 12.58 \\
		Our full pipeline & \textbf{6.27} & \textbf{9.18} & \textbf{16.59} & 22.57 & \textbf{6.61} & \textbf{7.02} & \textbf{11.37} \\
		\bottomrule[1pt]
	\end{tabular}
	\label{table:pcpnet}
\end{table*}

\subsection{Fitting patch selection}
Since $S_i$ may contain patches sampled on different sides of the intersection, a decision should be made to abandon undesired ones (Fig.~\ref{fig:fps:ani1}).
To this end, we first select several anisotropic patches from $S_i$ and pick the one that is fitting to the candidate point $p_i$ (Fig.~\ref{fig:fps:ani2}). Specifically, starting from the largest $\mathcal D(Q_j^t)$, our method adds a new patch $Q^* \in S_i$ and its corresponding normal $n^*$ into the anisotropic patch set $A_i = \{(Q_1^{\textbf{ani}}, n_1^{\textbf{ani}}), (Q_2^{\textbf{ani}}, n_2^{\textbf{ani}}), ...\}$, if $D(n^*, n_j^{\textbf{ani}}) > w_t, \forall j$ where the metric $D$ is calculated as the angular error of two normals. $w_t$ is an angular threshold, and we set $w_t$ to $60^{\circ}$ in our experiments.

The final step is to choose one patch from $A_i$, which can best describe the underlying surface of $p_i$. Following \cite{sanchez2020robust}, for each  $Q_j^{\textbf{ani}} \in A_i$, we find a reference point $p^{\textbf{ref}}_j$ by projecting $p_i$ onto its plane $\theta_{Q_j^{\textbf{ani}}}^*$ (see the black arrows in Fig.~\ref{fig:fps:ani1} and Fig.~\ref{fig:fps:ani2}).
Second, each normal $n_j^{\textbf{ani}}$ is reoriented to the exterior of the surface, which means that it must satisfy the following condition:
\begin{equation}
\sum_{p_k \in \mathcal{N}_i} n_j^{\textbf{ani}} \cdot (p^{\textbf{ref}}_j - p_k) < 0, \end{equation}
where $\mathcal{N}_i$ is the neighborhood of $p_i$. 
The values $\{n_j^{\textbf{ani}} \cdot (p^{\textbf{ref}}_j - p_i)\}(j=1,2,...)$ are computed, and the minimum value corresponds to the initial normal of the candidate point $p_i$. Note that, $p^{\textbf{ref}}_j$ is only used for patch selection, which is not a new point in the point cloud. We refer the reader to \cite{sanchez2020robust} for a detailed explanation.



Although they may be sensitive to high-level noise, geometric estimators are often more effective to predict point normals around tiny details. Such initial normals which perform better in sharp feature regions improve the final Refine-Net results significantly by preserving geometric information that is difficult to recover from learning. We give a better understanding of our geometric method in Sec.~\ref{sec:insight}, compared with those learning-based solutions.

\label{SECgeometry}

\begin{table*}
	\centering
	\footnotesize
	\setlength{\tabcolsep}{3.5mm}
	\caption{Comparison of normal estimation errors on the synthetic dataset \cite{wang2016mesh}, evaluated as mean angular error (mean) and root mean square error (rmse).}
	\begin{tabular}{c |cc|cc|cc|cc|cc} 
		\toprule[1pt]
		Category & \multicolumn{2}{|c|}{BigNoise} & \multicolumn{2}{|c|}{SharpFeature} & \multicolumn{2}{|c|}{RichFeature} & \multicolumn{2}{|c|}{SmoothSurface} & \multicolumn{2}{c}{Average}  \\
		& mean & rmse & mean & rmse & mean & rmse & mean & rmse & mean & rmse\\ 
		\midrule[0.3pt]
		\midrule[0.3pt]
		PCA \cite{hoppe1992surface} 					& 8.53  & 13.80 & 8.70  & 13.61 & 6.94  & 9.84  & 8.04  & 6.83  & 8.05  & 11.02 \\
		HF \cite{boulch2012fast} 						& 11.73 & 16.23 & 5.26  & 9.75  & 5.81  & 8.34  & 3.87  & 5.54  & 6.67  & 9.96  \\
		HoughCNN1s \cite{boulch2016deep} 				& 8.76  & 15.68 & 5.40  & 11.20 & 6.00  & 9.05  & 5.05  & 7.44  & 6.30  & 10.84 \\
		HoughCNN3s \cite{boulch2016deep}				& 8.61  & 15.16 & 5.52  & 11.23 & 5.61  & 8.49  & 5.89  & 8.55  & 6.41  & 10.86 \\
		HoughCNN5s \cite{boulch2016deep}				& 10.78 & 16.39 & 6.48  & 12.22 & 6.46  & 9.51  & 7.40  & 10.24 & 7.78  & 12.09 \\
		PCPNet-ss \cite{GuerreroEtAl:PCPNet:EG:2018}	& 9.61  & 14.26 & 12.84 & 16.23 & 6.84  & 9.78  & 12.84 & 15.87 & 10.53 & 14.04 \\
		PCPNet-ms \cite{GuerreroEtAl:PCPNet:EG:2018}	& 9.09  & 13.52 & 7.90  & 11.09 & 5.97  & 8.15  & 5.27  & 6.97  & 7.06  & 9.93  \\
		LRR \cite{zhang2013point} 						& 5.77  & 12.08 & 4.39  & 8.37  & 4.86  & 7.18  & 4.66  & 8.22  & 4.92  & 8.96  \\
		PCV \cite{zhang2018multi} 						& 5.89  & 11.92 & 4.52  & 8.50  & 4.80  & 6.87  & 3.97  & 6.24  & 4.80  & 8.38  \\
		Nesti-Net \cite{Ben-ShabatLF19} 				& 5.10  & 10.86 & 4.28  & 7.89  & 4.62  & 6.37  & 4.20  & 5.76  & 4.55  & 7.72  \\
		MFPS 											& 5.72  & 11.72 & 4.36  & 7.86  & 4.82  & 6.97  & 4.10  & 6.32  & 4.75  & 8.22  \\
		\midrule[0.3pt]
		\midrule[0.3pt]
		PCPNet-ms + Refine-Net                          & 7.11  & 12.08 & 6.08  & 8.91  & 5.23  & 7.02  & 4.13  & 5.15  & 5.64  & 8.29  \\
		Nesti-Net + Refine-Net 							& 4.79  & \textbf{10.63} & 3.76  & 6.93  & 4.25  & 5.74  & 3.64  & 4.78  & 4.11  & 7.02  \\
		Our full pipeline 								& \textbf{4.74}  & 10.73 & \textbf{3.37}  & \textbf{6.39}  & \textbf{4.07}  & \textbf{5.65}  & \textbf{3.28}  & \textbf{4.56}  & \textbf{3.87}  & \textbf{6.83}  \\
		\bottomrule[1pt]
	\end{tabular}
	\label{table:synthetic_result}
\end{table*}

\begin{table*}
	\centering
	\footnotesize
	\setlength{\tabcolsep}{3.2mm}
	\caption{Comparison of normal estimation accuracy using PGP5 and PGP10 on the synthetic dataset \cite{wang2016mesh}. Higher is better.}
	\begin{tabular}{c |cc|cc|cc|cc|cc} 
		\toprule[1pt]
		Category & \multicolumn{2}{|c|}{BigNoise} & \multicolumn{2}{|c|}{SharpFeature} & \multicolumn{2}{|c|}{RichFeature} & \multicolumn{2}{|c|}{SmoothSurface} & \multicolumn{2}{c}{Average}  \\
		& PGP5 & PGP10 & PGP5 & PGP10 & PGP5 & PGP10 &PGP5 & PGP10 & PGP5 & PGP10\\ 
		\midrule[0.3pt]
		\midrule[0.3pt]
		PCA \cite{hoppe1992surface} 					& 0.591 & 0.754 & 0.564  & 0.711 & 0.533  & 0.764  & 0.648  & 0.889  & 0.584  & 0.780 \\
		HF \cite{boulch2012fast} 						& 0.316 & 0.608 & 0.739  & 0.893 & 0.601  & 0.853  & 0.757  & 0.937  & 0.603  & 0.823 \\
		HoughCNN1s \cite{boulch2016deep} 				& 0.624 & 0.782 & 0.774  & 0.882 & 0.618  & 0.828  & 0.682  & 0.875  & 0.675  & 0.842 \\
		HoughCNN3s \cite{boulch2016deep}				& 0.610 & 0.784 & 0.756  & 0.868 & 0.637  & 0.847  & 0.634  & 0.825  & 0.659  & 0.831 \\
		HoughCNN5s \cite{boulch2016deep}				& 0.468 & 0.691 & 0.695  & 0.835 & 0.576  & 0.806  & 0.530  & 0.748  & 0.567  & 0.770 \\
		PCPNet-ss \cite{GuerreroEtAl:PCPNet:EG:2018}	& 0.473 & 0.732 & 0.381  & 0.620 & 0.527  & 0.791  & 0.261  & 0.537  & 0.410  & 0.670 \\
		PCPNet-ms \cite{GuerreroEtAl:PCPNet:EG:2018}	& 0.464 & 0.737 & 0.508  & 0.772 & 0.564  & 0.840  & 0.613  & 0.887  & 0.537  & 0.809  \\
		LRR \cite{zhang2013point} 						& 0.732 & 0.885 & 0.793  & 0.901 & 0.675  & 0.895  & 0.744  & 0.902  & 0.736  & 0.896  \\
		PCV \cite{zhang2018multi} 						& 0.725 & 0.882 & 0.767  & 0.919 & 0.672  & 0.895  & 0.765  & 0.930  & 0.732  & 0.906  \\
		Nesti-Net \cite{Ben-ShabatLF19} 				& 0.767 & 0.888 & 0.764  & 0.901  & 0.682 & 0.911  & 0.730  & 0.921  & 0.736  & 0.905   \\
		MFPS 											& 0.732 & 0.884 & 0.779  & 0.918  & 0.674 & 0.896  & 0.751  & 0.919  & 0.734  & 0.904  \\
		\midrule[0.3pt]
		\midrule[0.3pt]
		PCPNet-ms + Refine-Net                          & 0.623  & 0.808 & 0.608  & 0.852  & 0.616  & 0.880  & 0.703  & 0.945  & 0.637  & 0.871  \\
		Nesti-Net + Refine-Net 							& 0.783  & 0.895 & 0.796  & 0.923  & 0.709  & 0.927  & 0.763 & 0.948  & 0.763  & 0.923  \\
		Our full pipeline 								& \textbf{0.788}  & \textbf{0.903} & \textbf{0.842}  & \textbf{0.948}  & \textbf{0.722}  & \textbf{0.931}  & \textbf{0.802}  & \textbf{0.959}  & \textbf{0.788}  & \textbf{0.935}  \\
		\bottomrule[1pt]
	\end{tabular}
	\label{table:synthetic_PGP}
\end{table*}

\begin{figure*}
	\centering
	\includegraphics[width=0.99\linewidth]{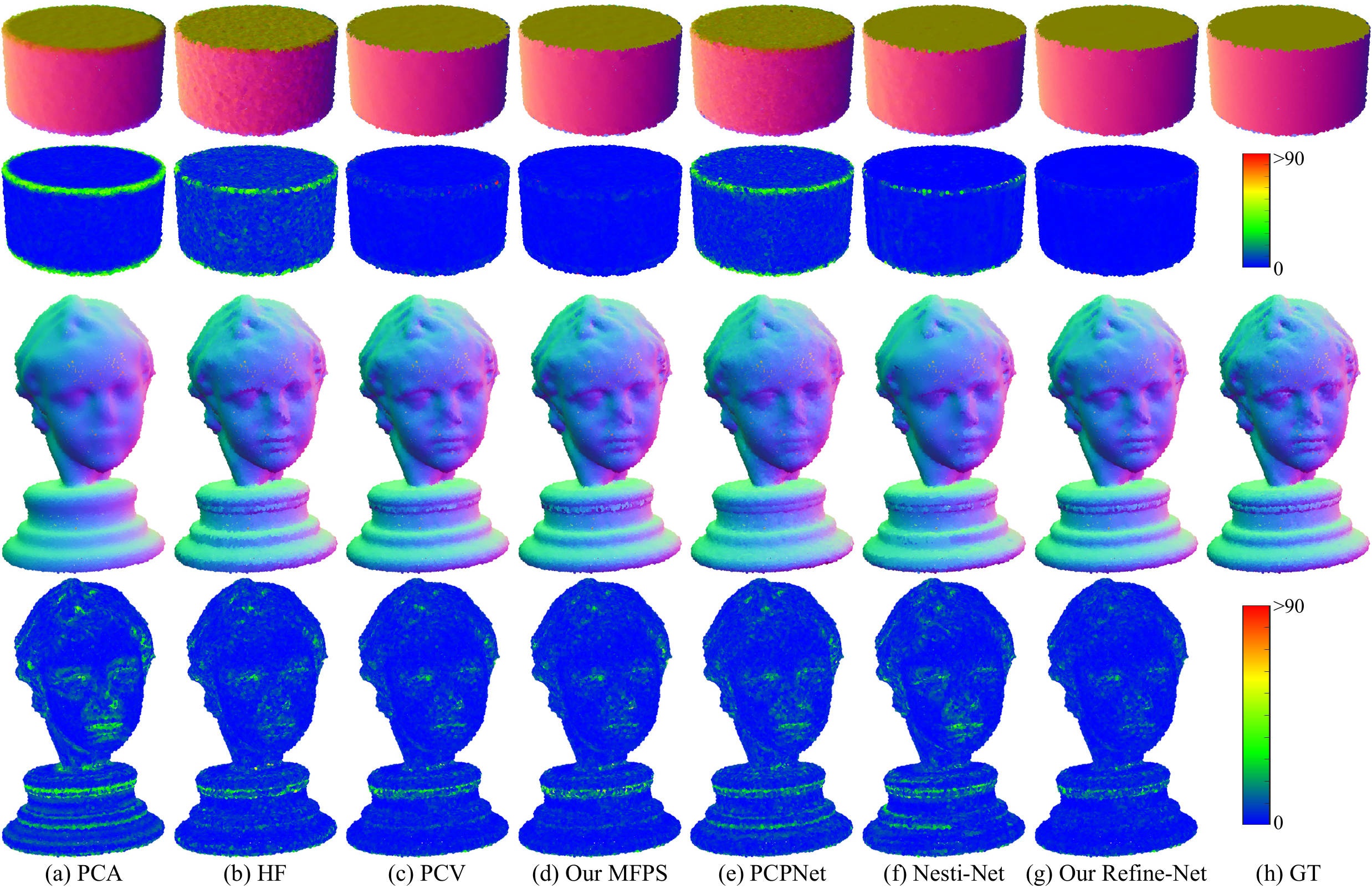}
	
	\caption{Visual comparison of estimated normals with other methods: PCA \cite{hoppe1992surface}, HF \cite{boulch2012fast}, PCV \cite{zhang2018multi}, PCPNet \cite{GuerreroEtAl:PCPNet:EG:2018} and Nesti-Net \cite{Ben-ShabatLF19}. We also give our MFPS and Refine-Net results. The 1-st and 3-rd rows denote the normal results rendered in RGB colors. The average normal angular errors (mean) are: (1-st row) 4.34, 6.37, 1.63, 1.49, 5.37, 1.66, \textbf{0.88}; (3-rd row) 8.14, 6.27, 5.60, 5.71,  6.22, 6.16, \textbf{4.99}. The 2-nd and 4-th rows denote the normal errors, mapped to a heatmap according to the color bar on the right. It can be seen that more errors occur near sharp edges (2-nd row) and facial details (4-th row). Our method suffers the least from these challenges. }
	\label{fig:syn}
\end{figure*}

\section{Evaluation}
\label{sec:eval}
In this section, we evaluate our Refine-Net on both synthetic and real-scanned datasets. Detailed network architectures and comparisons are provided. 

\subsection{Experimental setup}
\label{sec:eval:setup}
\textbf{Network architecture.} The network architecture of Refine-Net is illustrated in Fig.~\ref{fig:architecture}. In the first refinement, following \cite{GuerreroEtAl:PCPNet:EG:2018}, the patch radius of input points is set related to the length of the bounding box of the point cloud. The size is 0.05 and a maximum of 300 points is used. For neighborhoods with more than 300 points, we apply a random sampling, and for those with fewer points, we pad the rest with zeros (the patch center). Here, we use the PointNet architecture \cite{QiSMG17}. The T-Net module in PointNet is not included since the input points are already reoriented according to their initial normals. First, a shared MLP (64,64,64,128,1024) is applied to get a 1024 dimensional feature per point. Then, a max-pooling layer is used to get the global feature for the entire patch, after which fully-connected layers (256,128) are applied for the output $d_1$-dimensional vector. Dropout with a keep probability of $0.3$ is introduced in the last two fully-connected layers. The output of the point module follows the connection designs we discussed in Sec.~\ref{sec:refnet:connect}.

In the second refinement, for the HMP construction, we use a $7\times7$ ($m = 7$) height-map grid, and the neighborhood size is the same as the patch radius above. Then, we process these HMPs using several convolutional layers followed by fully-connected layers (see Tab.~\ref{table:network_architecture} left). In connection modules, we use the weight matrix option, i.e., $d_1 = 64\times3$ and $d_2 = 64\times64$ as discussed in Sec.~\ref{sec:refnet:connect}. Other connection choices will be evaluated later in Sec.~\ref{sec:eval:ablation}. As a consequence, two normal features (64) and the original filtered normal, extended by FC (64,64), are concatenated as the output of each branch. Finally, for the output module, fully-connected layers (512,256,3) are used to obtain the predicted normal. Dropout is included similarly in all fully-connected layers. All layers use the batch normalization and ReLU.  

\vspace{5pt}
\noindent\textbf{Initial normal and filtering.} The initial normal is computed for each point (Sec.~\ref{sec:geometric}) before the refinement steps. The proposed geometric normal estimation method plays an important role in capturing detailed structures and sharp features. We will also evaluate these normal results and demonstrate the effectiveness of Refine-Net based on several different initial normals. In MFPS, the multi-scale neighborhood size $K$ is the most practical parameter that could be tuned. We set $K = 50, 100, 150$ empirically for the synthetic models.
The parameters used in bilateral filtering are set as $P_s = \{0.025\overline{l}_d, 0.05\overline{l}_d\}$ and $P_r = \{0.1, 0.2, 0.35, 0.5\}$, where $\overline{l}_d$ is the diagonal length of the bounding box. Thus, together with the initial normal (non-filtered normal), Refine-Net contains 9 individual branches.

\subsection{Experiments on the PCPNet dataset}
\label{sec:eval:pcpnet}
\textbf{Dataset.} First, we compare our model with state-of-the-art deep networks on the PCPNet \cite{GuerreroEtAl:PCPNet:EG:2018} dataset. This dataset consists of 30 shapes from a mixture of figurines, man-made objects and differentiable surfaces, with 8 shapes for training and 22 for testing. All shapes are uniformly sampled with 100k points. We follow the experimental settings of \cite{GuerreroEtAl:PCPNet:EG:2018, Ben-ShabatLF19}. Each shape is augmented by Gaussian noise with standard deviations of $0.12\%$, $0.6\%$ and $1.2\%$ of the diagonal length of the bounding box. In addition, two categories with different sampling densities (Gradient and Stripes) are generated for each shape. For evaluation, we use the same subset of 5k points from each point cloud, following the protocol of Guerrero et al. \cite{GuerreroEtAl:PCPNet:EG:2018}.

\vspace{5pt}
\noindent\textbf{Training.} We use the SGD optimizer with a learning rate of $0.0001$, and we do not use the learning rate decay in all epochs. The batch size is 512 for all training clusters and the momentum for batch normalization is $0.9$. We implement Refine-Net using PyTorch and train it on a RTX 2080 Ti GPU.

\vspace{5pt}
\noindent\textbf{Results.} We report the root mean square error (rmse) in Tab.~\ref{table:pcpnet}. Here, we show both the initial normal estimation (MFPS) and Refine-Net results with MFPS initial normals (our full pipeline). Following \cite{Ben-ShabatLF19}, we give the single-scale (ss) and multi-scale (ms) versions for HoughCNN \cite{boulch2016deep} and PCPNet \cite{GuerreroEtAl:PCPNet:EG:2018}. For traditional PCA \cite{hoppe1992surface} and Jet \cite{cazals2005estimating}, results for medium neighborhood sizes are shown. It can be seen that our method achieves state-of-the-art performances on most of the categories. For shapes with high-level noise, general directions of point normals are predicted correctly, while fine details can hardly be observed for all of the methods. Also, we can see that our geometric method already performs promisingly against other models on point clouds with varying densities, thanks to the multi-scale scheme for patch selection.

\subsection{Experiments on more synthetic data}
\label{sec:eval:synthetic}
\textbf{Dataset.} In order to conduct a more comprehensive evaluation with both the traditional geometric and deep learning-based methods, we use another synthetic dataset from Wang et al. \cite{wang2016mesh}.
The training set contains 21 synthetic triangular mesh models with varying sampling densities and is manually divided into three categories: CAD-like models, smooth models and feature-rich models. Each type is collected from shapes representing typical challenging geometric features. We simply extract the mesh vertices as point samples and the mesh normals as ground-truth normals in order to preserve the original features for training. To generate noisy inputs, we introduce Gaussian noise for each point cloud with standard deviations of $0.1\%$, $0.2\%$ and $0.3\%$ of the diagonal length of the bounding box. The final training dataset contains 1.5M points from 63 noisy point clouds, in which $20\%$ of point samples are used for validation.

To test the normal results on synthetic data, we use the test set from \cite{wang2016mesh,zhang2018multi}. Similarly, the test set includes 4 categories: SharpFeature, SmoothSurface, RichFeature and BigNoise, in which there are 11, 8, 8 and 8 models respectively. For data augmentation, each point cloud in the SharpFeature and SmoothSurface categories is perturbed by Gaussian noise with a standard deviation of $0.05\%$, $0.1\%$ and $0.15\%$ of the diagonal length of the bounding box. For the RichFeature category, $0.05\%$, $0.1\%$, $0.15\%$ and $0.2\%$ noise are introduced and $0.2\%$, $0.3\%$, $0.4\%$ and $0.5\%$ for the BigNoise category. Our test set contains totally 121 point clouds with over 3M point samples.

\vspace{5pt}
\noindent\textbf{Results.} The evaluation metrics \cite{Ben-ShabatLF19} for normal estimation are mean angular error (mean), root mean square error (rmse) and proportion of good points metric PGP$\alpha$ (percentage of points with angular error below threshold $\alpha$) where $\alpha \in [5^{\circ}, 10^{\circ}]$.

Tab.~\ref{table:synthetic_result} and Tab.~\ref{table:synthetic_PGP} show the results on the synthetic models. We compare our method, including MFPS initial normal results and the full pipeline, with several traditional and deep-learning state-of-the-arts: PCA \cite{hoppe1992surface}, HF \cite{boulch2012fast}, LRR \cite{zhang2013point}, PCV \cite{zhang2018multi}, HoughCNN \cite{boulch2016deep}, PCPNet \cite{GuerreroEtAl:PCPNet:EG:2018} and Nesti-Net \cite{Ben-ShabatLF19}. We retrain HoughCNN, PCPNet and Nesti-Net on this dataset for a fair comparison. In addition, HoughCNN has three versions with different scales and PCPNet has the single-scale and multi-scale versions. We evaluate all these versions. For the above methods, we set a same neighborhood size $K=100$ for the three categories: SharpFeature, RichFeature and SmoothSurface.
In particular, the HoughCNN3s considers 3 scales, $K = 50,100,200$, which is recommended in their paper. For the BigNoise category, we double the neighborhood size, i.e., $K = 100,200,400$ for HoughCNN3s, and $K=200$ for other single-scale methods. Besides, HoughCNN5s considers 5 scales, $K = 32,64,128,256,512$, for all categories. All other parameters are set using default values. 
From the above comparison, our method outperforms others in all categories, especially on models with high-level noise.

For visual comparisons, we show normal results and corresponding angular errors (mapped to a heatmap) of different methods in Fig.~\ref{fig:syn}. For PCPNet, we display the best results out of its two versions. We can see that Refine-Net produces more accurate normals on sharp edges and facial details in the 1-st and 3-rd rows respectively.

\begin{table}[t]
	\centering
	\footnotesize
	\setlength{\tabcolsep}{3.5mm}
	\caption{Comparison of normal estimation results on real-scanned dataset from Wang et al. \cite{wang2016mesh}.}
	\begin{tabular}{c|cc|cc} 
		\toprule[1pt]
		& \multicolumn{2}{|c|}{Error} & \multicolumn{2}{|c}{Accuracy}  \\
		& mean & rmse & PGP5 & PGP10 \\ 
		\midrule[0.3pt]
		\midrule[0.3pt]
		PCA \cite{hoppe1992surface} 					& 8.66  & 11.29 & 0.34  & 0.69  \\
		HF \cite{boulch2012fast} 						& 8.14  & 11.99 & 0.46  & 0.75    \\
		HoughCNN \cite{boulch2016deep} 					& 12.30 & 15.58 & 0.24  & 0.49  \\
		PCV \cite{zhang2018multi} 						& 8.03  & 11.36 & 0.42  & 0.75    \\
		PCPNet \cite{GuerreroEtAl:PCPNet:EG:2018}		& 7.90  & 10.72 & 0.42  & 0.75  \\
		Nesti-Net \cite{Ben-ShabatLF19} 				& 6.93  & 9.68  & 0.50  & 0.79    \\
		\midrule[0.3pt]
		MFPS & 8.22 & 11.62& 0.41& 0.74 \\
		Our full pipeline								& \textbf{6.76}  & \textbf{9.58} & \textbf{0.52}  & \textbf{0.81}  \\
		\bottomrule[1pt]
	\end{tabular}
	\label{table:Kinect_results}
\end{table}

\begin{figure*}
	\centering
	\includegraphics[width=0.99\linewidth]{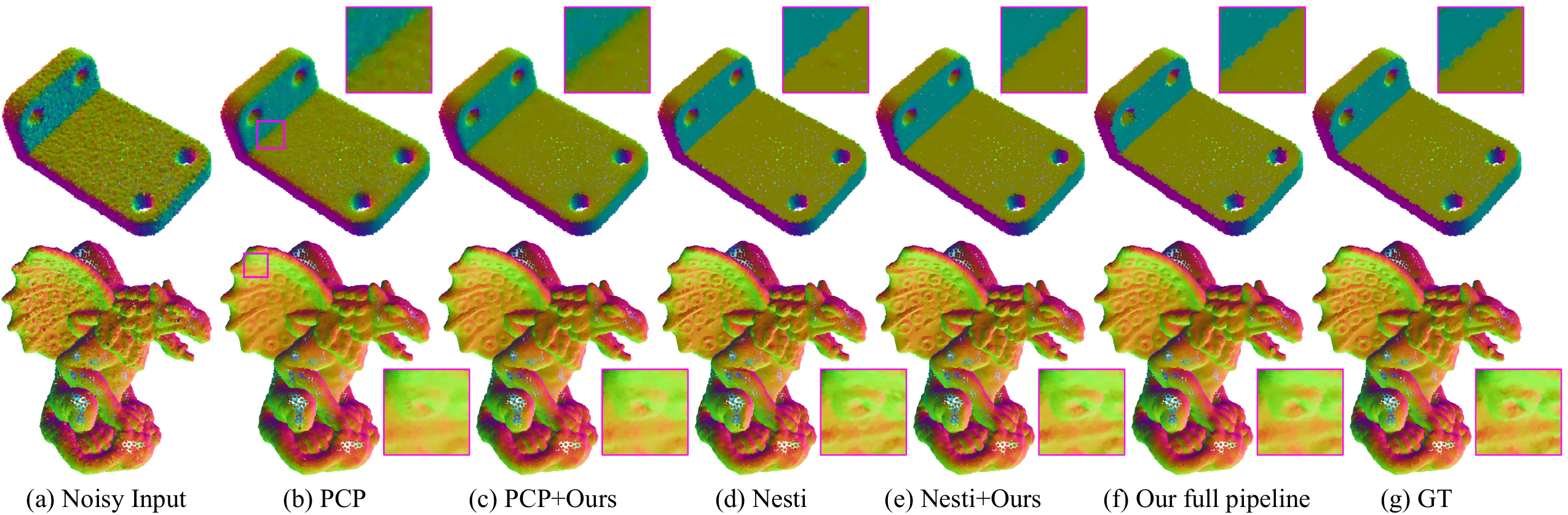}
	
	\caption{Visual comparison of estimated normals on synthetic models. ``PCP+Ours'' indicates that Refine-Net takes PCPNet \cite{GuerreroEtAl:PCPNet:EG:2018} results as initial normals. ``Nesti+Ours'' indicates the Nesti-Net \cite{Ben-ShabatLF19} results as initial normals. By applying our normal refinement system, the over-smoothed sharp edge (1-st row) is recovered and tiny details on the wings of gargoyle (2-nd row) are clearer to observe. }
	\label{fig:syn_refine}
\end{figure*}

\begin{figure*}
	\newlength{\unit}
	\setlength{\unit}{0.142\linewidth}
	\centering
	\includegraphics[width=\unit]{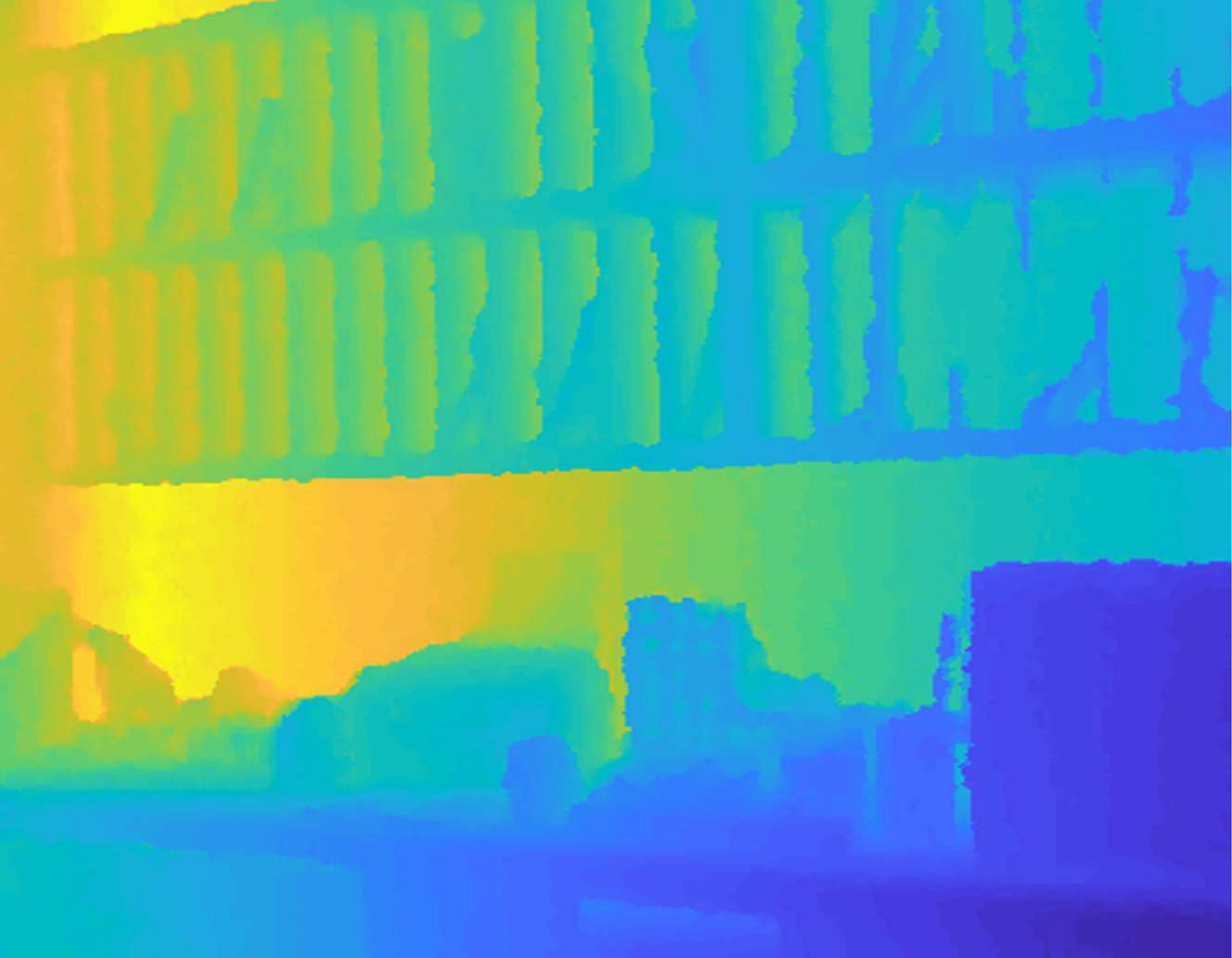}%
	\includegraphics[width=\unit]{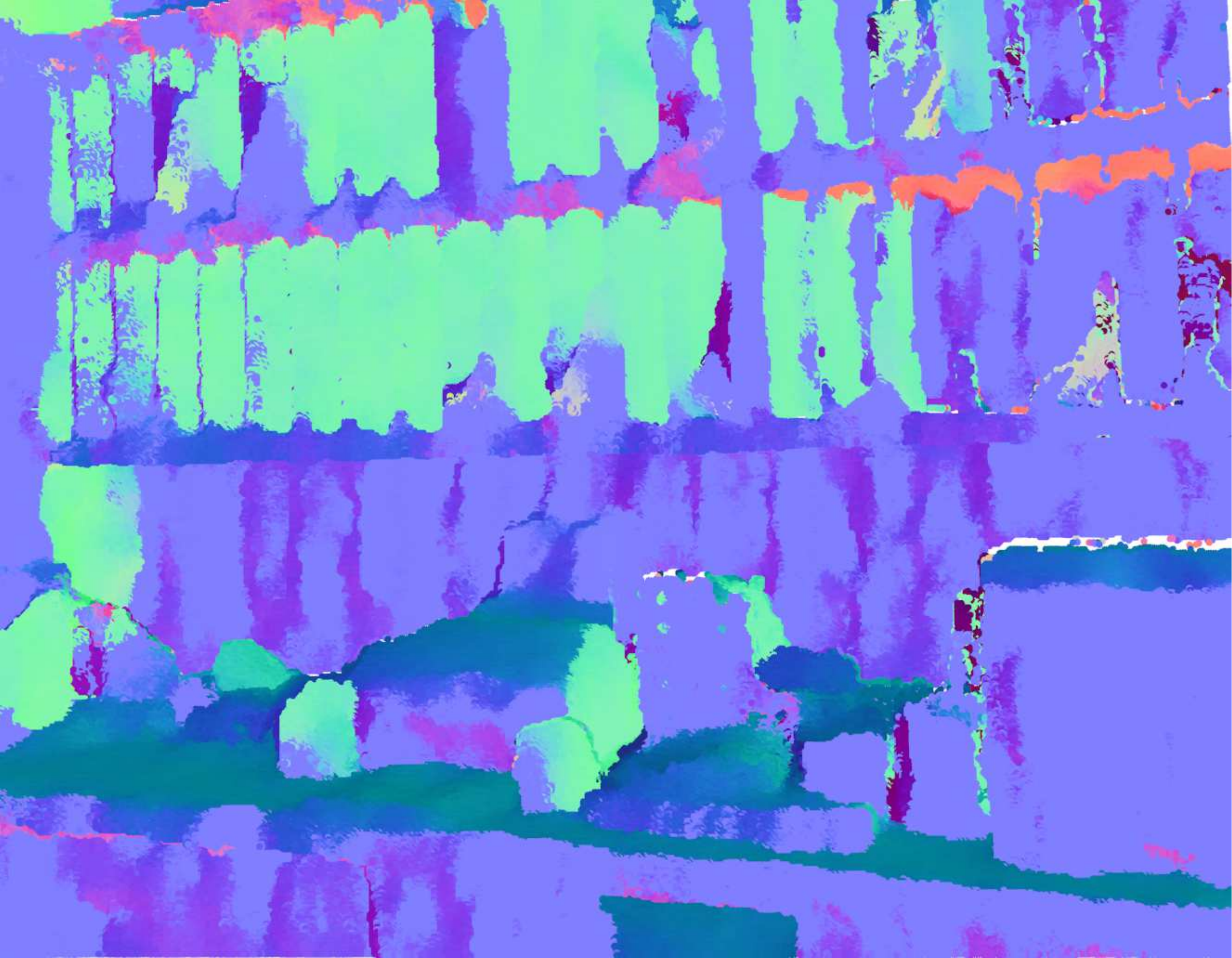}%
	\includegraphics[width=\unit]{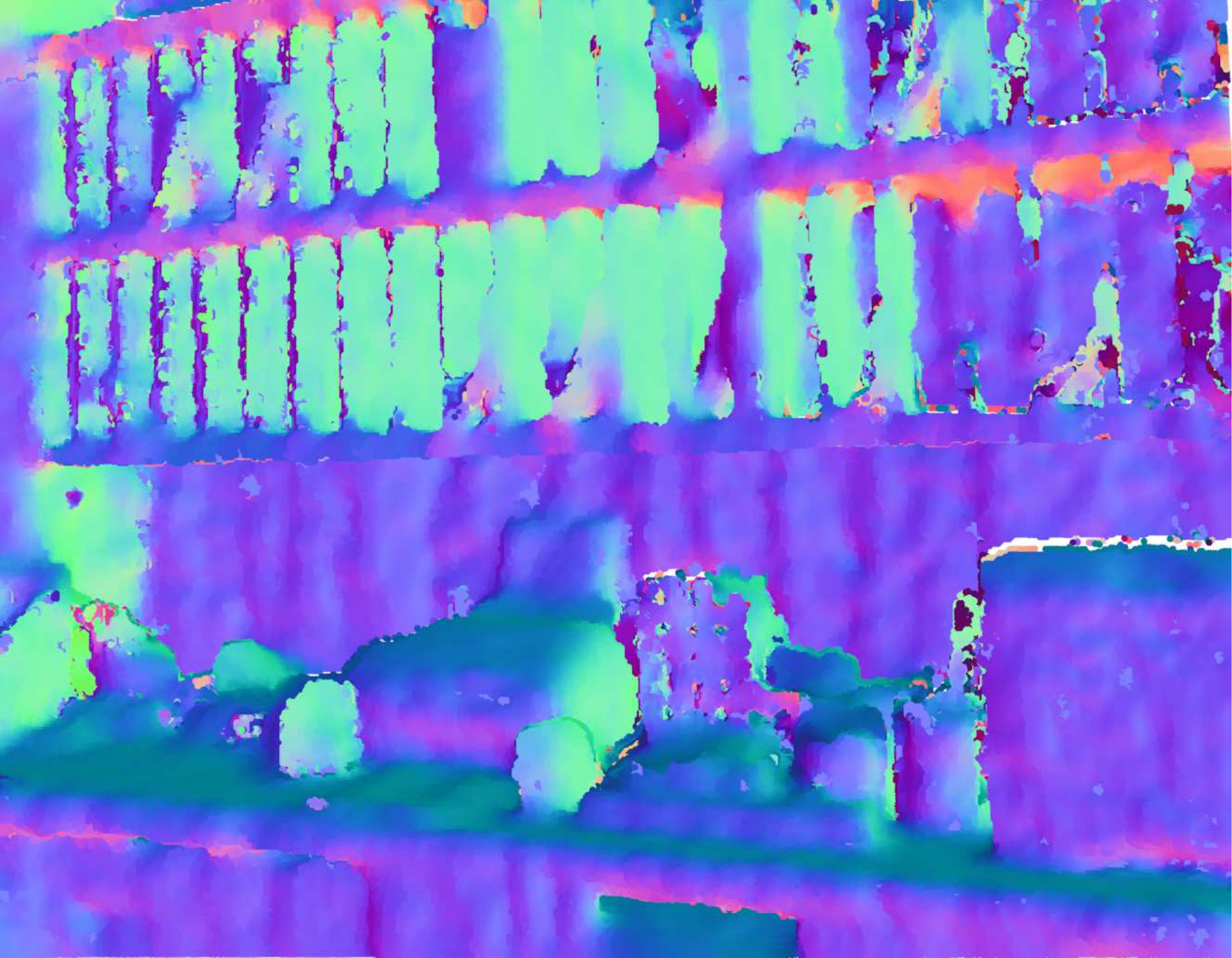}%
	\includegraphics[width=\unit]{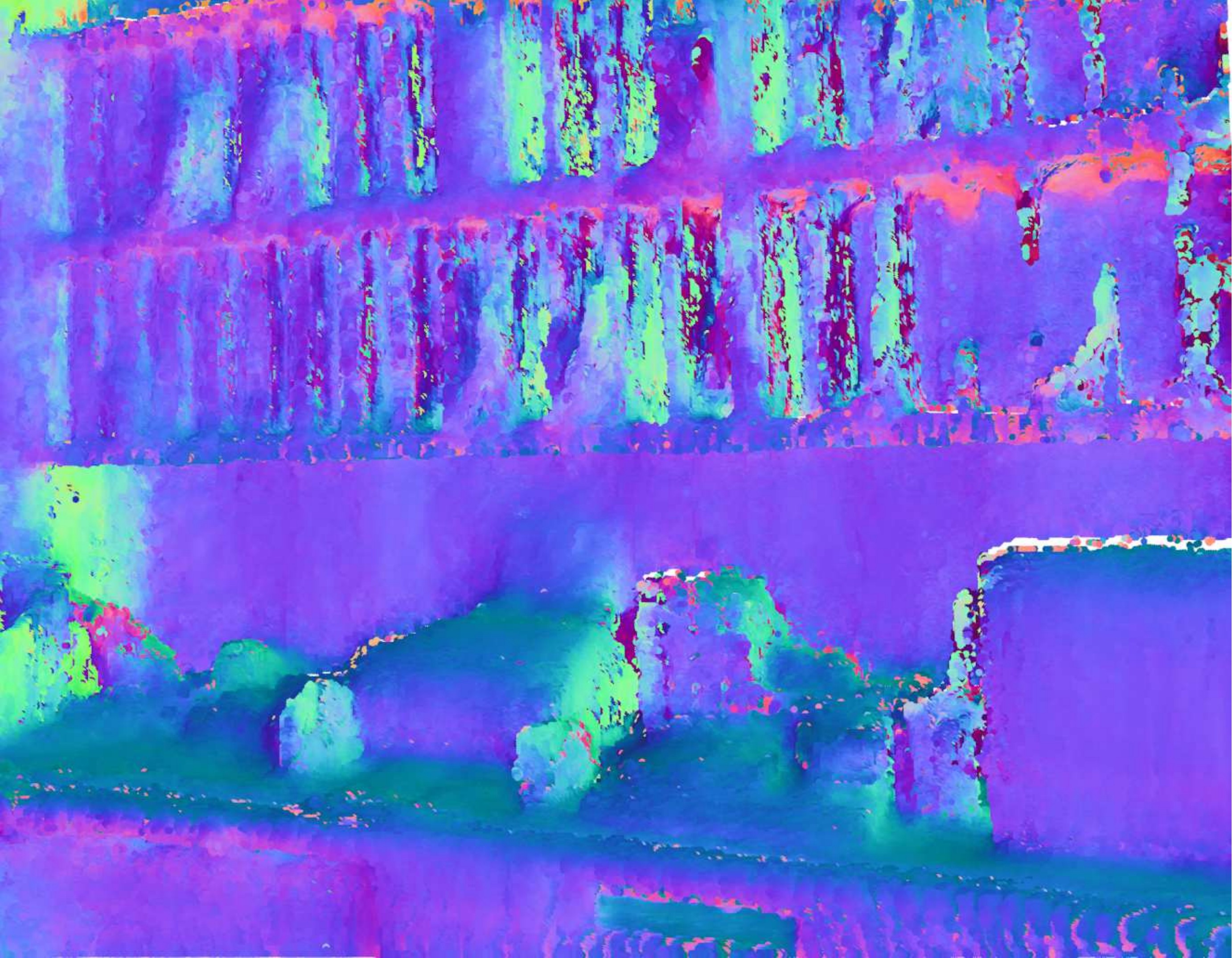}%
	\includegraphics[width=\unit]{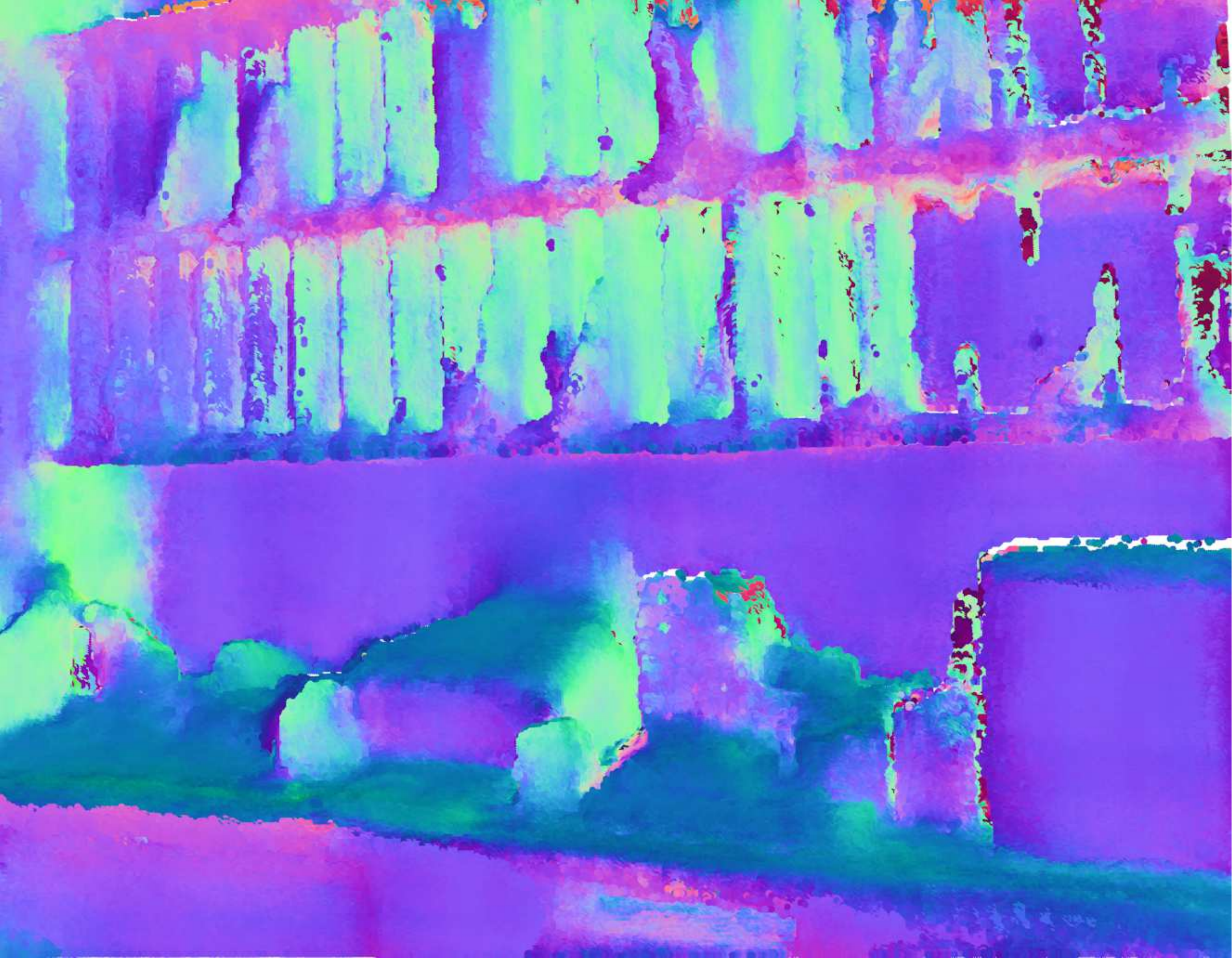}%
	\includegraphics[width=\unit]{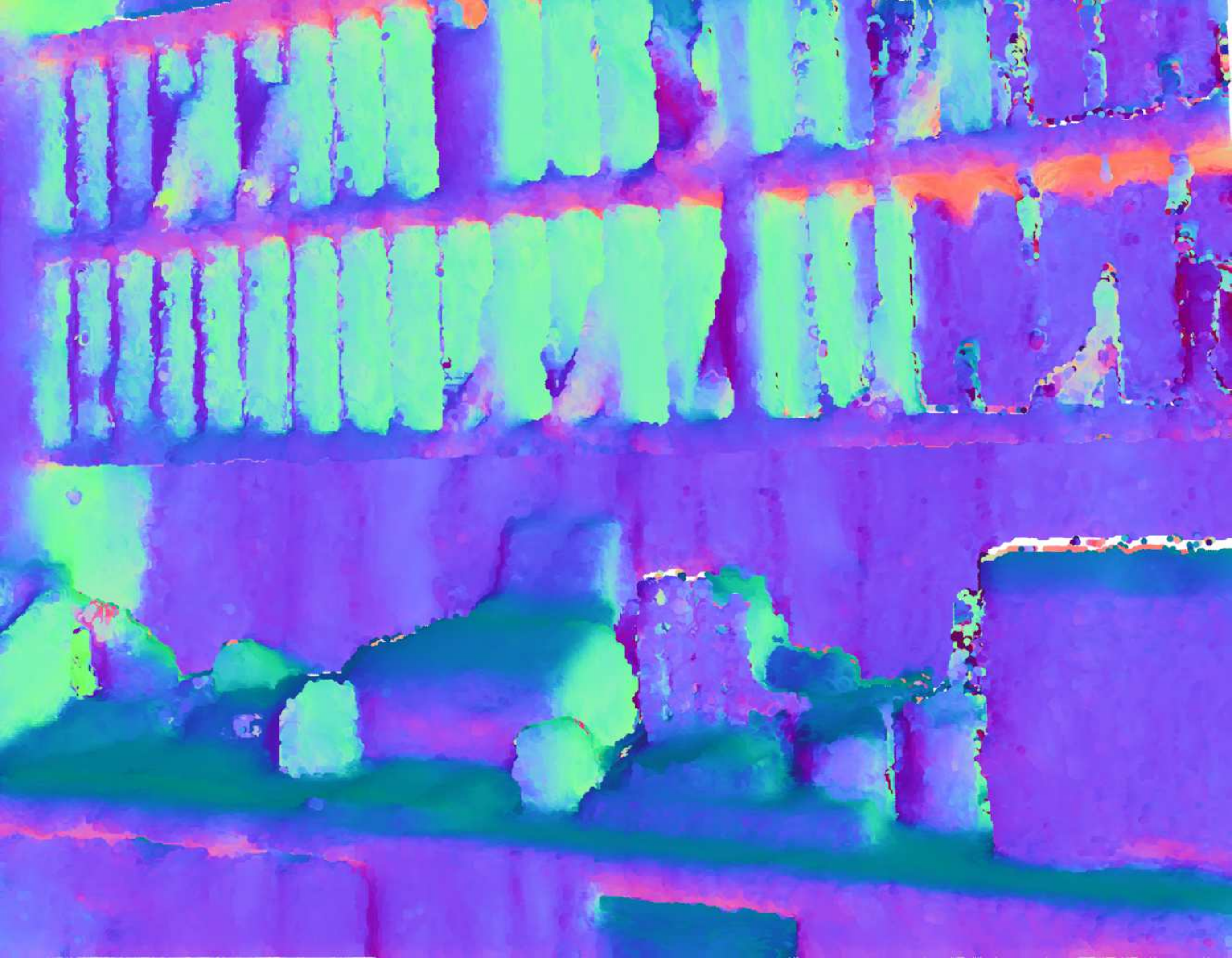}%
	\includegraphics[width=\unit]{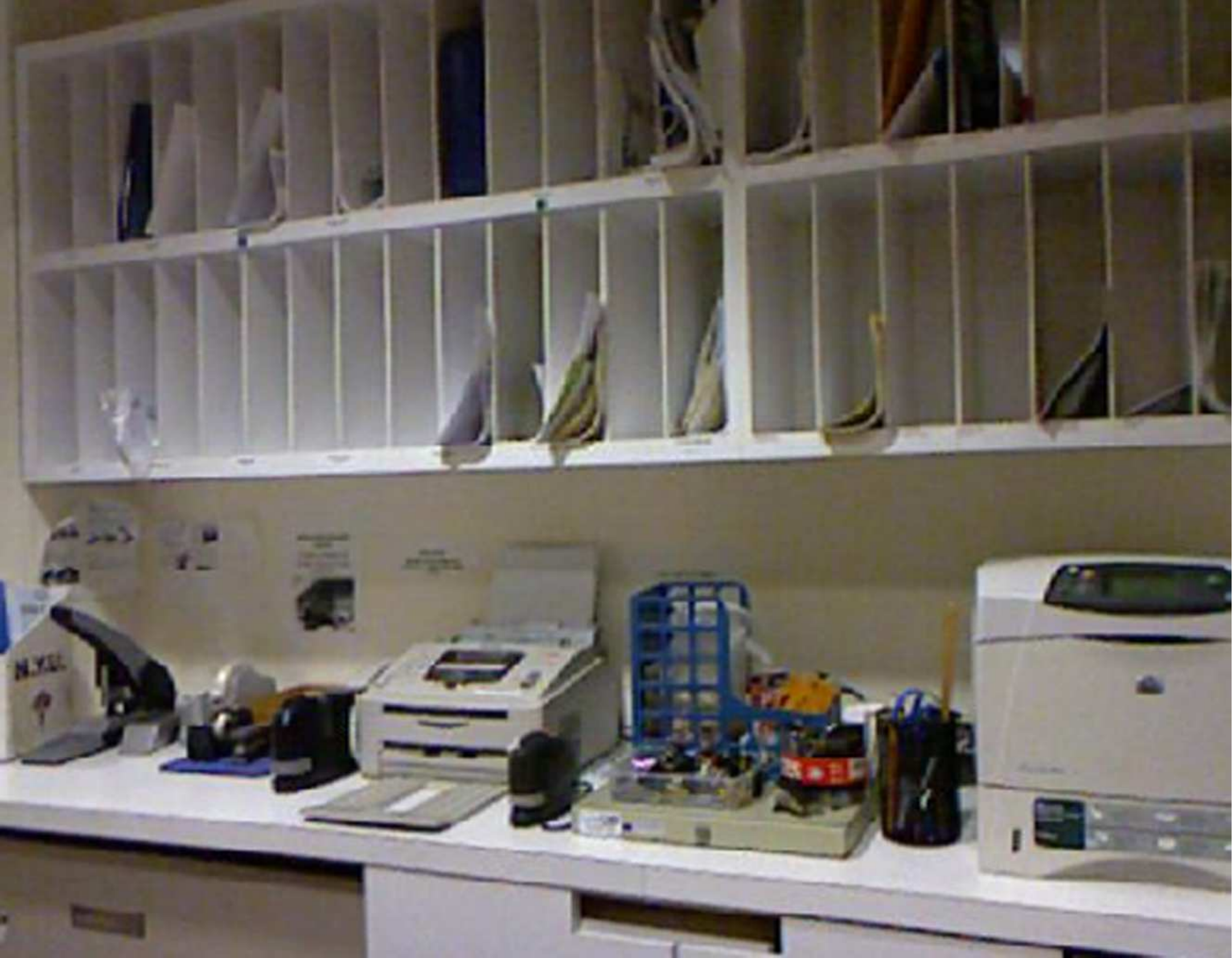}
	
	\vspace{3pt}
	
	\includegraphics[width=\unit]{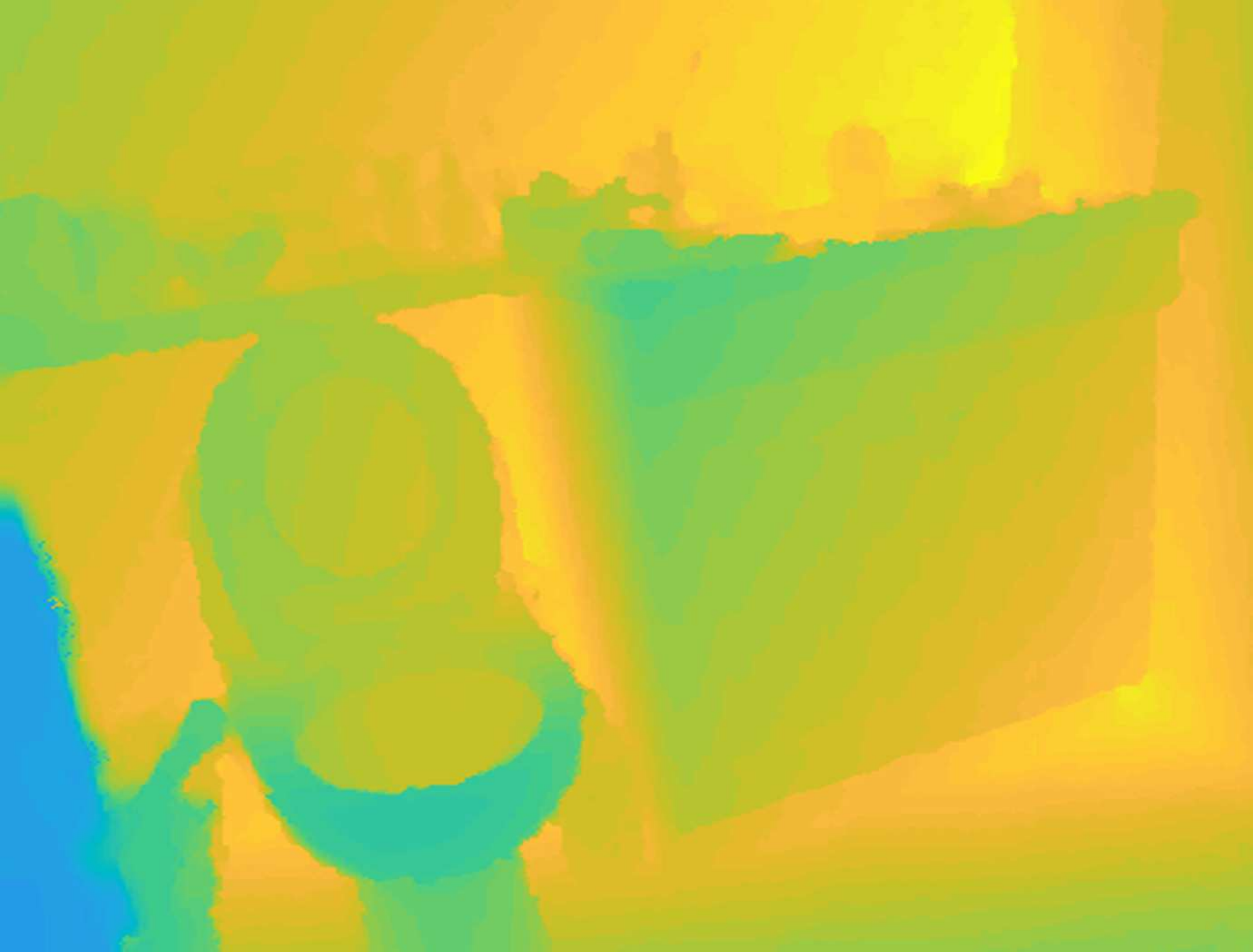}%
	\includegraphics[width=\unit]{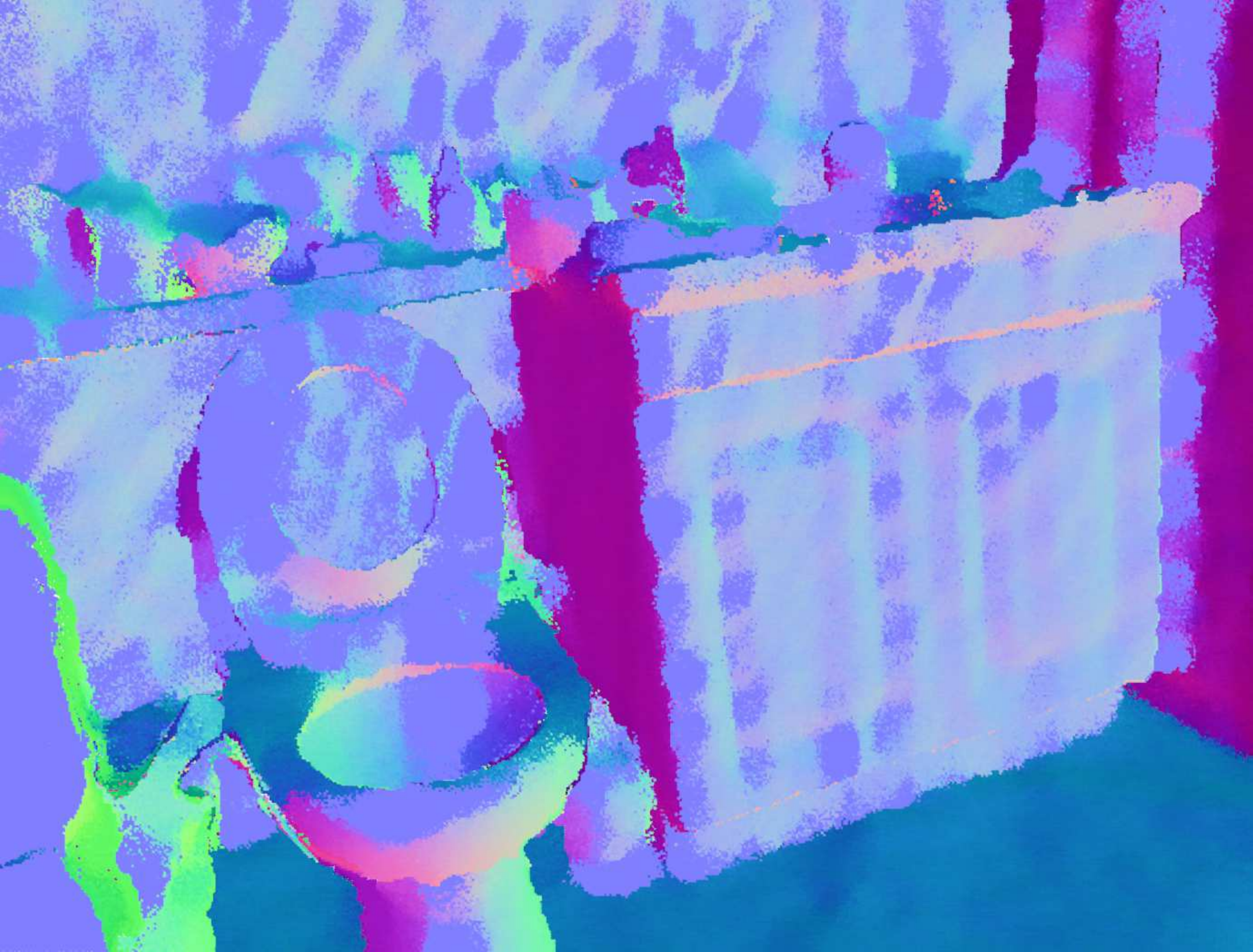}%
	\includegraphics[width=\unit]{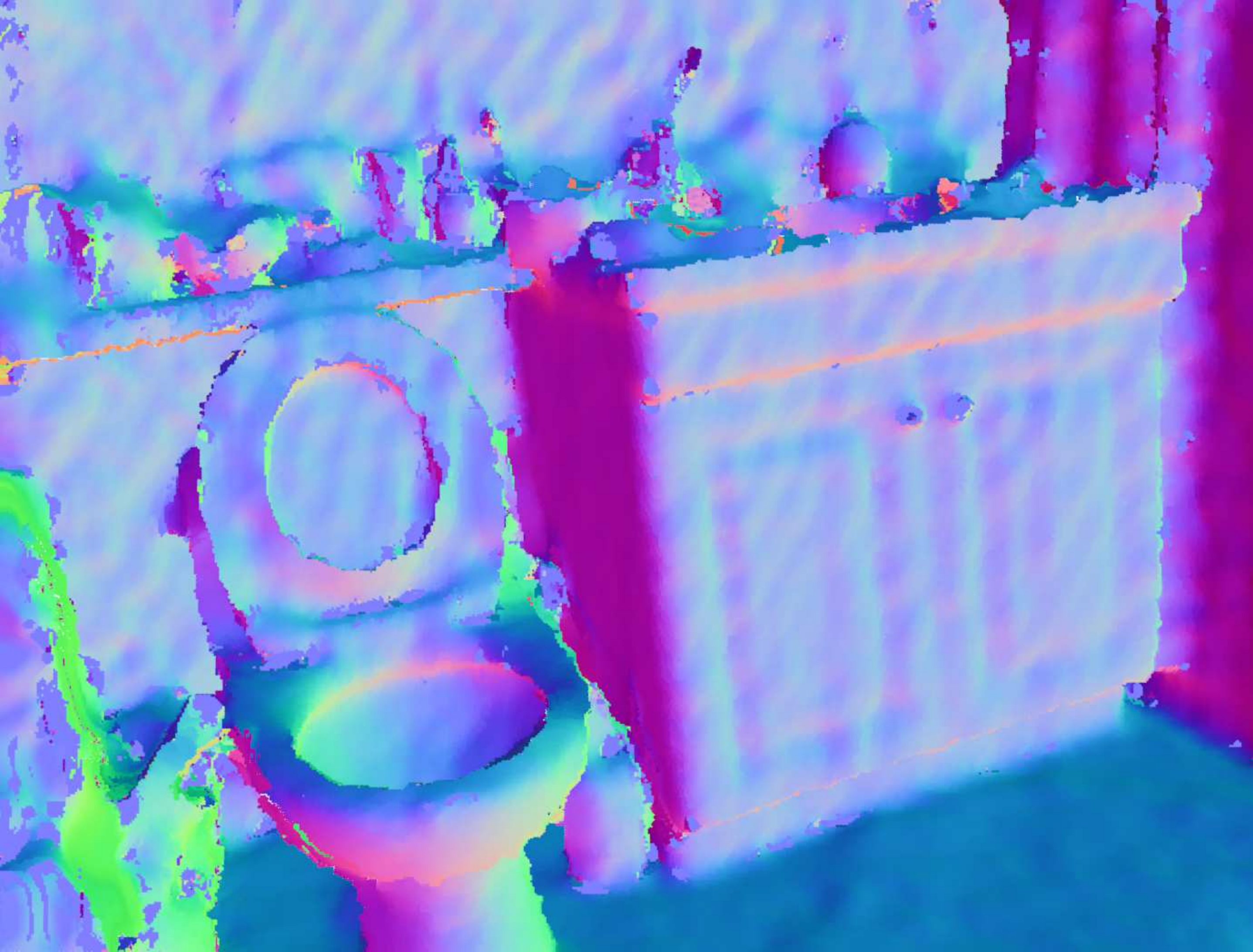}%
	\includegraphics[width=\unit]{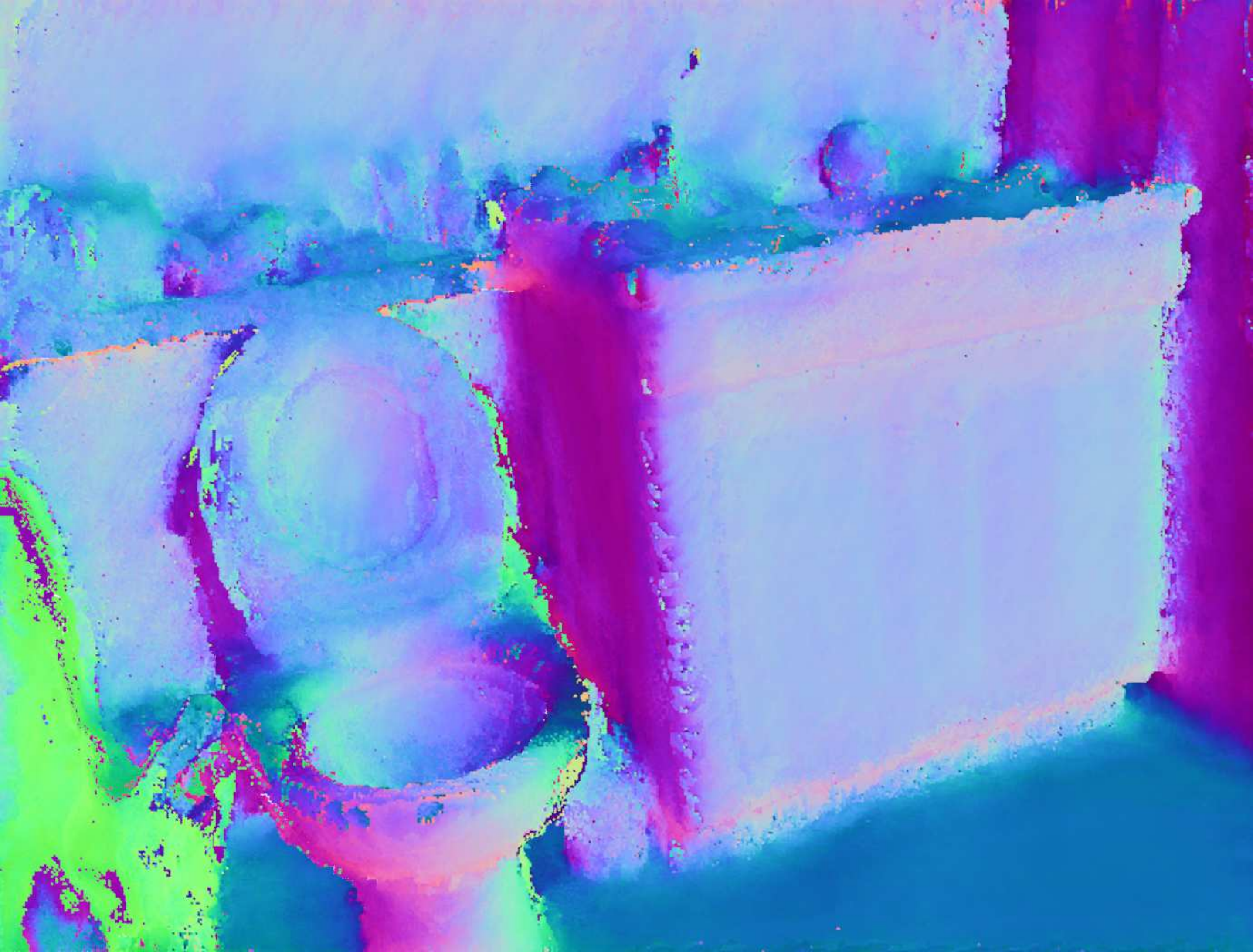}%
	\includegraphics[width=\unit]{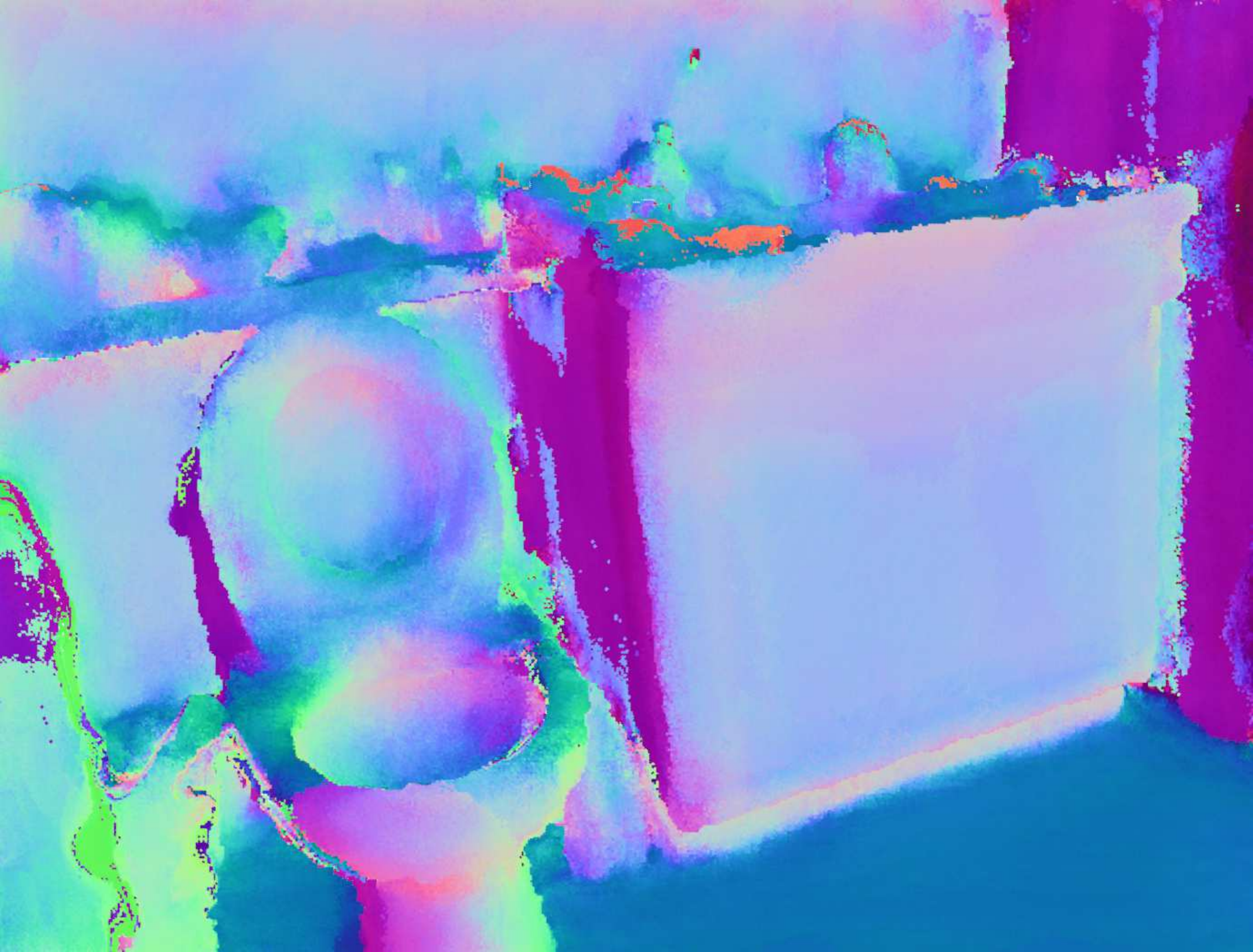}%
	\includegraphics[width=\unit]{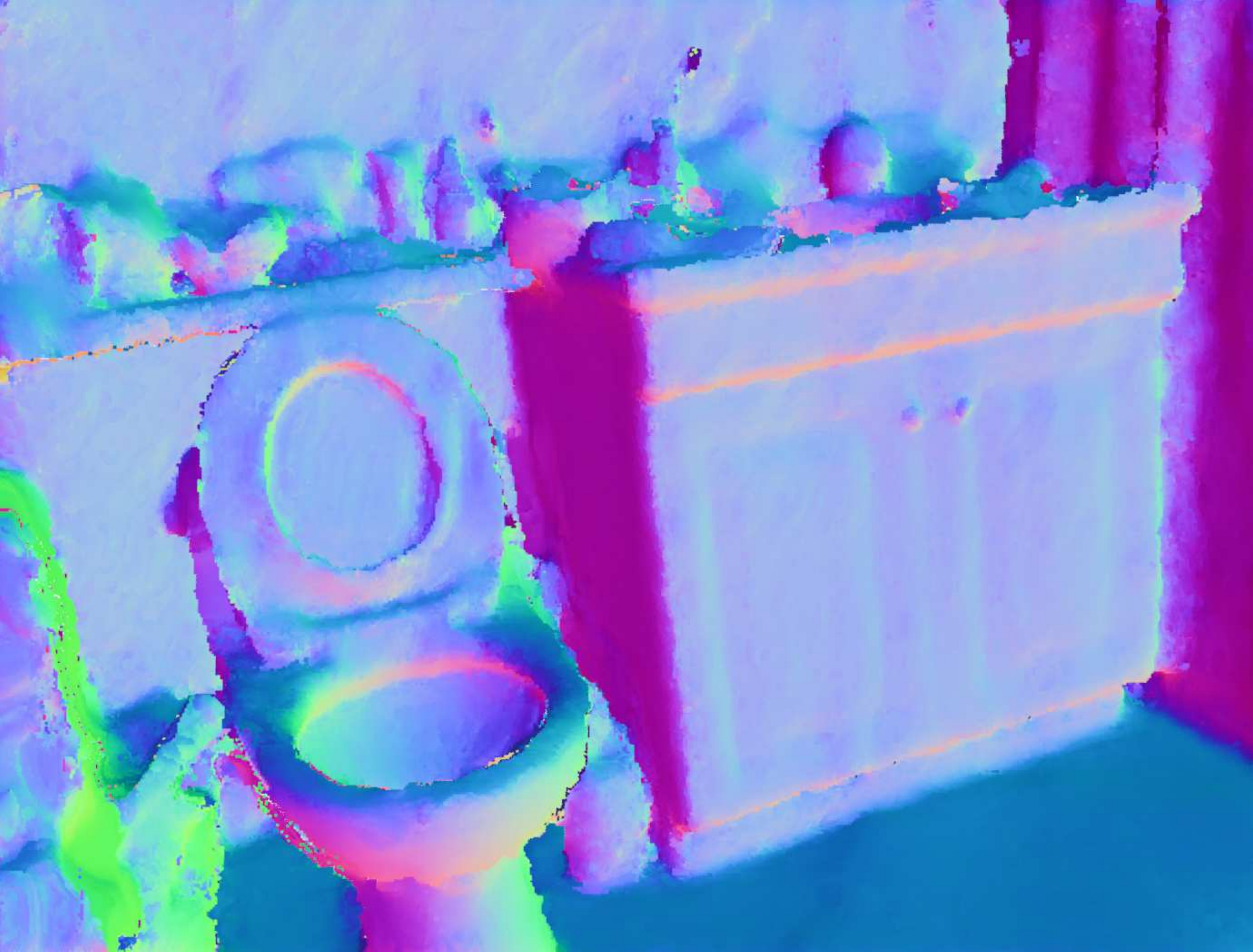}%
	\includegraphics[width=\unit]{fig/indoor/scene45f/rgb.eps}
	
	\vspace{3pt}
	
	\includegraphics[width=\unit]{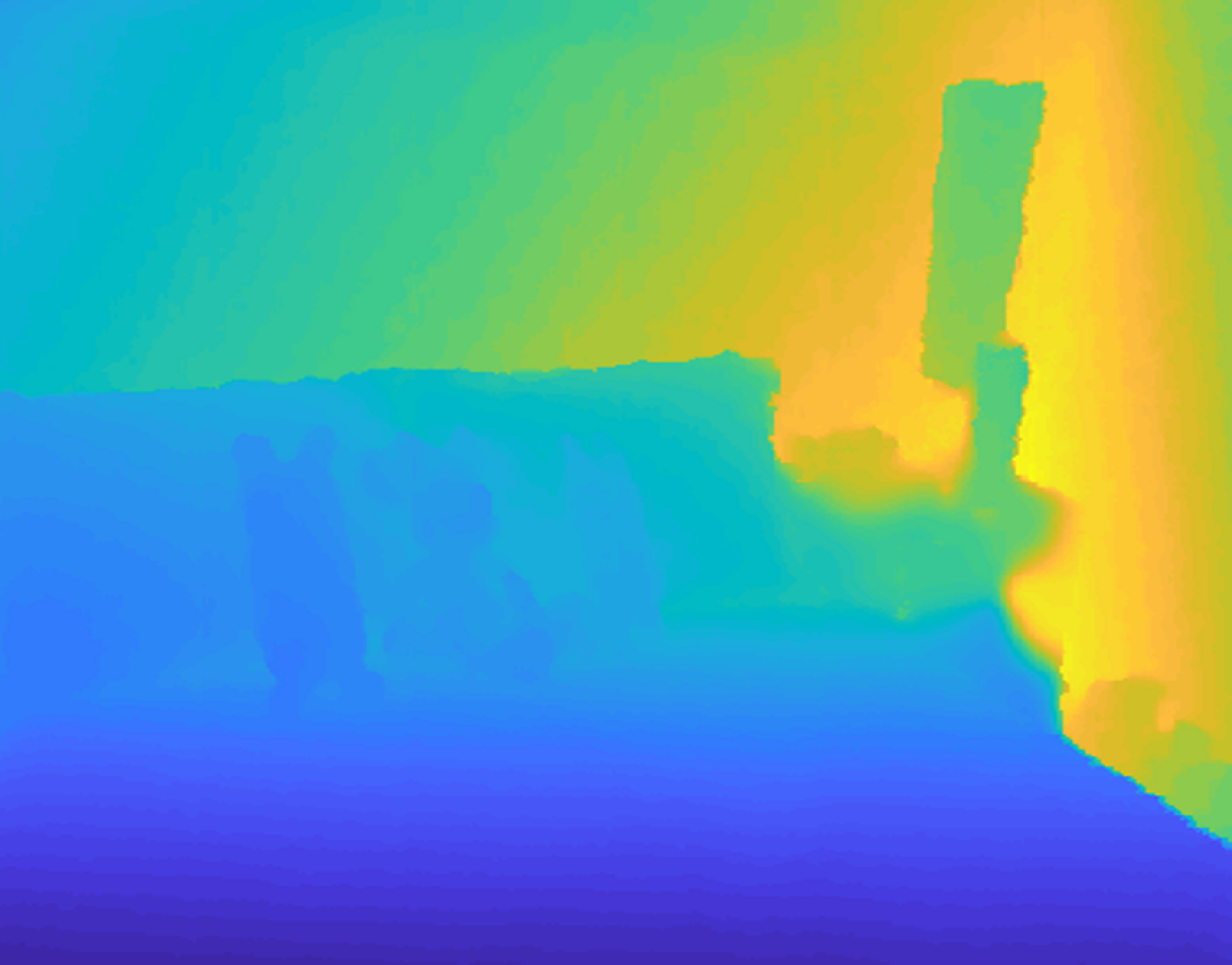}%
	\includegraphics[width=\unit]{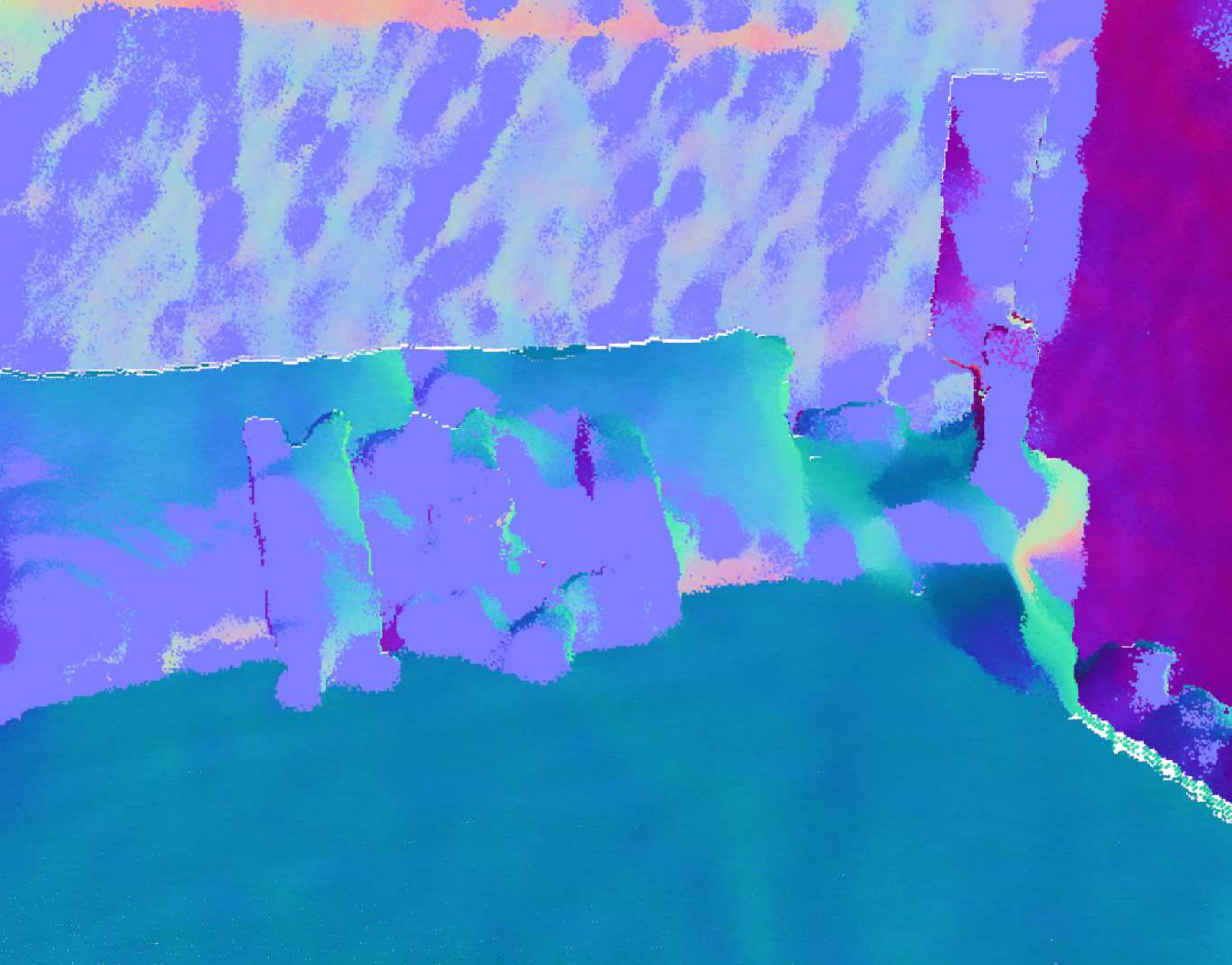}%
	\includegraphics[width=\unit]{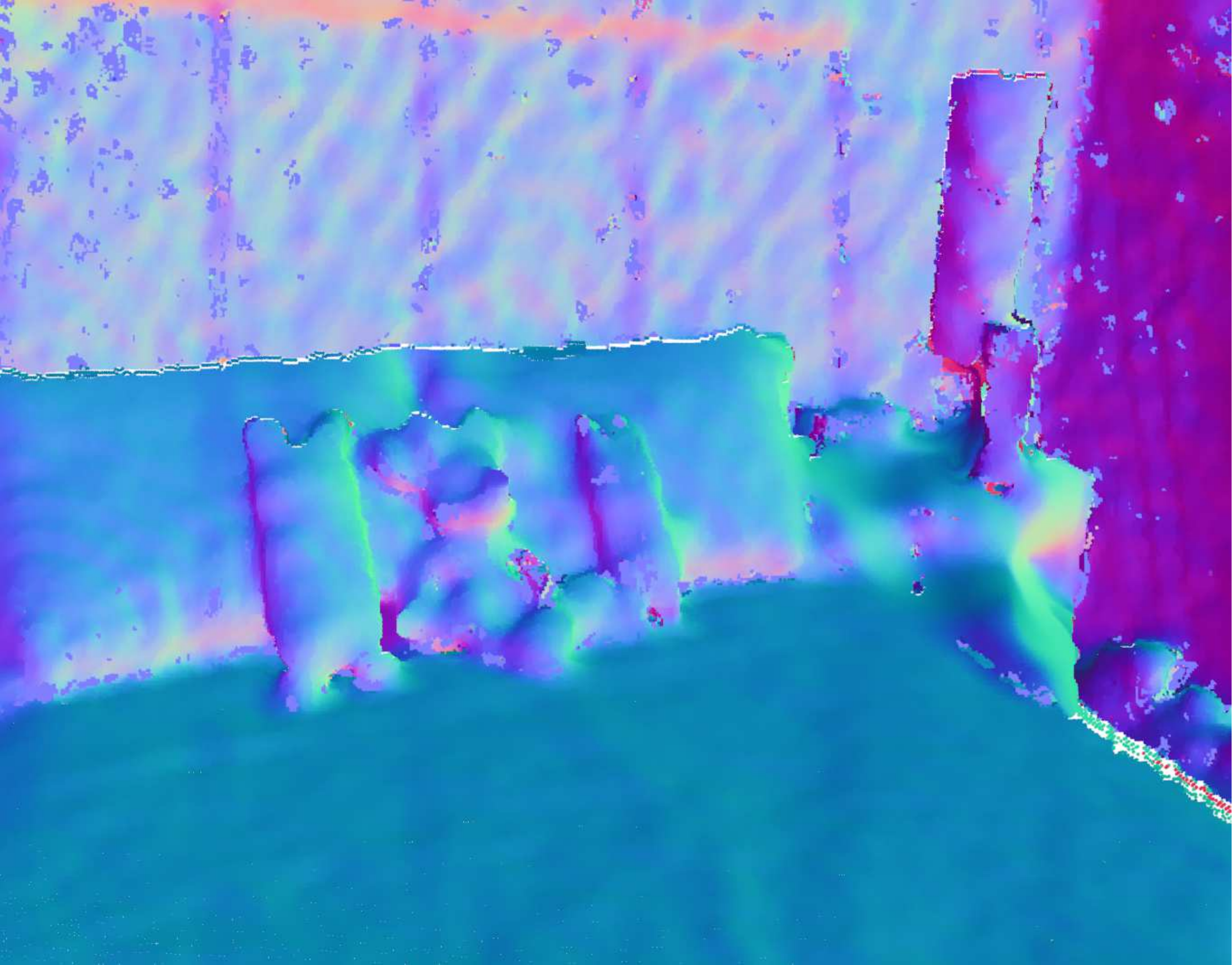}%
	\includegraphics[width=\unit]{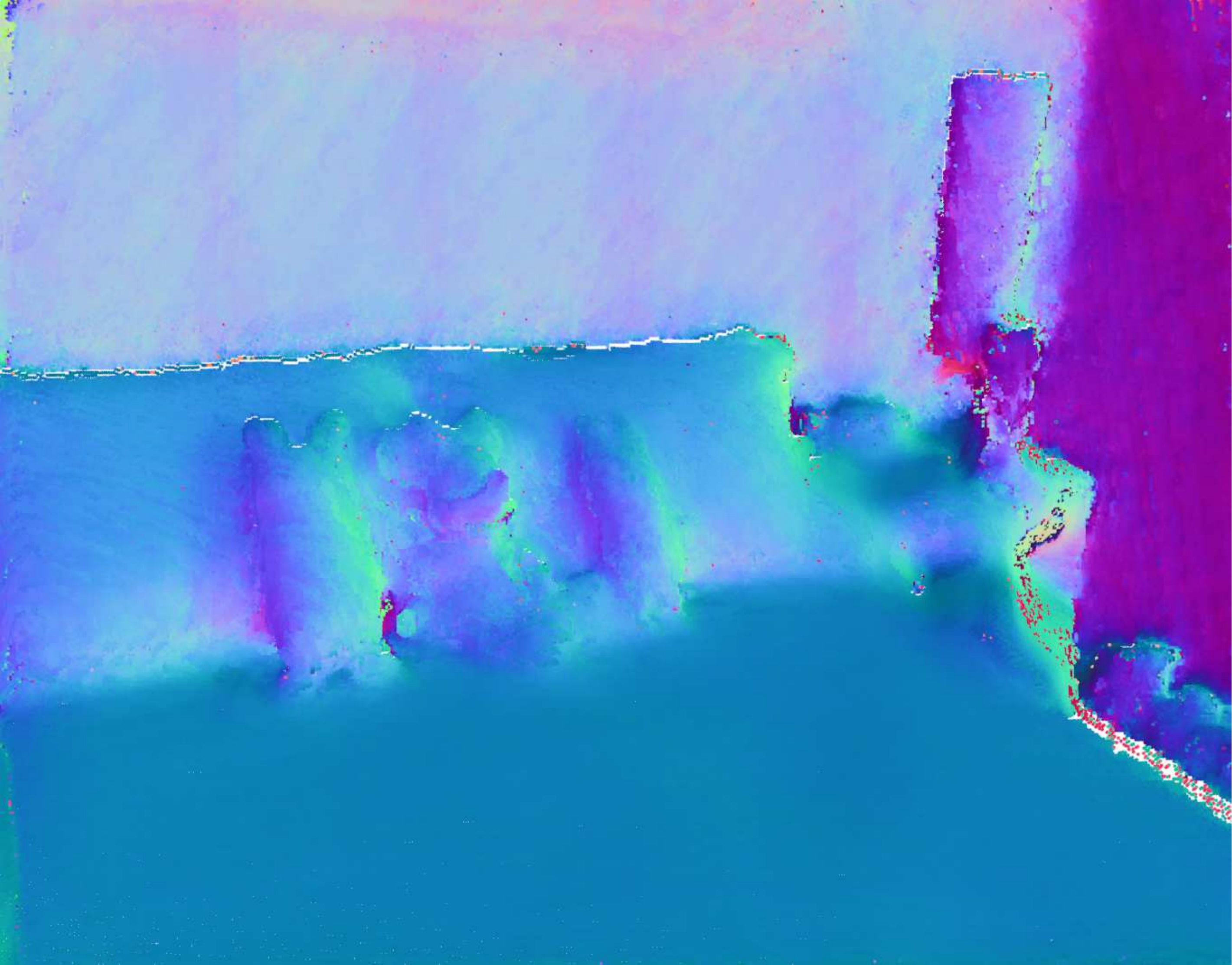}%
	\includegraphics[width=\unit]{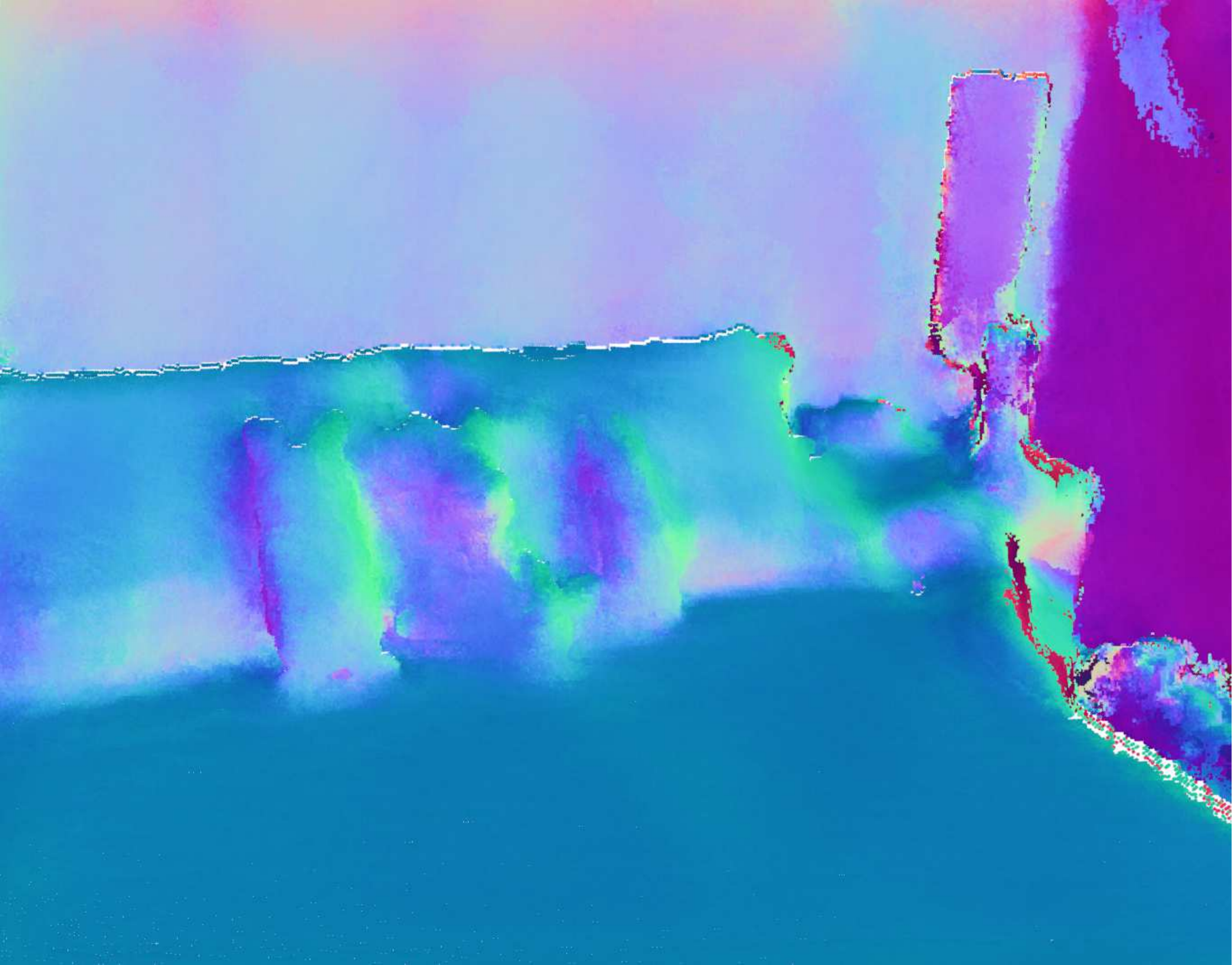}%
	\includegraphics[width=\unit]{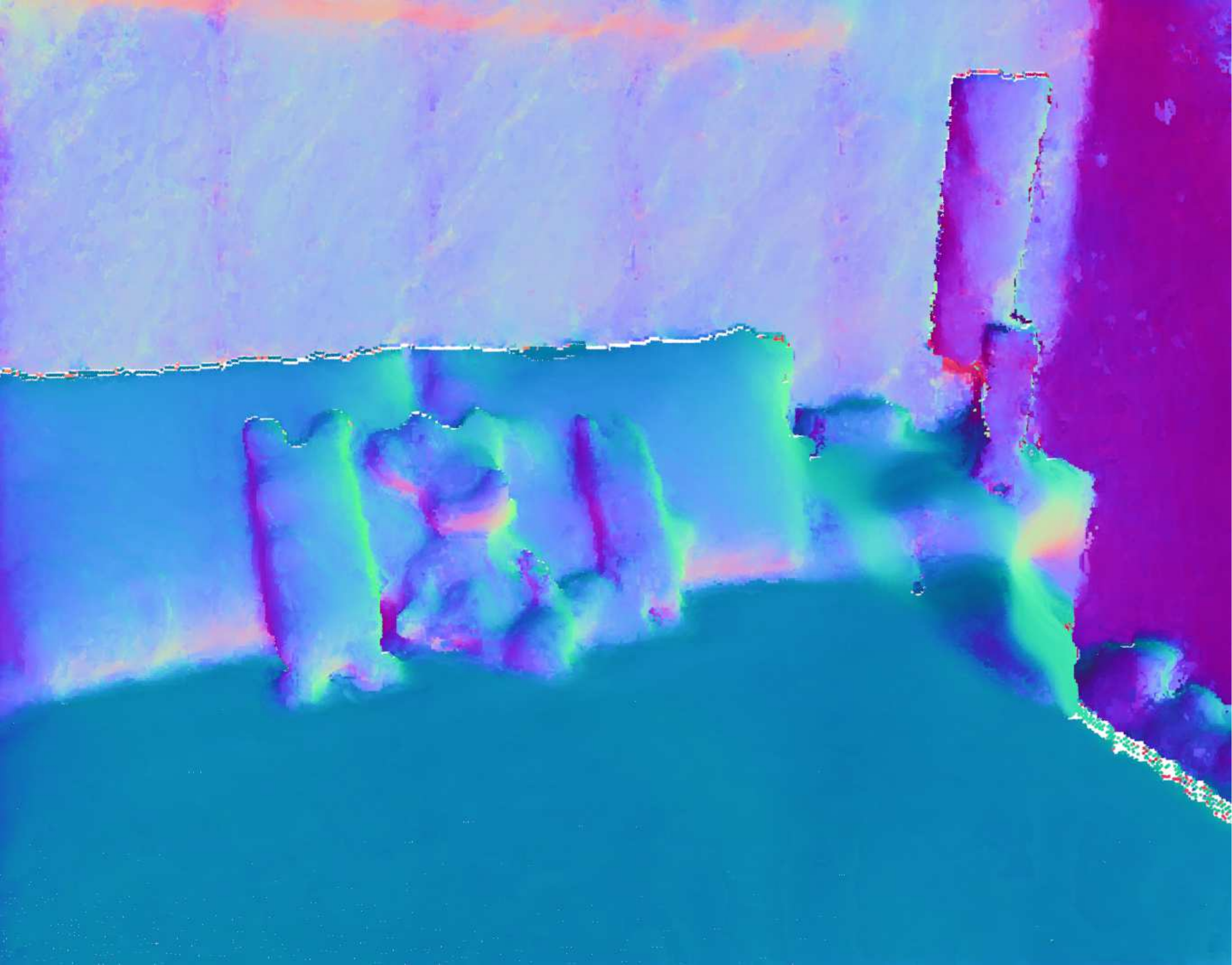}%
	\includegraphics[width=\unit]{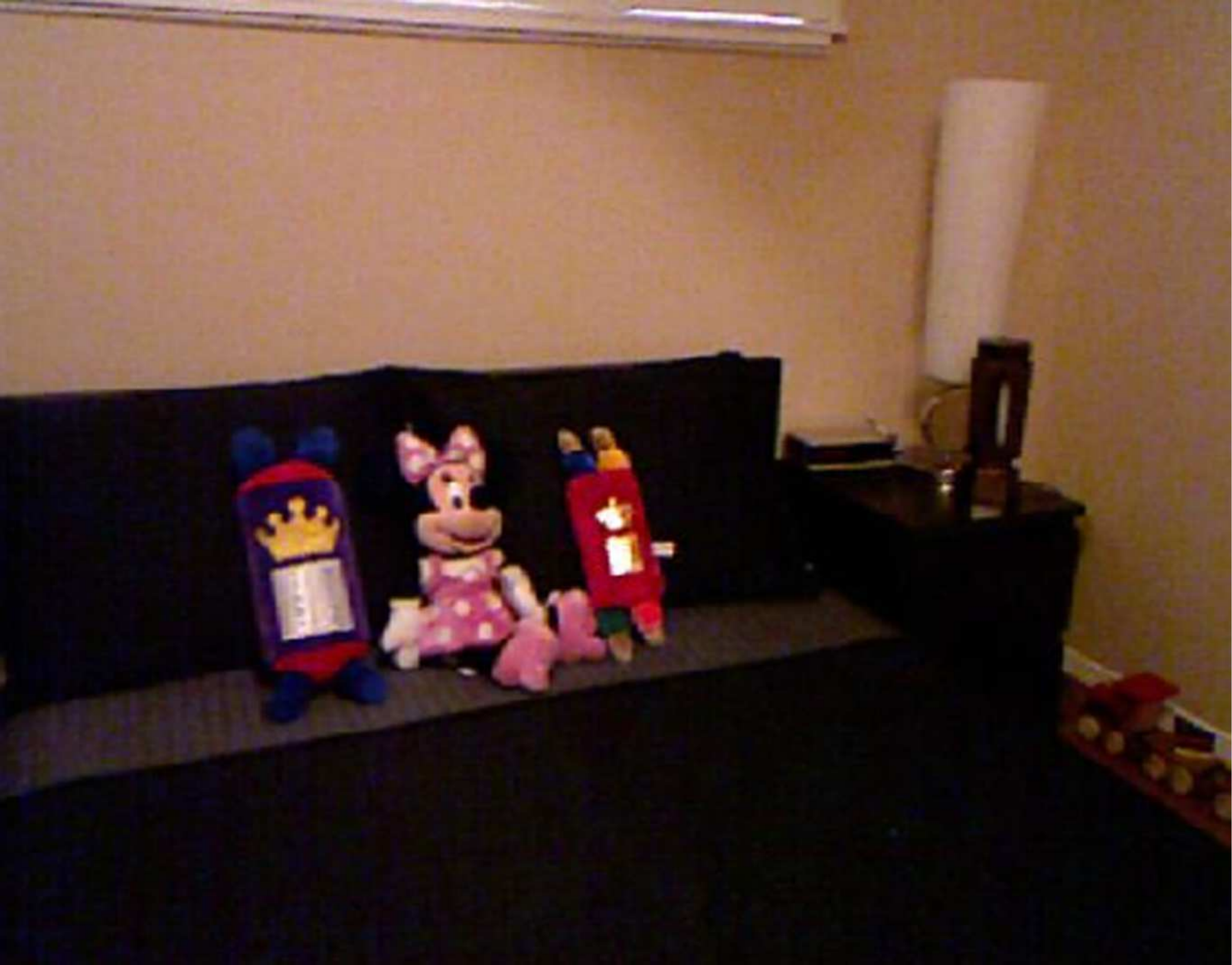}
	
	\vspace{-2pt}
	
	\subfigure[Depth image]{\label{fig:indoor:depth}\includegraphics[width=\unit]{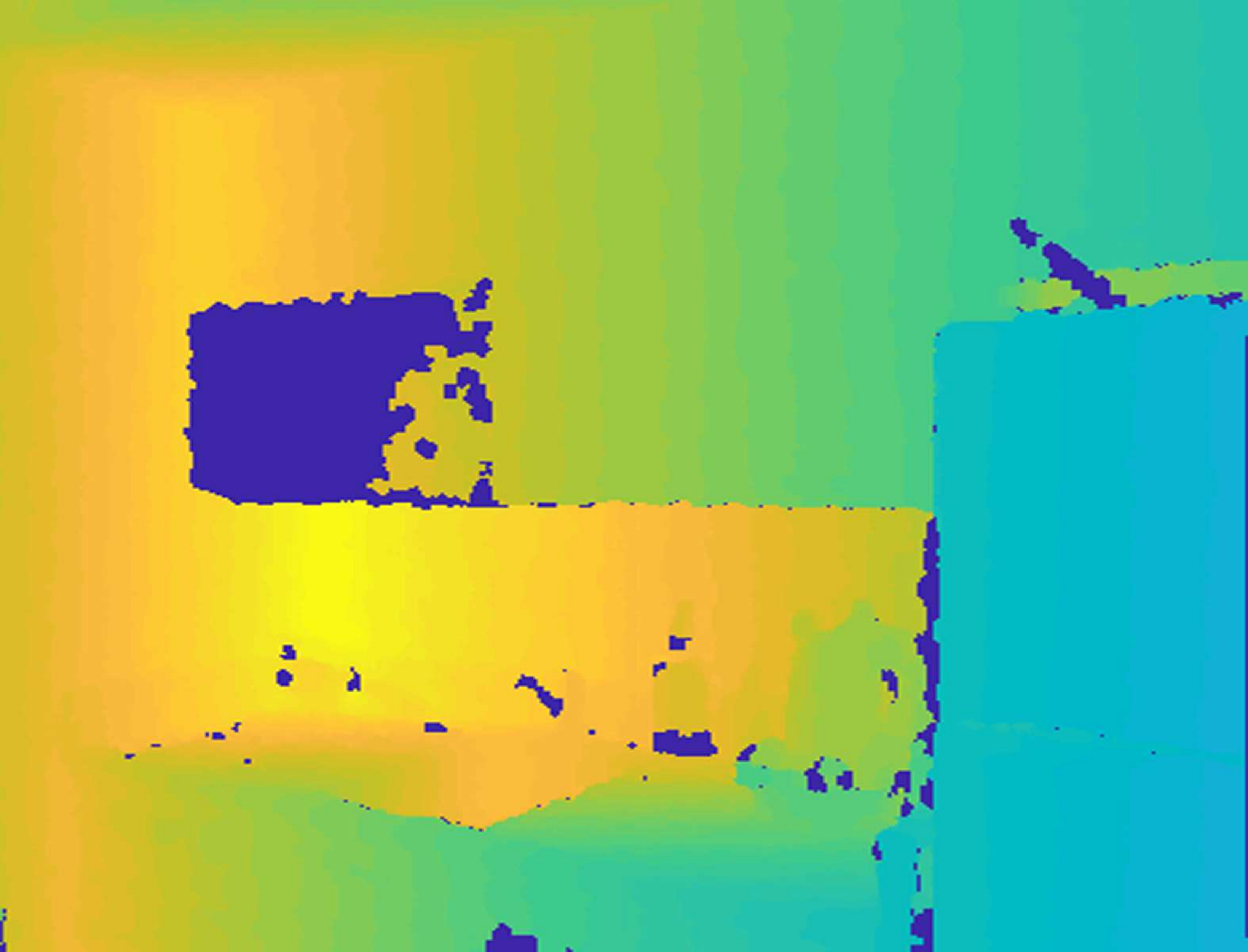}}%
	\subfigure[HF \cite{boulch2012fast}]{\label{fig:indoor:HF}\includegraphics[width=\unit]{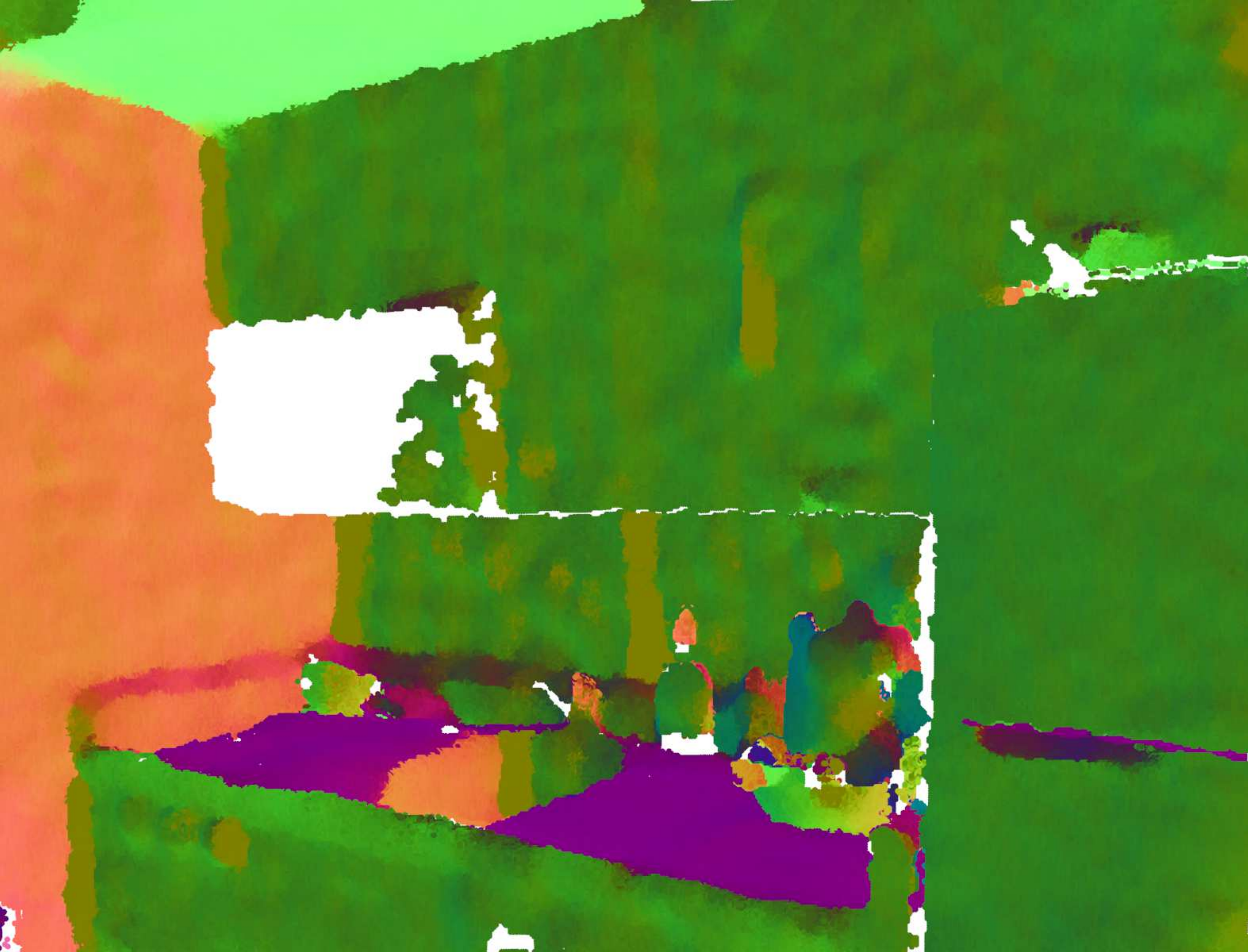}}%
	\subfigure[Our MFPS]{\label{fig:indoor:MFPS}\includegraphics[width=\unit]{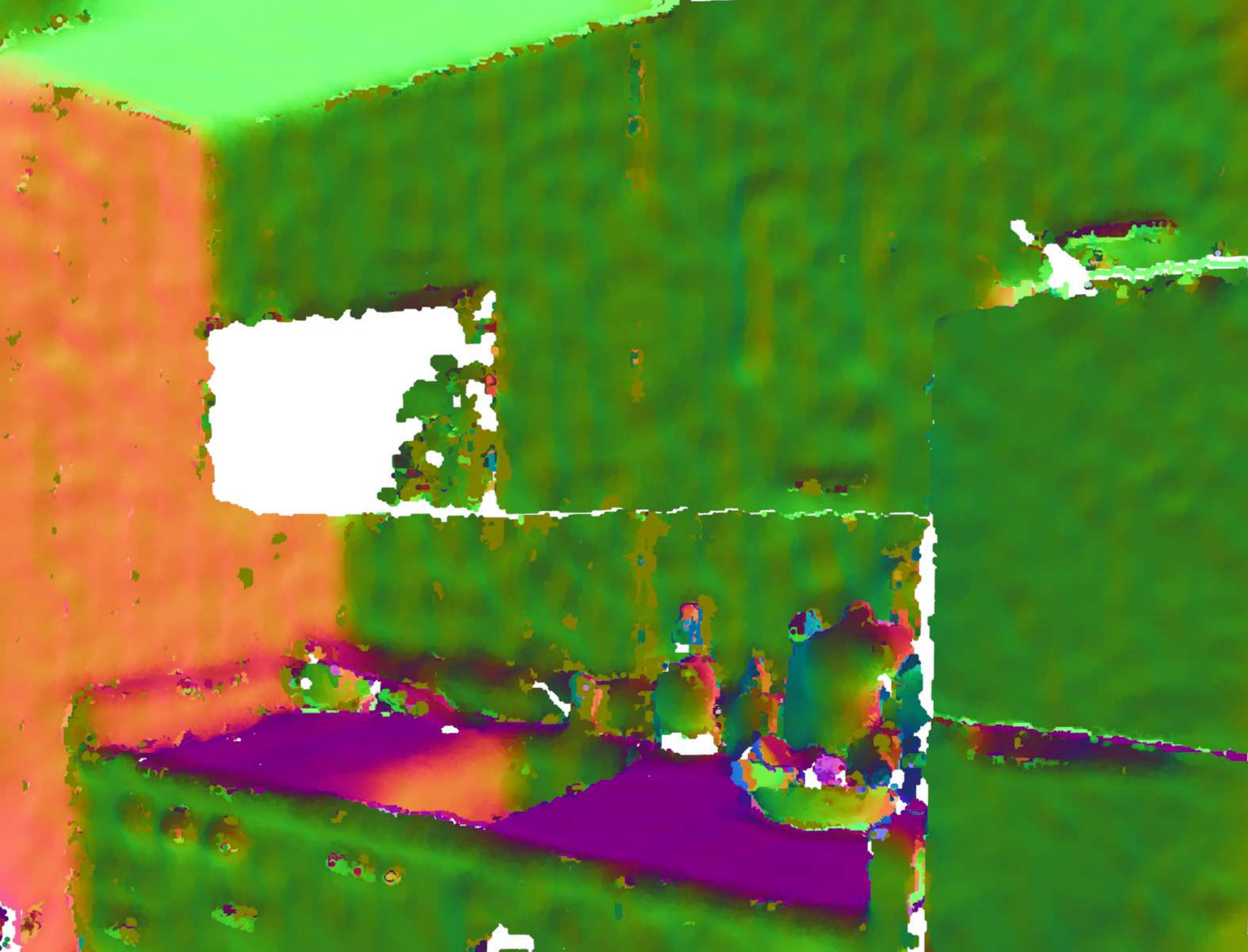}}%
	\subfigure[PCPNet \cite{GuerreroEtAl:PCPNet:EG:2018}]{\label{fig:indoor:pcp}\includegraphics[width=\unit]{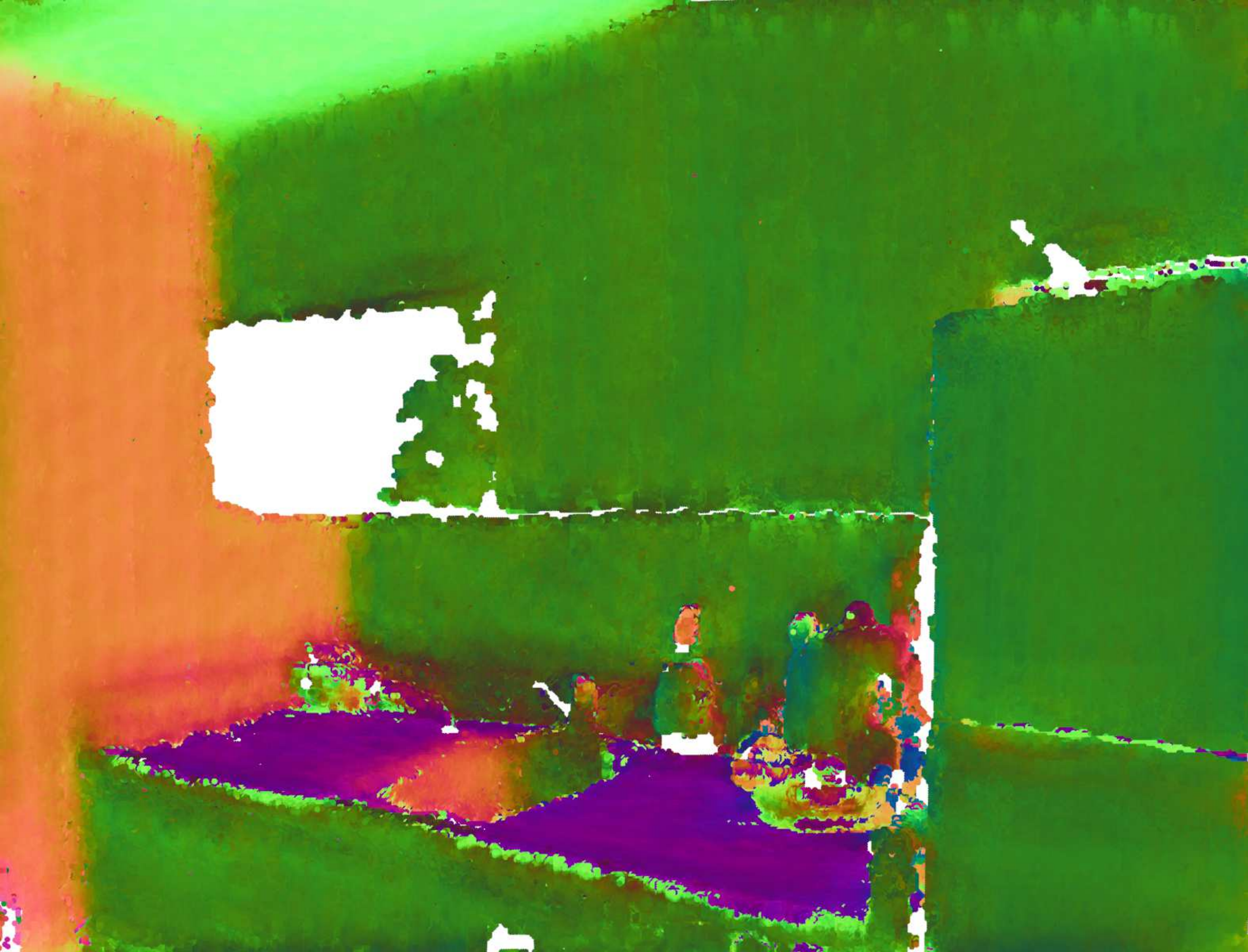}}%
	\subfigure[Nesti-Net \cite{Ben-ShabatLF19}]{\label{fig:indoor:nesti}\includegraphics[width=\unit]{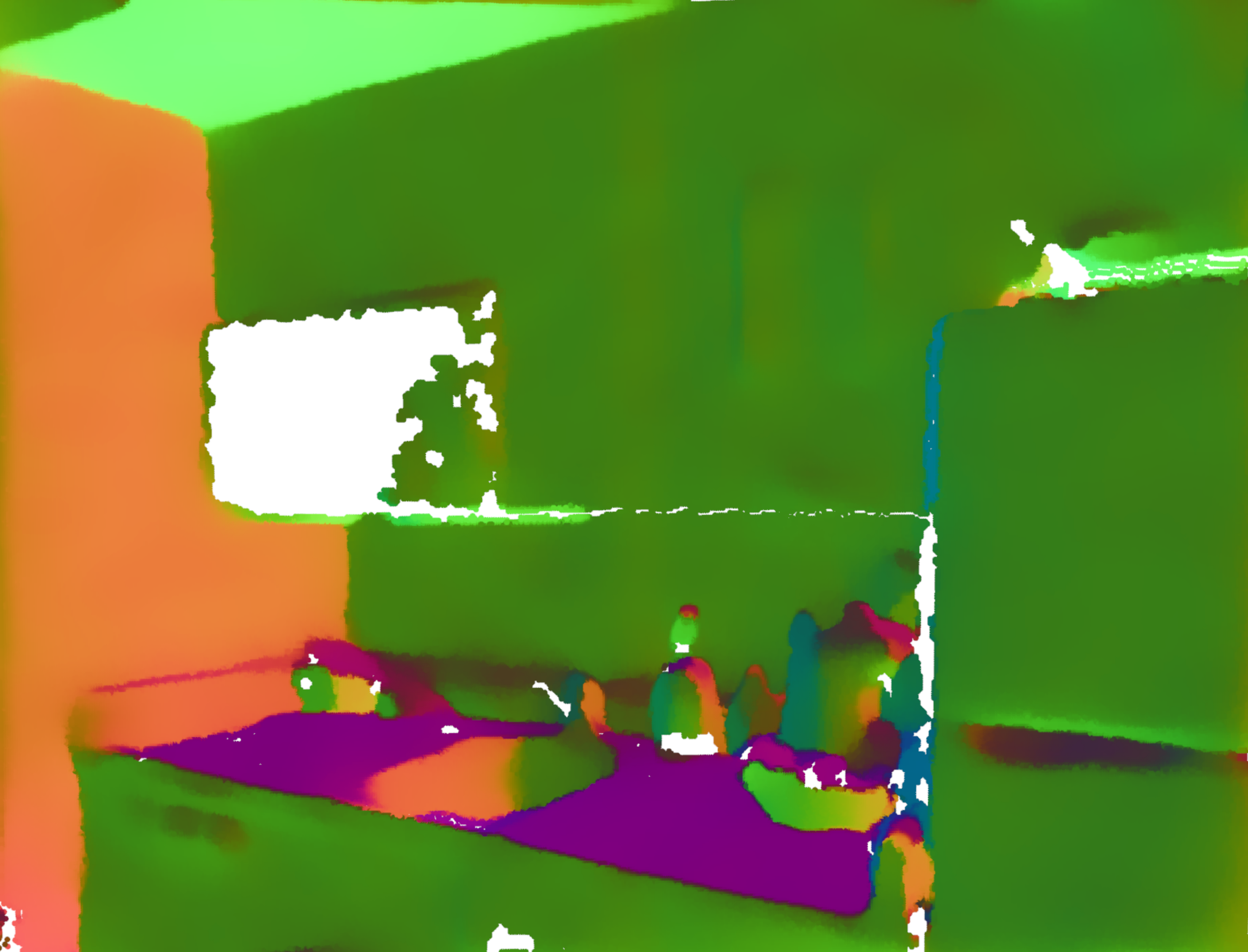}}%
	\subfigure[Our Refine-Net]{\label{fig:indoor:ours}\includegraphics[width=\unit]{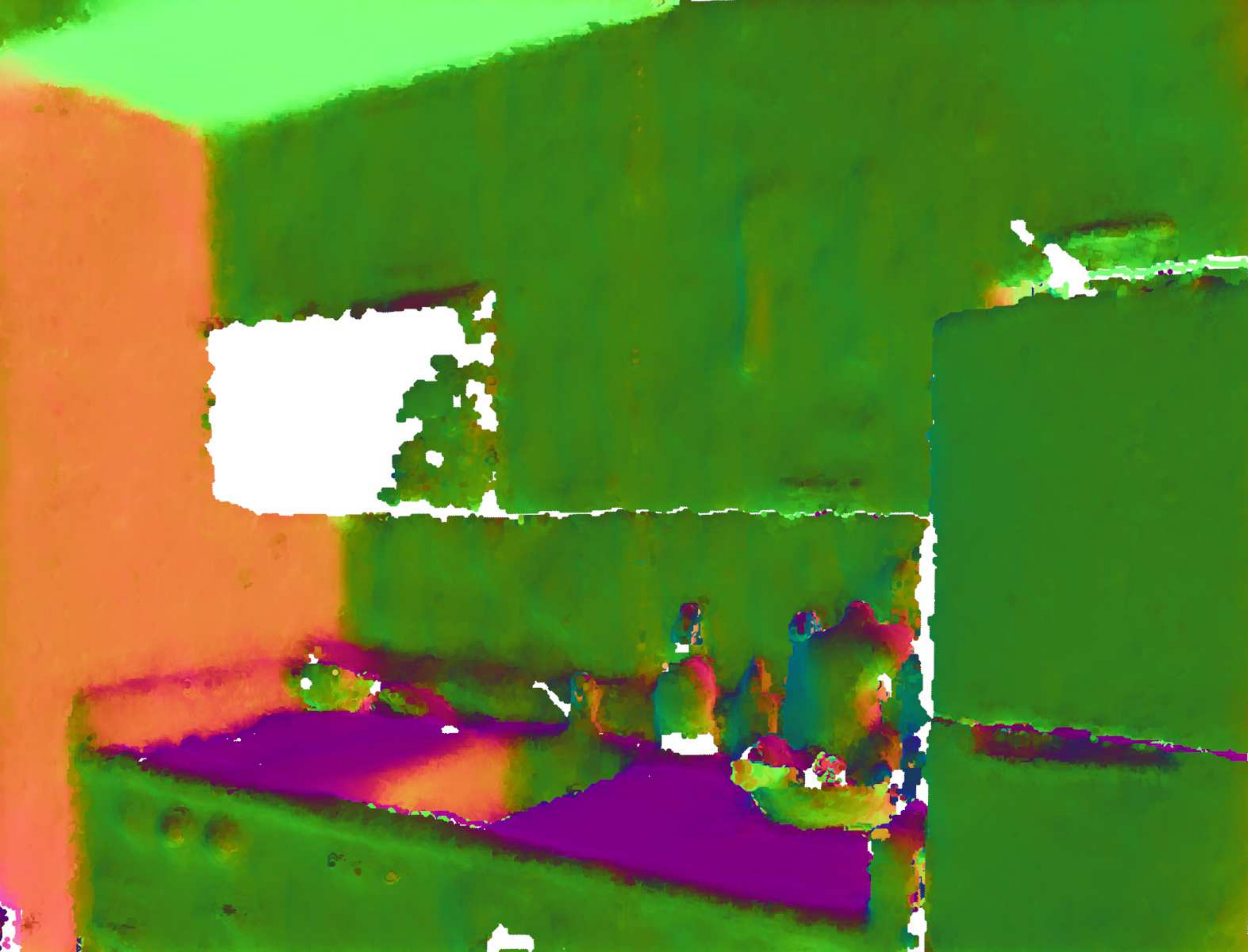}}%
	\subfigure[RGB image]{\includegraphics[width=\unit]{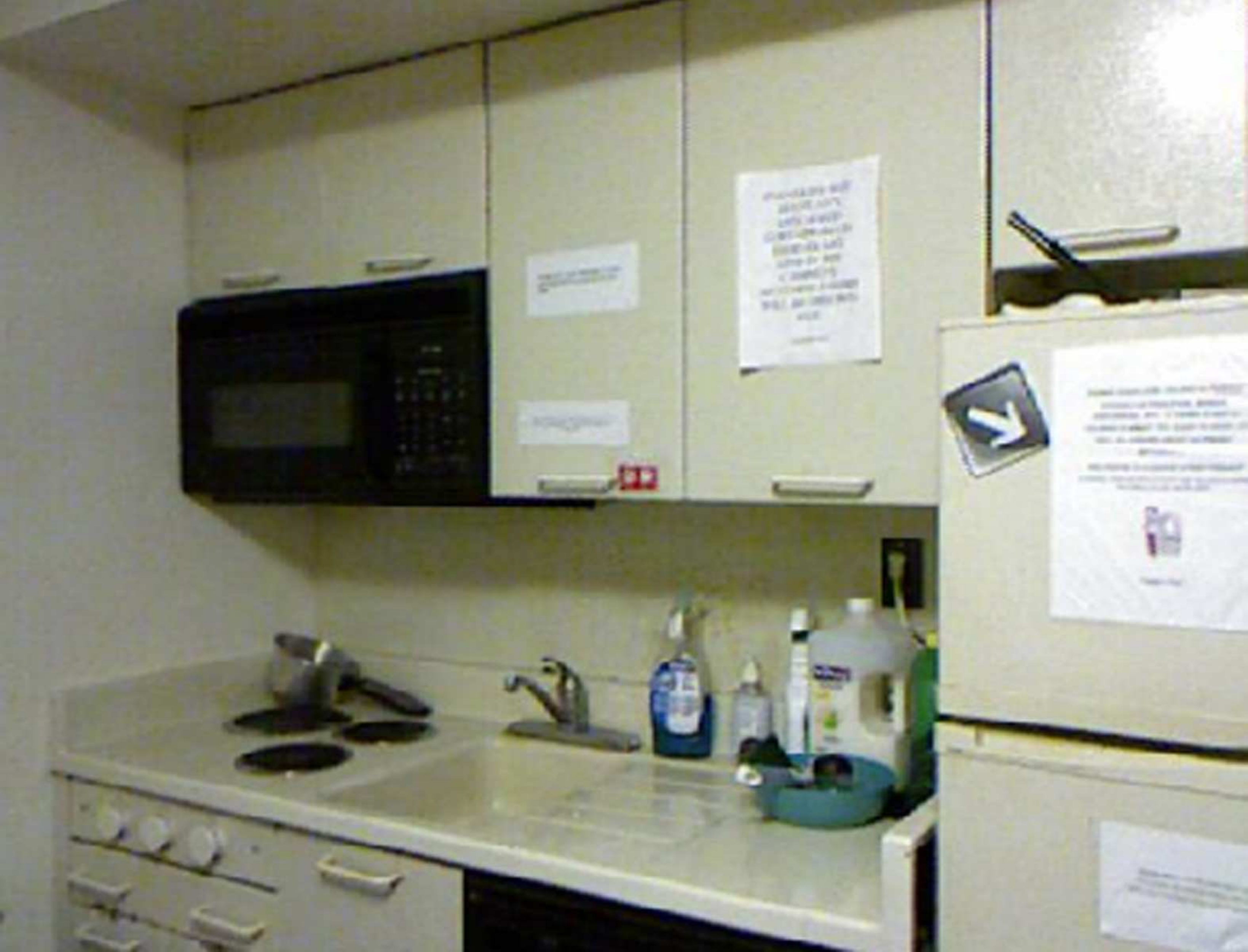}}
	
	\caption{Visual comparison of normal estimation results on the scanned point clouds from the NYU Depth V2 dataset \cite{silberman2012indoor}. We show the scanned depth map (noisy input) in (a) and the corresponding RGB image in (g). PCPNet \cite{GuerreroEtAl:PCPNet:EG:2018} and Nesti-Net \cite{Ben-ShabatLF19} tend to smooth the captured small objects and edges. Our method clearly produces better normal results with nice geometric details. }
	\label{fig:indoor}
\end{figure*}

\begin{figure*}
	\newlength{\unitE}
	\setlength{\unitE}{0.122\linewidth}
	\centering
	\includegraphics[width=\unitE]{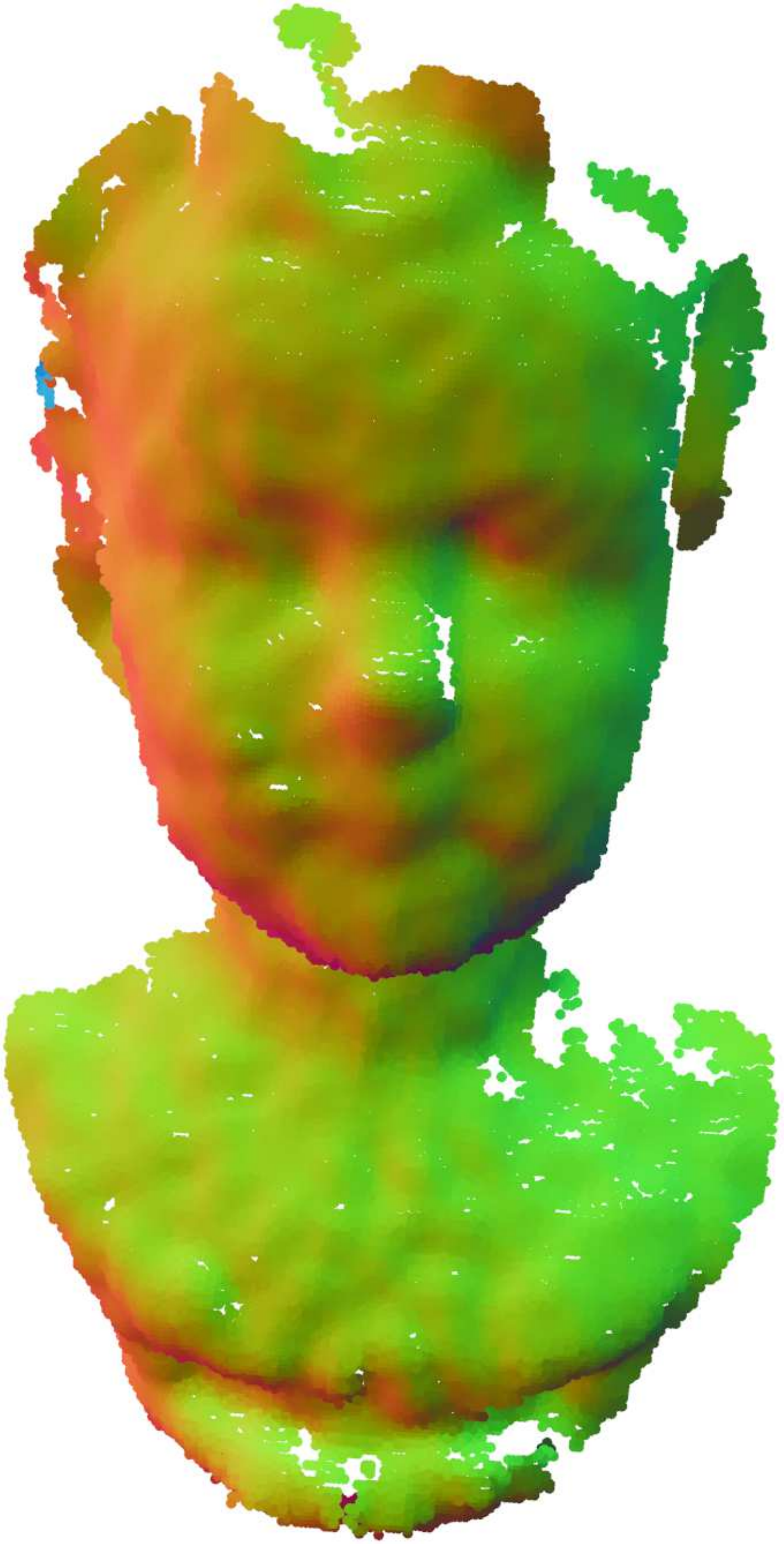}%
	\includegraphics[width=\unitE]{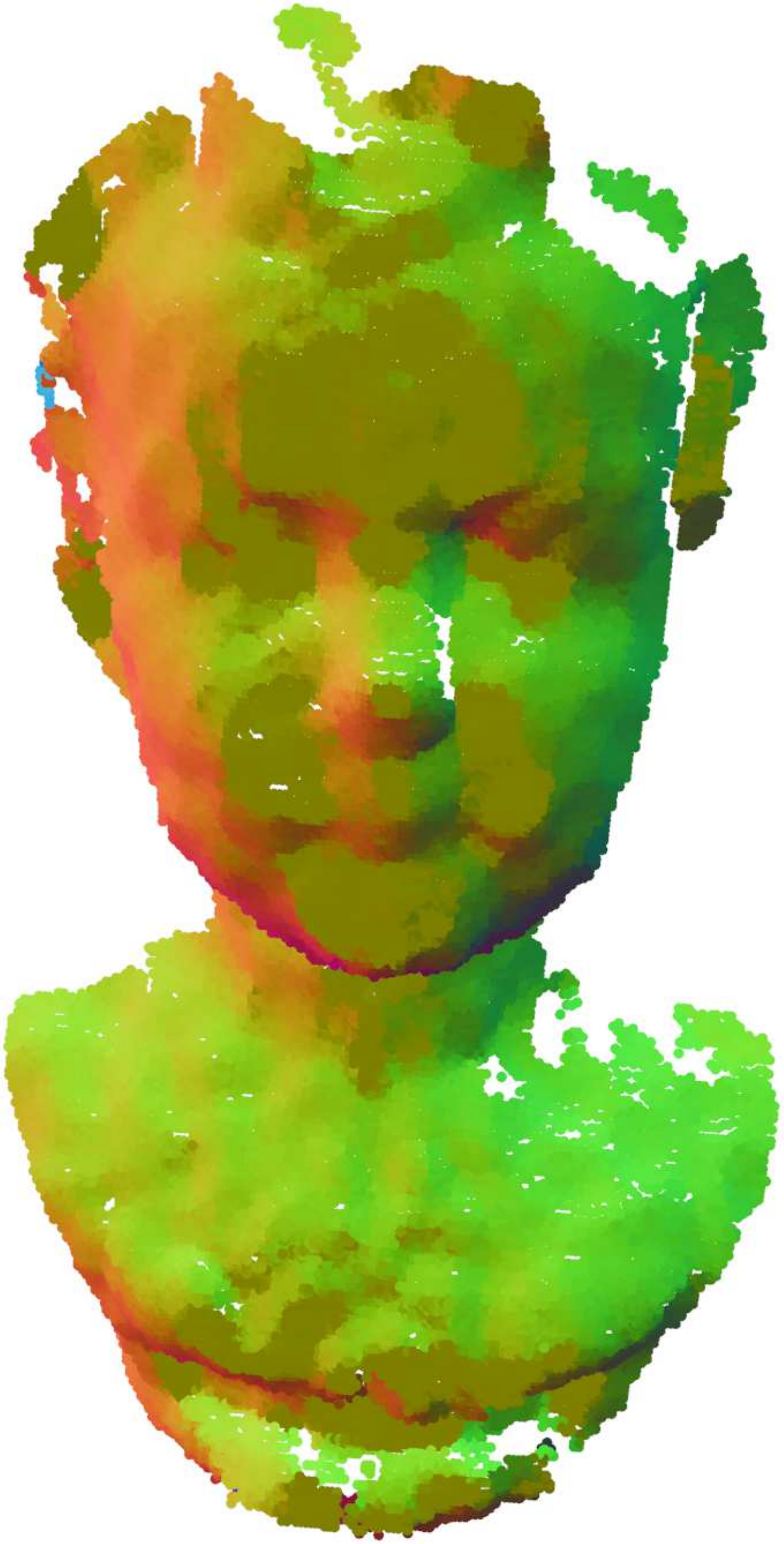}%
	\includegraphics[width=\unitE]{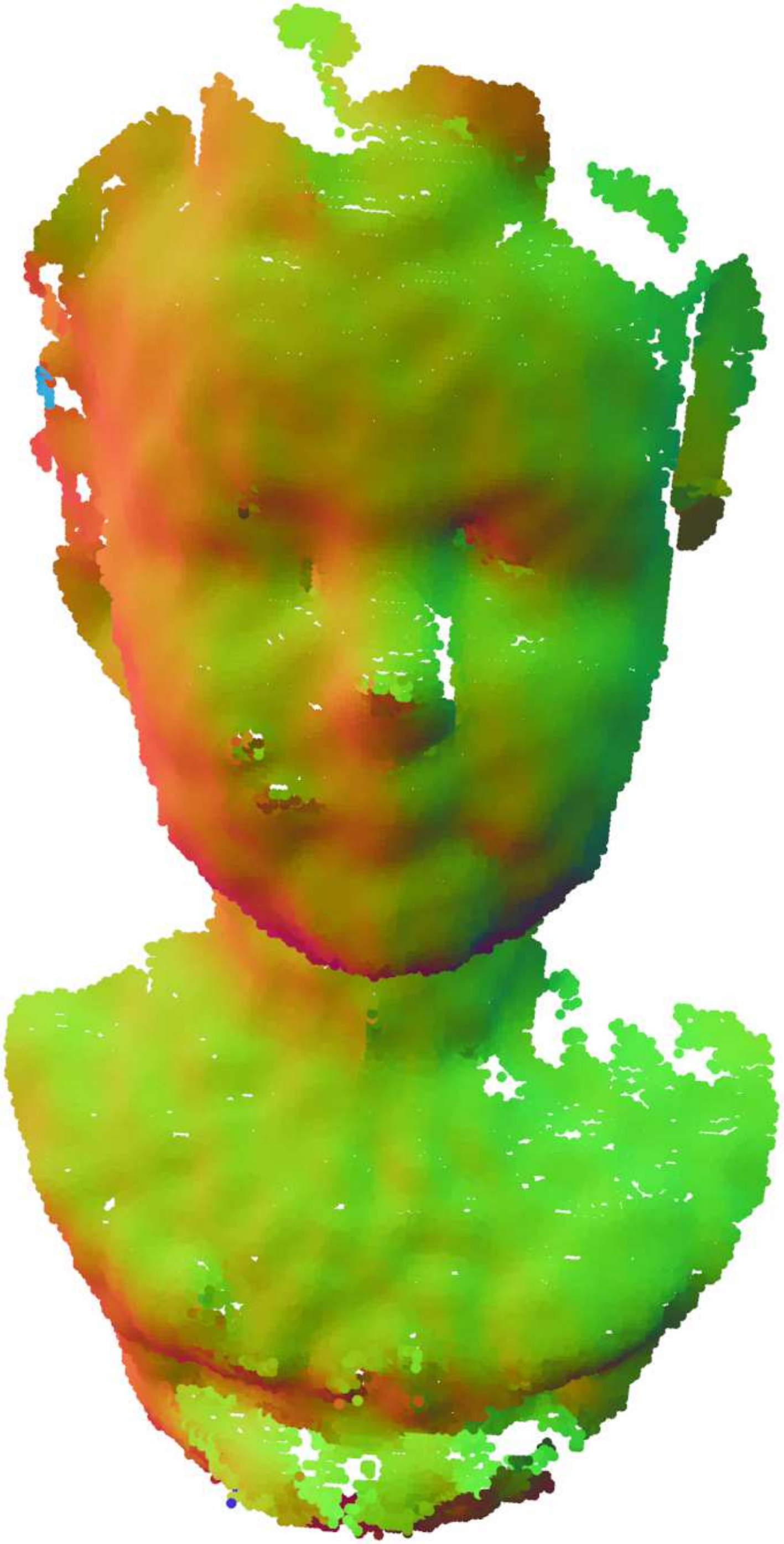}%
	\includegraphics[width=\unitE]{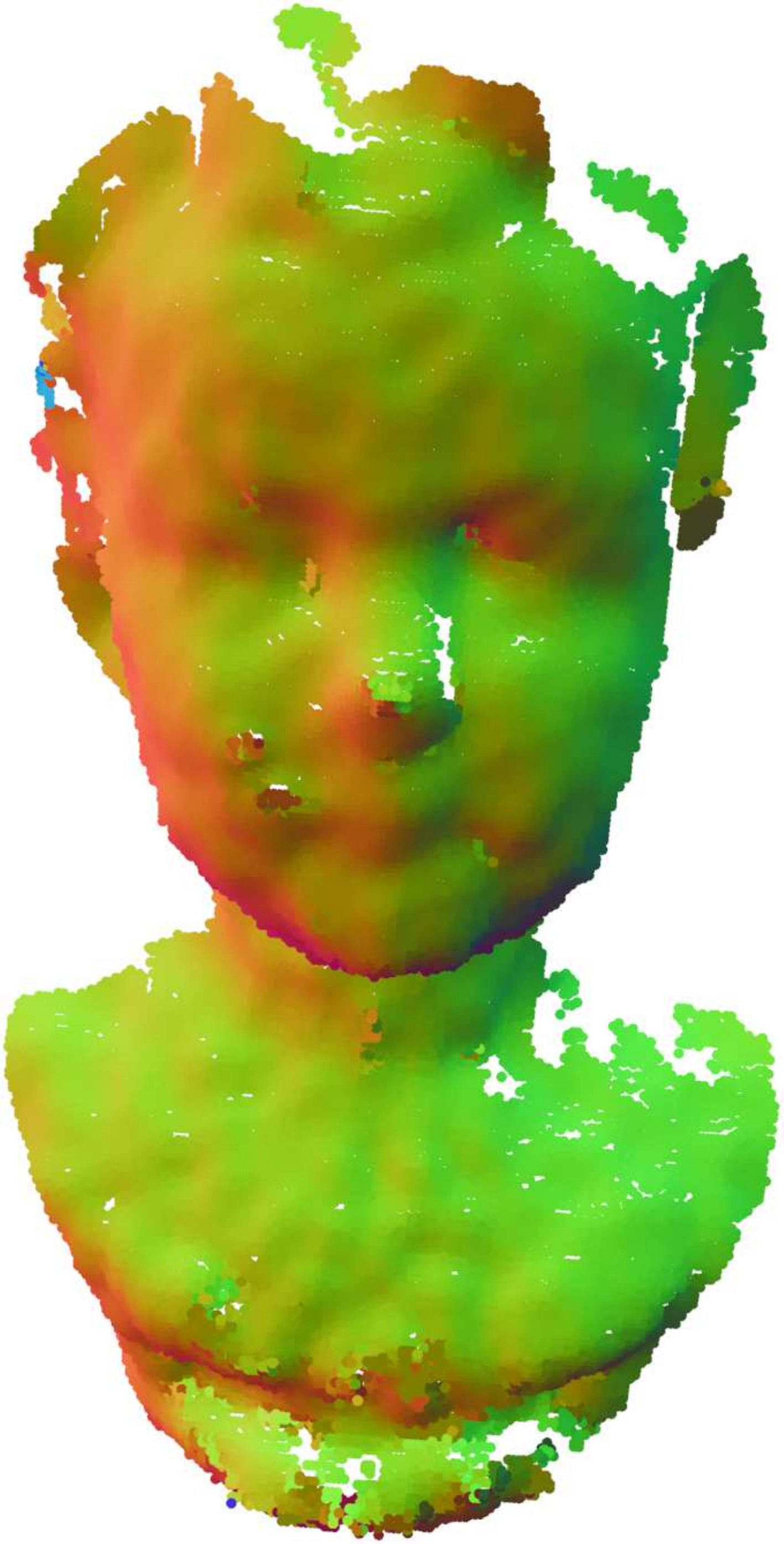}%
	\includegraphics[width=\unitE]{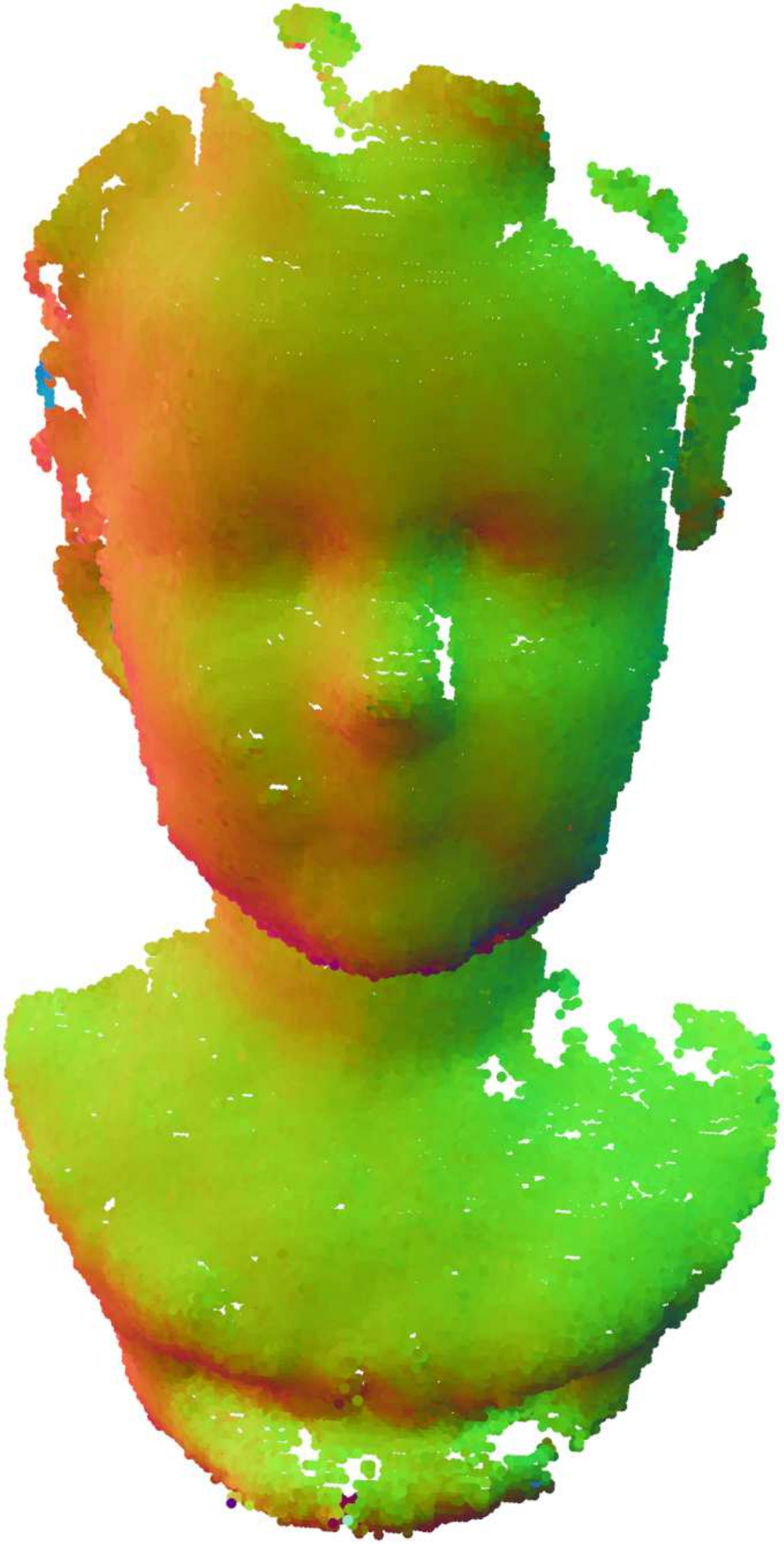}%
	\includegraphics[width=\unitE]{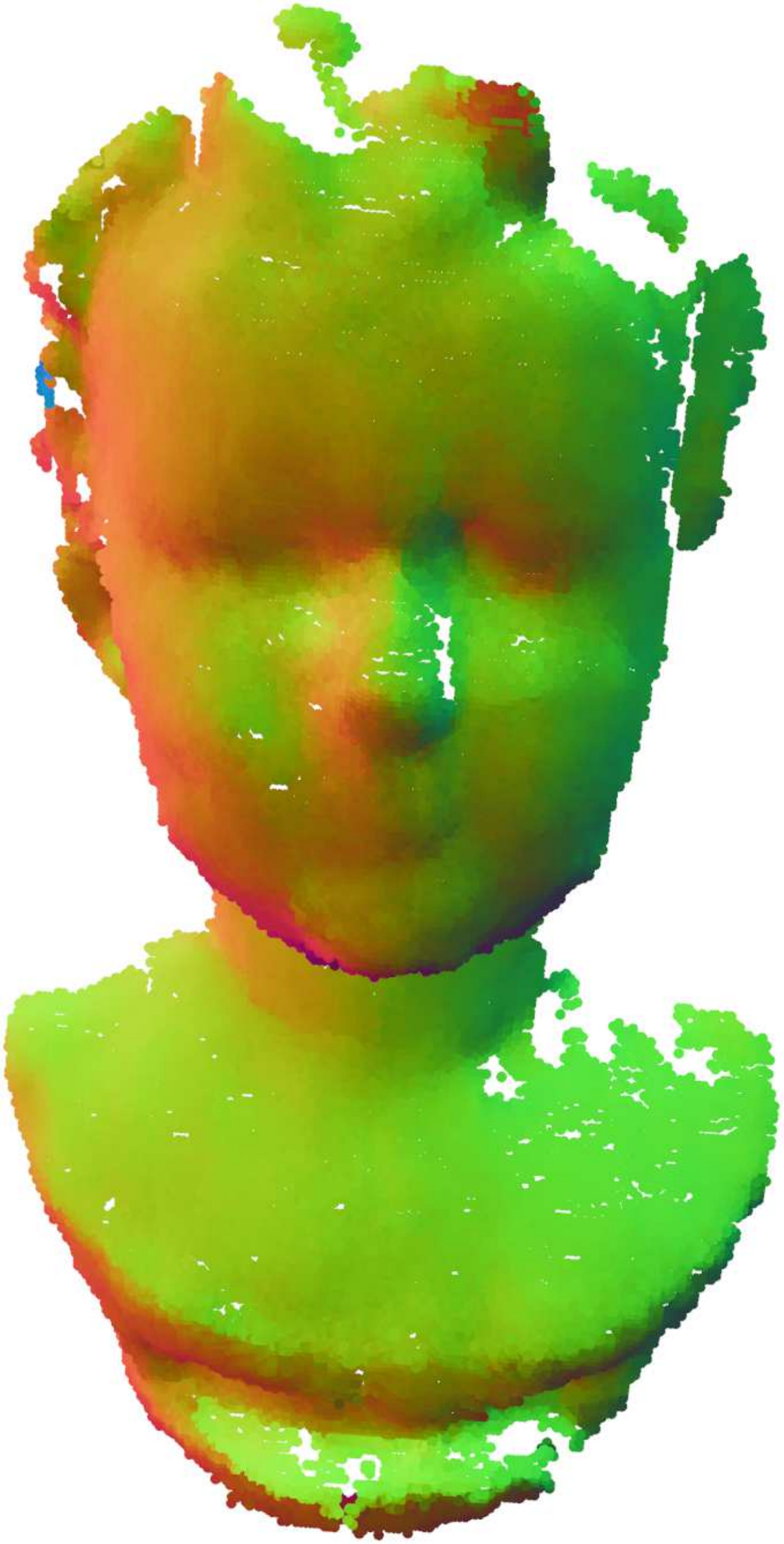}%
	\includegraphics[width=\unitE]{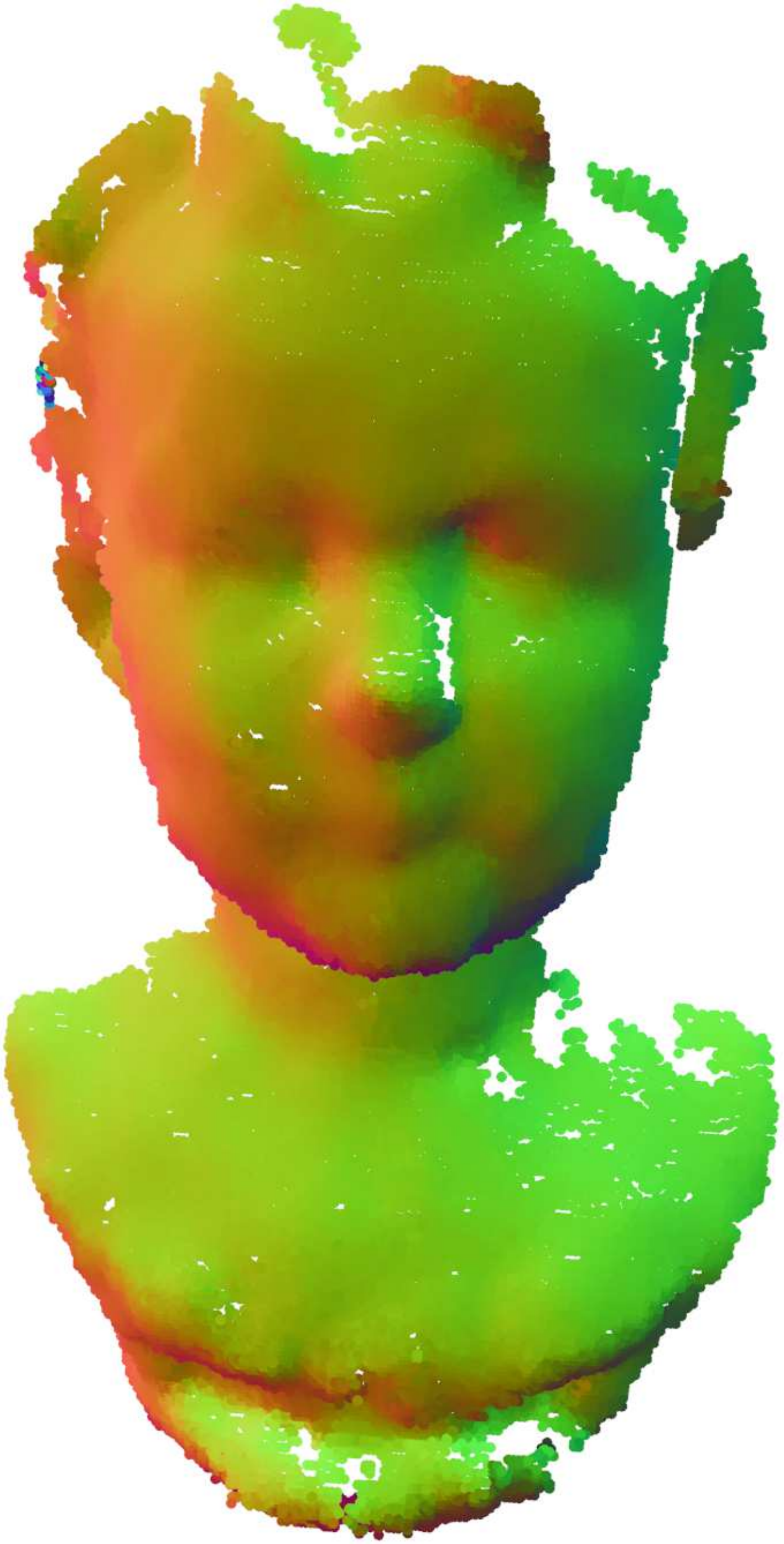}%
	\includegraphics[width=\unitE]{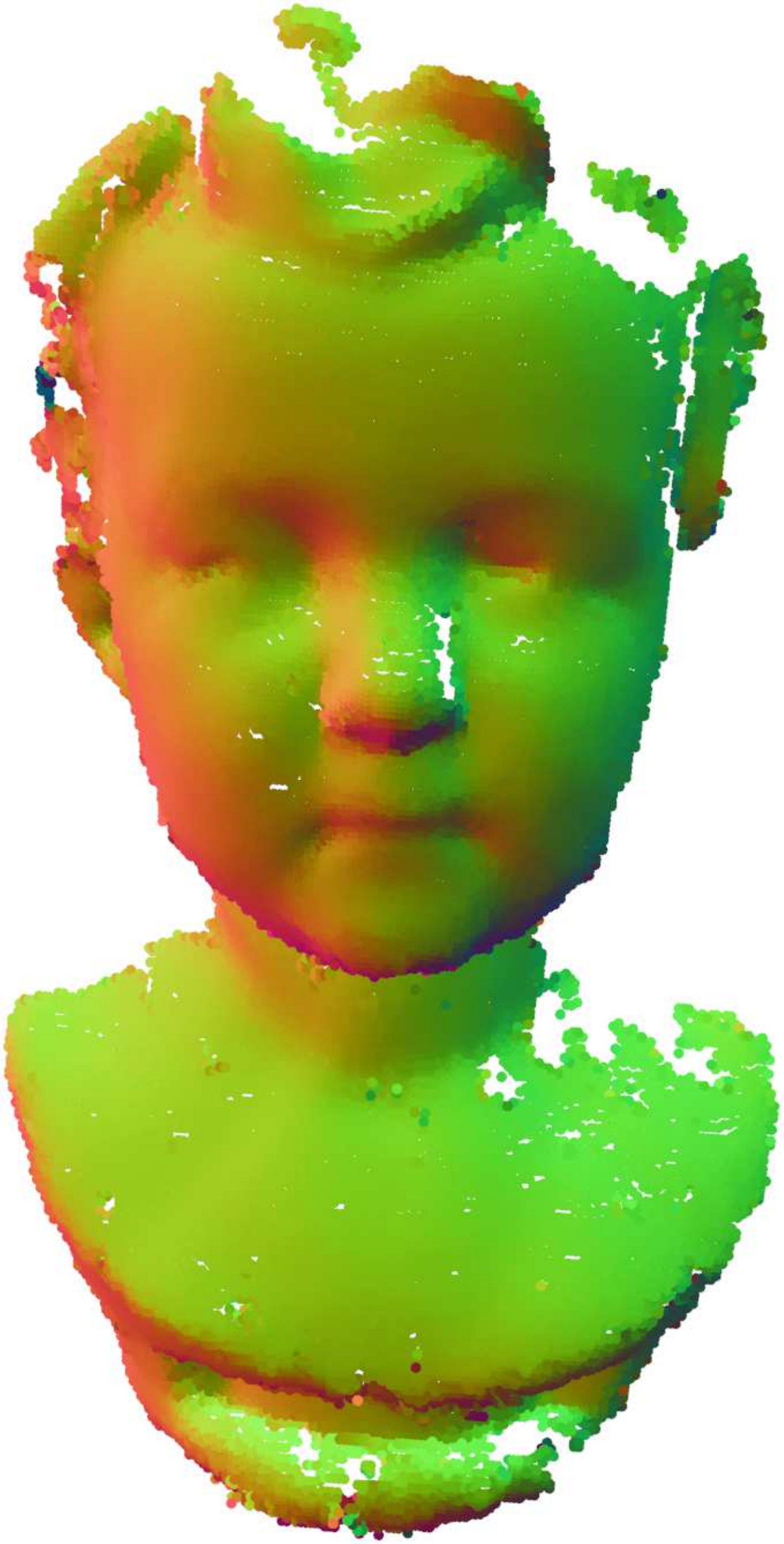}
	
	
	\subfigure[PCA \cite{hoppe1992surface}]{\includegraphics[width=\unitE]{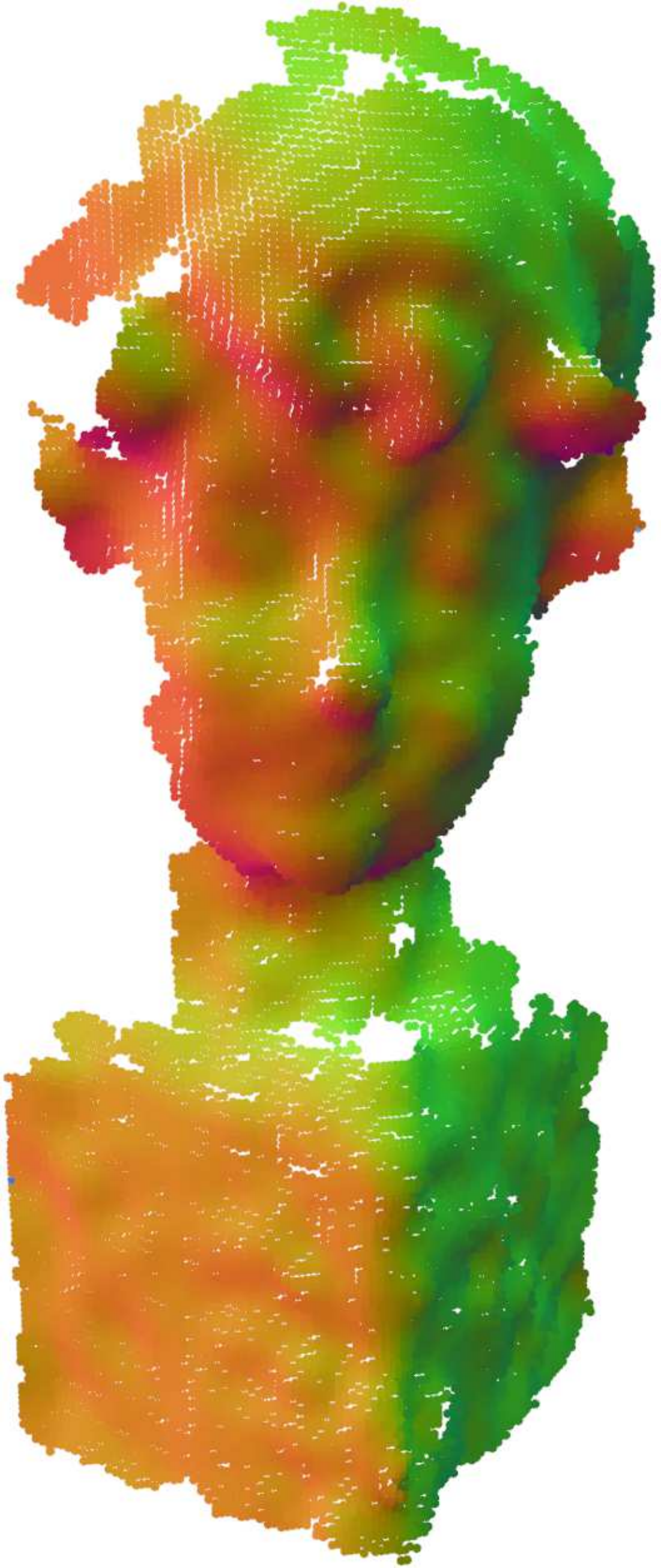}}%
	\subfigure[HF \cite{boulch2012fast}]{\includegraphics[width=\unitE]{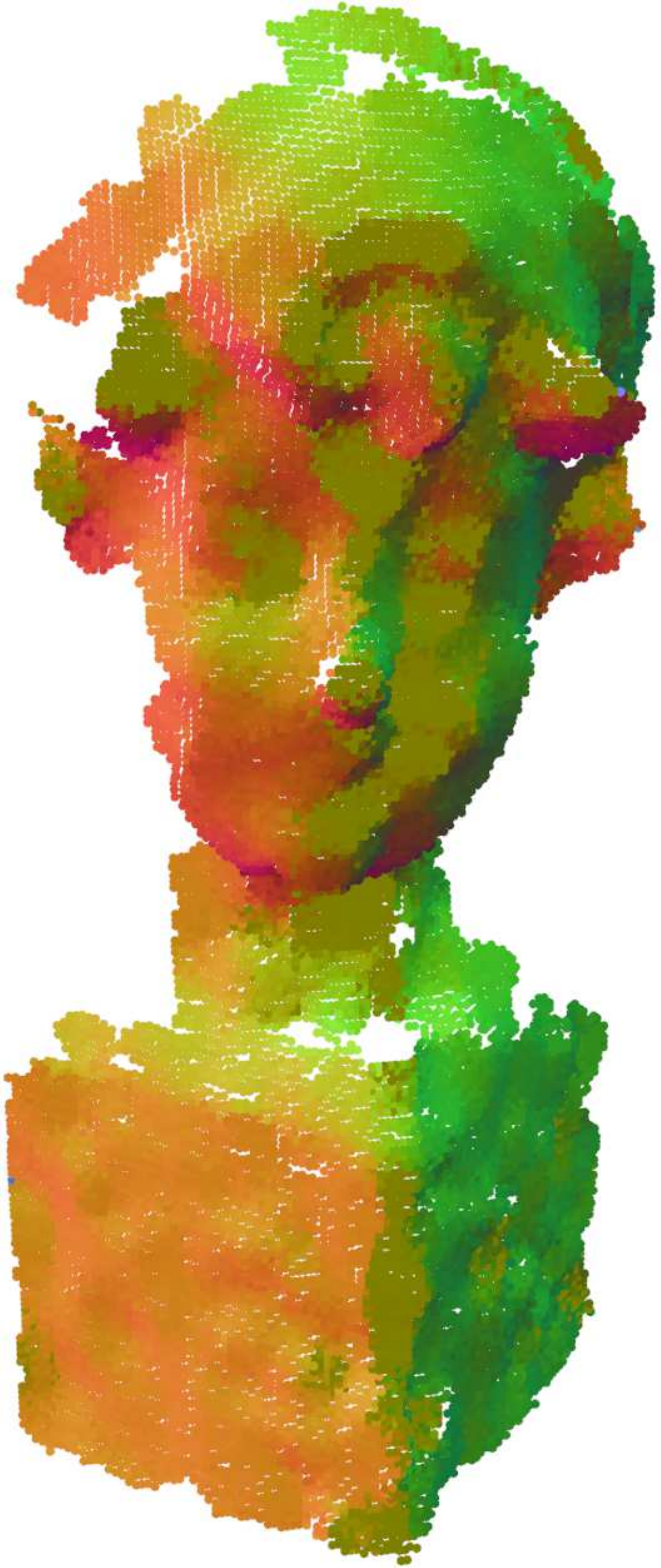}}%
	\subfigure[PCV \cite{zhang2018multi}]{\includegraphics[width=\unitE]{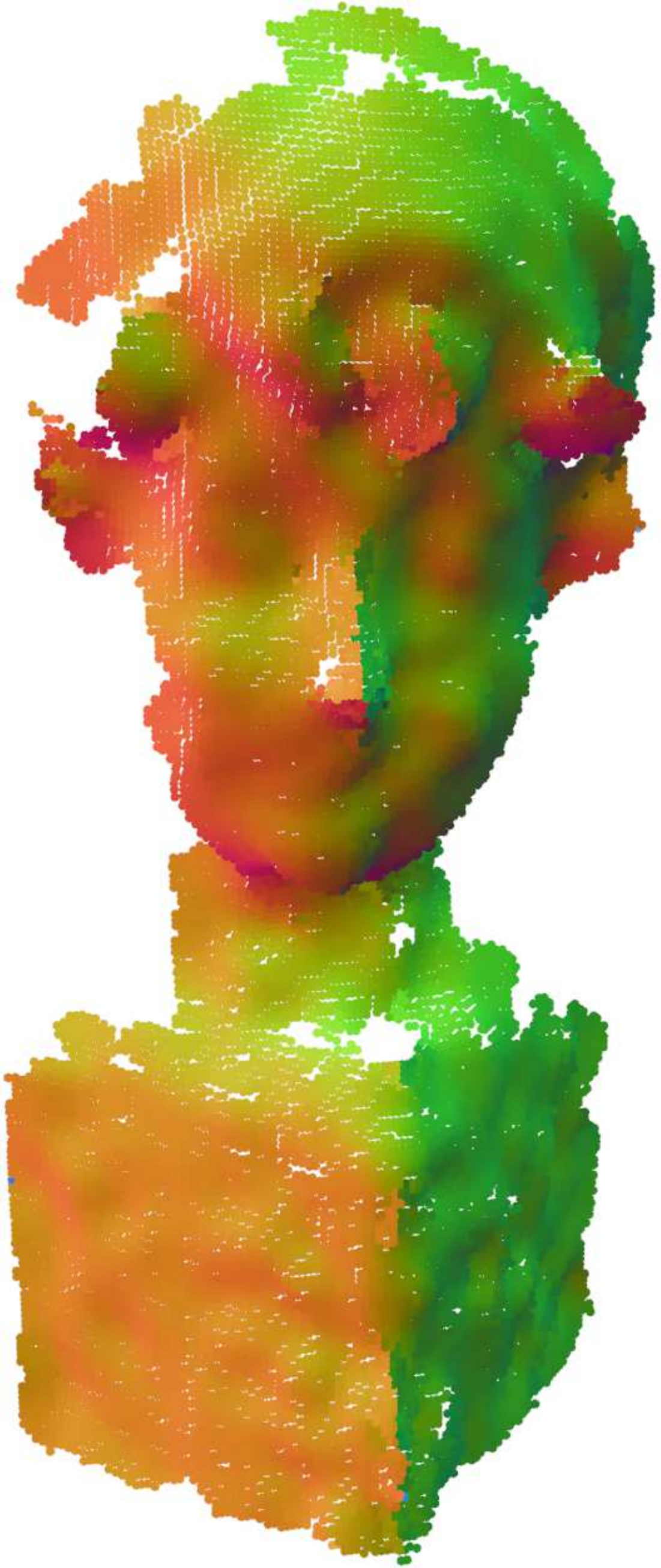}}%
	\subfigure[Our MFPS]{\includegraphics[width=\unitE]{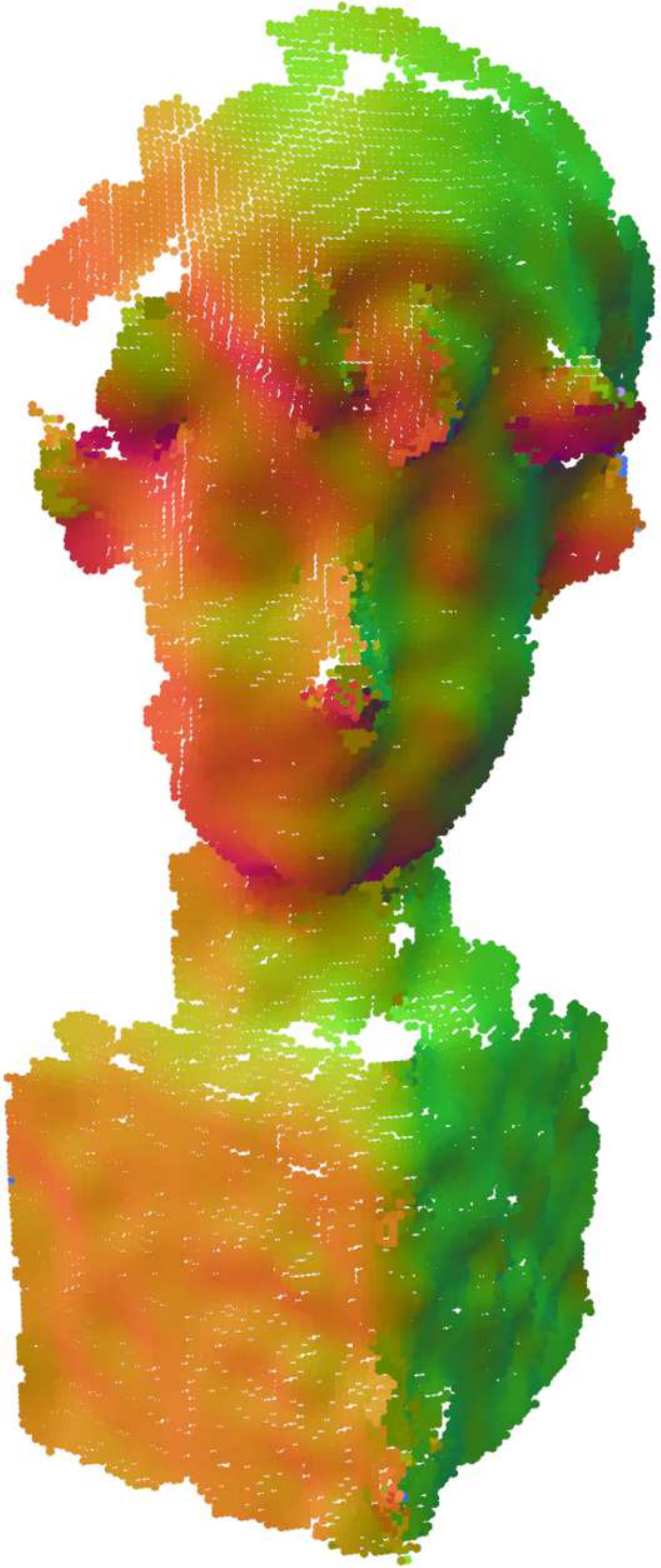}}%
	\subfigure[PCPNet \cite{GuerreroEtAl:PCPNet:EG:2018}]{\includegraphics[width=\unitE]{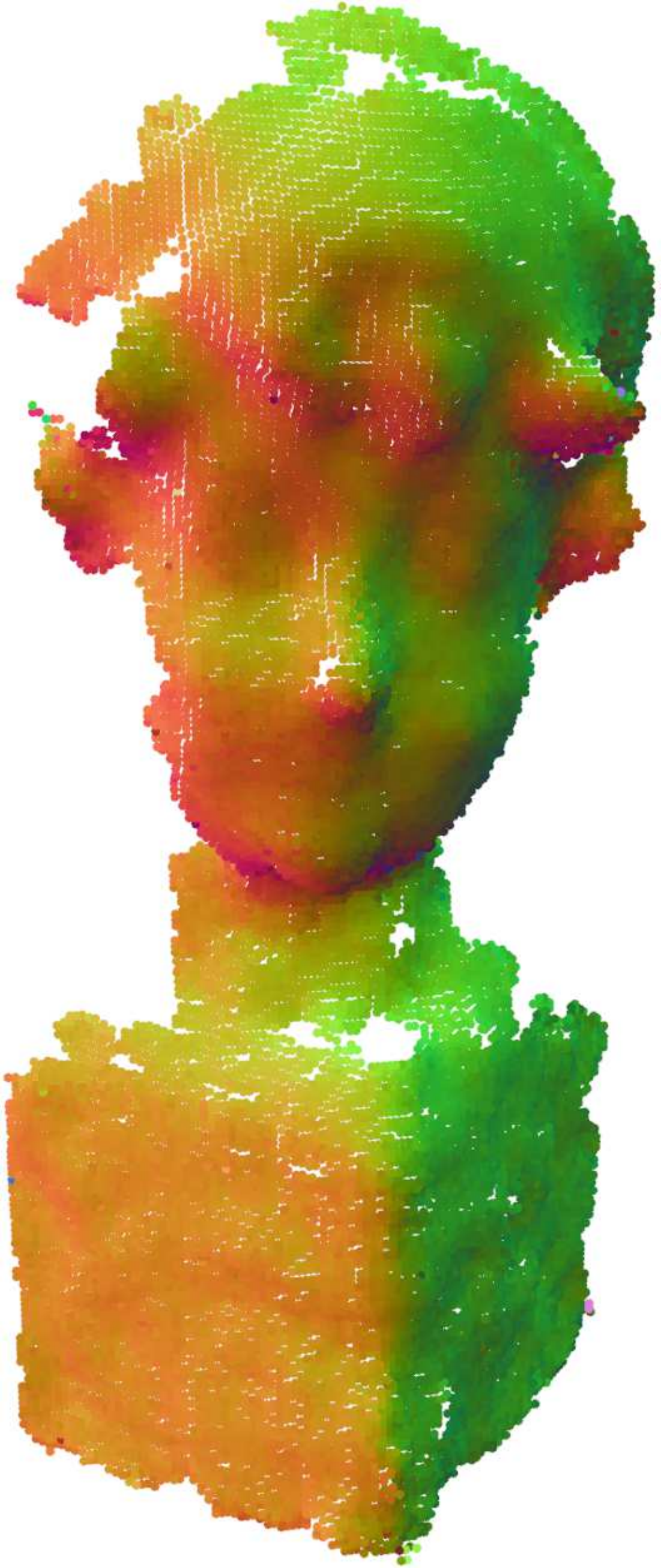}}%
	\subfigure[Nesti-Net \cite{Ben-ShabatLF19}]{\includegraphics[width=\unitE]{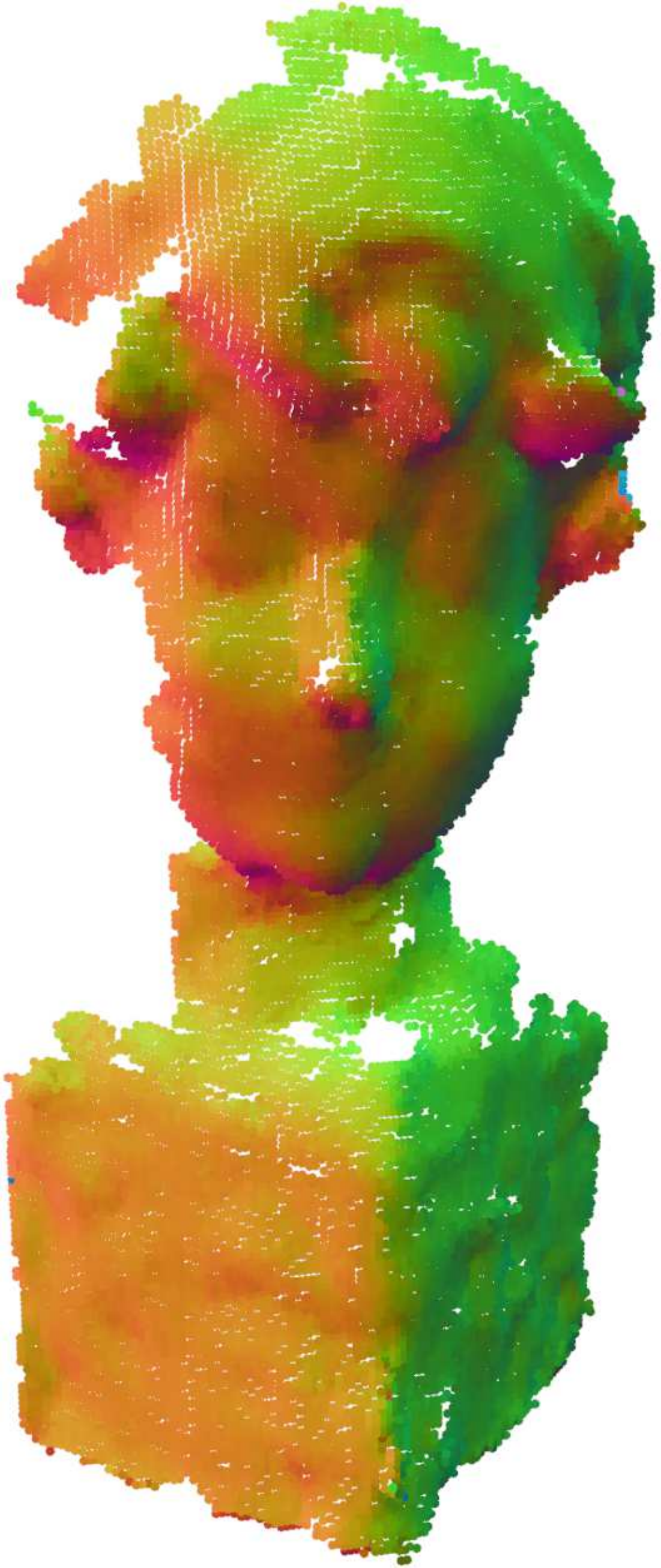}}%
	\subfigure[Our Refine-Net]{\includegraphics[width=\unitE]{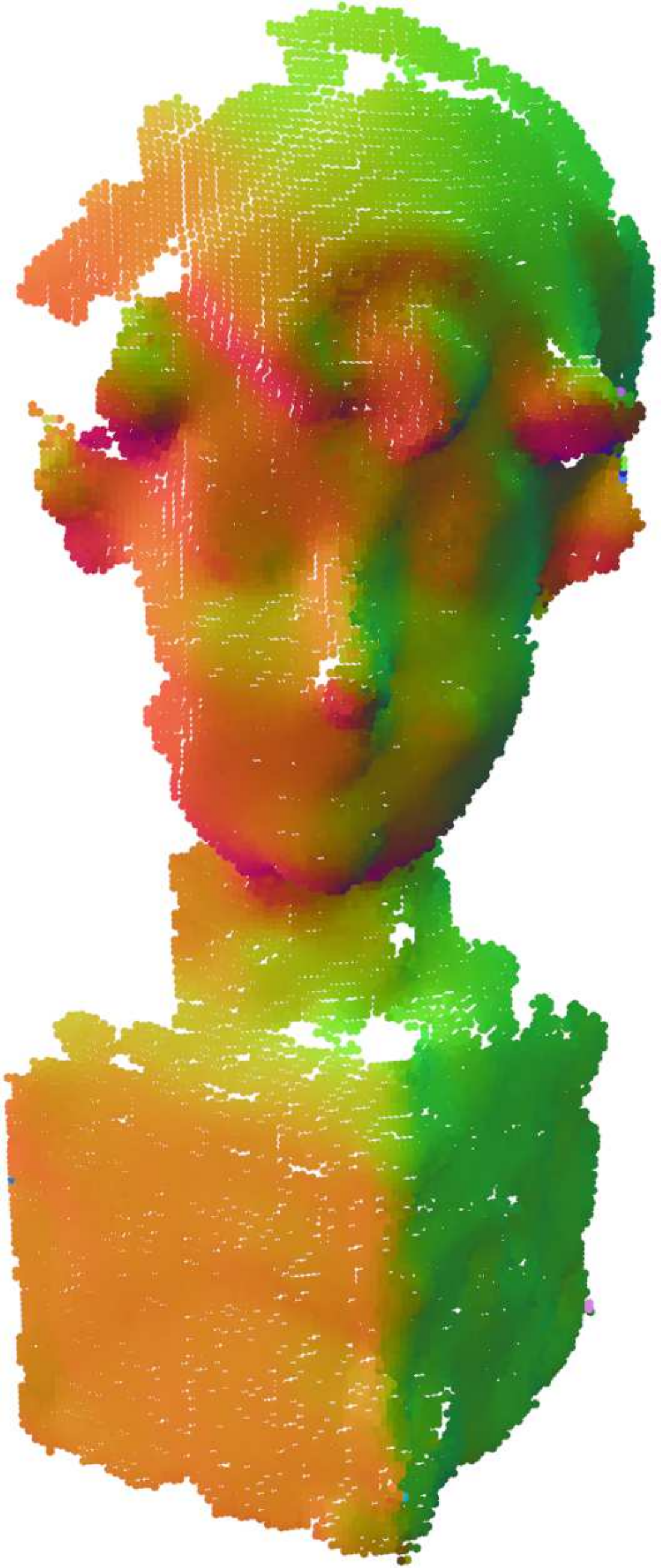}}%
	\subfigure[GT]{\includegraphics[width=\unitE]{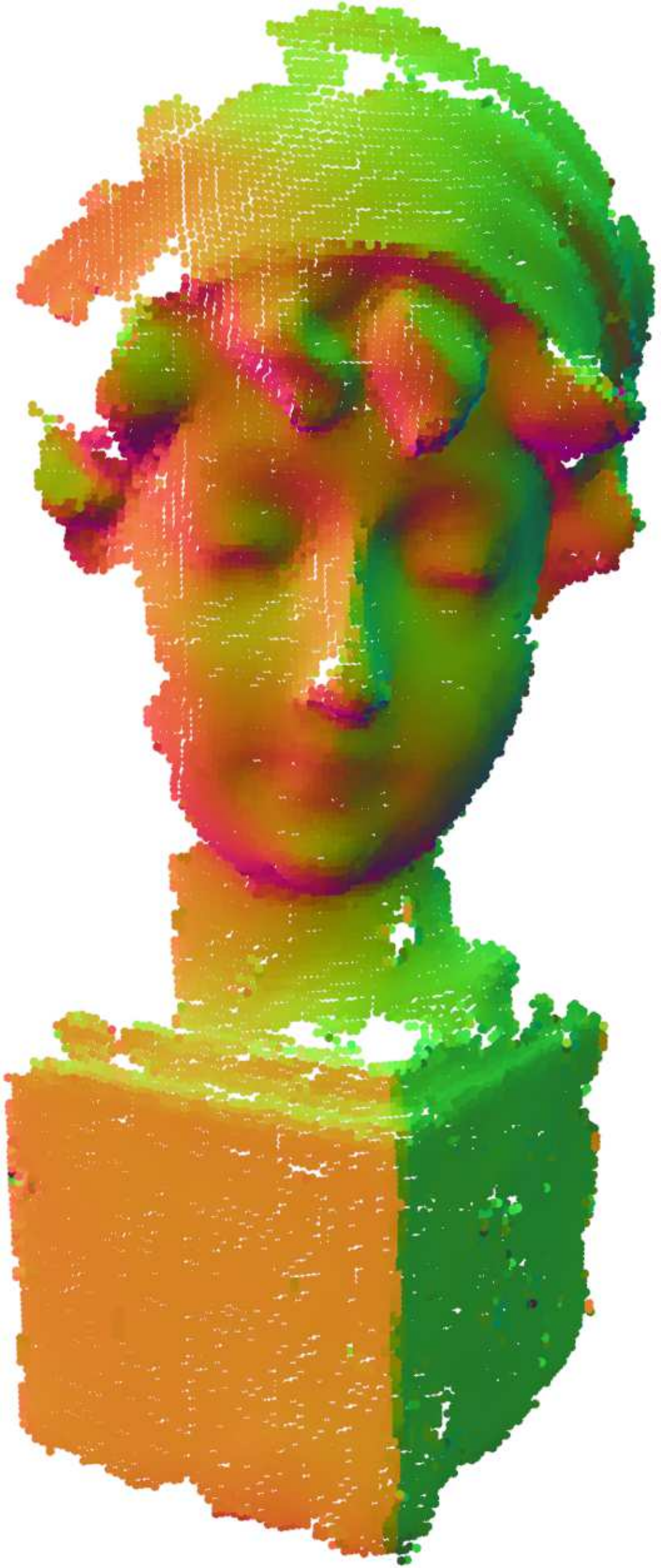}}
	
	\caption{Visual comparison of estimated normals on two real-scanned models. The average normal angular errors (mean) are: (1-st row) 8.3, 10.2, 8.7, 8.7, 7.9, 7.8, \textbf{7.0}; (2-nd row) 10.0, 11.9, 10.0, 10.2, 10.2, 9.5, \textbf{8.6}. Our Refine-Net is better at recovering facial details of the two models and is able to remove noise on flat areas.}
	\label{fig:real_boy}
\end{figure*}

\vspace{5pt}
\noindent\textbf{Using other initial normals.} On the other hand, Refine-Net is able to refine the results of other deep networks. We show in Tab.~\ref{table:synthetic_result} and Tab.~\ref{table:synthetic_PGP} that normals predicted by PCPNet \cite{GuerreroEtAl:PCPNet:EG:2018} and Nesti-Net \cite{Ben-ShabatLF19} can be improved significantly by replacing the initial normal in our system. 
In this experiment, we train the networks separately, and then Refine-Net takes the Nesti-Net/PCPNet outputs as initial normals and generates the final predicted normals. It is also practical to design an end-to-end network where, however, the filtering for multi-scale normals in the refinement should be abandoned since it requires complete normal results of the point cloud. We prefer to keeping this multi-scale design in the network, which considers different normal directions in the neighborhood and performs better on sharp features. Moreover, Fig.~\ref{fig:syn_refine} depicts a visual comparison by using Nesti-Net/PCPNet normal results as the initial normals in our framework. Clear improvements are seen from regions of sharp features and details in the bounding box.

\begin{figure*}
	\centering
	
	\subfigure[Input]{\label{fig:outdoor:input}\includegraphics[width=0.245\linewidth]{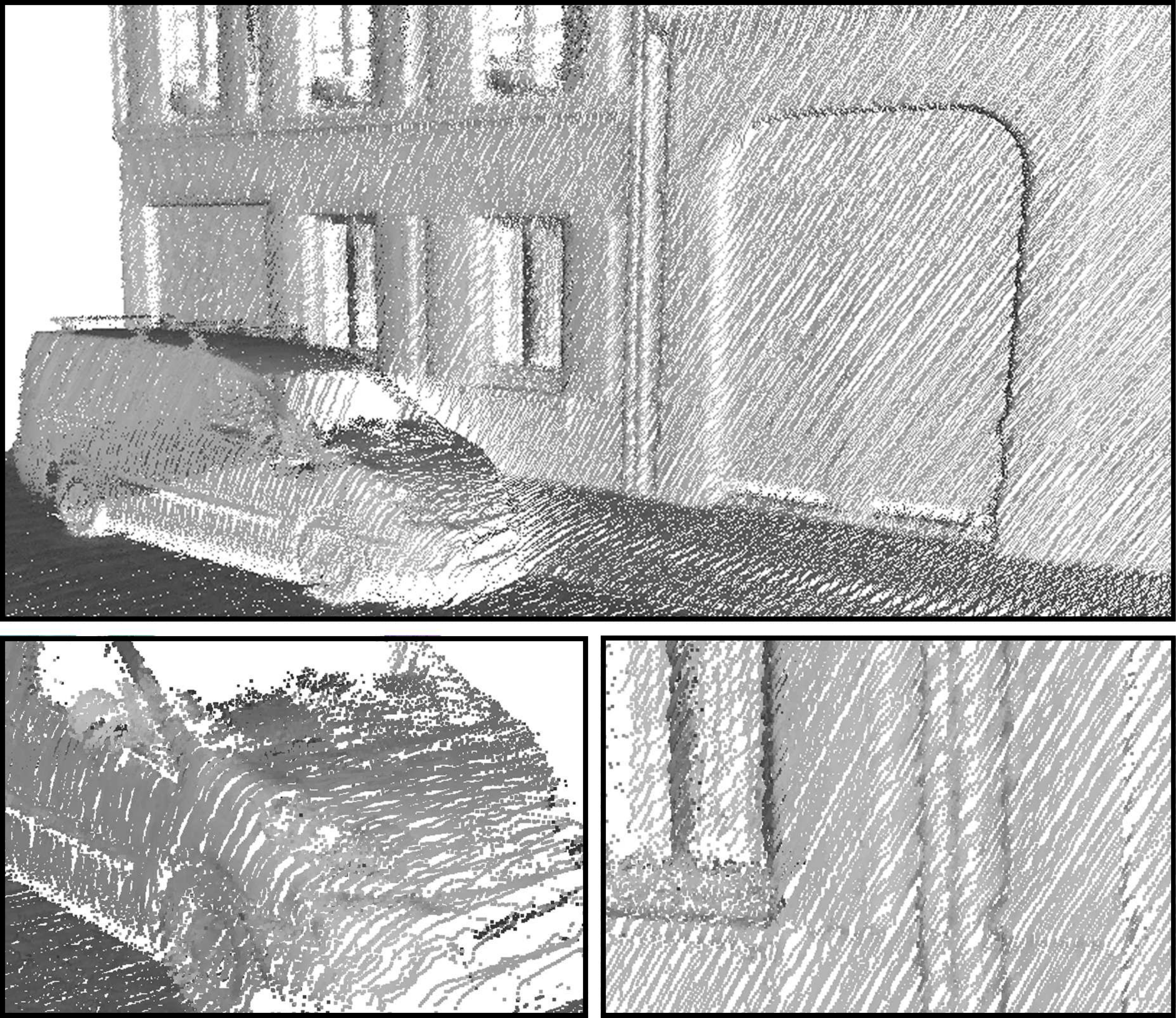}}%
	\hspace{3pt}\subfigure[PCPNet \cite{GuerreroEtAl:PCPNet:EG:2018}]{\label{fig:outdoor:pcp}\includegraphics[width=0.245\linewidth]{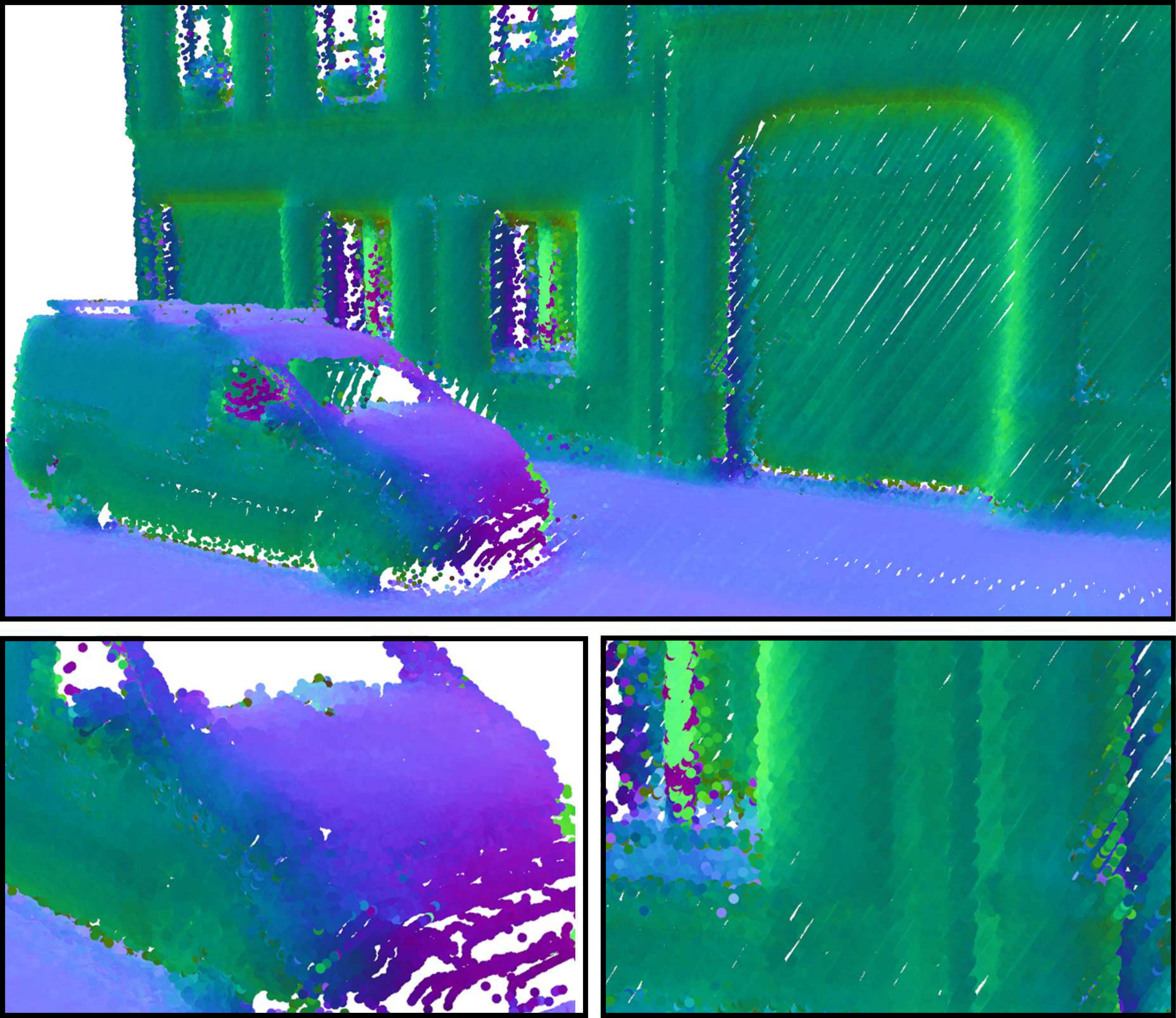}}%
	\hspace{3pt}\subfigure[Nesti-Net \cite{Ben-ShabatLF19}]{\label{fig:outdoor:nesti}\includegraphics[width=0.245\linewidth]{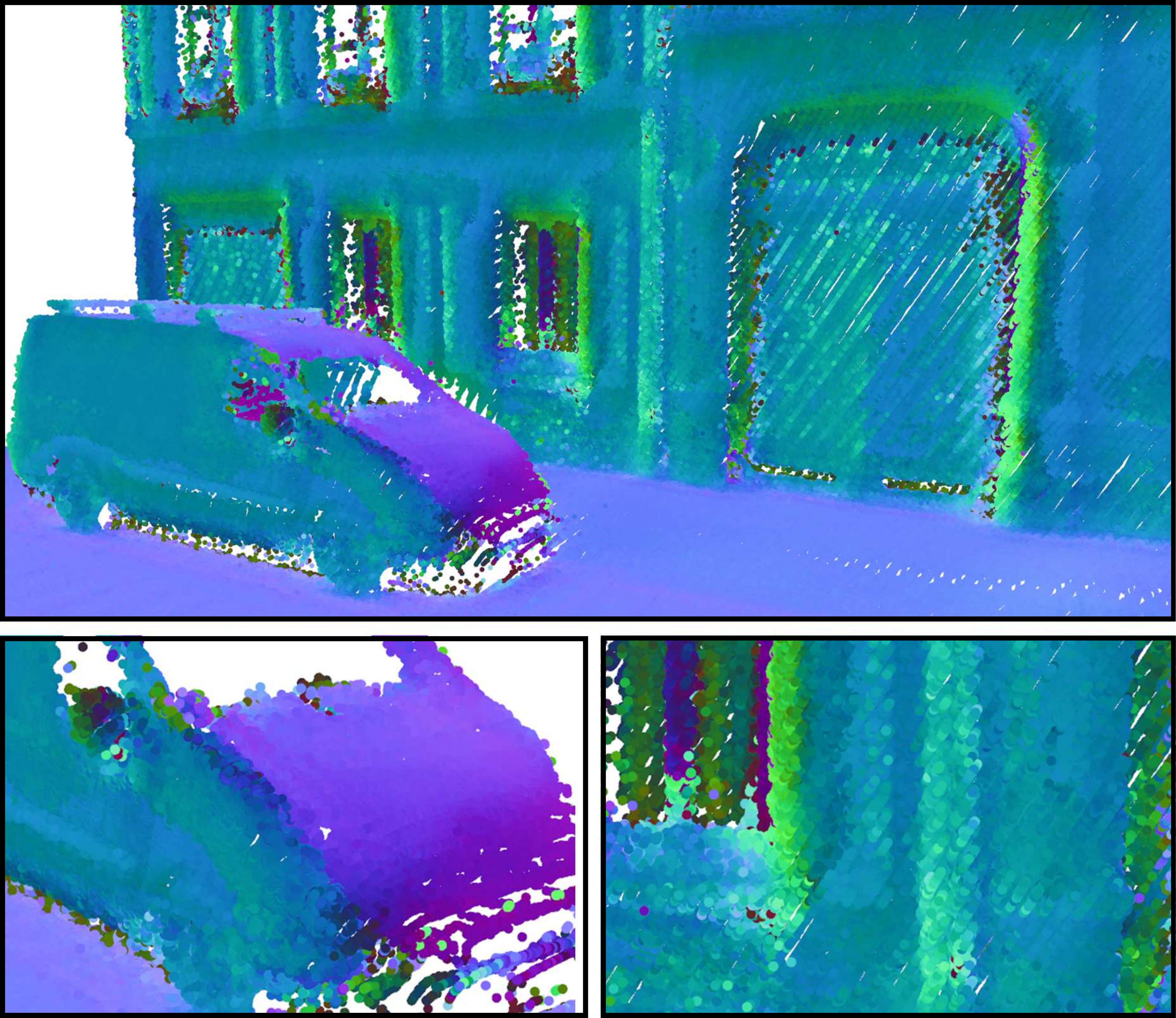}}%
	\hspace{3pt}\subfigure[Ours]{\label{fig:outdoor:ours}\includegraphics[width=0.245\linewidth]{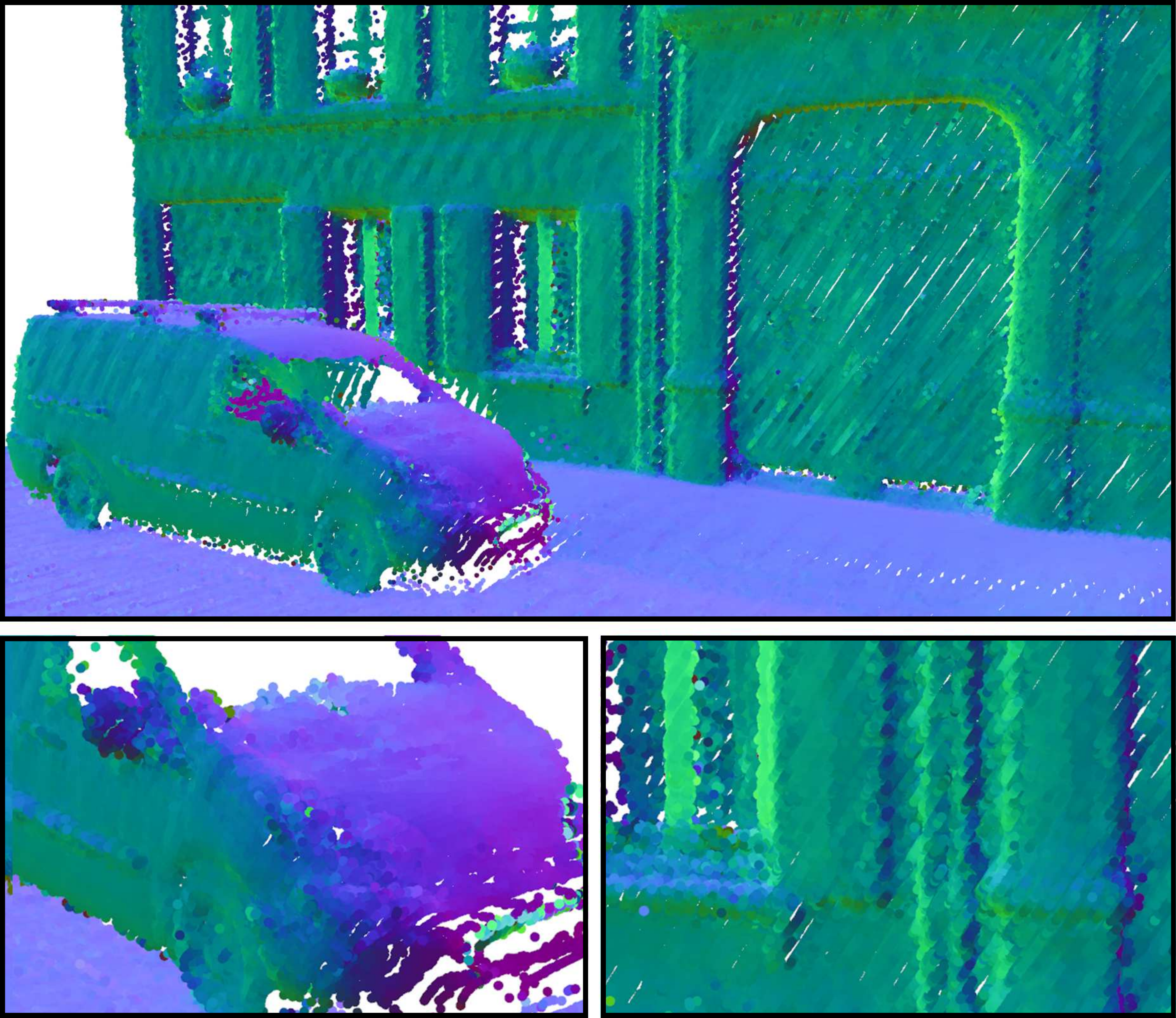}}
	
	\caption{Visual comparison of normal estimation results on the large-scale outdoor scenes from Paris-rue-Madame \cite{serna:hal-00963812} dataset. Our method produces more faithful results on the vehicle. Details on the wall are clearly recovered.}
	\label{fig:outdoor}
\end{figure*}

\subsection{Experiments on real scans}
\label{sec:eval:real}
\textbf{Dataset.} We extend our Refine-Net to recover truth surface normals on scanned point clouds. 
In this task, the real-scanned data reveals more challenges, such as the fluctuation on flat surface, originated from the projection process of 3D sensors. The non-Gaussian and discontinuous noise heavily interferes both traditional and learning-based estimators against recovering the underlying surface.

The training set comes from \cite{wang2016mesh}. This dataset contains scanned meshes of seven real models using Microsoft Kinect v1. There are 71 scans in the training set and the total number of samples is about 2.6M. Further, for each scan, they use another high-resolution scanned surface to help building ground truth normals. The benchmark test set of \cite{wang2016mesh} contains 73 scans and 930k samples with ground truth normals.

Moreover, we test our trained network on NYUD-V2 \cite{silberman2012indoor} and Paris-rue-Madame \cite{serna:hal-00963812} datasets. NYUD-V2 captures a variety of indoor scenes recorded by both the RGB and depth cameras. We extract 3D point clouds from the depth channel as the noisy input. Missing pixels of raw depth maps are filled in using \cite{levin2004colorization}. For large-scale outdoor scenes, we employ the Paris-rue-Madame dataset \cite{serna:hal-00963812}.

\vspace{5pt}
\noindent\textbf{Architecture.} We use a similar Refine-Net architecture as in our synthetic experiments. The filtering parameters are the same, but the non-filtered normal branch is not included. Thus, $8$ branches are developed in the network. Since the real-scanned point cloud tends to be dense and large, in the point module, the patch size is set to be $0.03$, and the maximum number of points is extended to $500$. The neighborhood in HMP construction is also resized similarly and the height-map grid size is $m = 7$.

\vspace{5pt}
\noindent\textbf{Results.} Quantitative comparisons of normal estimation results from \cite{wang2016mesh} are shown in Tab.~\ref{table:Kinect_results}. Please note that we train all networks on the same dataset for a fair comparison. Our method outperforms the state-of-the-arts according to all the metrics. Visual comparisons on two real models are illustrated in Fig.~\ref{fig:real_boy}. Refine-Net produces more faithful results on, e.g., the eyes and nose of the boy and girl model respectively while removing undesired noise on the model bases.

\begin{figure*}
	\centering
	\newlength{\unitS}
	\setlength{\unitS}{0.142\linewidth}
	\includegraphics[width=\unitS]{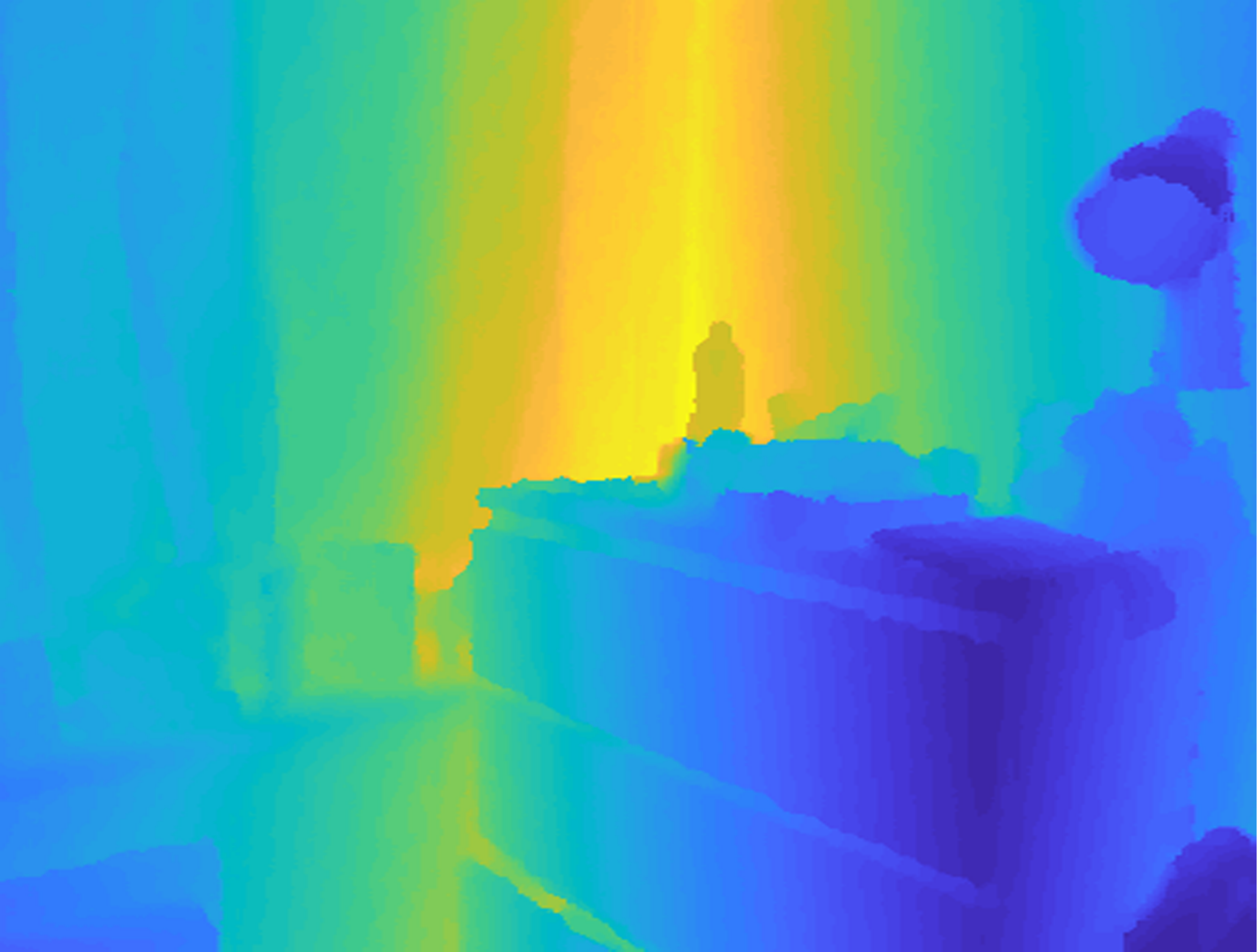}%
	\includegraphics[width=\unitS]{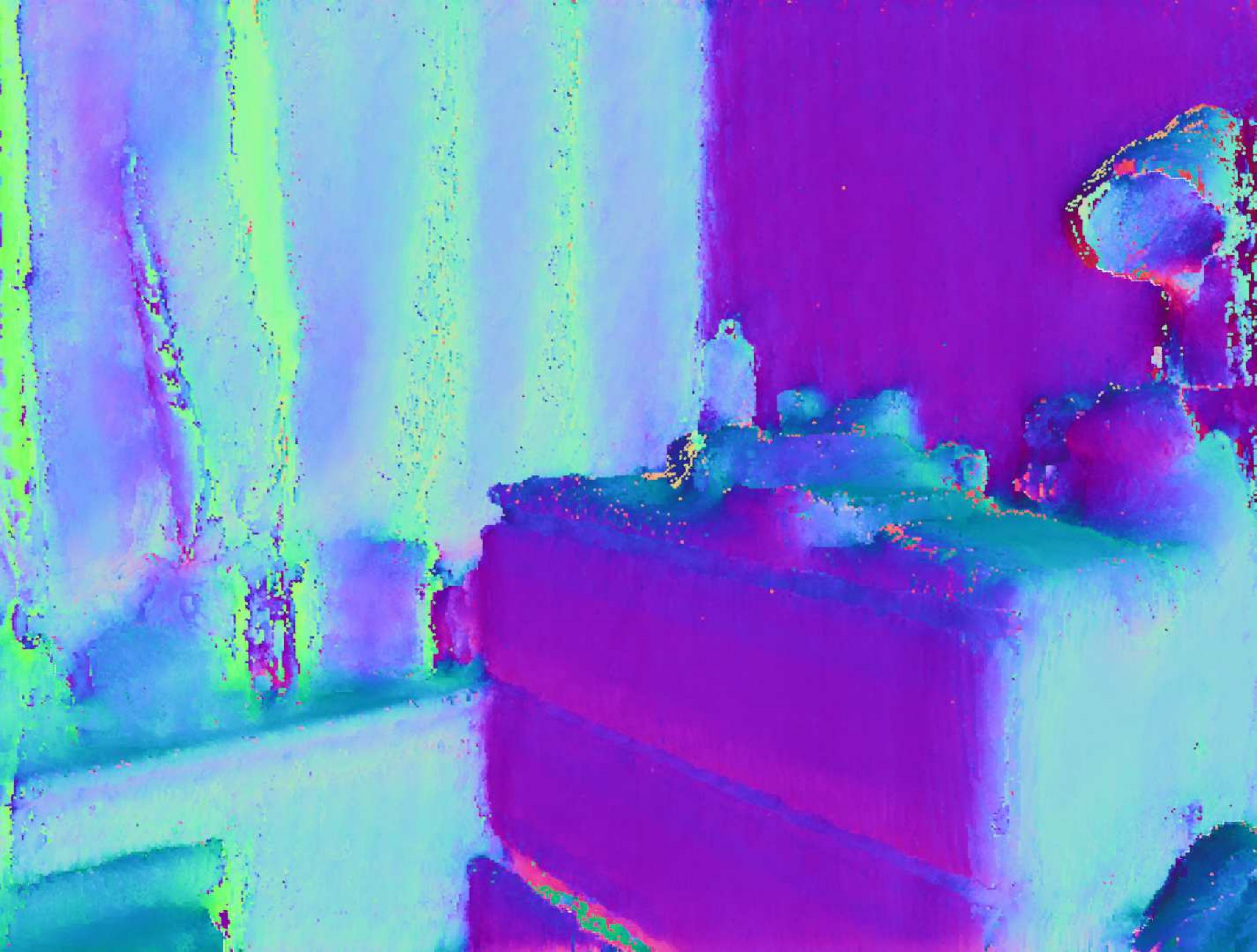}%
	\includegraphics[width=\unitS]{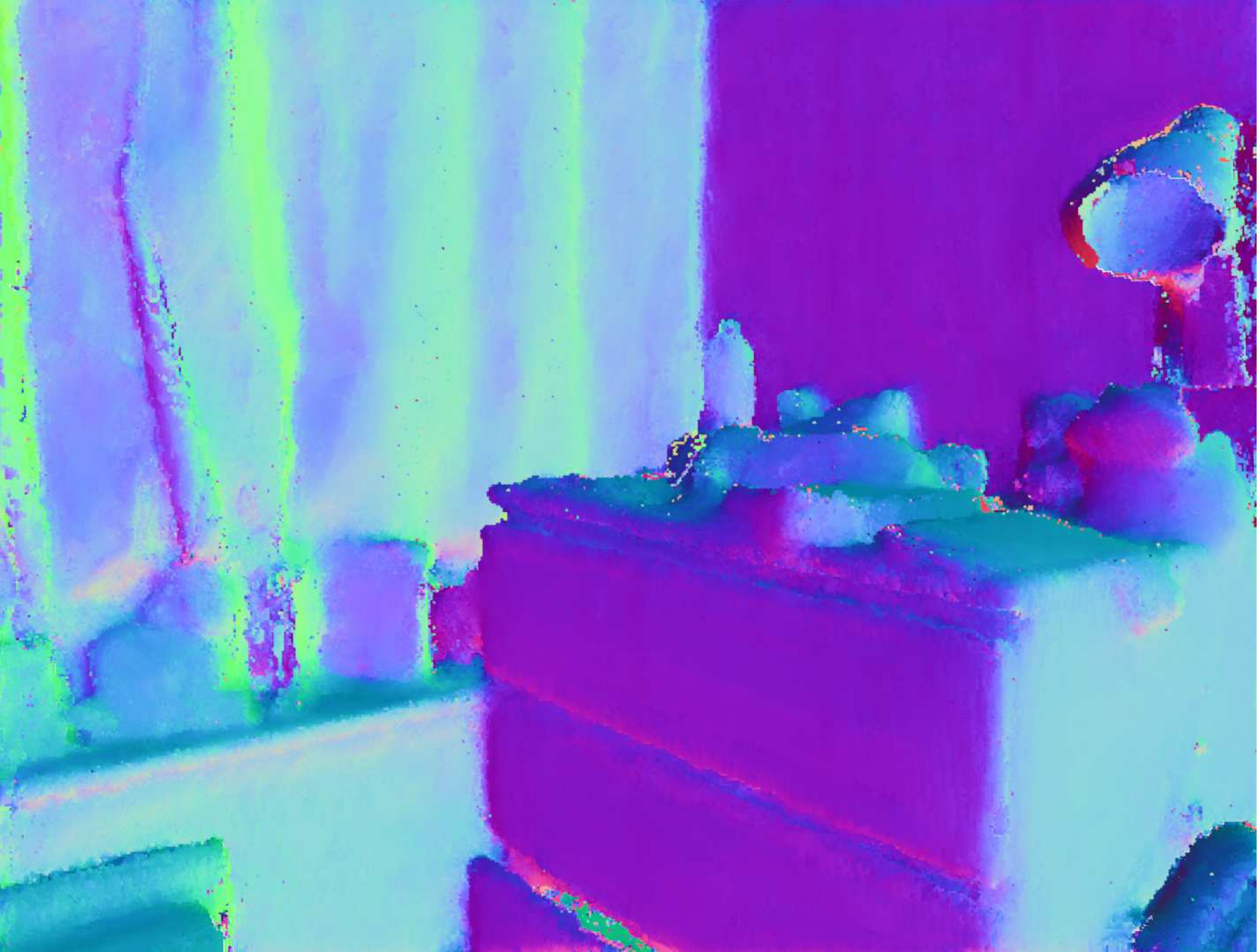}%
	\includegraphics[width=\unitS]{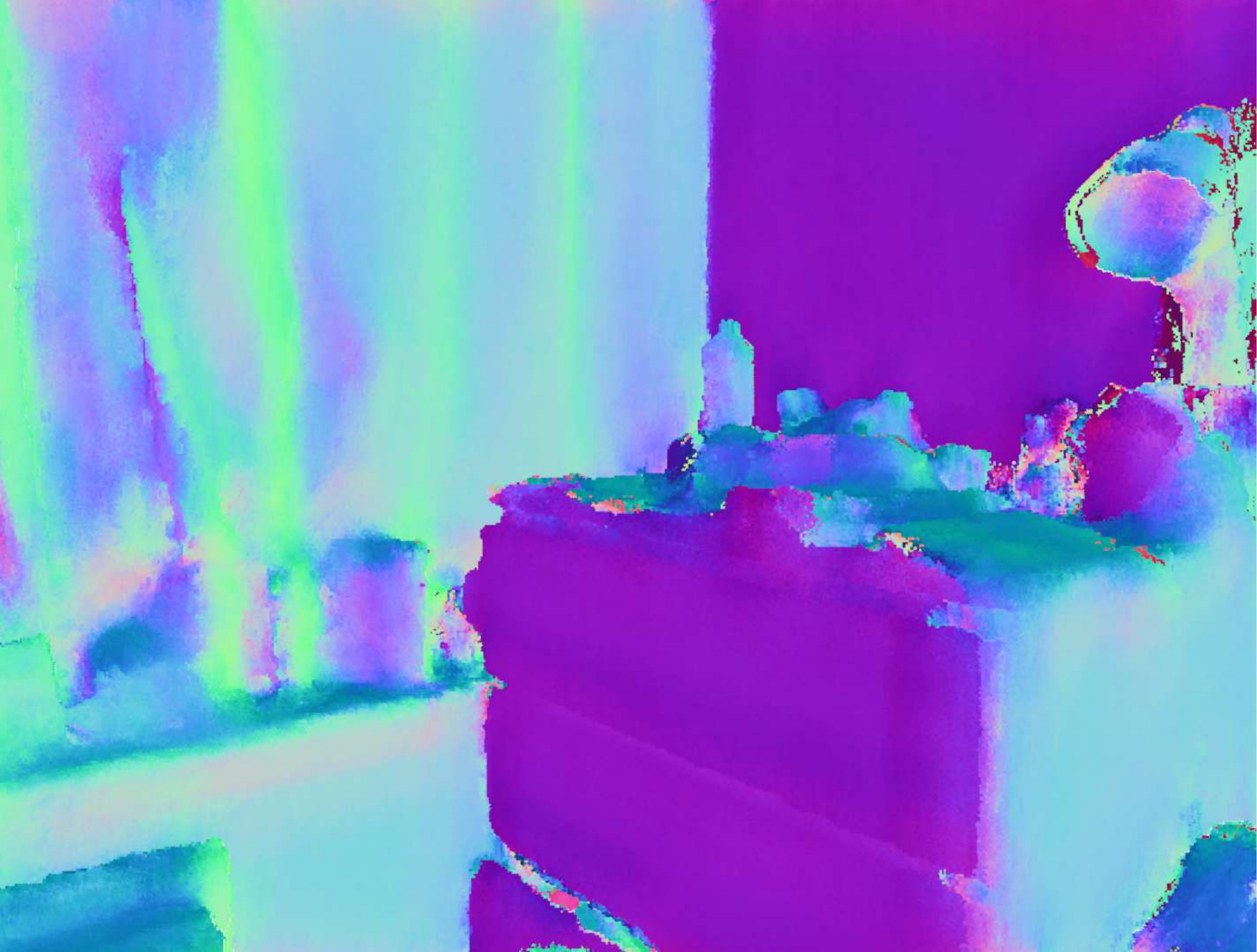}%
	\includegraphics[width=\unitS]{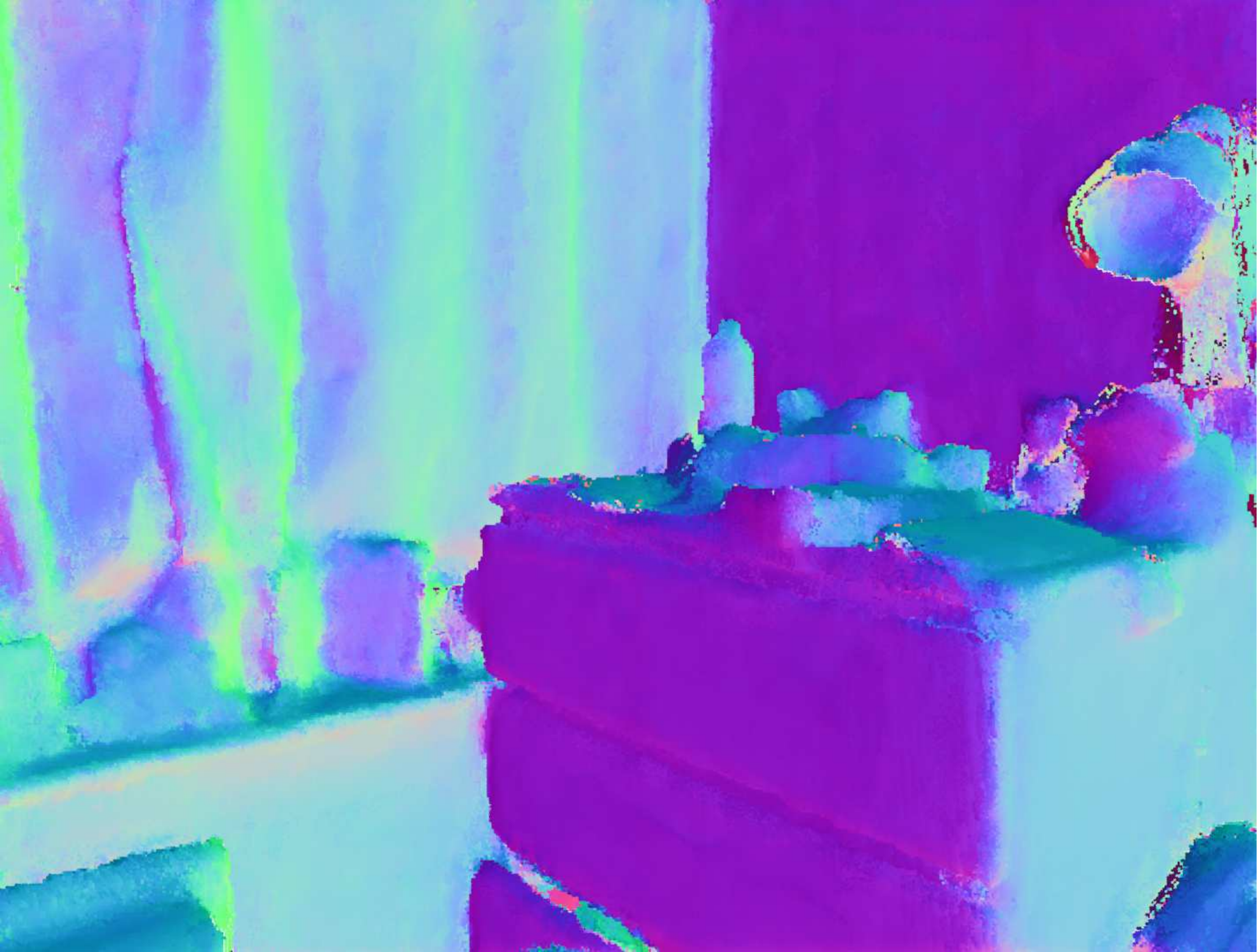}%
	\includegraphics[width=\unitS]{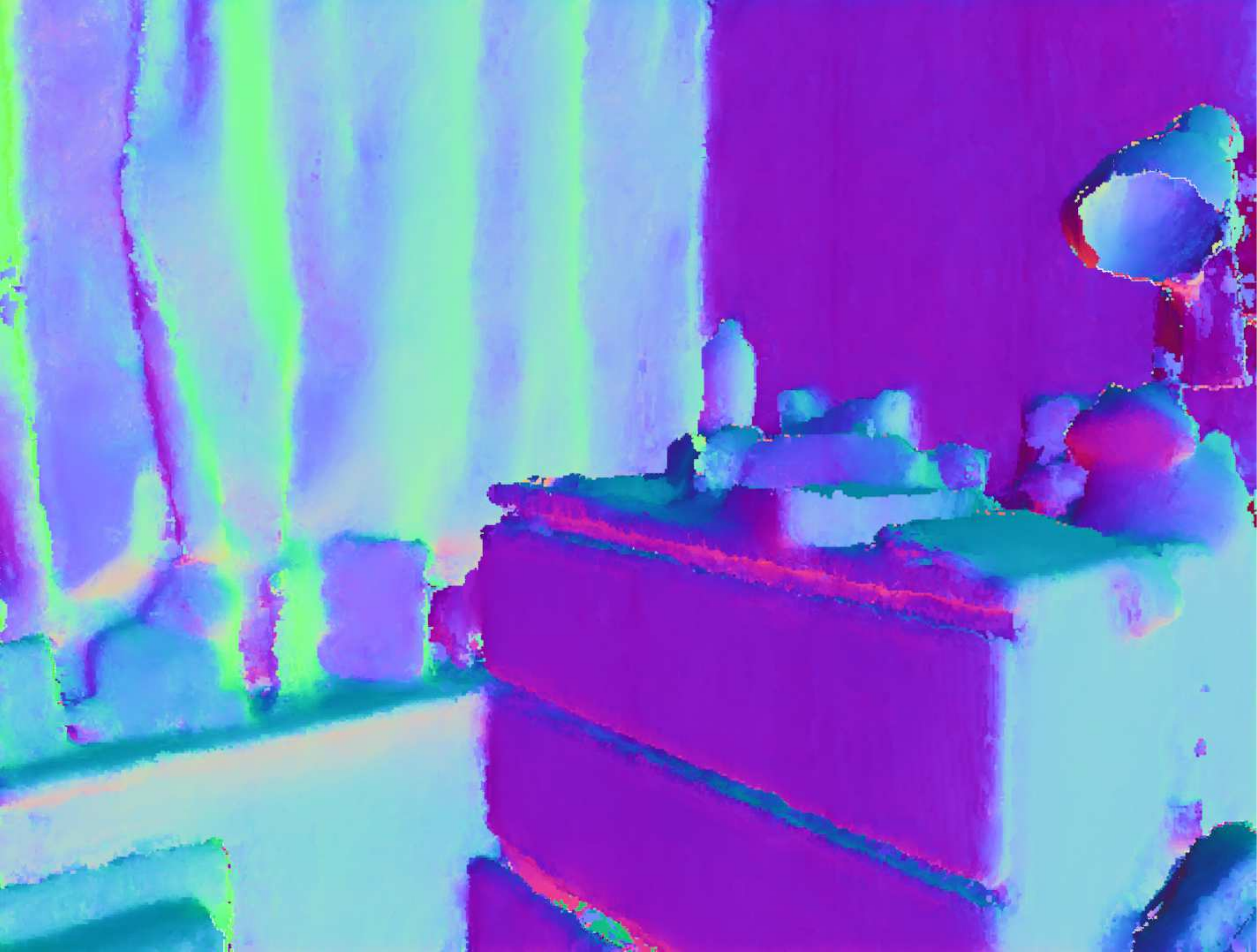}%
	\includegraphics[width=\unitS]{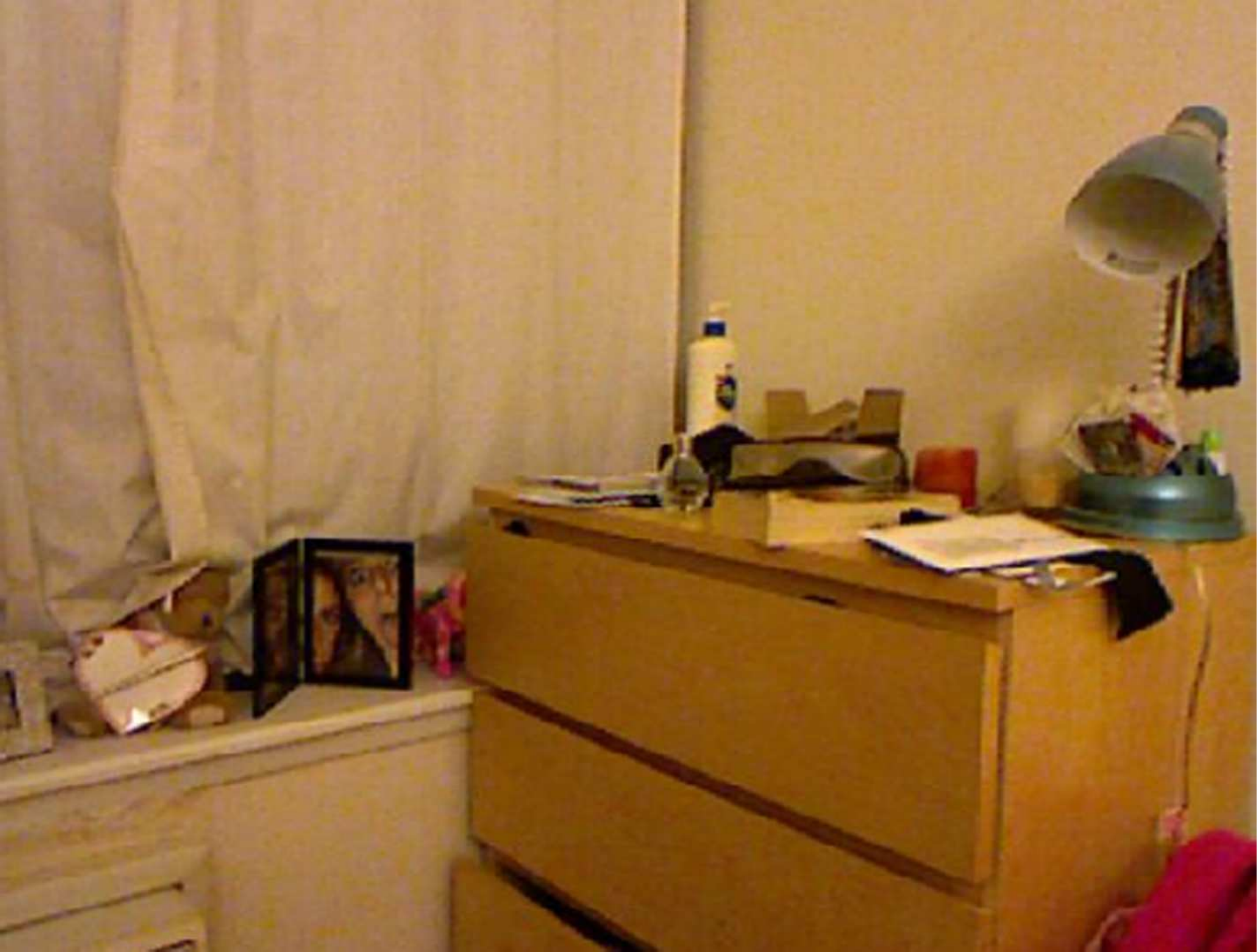}
	
	\vspace{3pt}
	
	\includegraphics[width=\unitS]{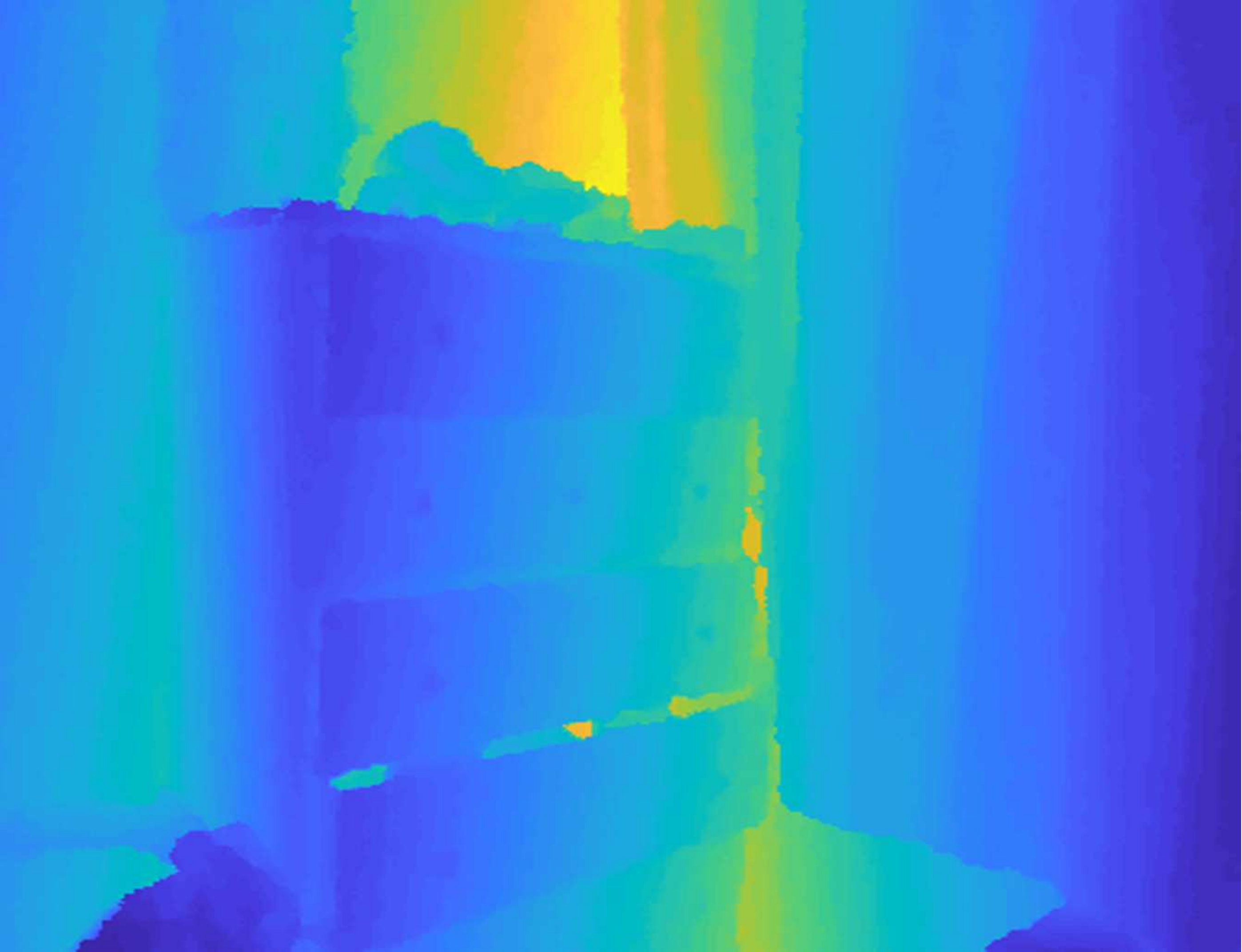}%
	\includegraphics[width=\unitS]{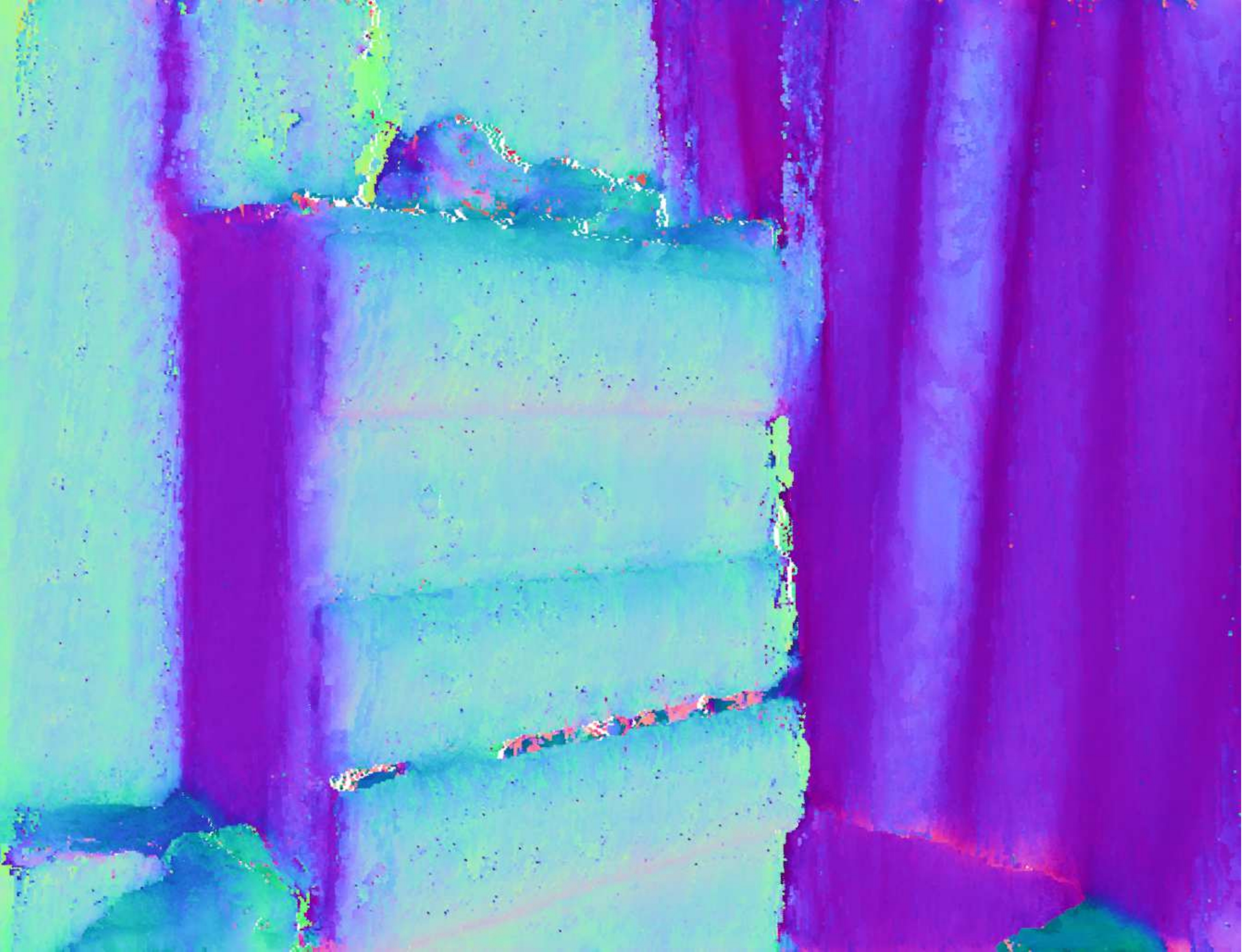}%
	\includegraphics[width=\unitS]{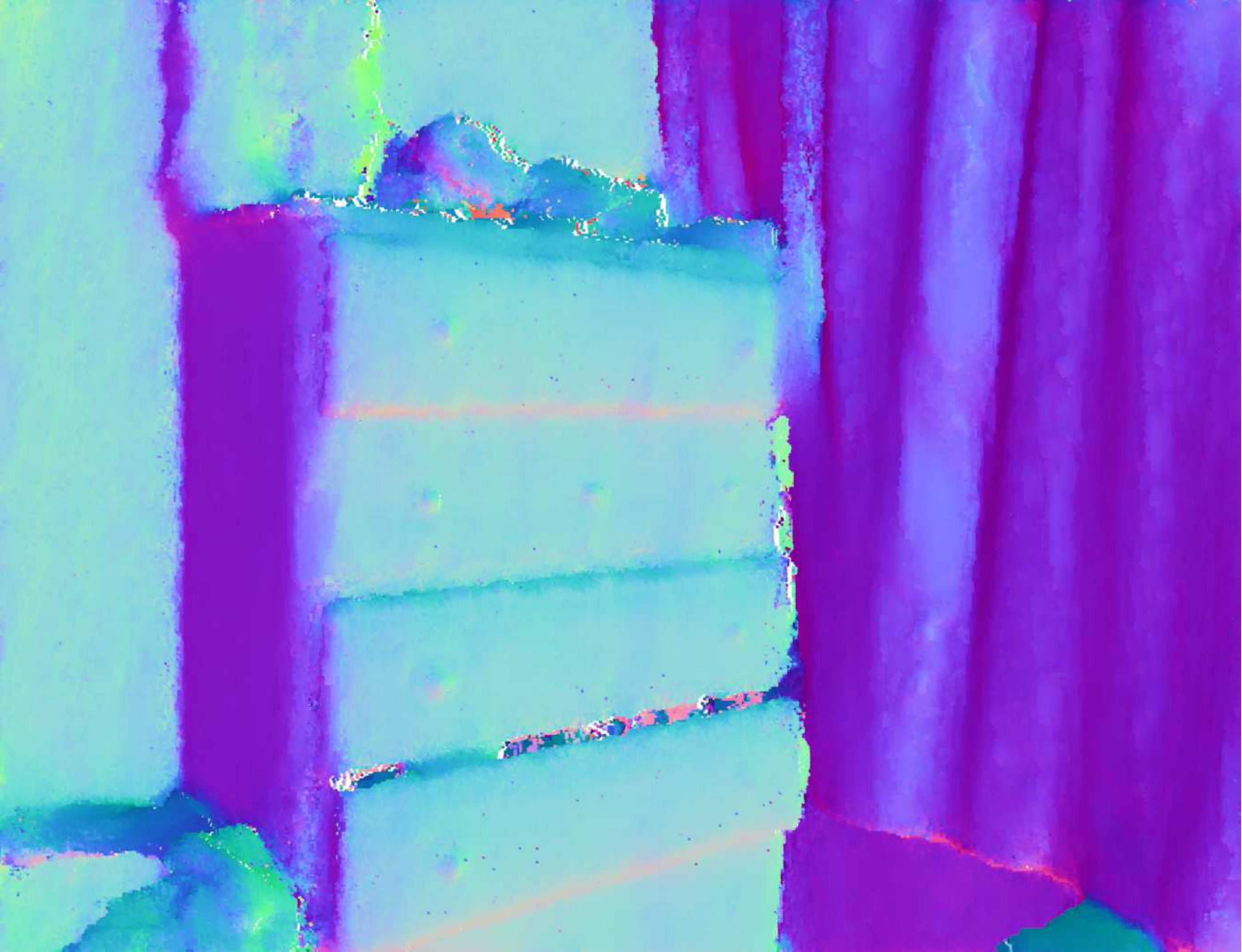}%
	\includegraphics[width=\unitS]{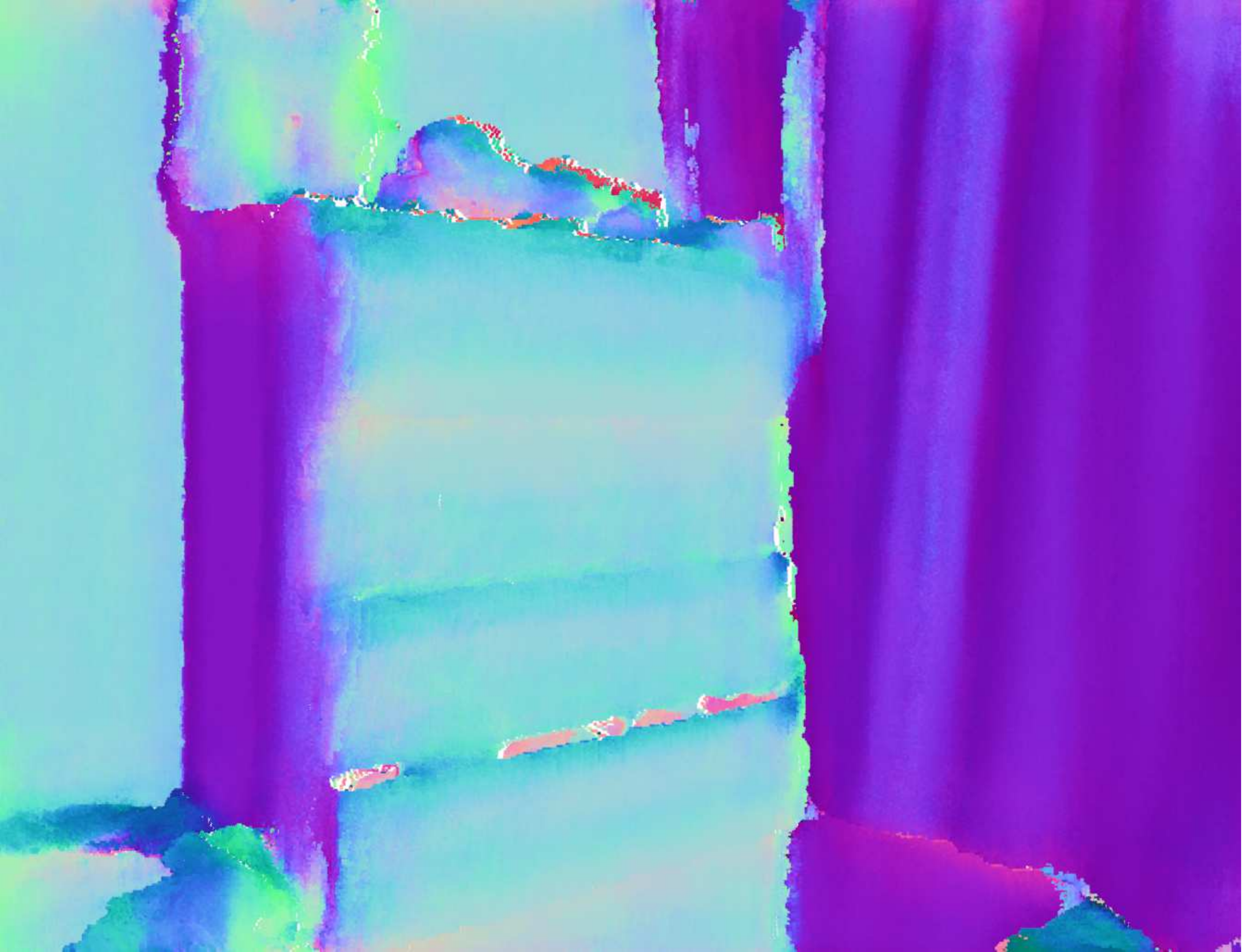}%
	\includegraphics[width=\unitS]{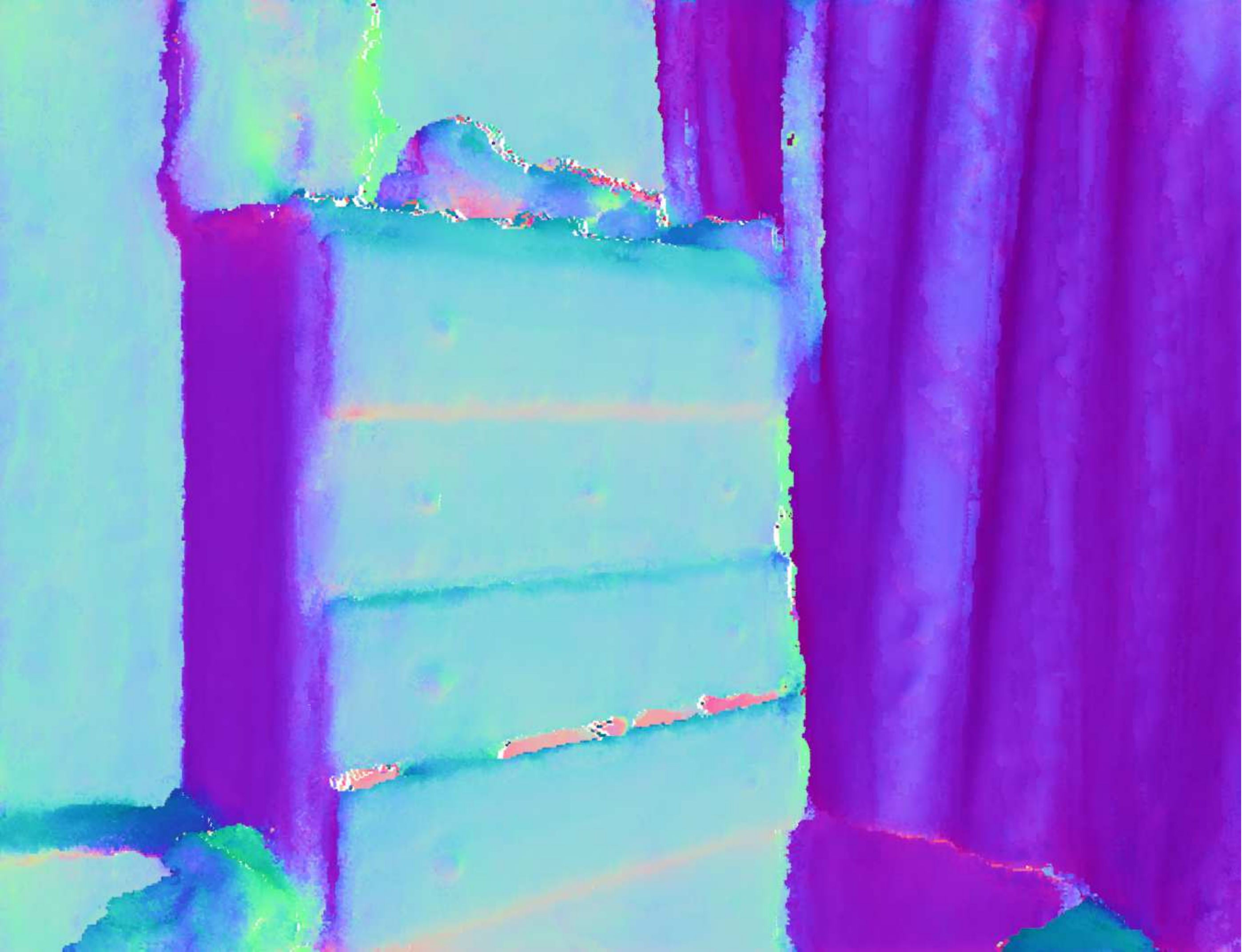}%
	\includegraphics[width=\unitS]{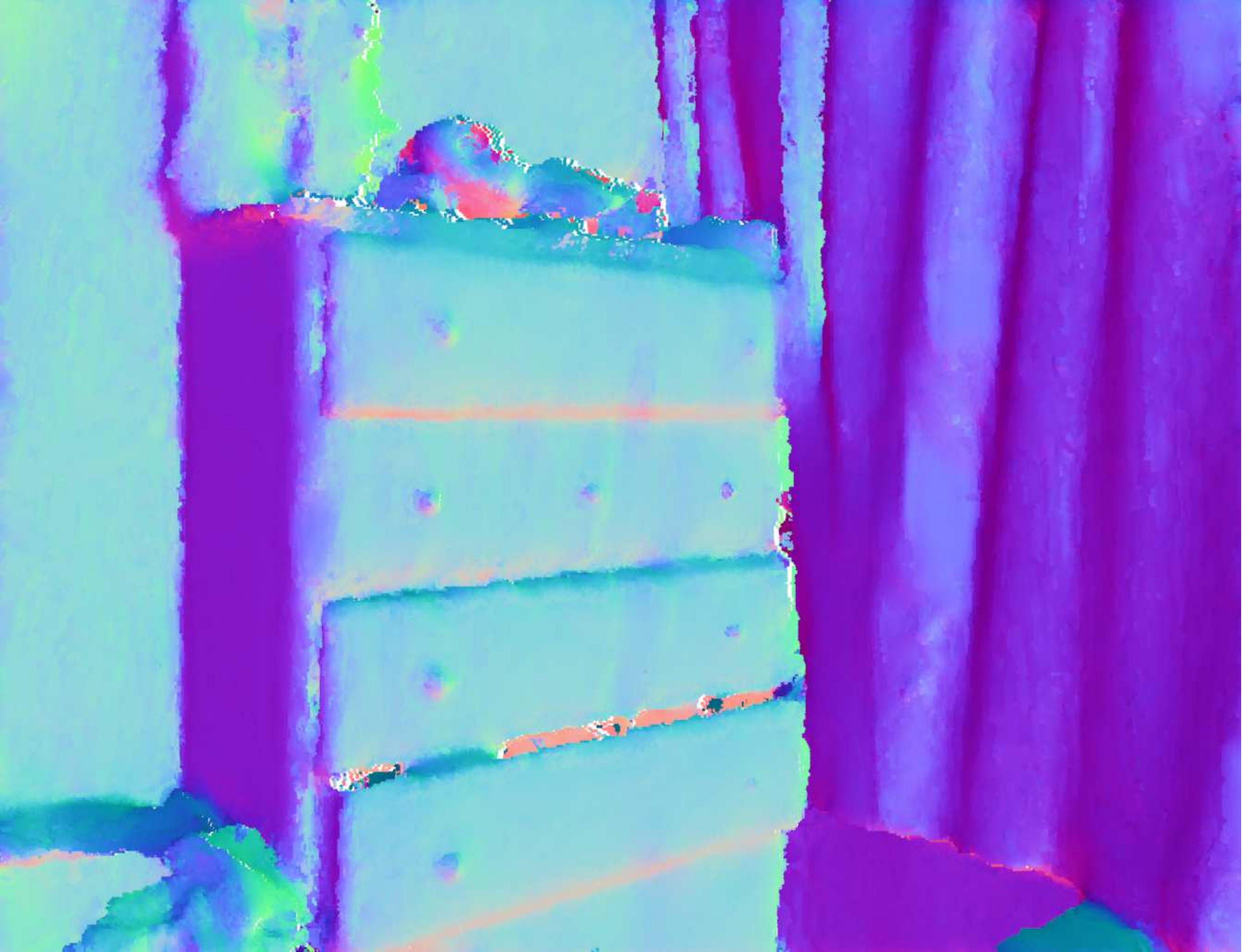}%
	\includegraphics[width=\unitS]{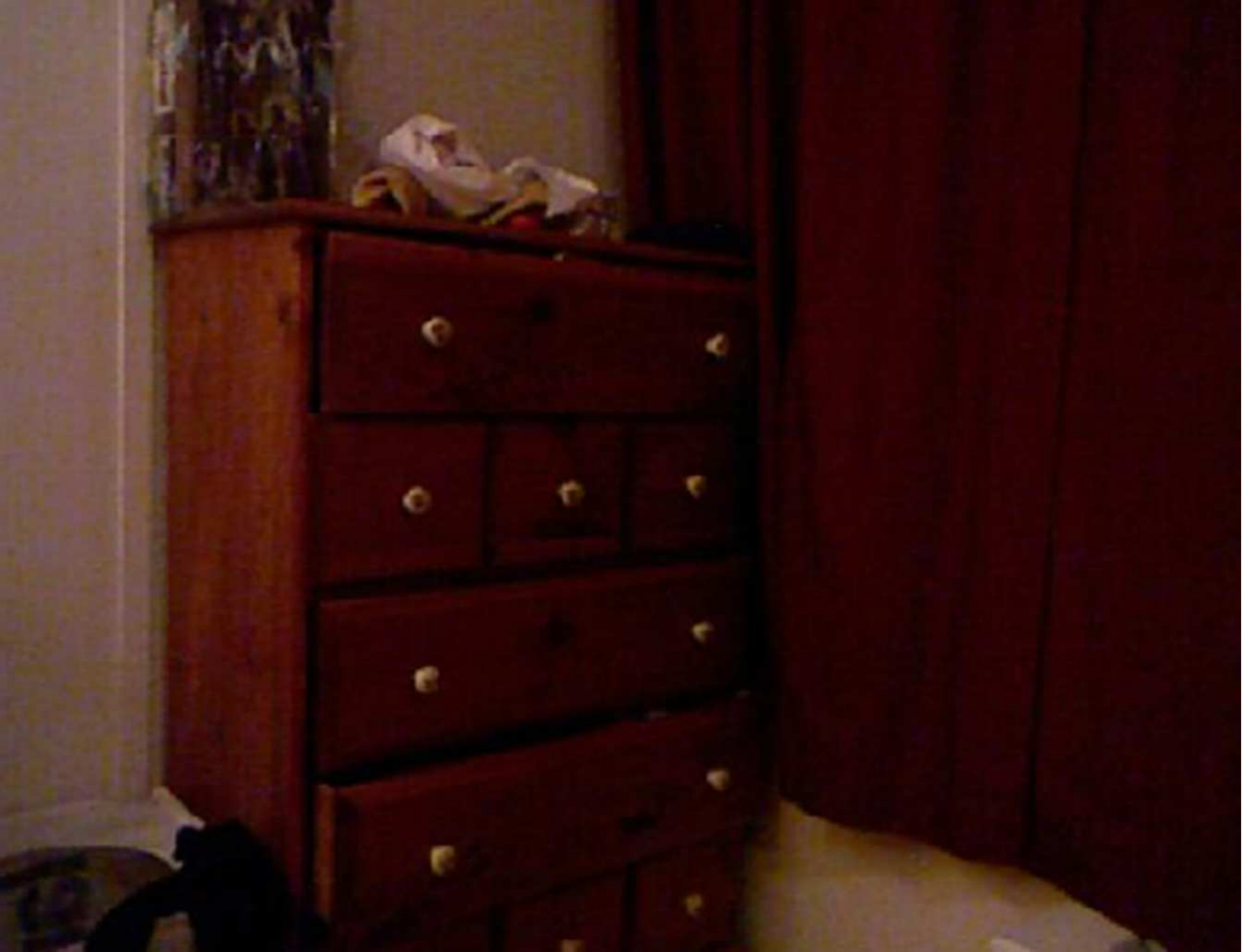}
	
	\vspace{3pt}
	
	\includegraphics[width=\unitS]{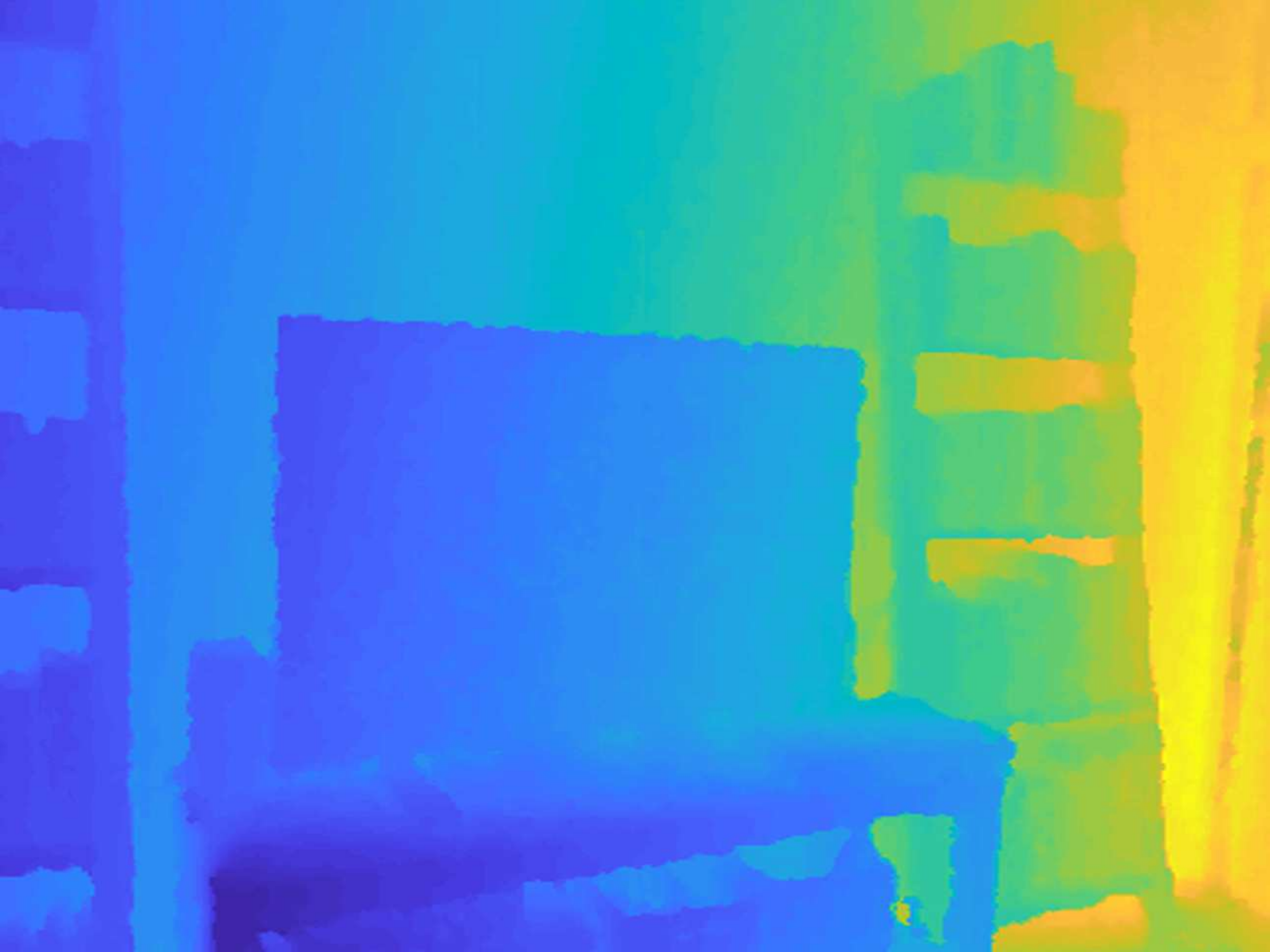}%
	\includegraphics[width=\unitS]{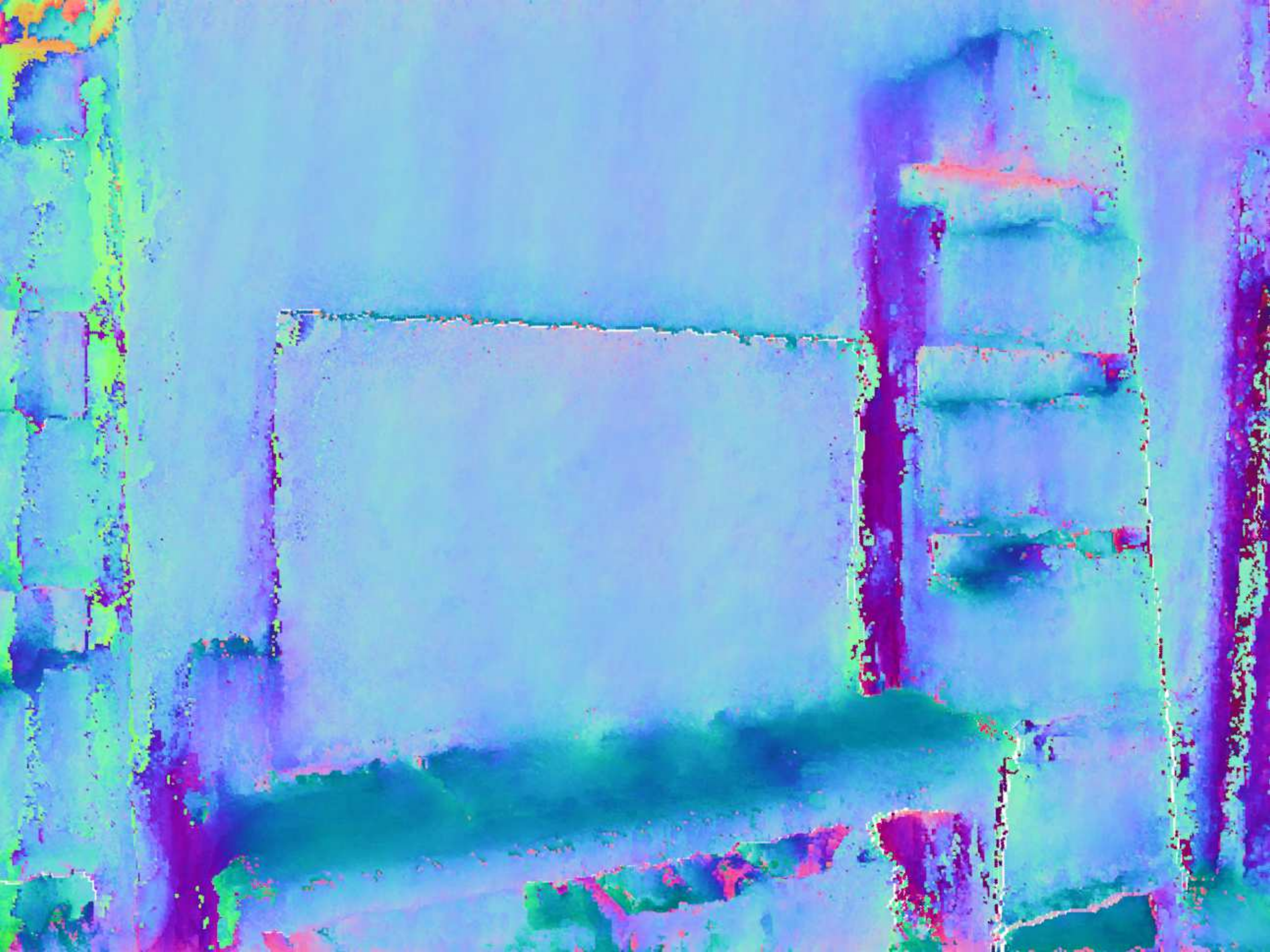}%
	\includegraphics[width=\unitS]{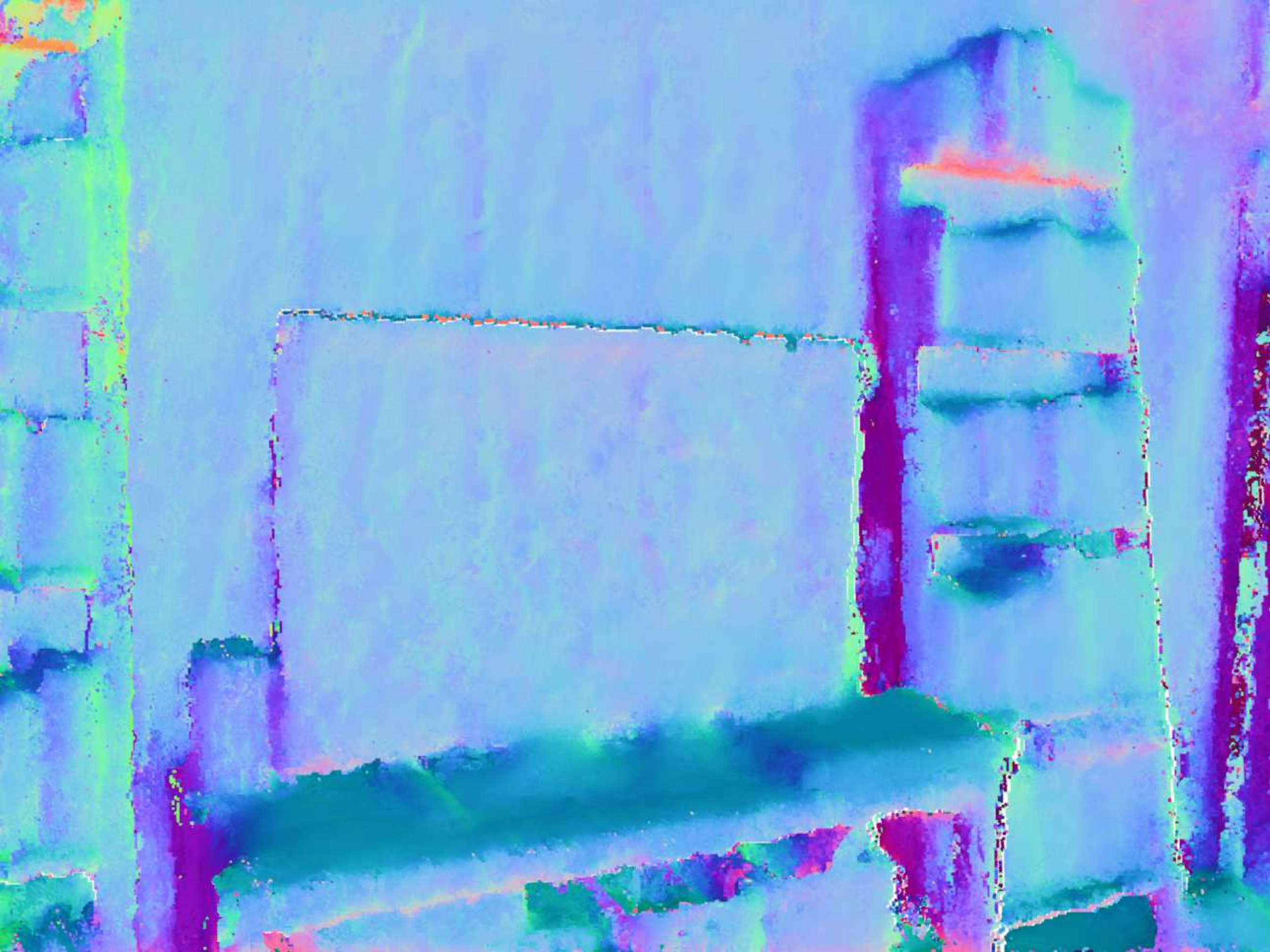}%
	\includegraphics[width=\unitS]{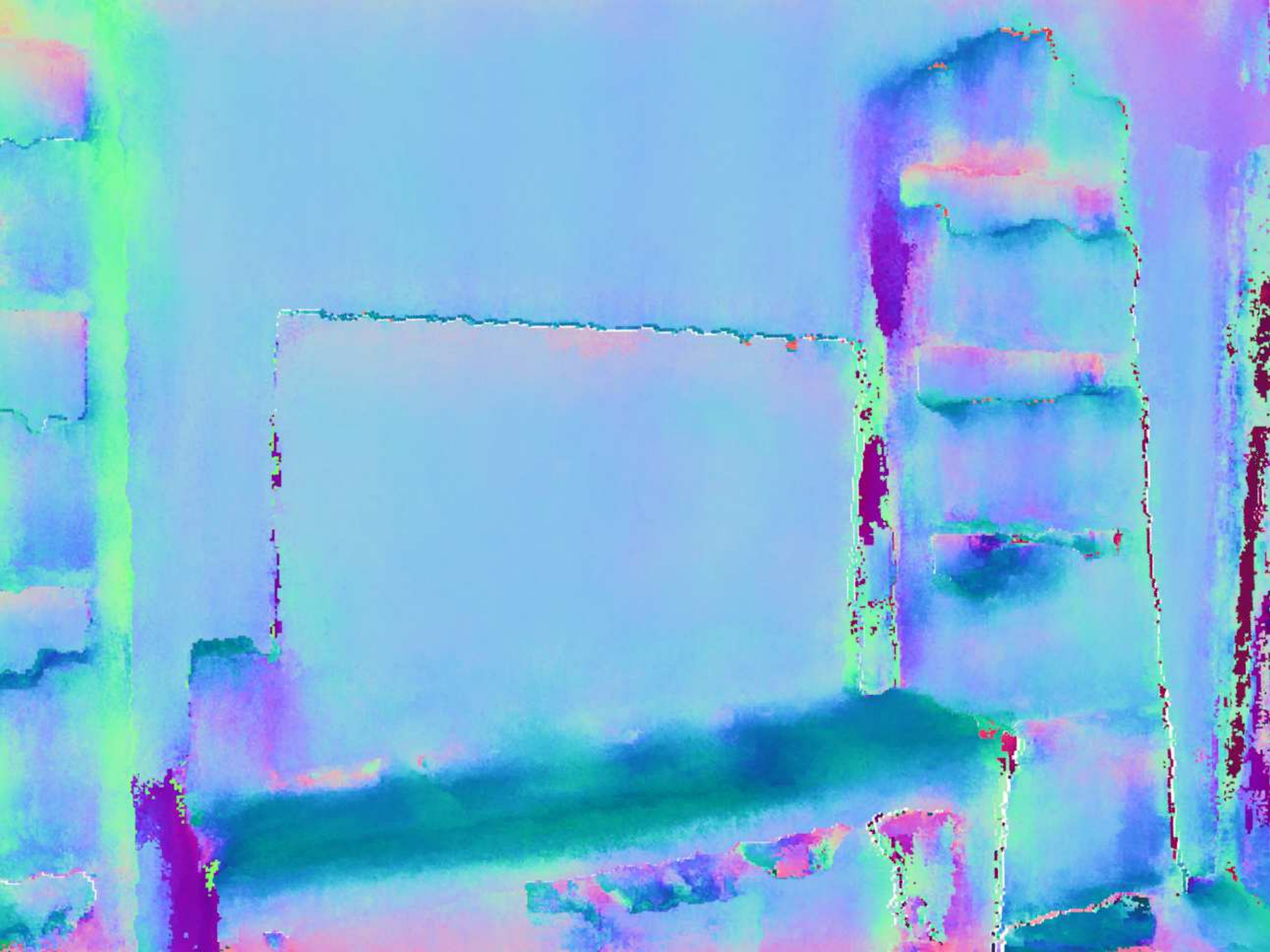}%
	\includegraphics[width=\unitS]{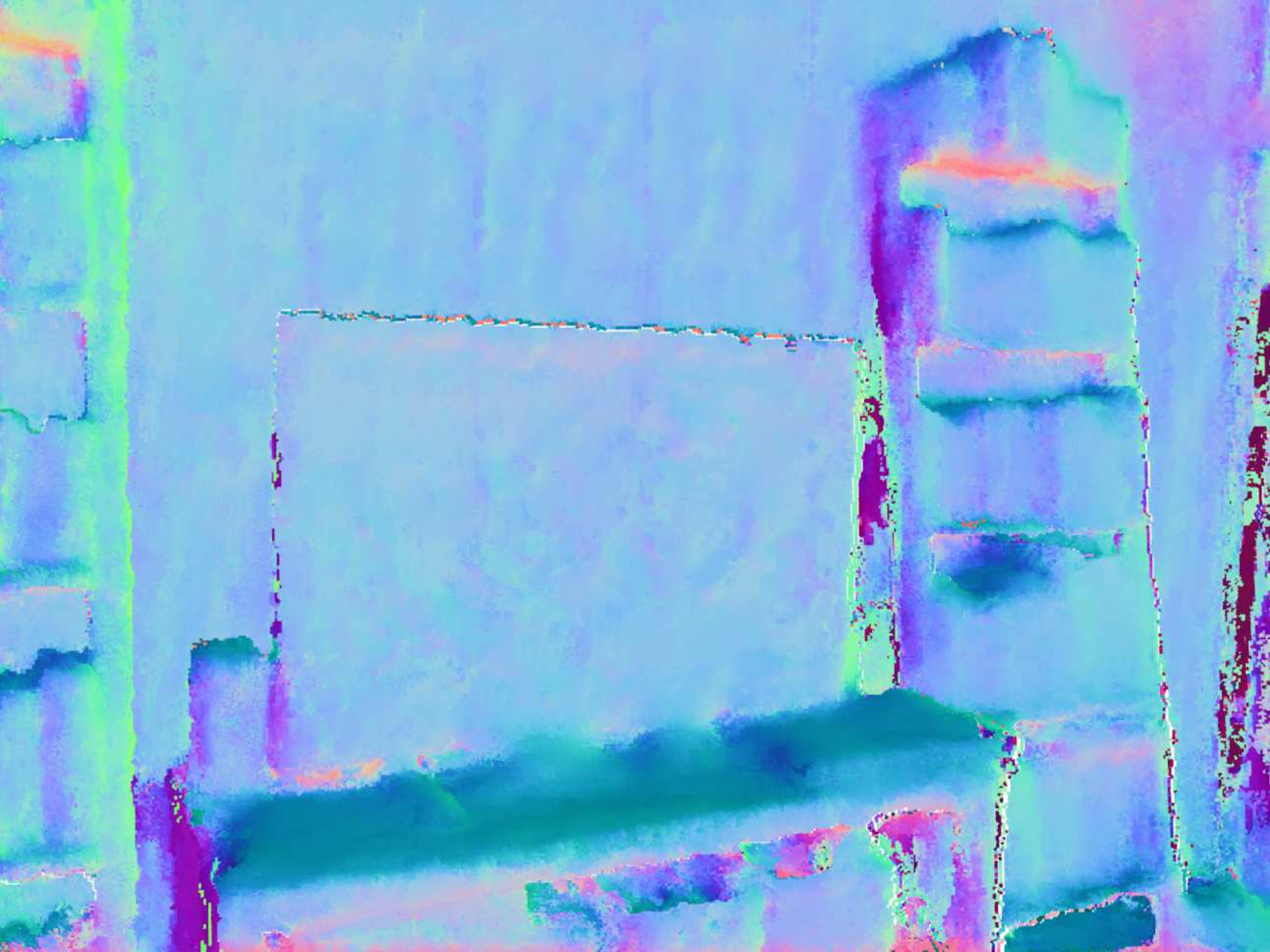}%
	\includegraphics[width=\unitS]{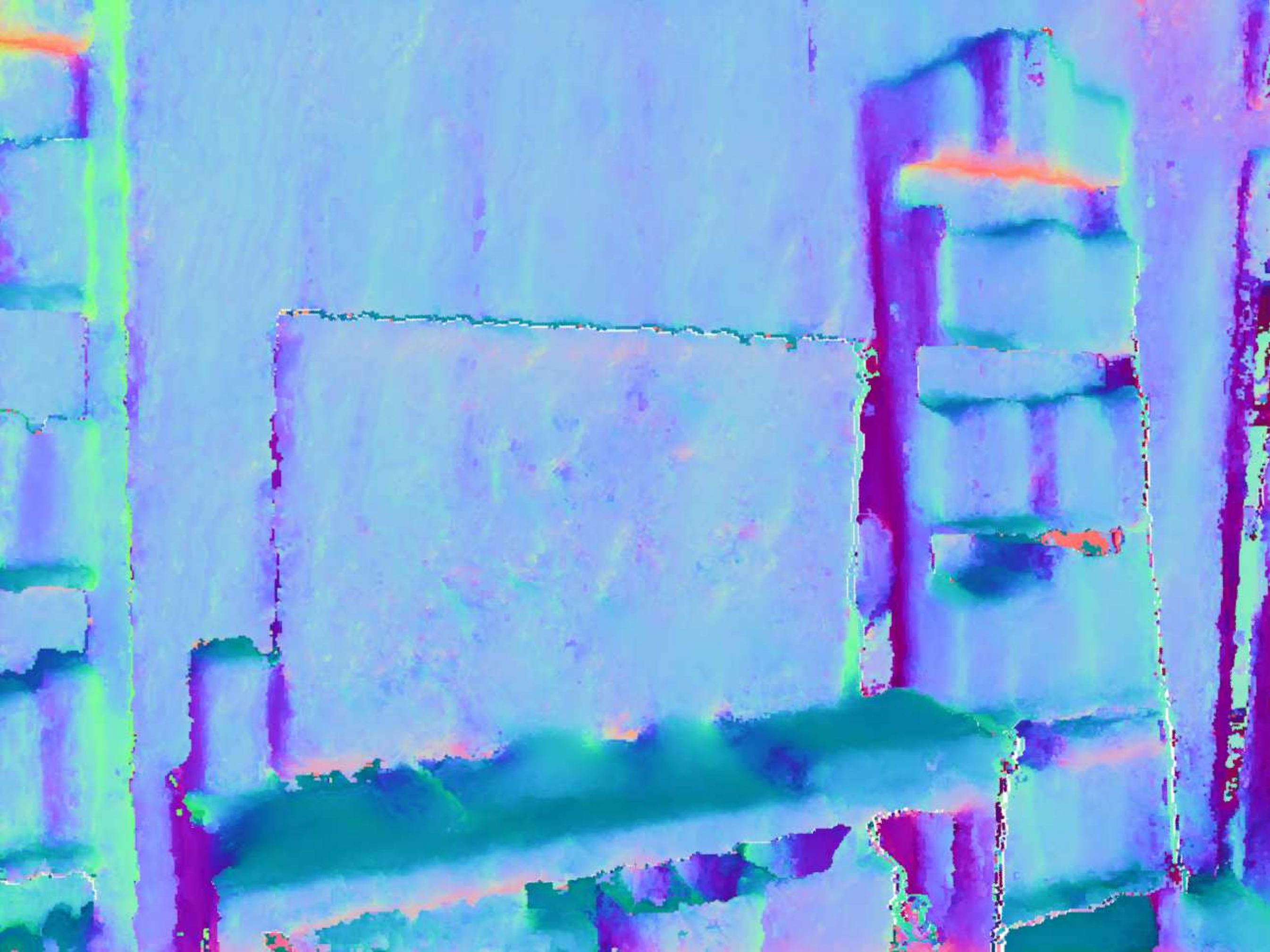}%
	\includegraphics[width=\unitS]{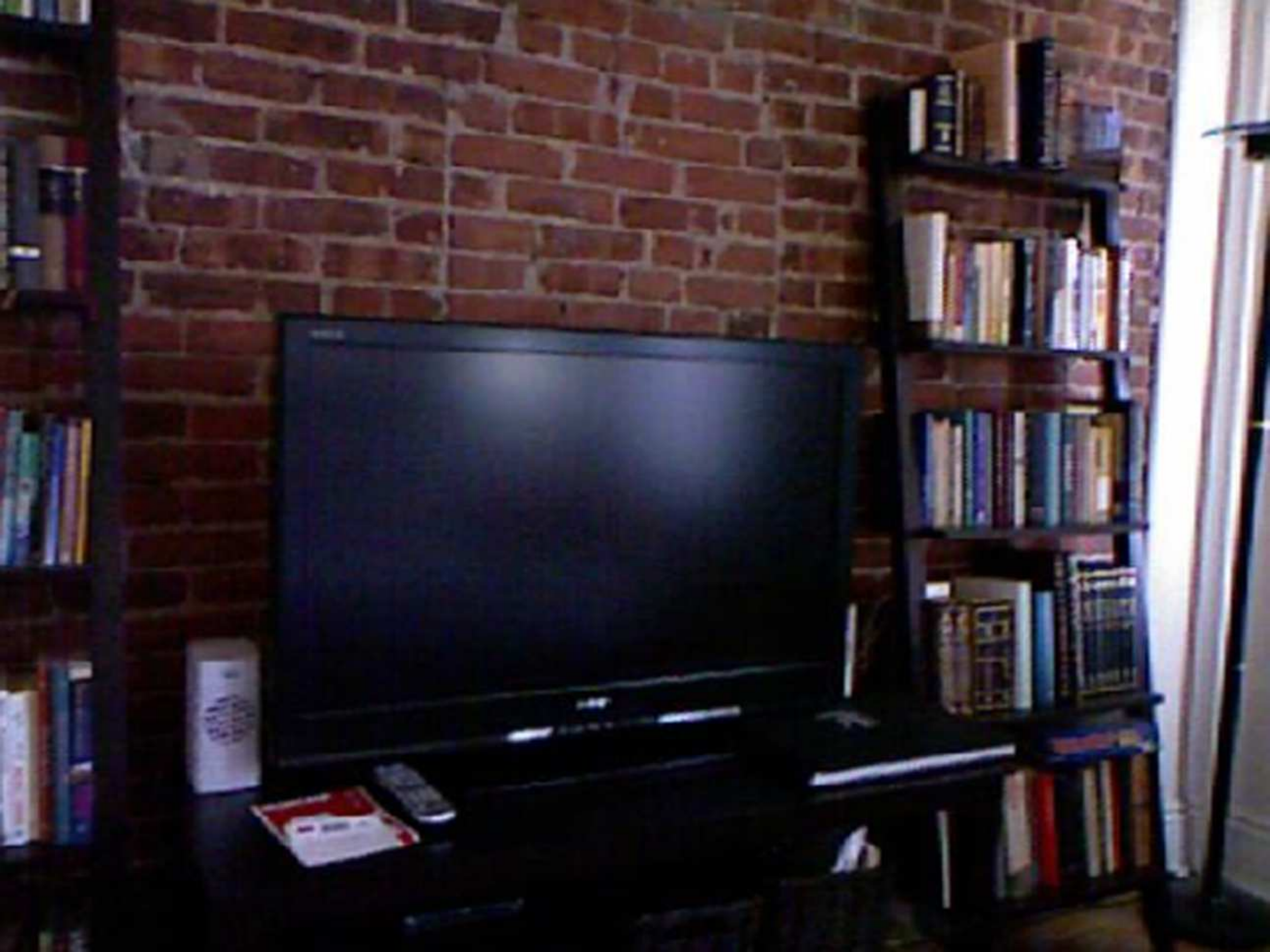}
	
	\vspace{-2pt}
	
	\subfigure[Depth image]{\includegraphics[width=\unitS]{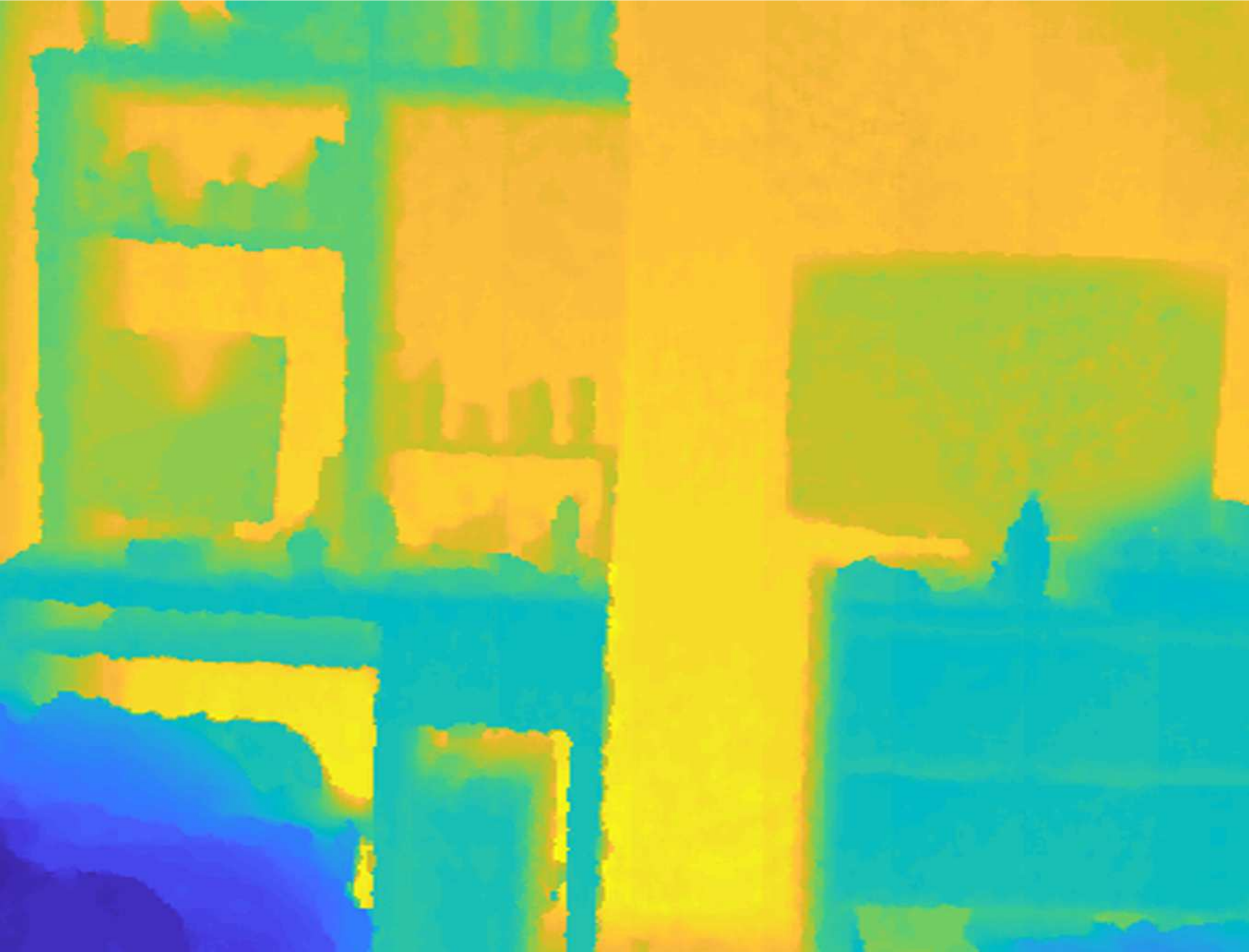}}%
	\subfigure[PCP \cite{GuerreroEtAl:PCPNet:EG:2018}]{\includegraphics[width=\unitS]{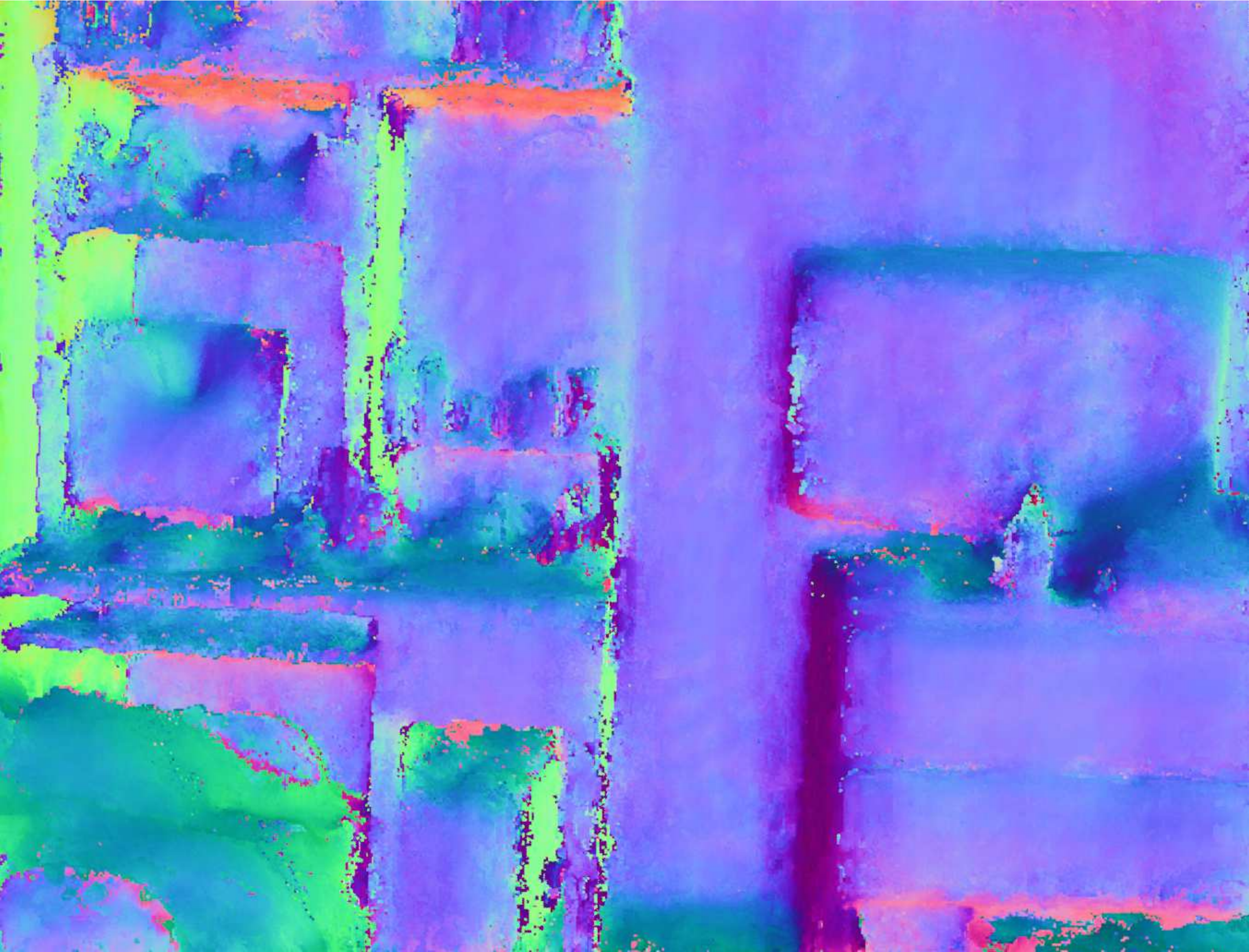}}%
	\subfigure[PCP \cite{GuerreroEtAl:PCPNet:EG:2018} + Ours]{\includegraphics[width=\unitS]{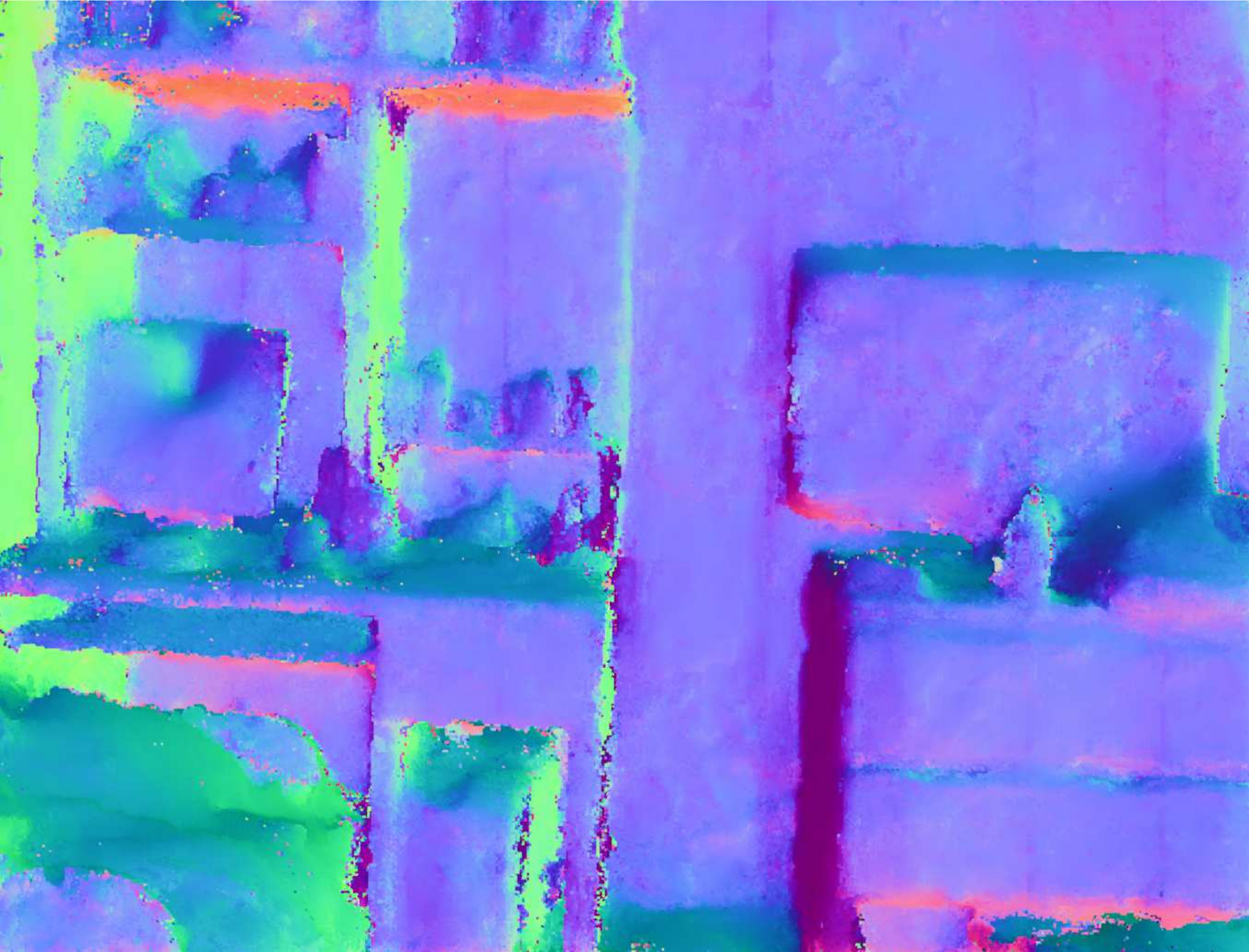}}%
	\subfigure[Nesti \cite{Ben-ShabatLF19}]{\includegraphics[width=\unitS]{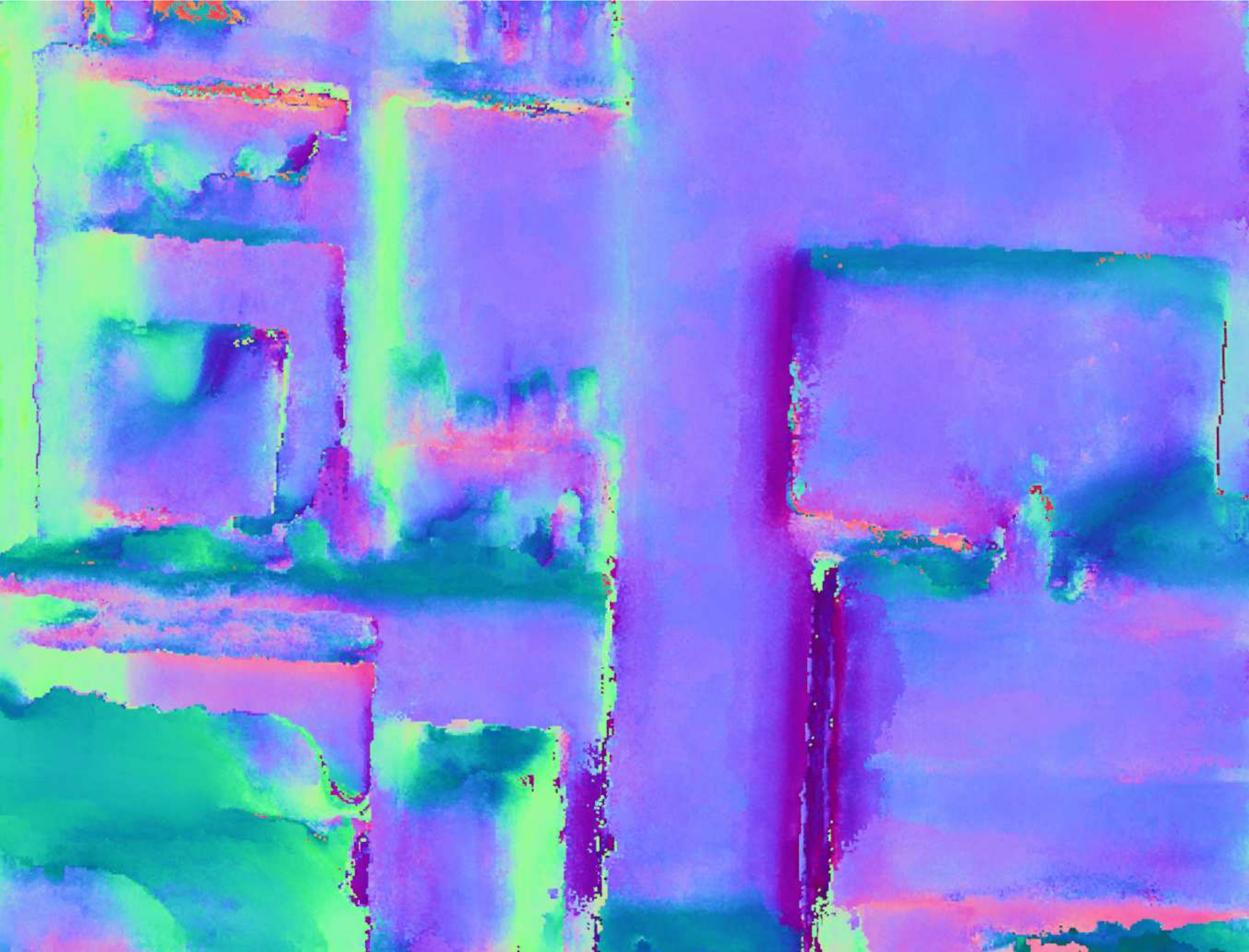}}%
	\subfigure[Nesti\cite{Ben-ShabatLF19} + Ours]{\includegraphics[width=\unitS]{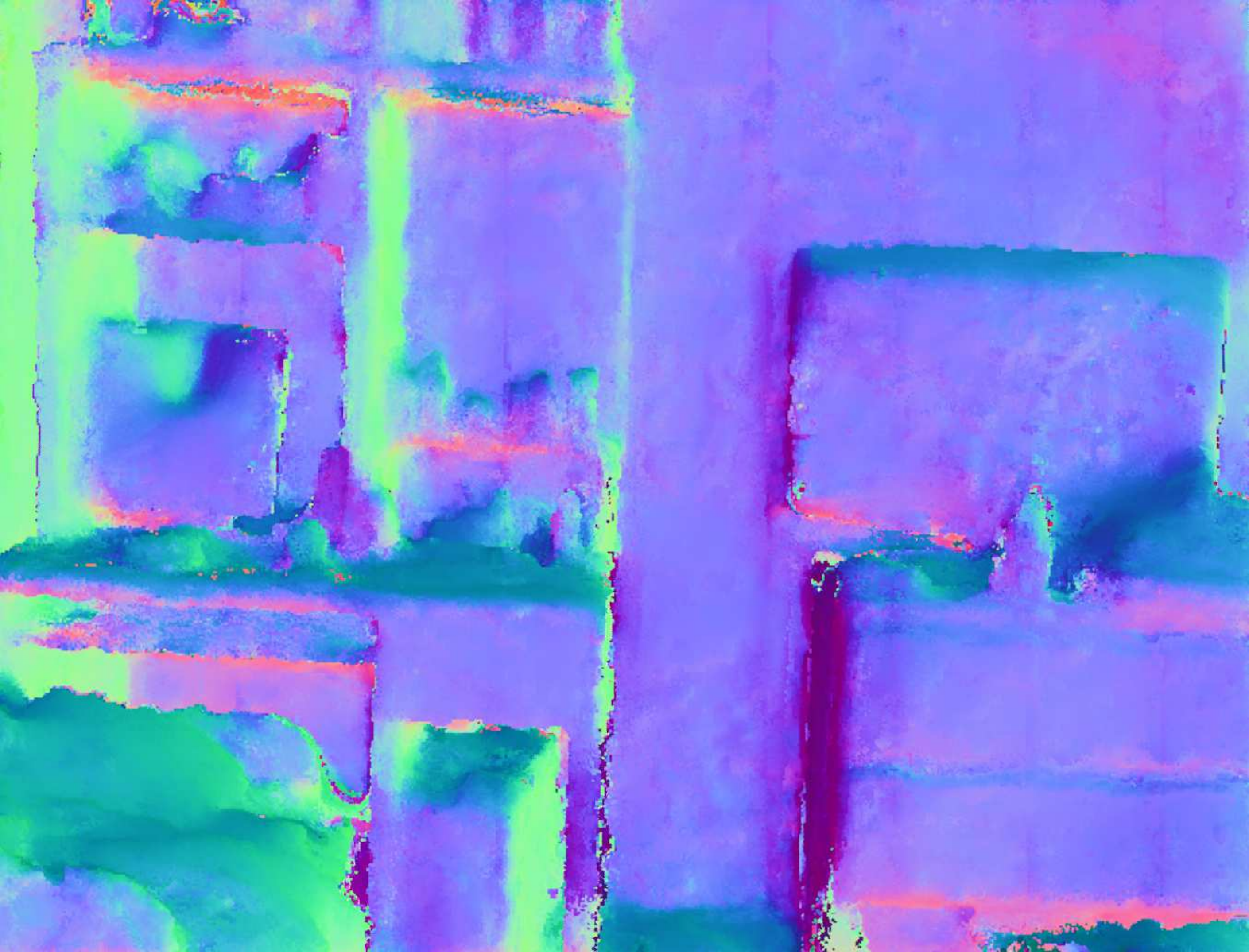}}%
	\subfigure[Our full pipeline]{\includegraphics[width=\unitS]{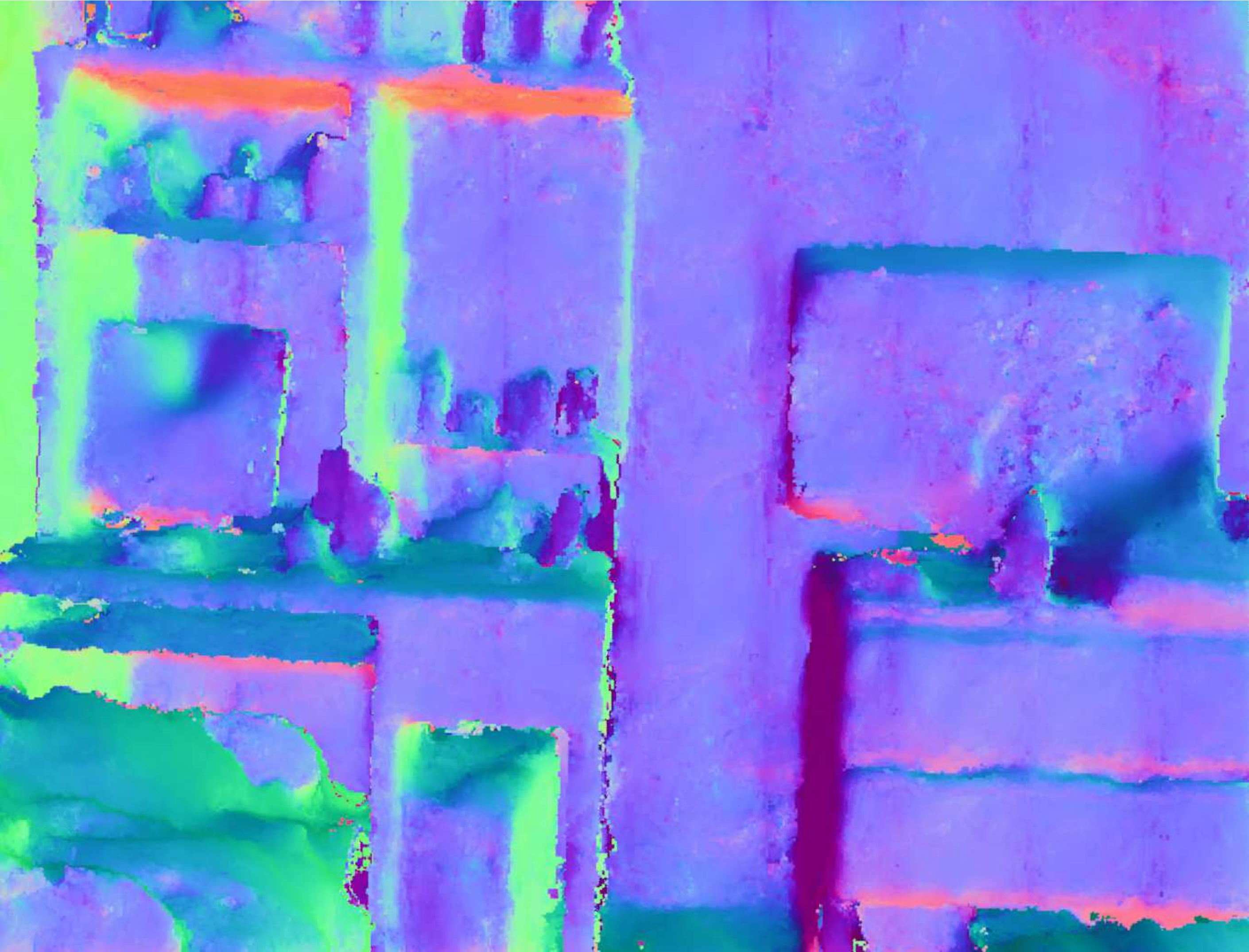}}%
	\subfigure[RGB image]{\includegraphics[width=\unitS]{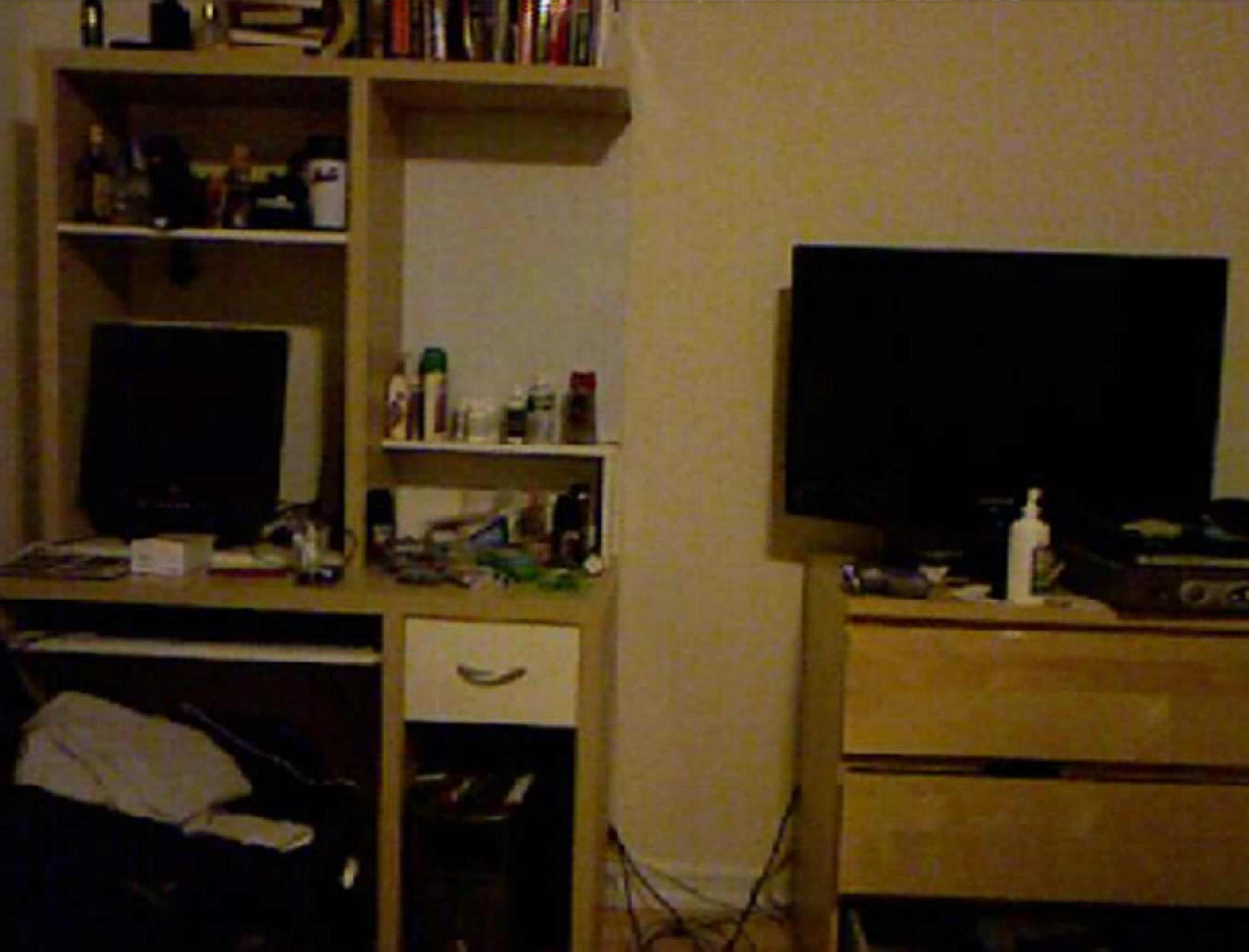}}
	
	\caption{Visual comparison of normal estimation results on the scanned point clouds from the NYU Depth V2 dataset \cite{silberman2012indoor}. ``PCP+Ours'' indicates that Refine-Net takes PCPNet \cite{GuerreroEtAl:PCPNet:EG:2018} results as initial normals. ``Nesti+Ours'' indicates the Nesti-Net \cite{Ben-ShabatLF19} results as initial normals.}
	\label{fig:indoor_refine}
\end{figure*}

For the NYUD-V2 dataset, we show visual comparisons of several real scans of indoor scenes in Fig.~\ref{fig:indoor}. Traditional geometric estimators (HF and our MFPS) can preserve tiny details but with the price of retaining noise (see Fig.~\ref{fig:indoor:HF} and Fig.~\ref{fig:indoor:MFPS}). Nesti-Net and PCPNet can well smooth the noisy surface, but over-smooth geometric features (see Fig.~\ref{fig:indoor:nesti} and Fig.~\ref{fig:indoor:pcp}). Our Refine-Net can well handle both challenges. For instance, the produced normal results of detailed objects are more faithful on, e.g., the toys and bottles from the 3-rd and 4-th rows respectively. Meanwhile, our method is able to remove the scanner noise on the flat surfaces.
Fig.~\ref{fig:indoor_refine} depicts more visual comparisons. It can be seen clearly that better normal results can be recovered when the predicted normals from other networks are refined in our framework. We refer the reader to look closely at the drawers and handles in the 1-st and 2-nd rows respectively. Other methods tend to smooth these details as their local regions are similar to planar areas. Refine-Net, however, produces results with nice details which are more faithful to the ground-truth surfaces. Our method can utilize additional information extracted from two feature modules and outperforms other networks on real-scanned point clouds.

Finally, we show more results on the Paris-rue-Madame \cite{serna:hal-00963812} dataset of large-scale scenes in Fig.~\ref{fig:outdoor}. Compared with PCPNet and Nesti-Net, our network produces better normal results on the vehicle in a real-world environment. Also, the details on the wall are clearly recovered, which are ignored by other methods.

\subsection{Efficiency}
We test our Refine-Net and other network models on a single RTX 2080 Ti GPU and report the execution times (per 1k points) and model sizes in Tab.~\ref{table:time}. The reported time of our method includes the initial normal estimation and Refine-Net forward time (aggregating all the clusters). The computing time of our geometric method depends largely on the shape geometric structures, with about 150s per shape on the PCPNet dataset (100k points for each shape). It is run on the CPU and can be further sped up by parallel computing. The forward pass time of Refine-Net is 0.8s (per 1k points), which is relatively fast among the related normal estimation networks. This can be contributed to the smaller network size and our efficient connection modules.

\begin{table}[t]
	\centering
	\small
	\setlength{\tabcolsep}{3.5mm}
	\caption{Comparison of complexity and execution times (1k points) of different normal estimation networks. The reported time of our method includes the initial normal estimation and Refine-Net forward time.}
	\begin{tabular}{c|cc}
		\toprule[1pt]
		 & Time, 1k p. & Num. parameters \\ 
		\midrule[0.3pt]
		\midrule[0.3pt]
		HoughCNN \cite{boulch2016deep} & 1.0 s & 9.7M \\
		PCPNet \cite{GuerreroEtAl:PCPNet:EG:2018} & 4.8 s & 22M \\
		Nesti-Net \cite{Ben-ShabatLF19} & 9.5 s & 179M \\
		Ours & 2.2 s & 10.4M \\
		\bottomrule[1pt]
	\end{tabular}
	\label{table:time}
\end{table}

\begin{table}[t]
	\centering
	\small
	\setlength{\tabcolsep}{3.5mm}
	\caption{Network architecture details of height-map modules for Refine-Net (left) and ablation network (right).}
	\begin{tabular}{c|c}
		\toprule[1pt]
		Height-map module & Height-map module (ablation) \\ 
		\midrule[0.3pt]
		\midrule[0.3pt]
		Conv(3, 3, 64, P=1) + Relu & Conv(3, 3, 64, P=1) + Relu \\
		Conv(3, 3, 64, P=1) + Relu & Conv(3, 3, 64, P=1) + Relu \\
		Maxpool(3, 3, S=1) & Maxpool(3, 3, S=1) \\
		Conv(3, 3, 128, P=1) + Relu & Conv(3, 3, 128, P=1) + Relu \\
		Conv(3, 3, 128, P=1) + Relu & Conv(3, 3, 128, P=1) + Relu \\
		Maxpool(3, 3, S=1) & Maxpool(3, 3, S=1) \\
		Conv(3, 3, 128, P=1) + Relu & Conv(3, 3, 256, P=1) + Relu \\
		FC(256) & FC(512) \\
		FC(128) & FC(256) \\
		FC($d$) & FC($d$) \\
		\bottomrule[1pt]
	\end{tabular}
	\label{table:network_architecture}
\end{table}

\begin{table*}
	\centering
	\footnotesize
	\setlength{\tabcolsep}{3.5mm}
	\caption{The normal estimation errors of ablation networks on the synthetic dataset \cite{wang2016mesh}.}
	\begin{tabular}{c |cc|cc|cc|cc|cc} 
		\toprule[1pt]
		Category & \multicolumn{2}{|c|}{BigNoise} & \multicolumn{2}{|c|}{SharpFeature} & \multicolumn{2}{|c|}{RichFeature} & \multicolumn{2}{|c|}{SmoothSurface} & \multicolumn{2}{c}{Average}  \\
		& mean & rmse & mean & rmse & mean & rmse & mean & rmse & mean & rmse\\ 
		\midrule[0.3pt]
		\midrule[0.3pt]
		MFPS (initial normal)  				& 5.72  & 11.72 & 4.36  & 7.86  & 4.82  & 6.97  & 4.10  & 6.32  & 4.75  & 8.22  \\
		simple MFPS & 7.28 &16.75& 5.86& 12.47& 5.01& 7.91& 4.89& 8.49& 5.76& 11.40 \\
		\midrule[0.3pt]
		normals  						& 5.51  & 12.53 & 3.86 & 7.46 & 4.52 & 6.55 & 3.72 & 5.64 & 4.40 & 8.04  \\
		normals\&HMPs					& 5.50  & 12.48 & 3.74  & 7.26 & 4.40  & 6.32  & 3.61  & 5.38  & 4.31  & 7.86 \\
		normals\&points					& 4.83  & 10.86 & 3.55  & 6.67 & 4.14  & 5.79  & 3.37  & 4.78  & 3.97  & 7.03 \\
		\midrule[0.3pt]
		concat							& 4.83  & 10.98 & 3.61  & 6.91 & 4.23  & 5.99  & 3.48  & 5.06 & 4.04  & 7.23 \\
		residual						& 4.93  & 11.10 & 3.59 & 6.77 & 4.19  & 5.87  & 3.40 & 4.88 & 4.03 & 7.15 \\
		Refine-Net-Rot					& 4.84  & 11.03 & 3.56  & 6.78 & 4.19  & 5.90  & 3.43  & 4.93  & 4.00  & 7.16  \\
		Refine-Net-Trans 				& 4.87  & \textbf{10.71} & 3.54  & 6.49  & 4.10  & 5.67  & 3.29  & 4.59  & 3.95  & 6.87  \\
		Refine-Net-Weight				& \textbf{4.74}  & 10.73 & \textbf{3.37}  & \textbf{6.39}  & \textbf{4.07}  & \textbf{5.65}  & \textbf{3.28}  & \textbf{4.56}  & \textbf{3.87}  & \textbf{6.83}  \\
		\bottomrule[1pt]
	\end{tabular}
	\label{table:ablation_results}
\end{table*}

\begin{figure}[t]
	\centering
	
	\includegraphics[width=0.98\linewidth]{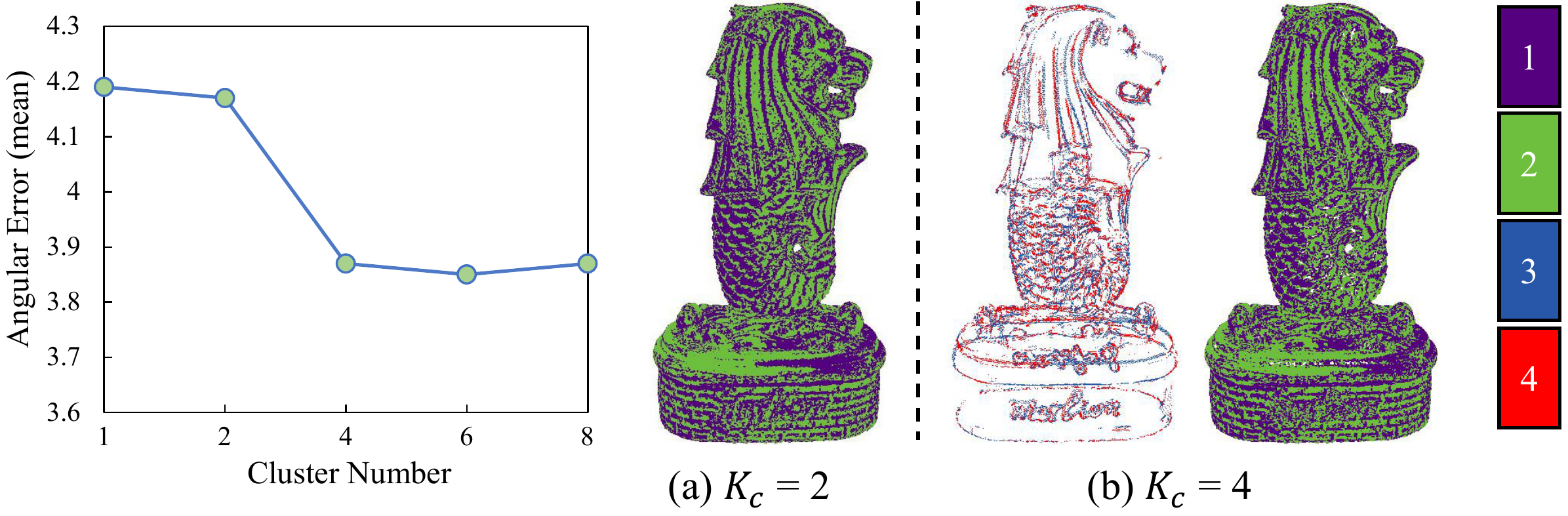}
	
	\caption{\textbf{Left:} The averaged angluar error (mean) improves when the cluster number $K_c$ increases. We use $K_c = 4$ by default and a larger number improves the results very slightly. \textbf{Right:} We show points divided into different clusters when using $K_c = 2$ (a) and $K_c = 4$ (b). In (b), the 4 clusters are visualized in two parts, in which the left part (with red and blue colors) can capture points on geometric features. Color coding is given on the right.}
	\label{fig:cluster}
\end{figure}

%
	
%

\begin{figure*}[t]
	\centering
	\includegraphics[width=0.99\linewidth]{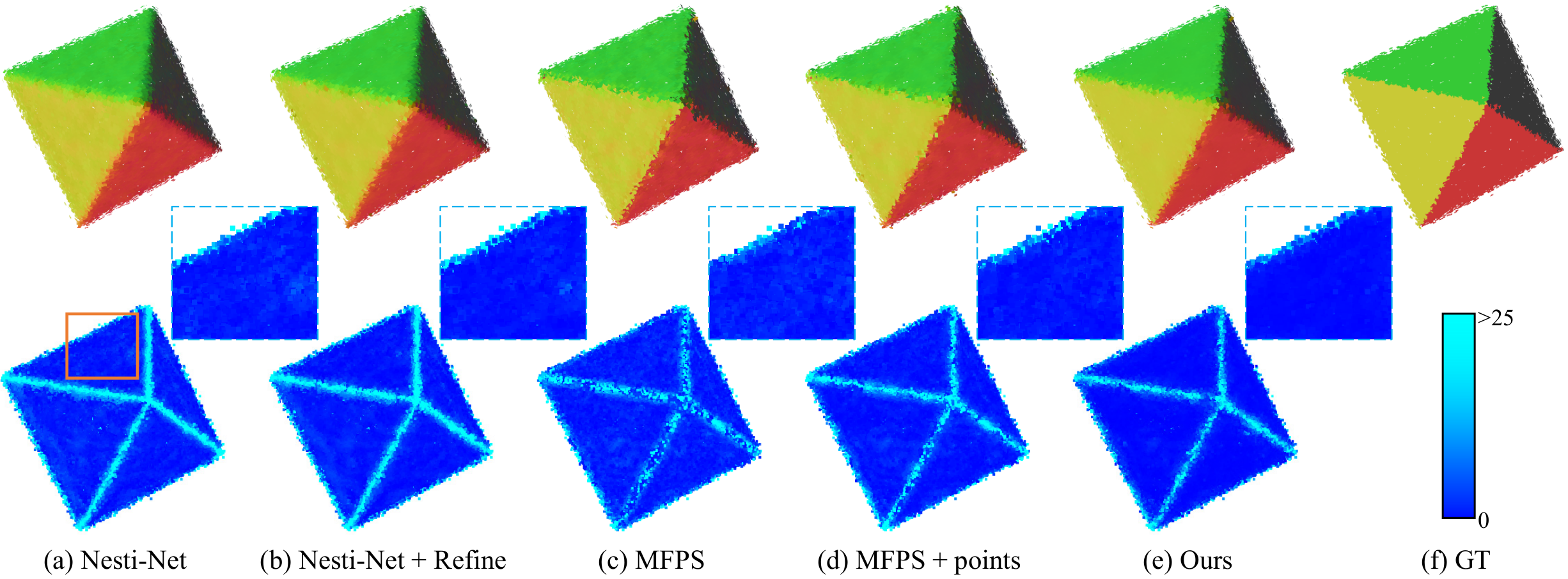}
	
	\caption{Visual comparison of normal results and error heatmaps between different methods. We show Nesti-Net and ``Nesti-Net + Refine-Net" results in (a) and (b) respectively. (c) denotes our initial normals computed by MFPS, (d) denotes the Refine-Net with only the point module, and (e) is the result from our full pipeline. Zoom in to see clearer.}
	\label{fig:insight}
\end{figure*}

	
	
	

\begin{figure}[t]
	\newlength{\unity}
	\setlength{\unity}{0.2\linewidth}
	\newlength{\unitz}
	\setlength{\unitz}{0.1\linewidth}
	\centering
	
	\includegraphics[width=\unity]{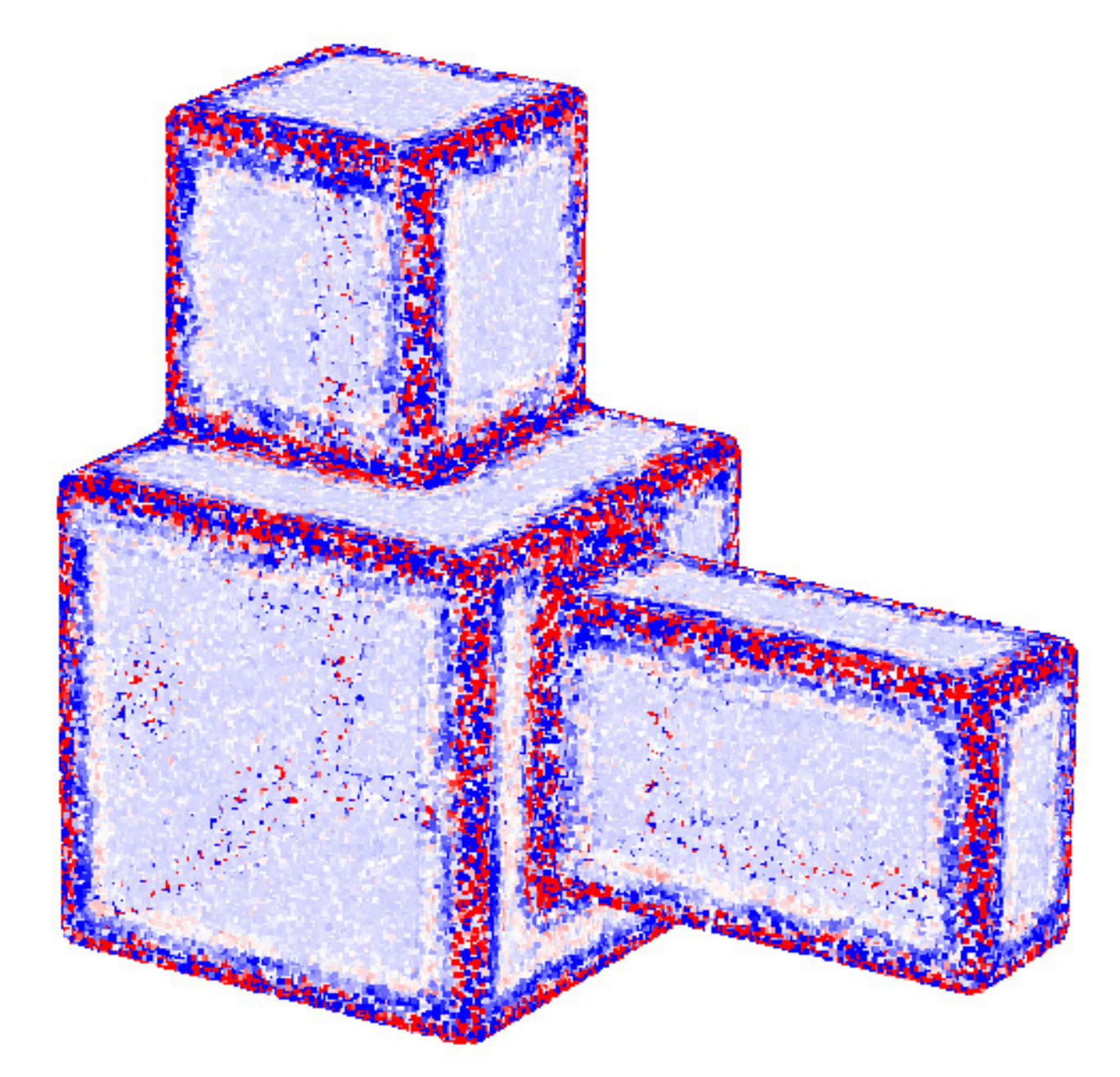}%
	\hspace{10pt}\includegraphics[width=\unity]{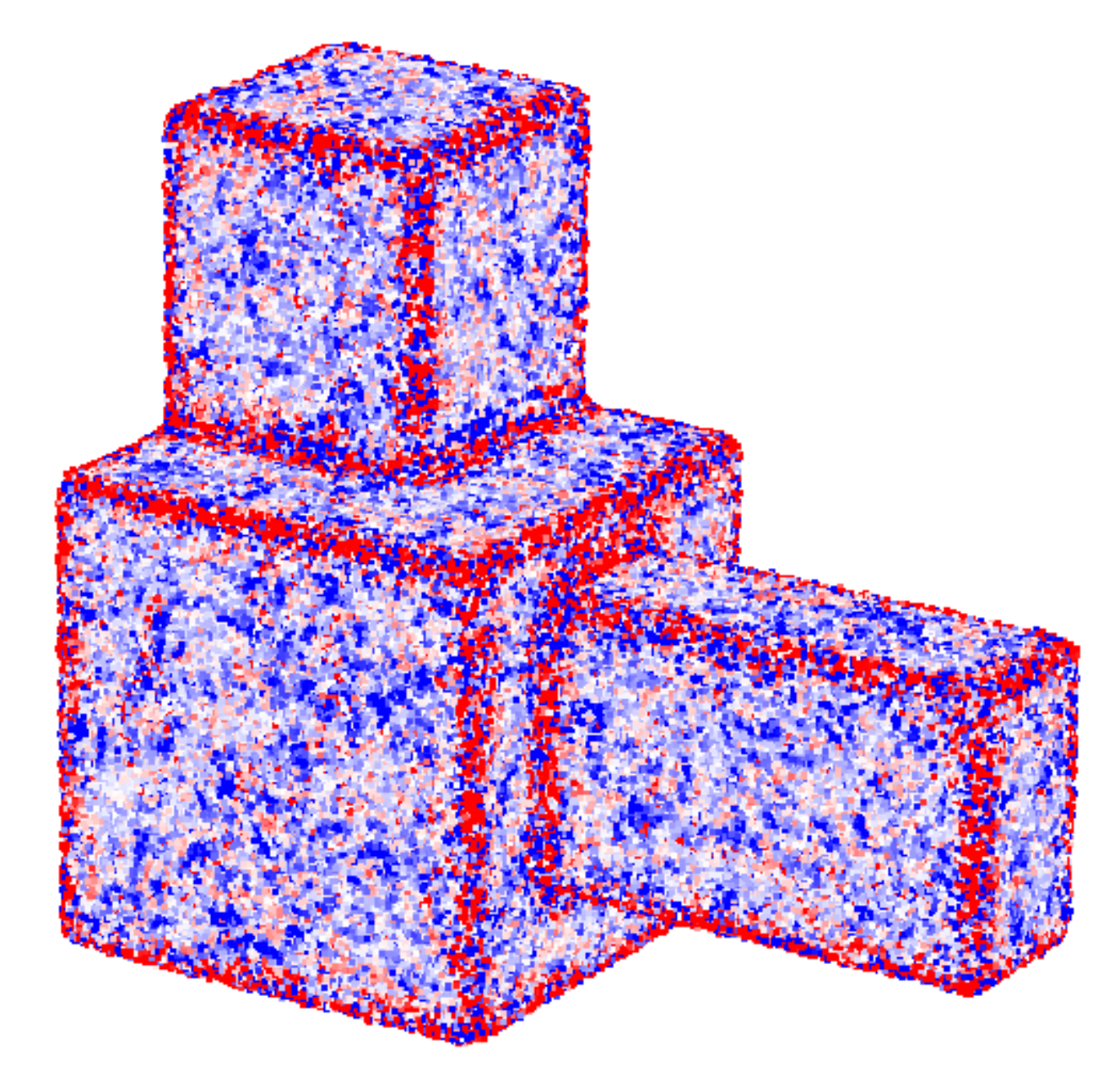}%
	\hspace{10pt}\includegraphics[width=\unity]{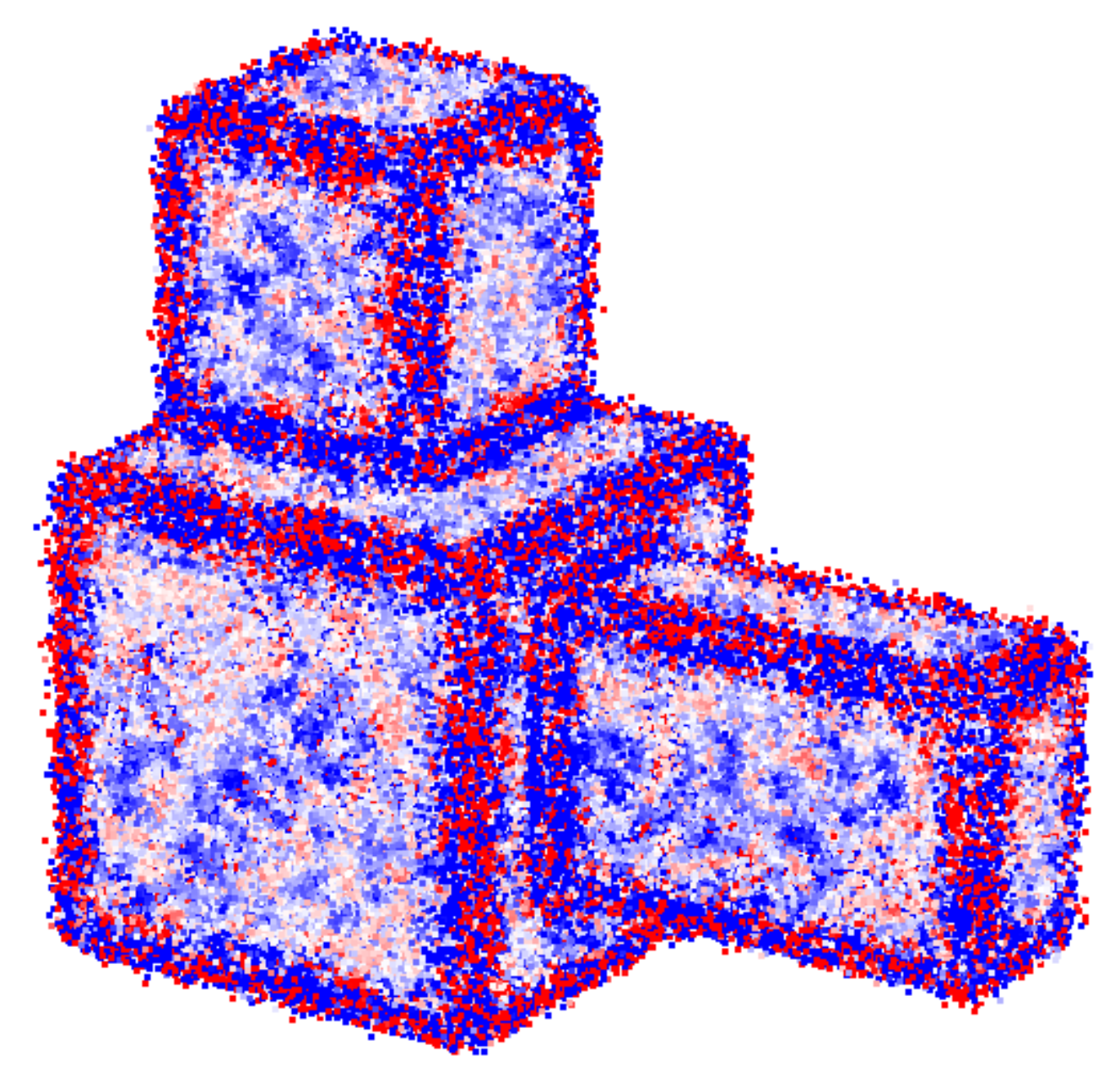}%
	\hspace{10pt}\includegraphics[width=\unity]{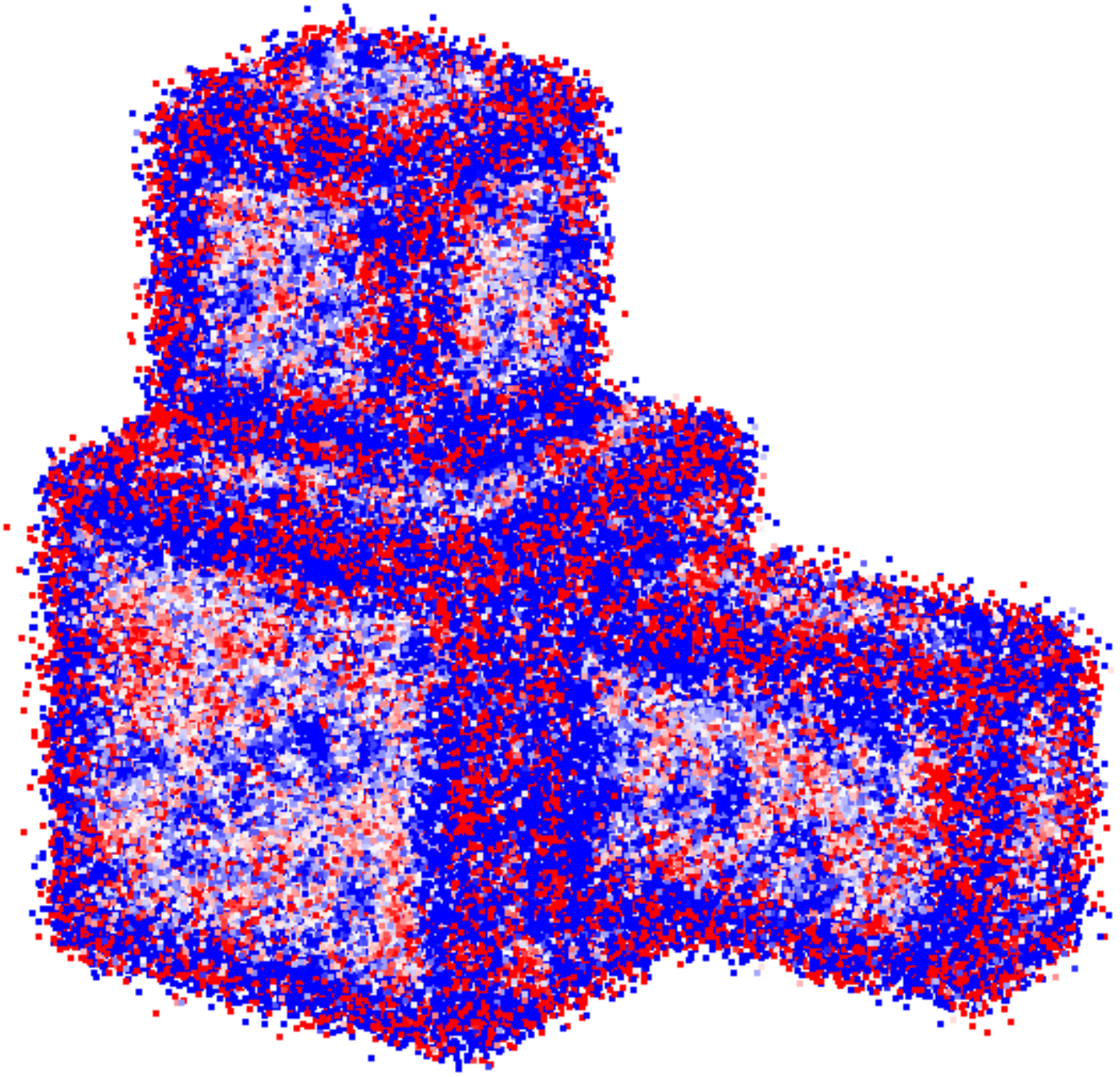}
	
	\includegraphics[width=\unitz]{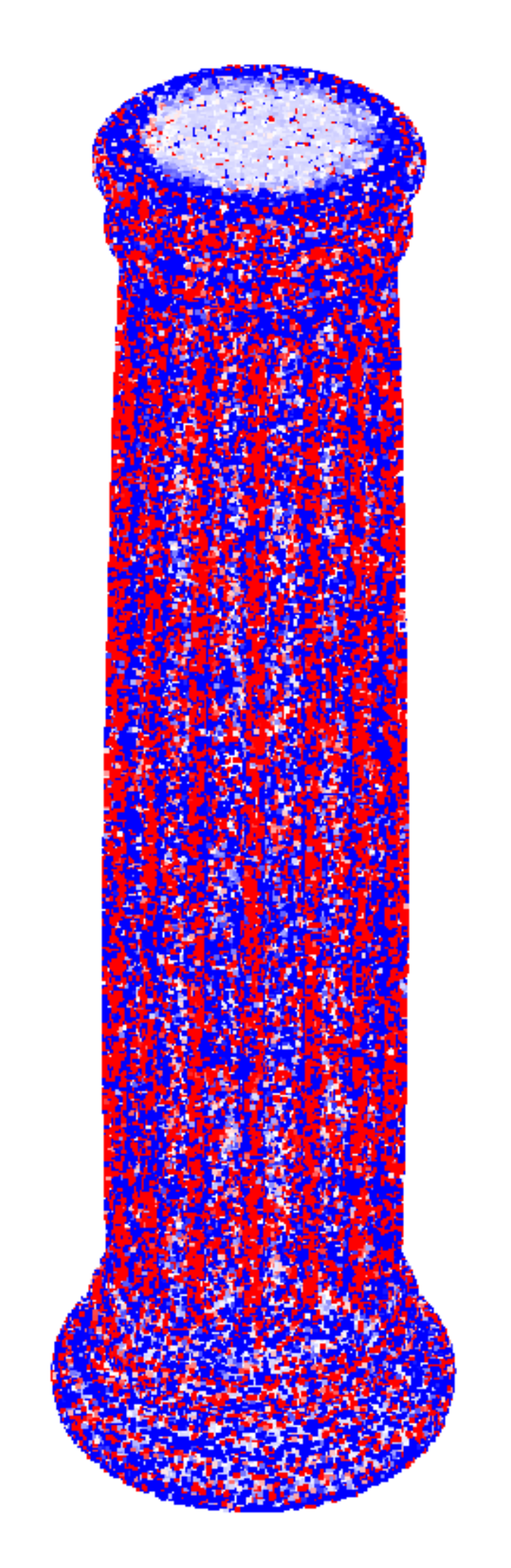}%
	\hspace{36pt}\includegraphics[width=\unitz]{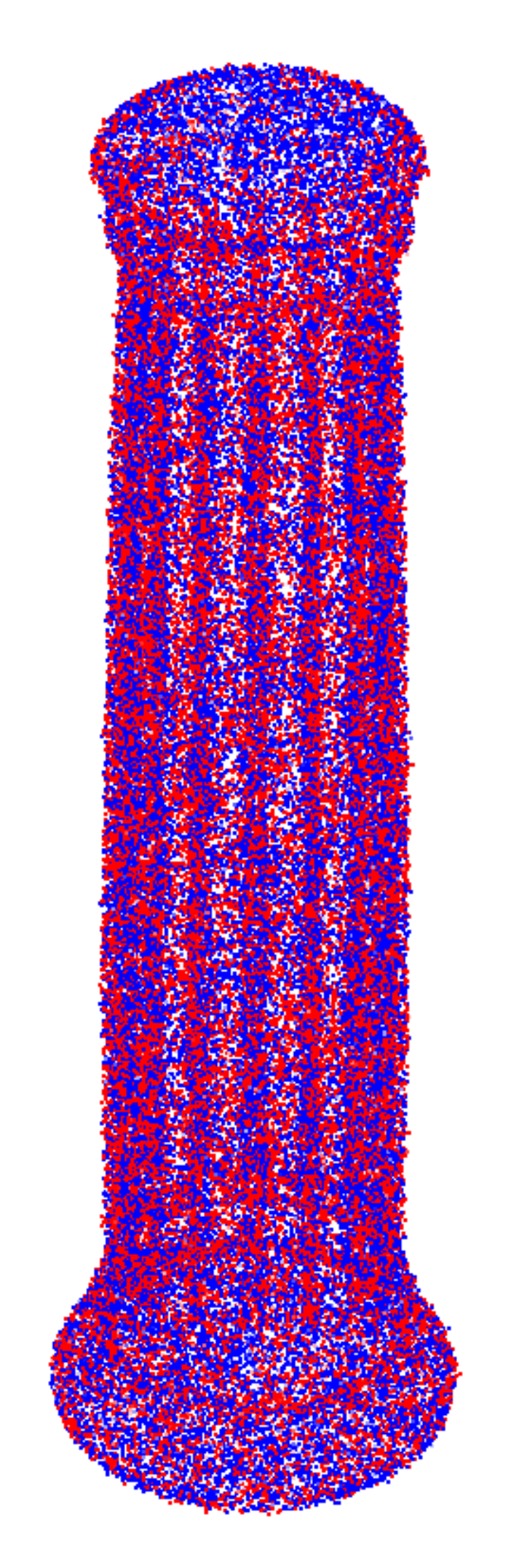}%
	\hspace{36pt}\includegraphics[width=\unitz]{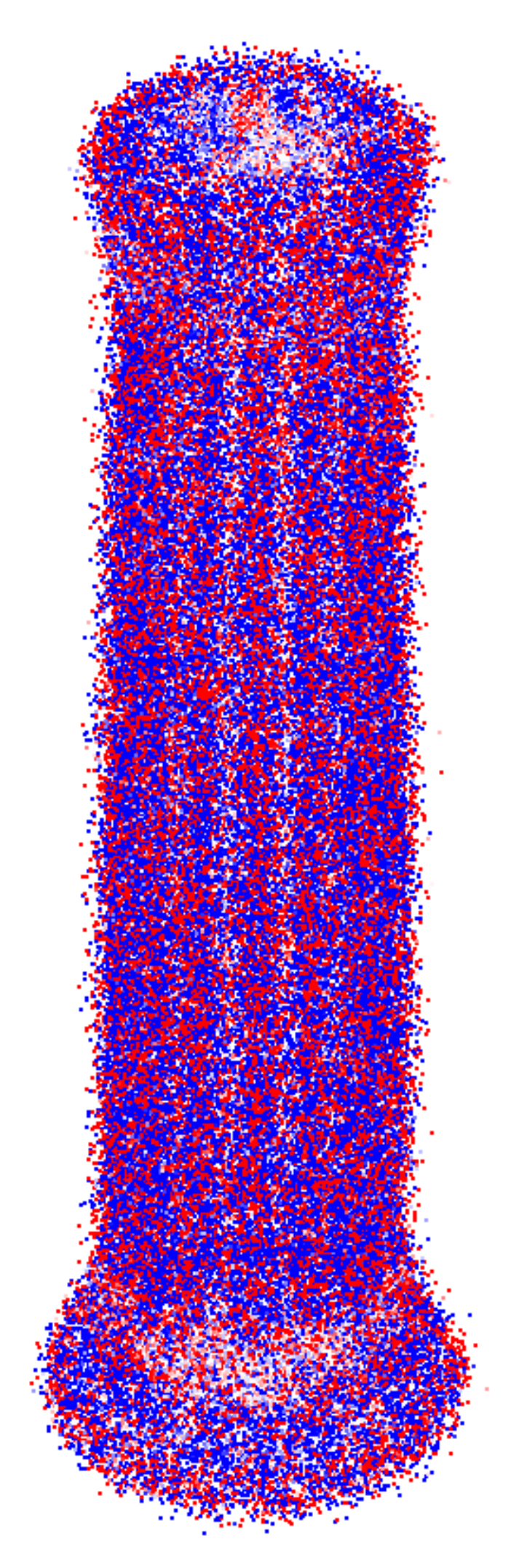}%
	\hspace{36pt}\includegraphics[width=\unitz]{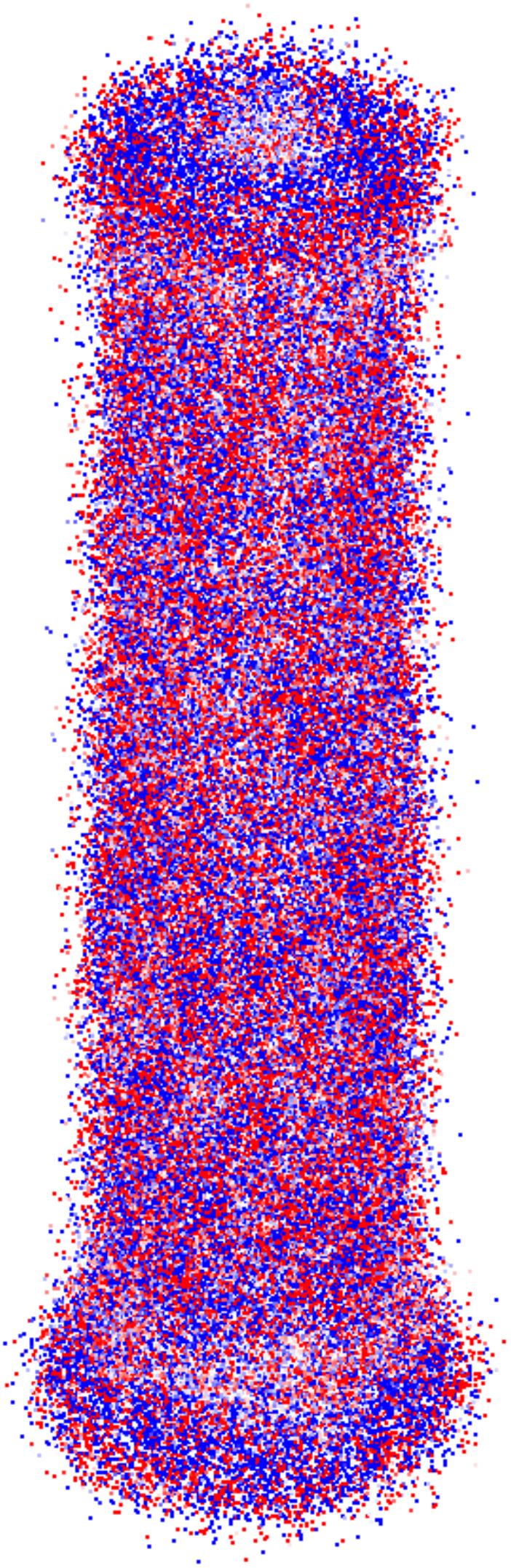}
	
	\vspace{5pt}
	\includegraphics[width=0.6\linewidth]{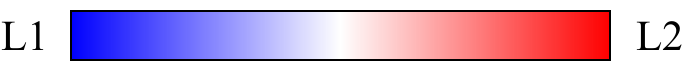}
	
	\caption{Comparison of Refine-Net results using the L1 loss and the L2 loss on shapes with increasing noise levels. A point is colored blue (red) when the corresponding angular error is smaller with the L1 (L2) loss. We can see that the L2 loss performs clearly better on sharp edges and tiny details as shown in the 1-st and 2-nd rows respectively.}
	\label{fig:loss2}
\end{figure}

\begin{figure*}
	\centering
	\includegraphics[width=0.122\linewidth]{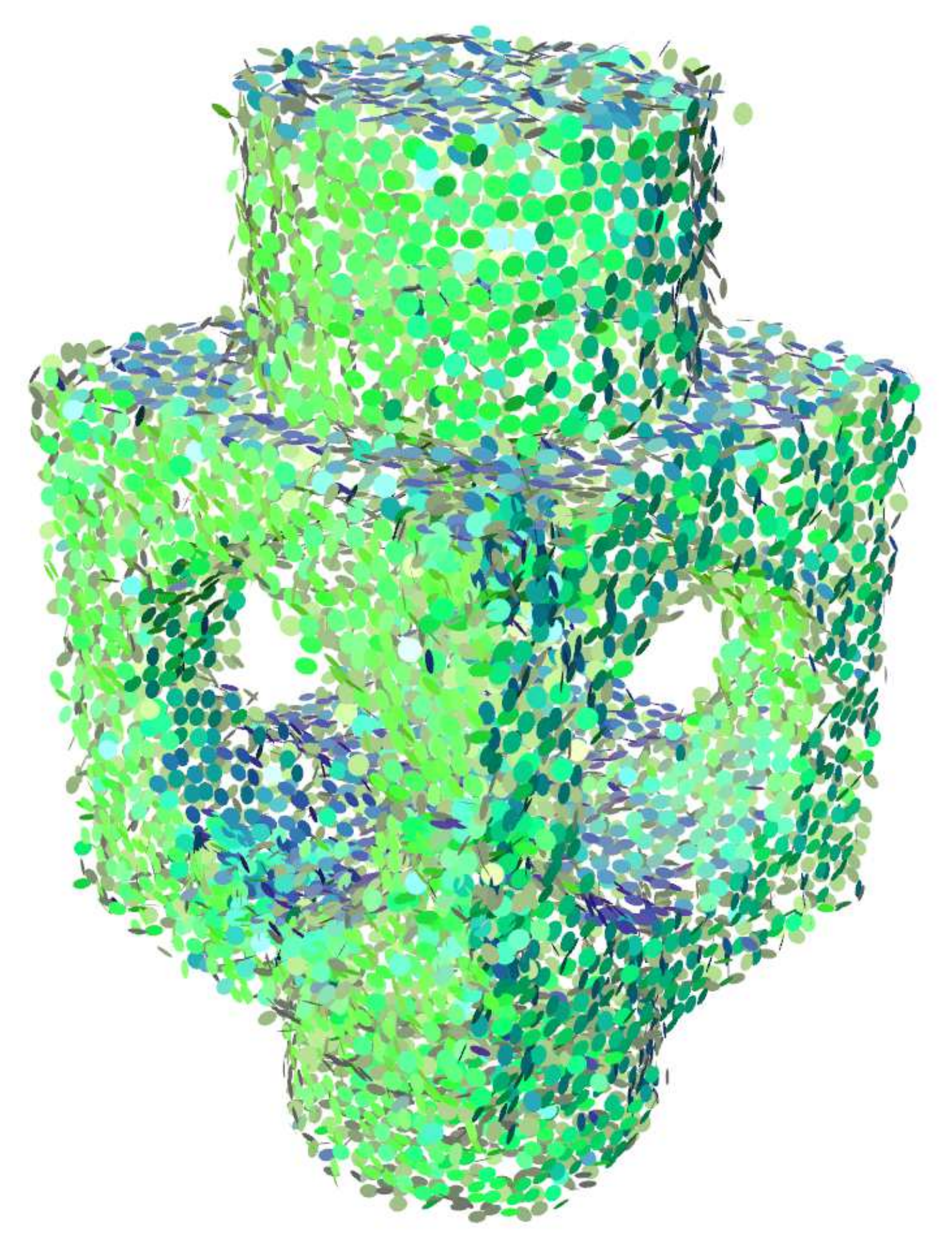}%
	\includegraphics[width=0.122\linewidth]{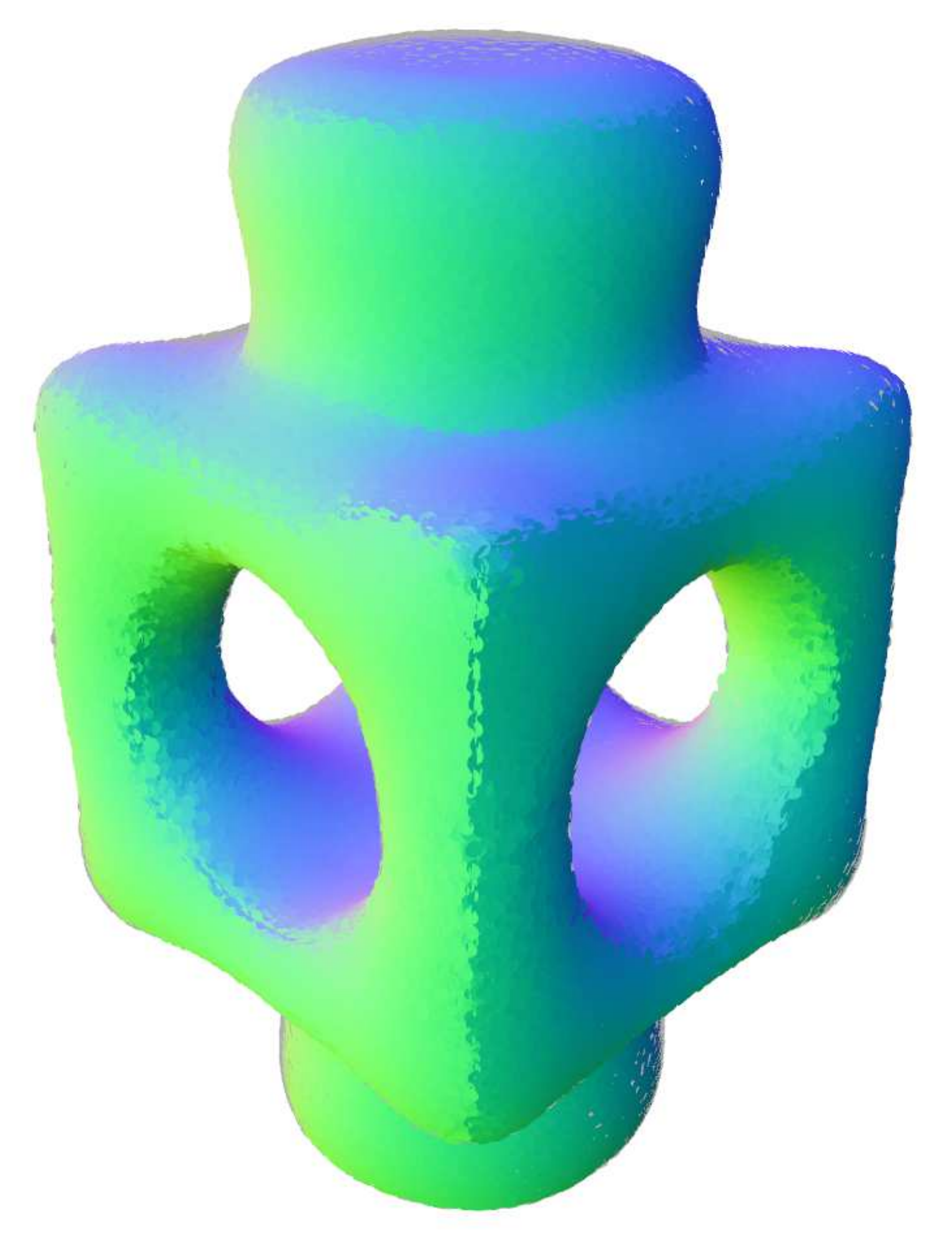}%
	\includegraphics[width=0.122\linewidth]{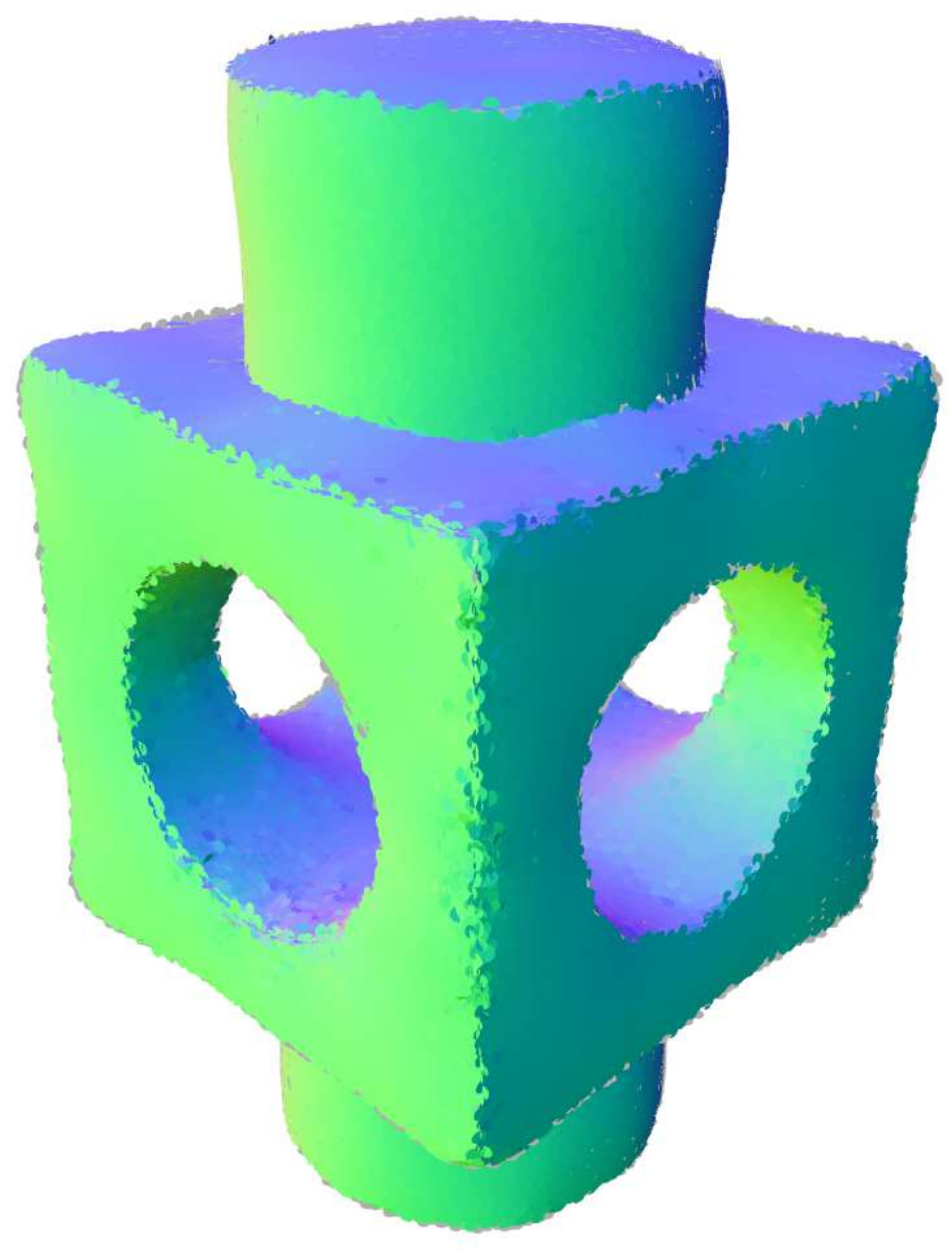}%
	\includegraphics[width=0.122\linewidth]{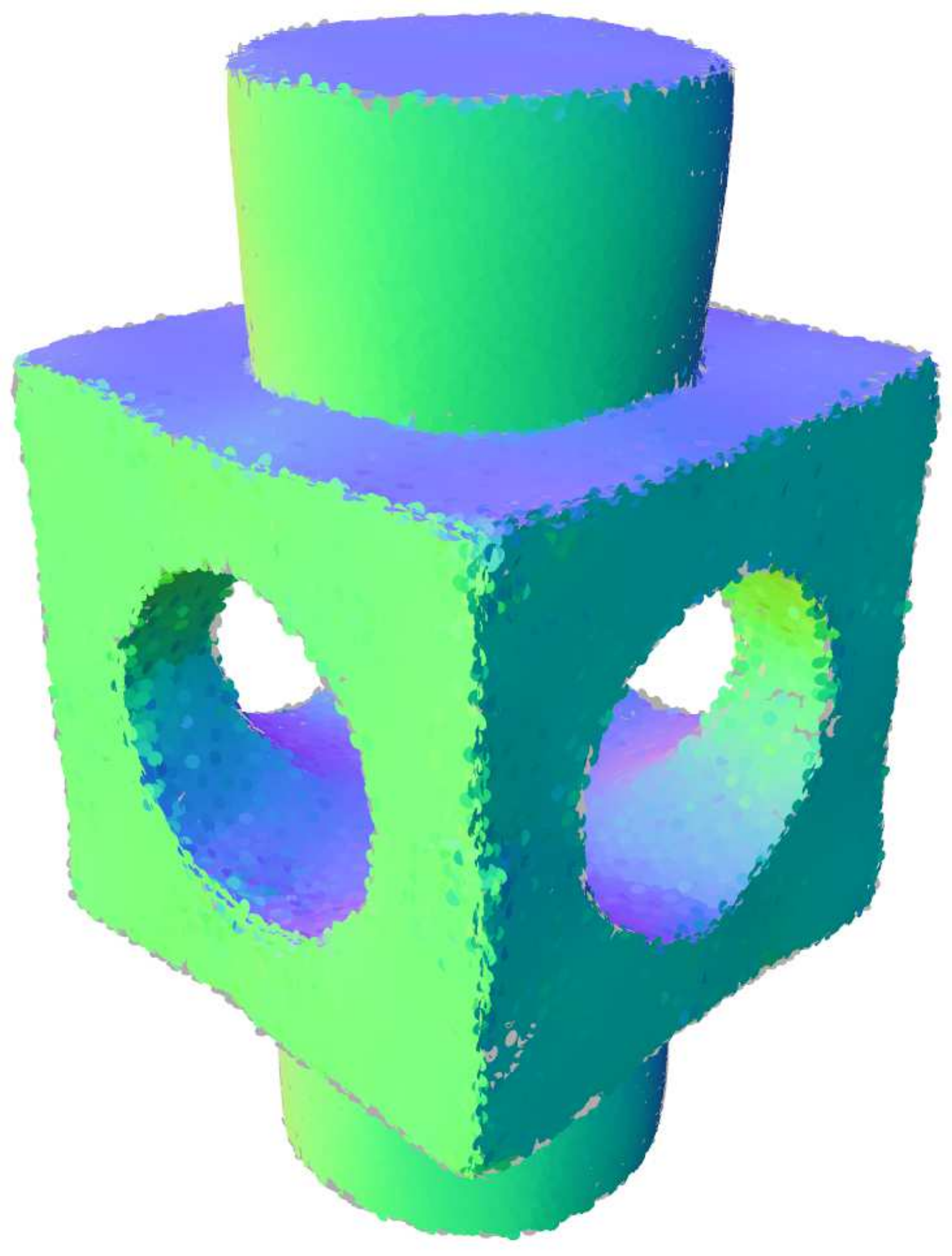}%
	\includegraphics[width=0.122\linewidth]{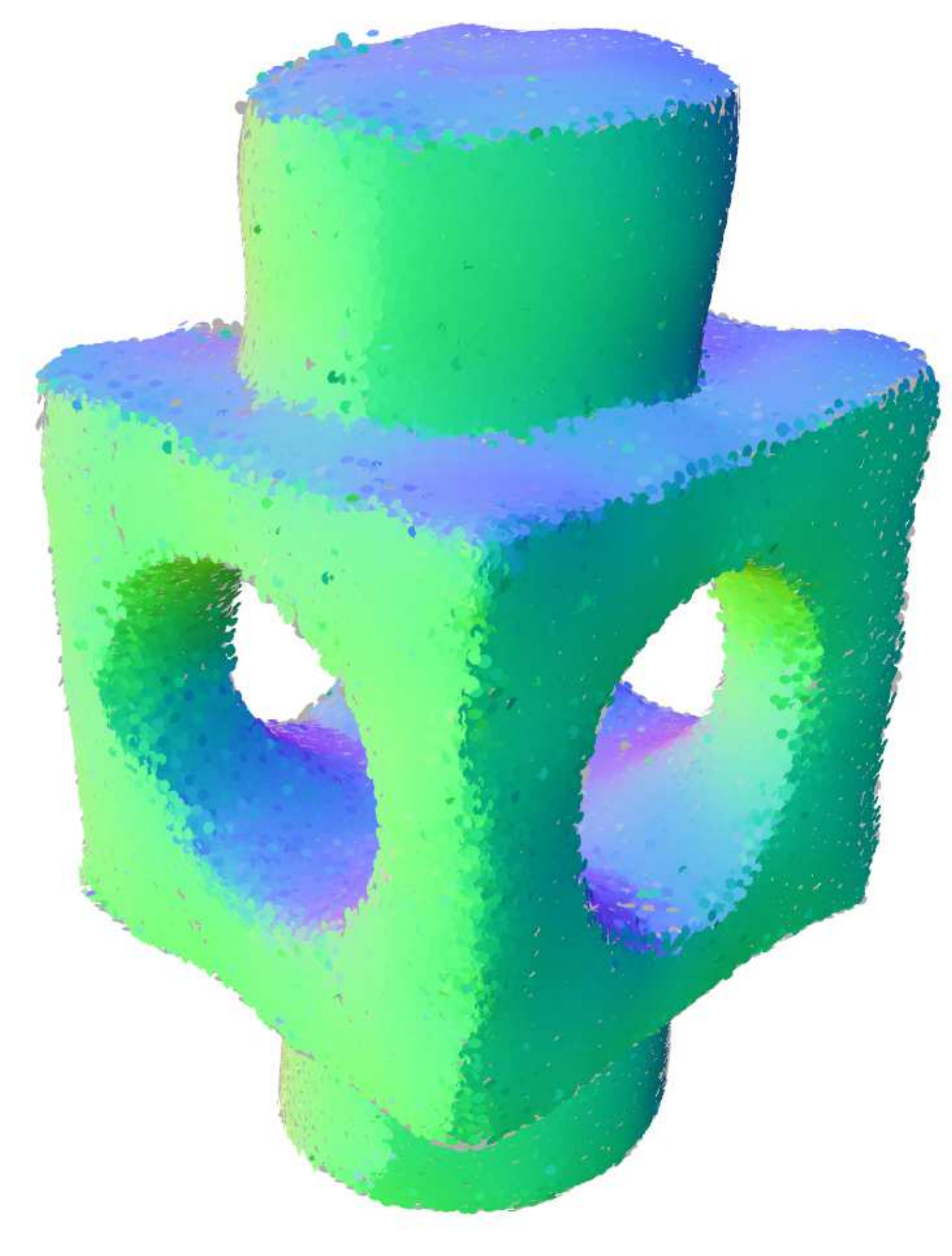}%
	\includegraphics[width=0.122\linewidth]{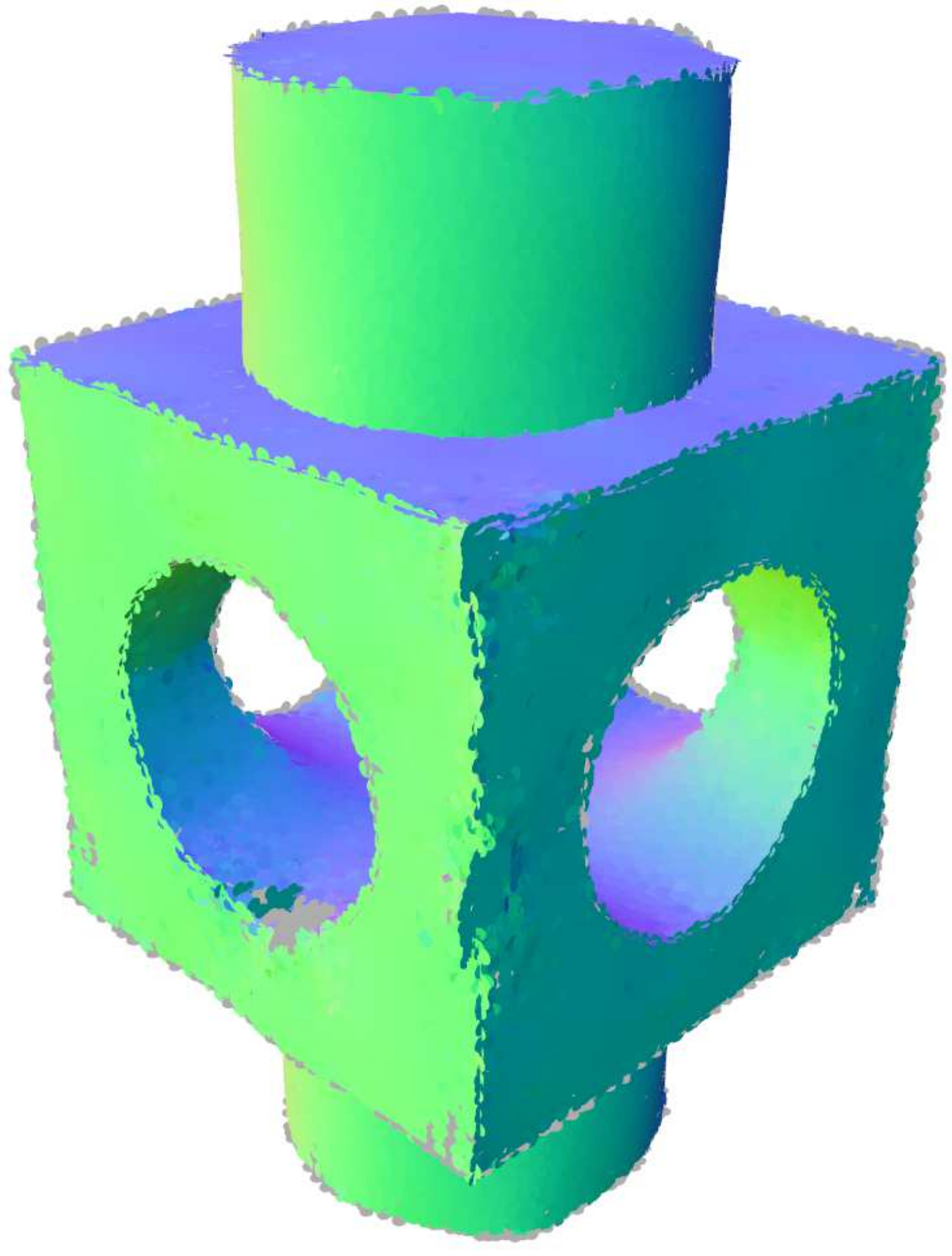}%
	\includegraphics[width=0.122\linewidth]{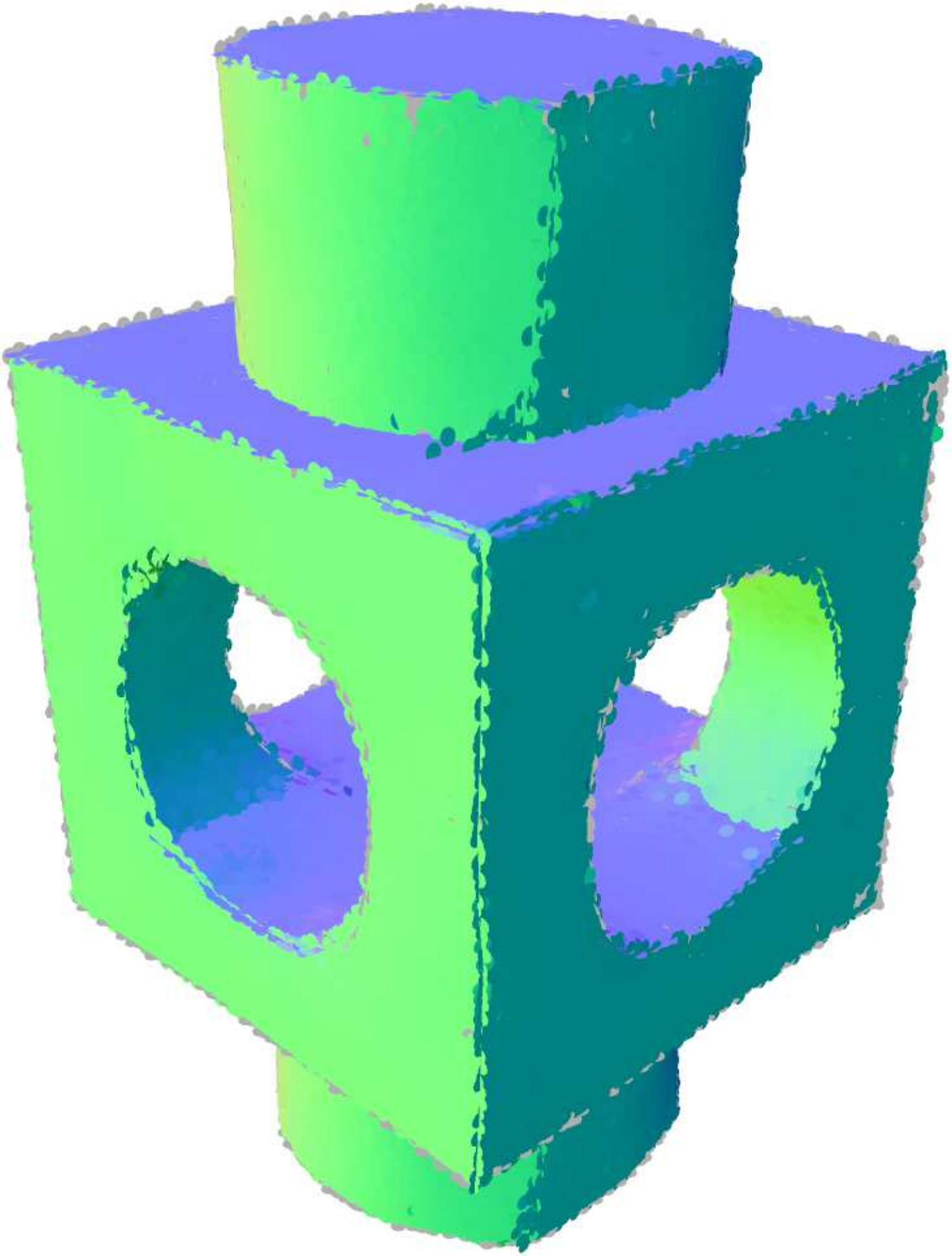}%
	\includegraphics[width=0.122\linewidth]{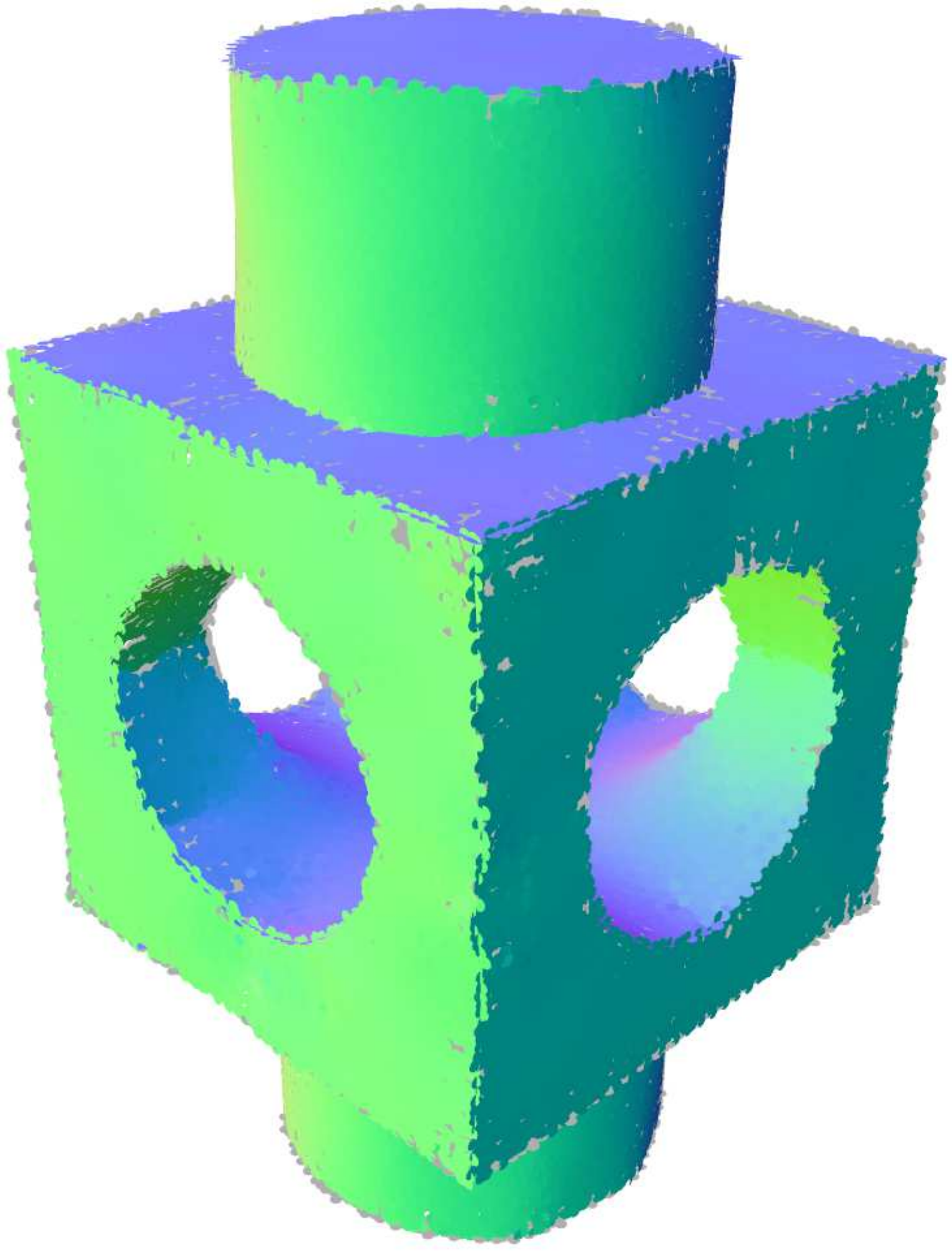}

	\includegraphics[width=0.122\linewidth]{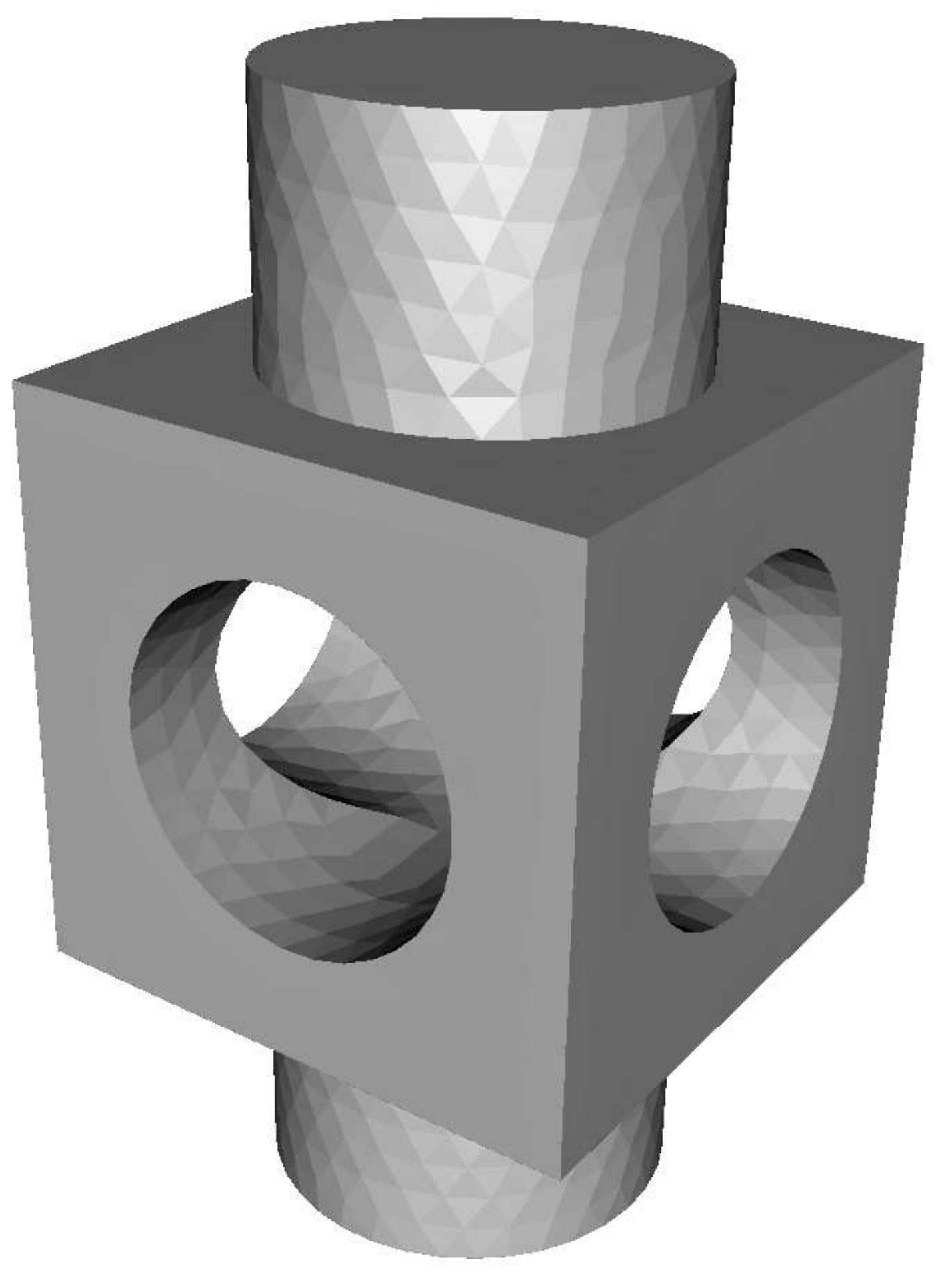}%
	\includegraphics[width=0.122\linewidth]{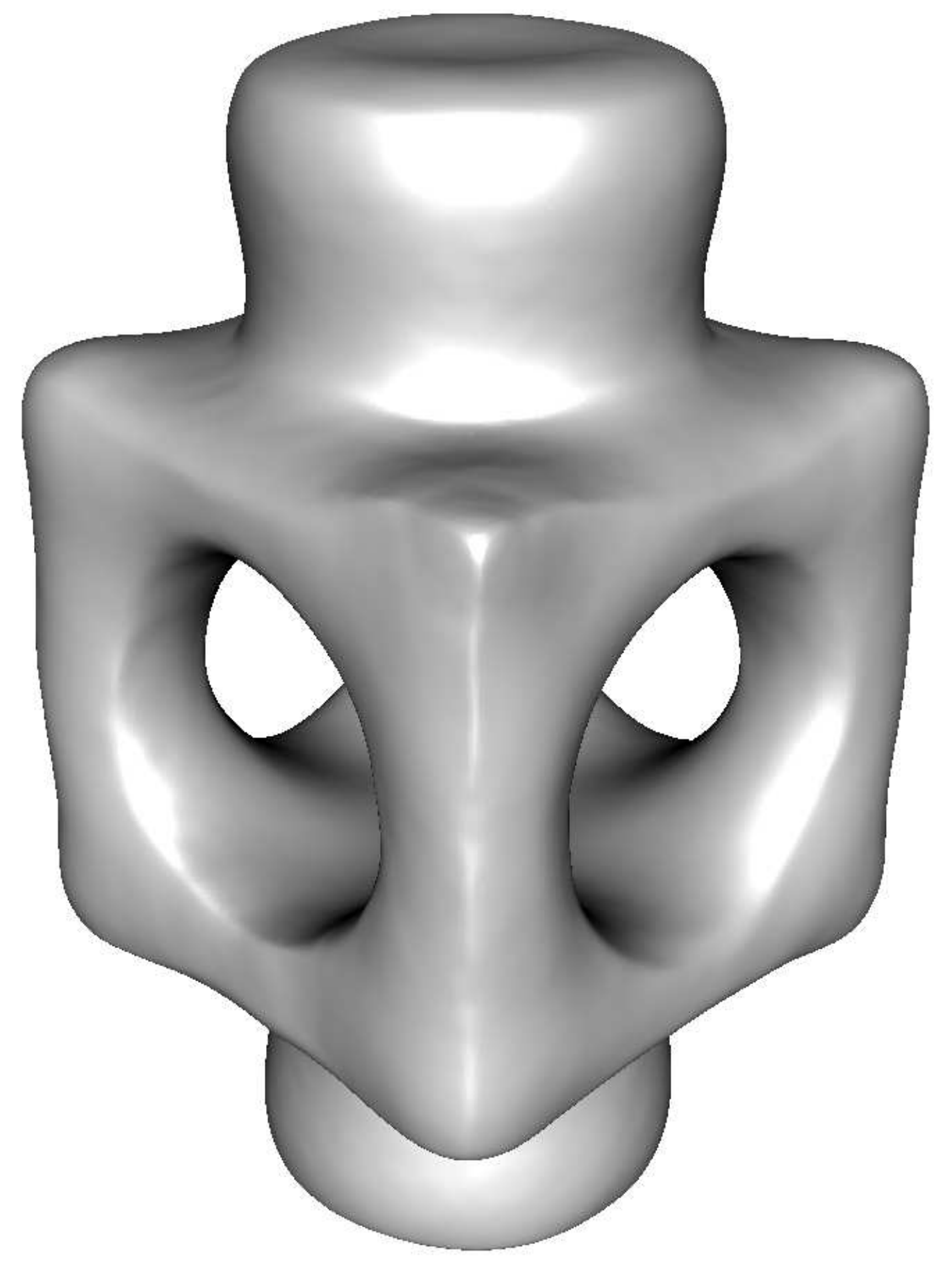}%
	\includegraphics[width=0.122\linewidth]{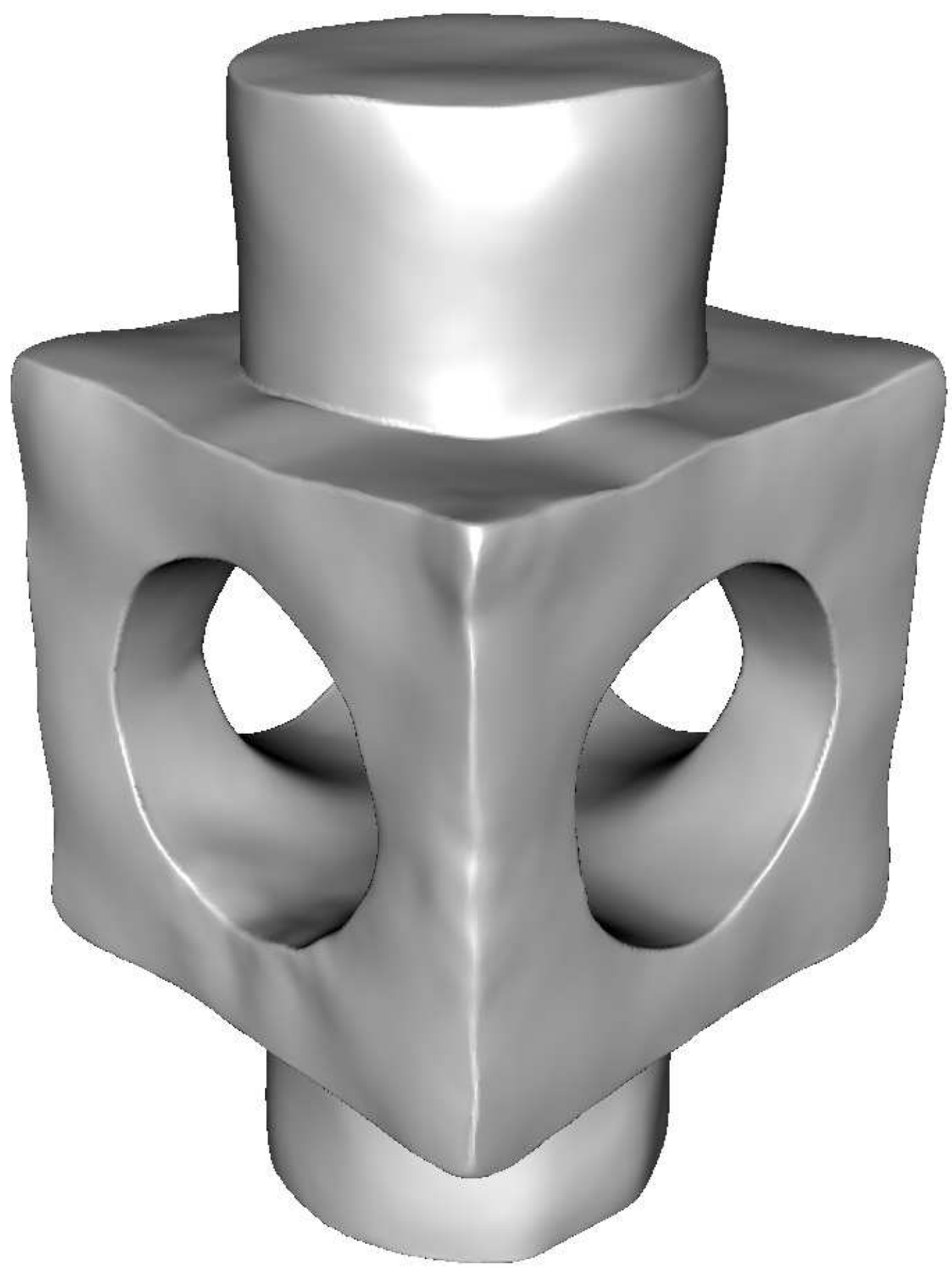}%
	\includegraphics[width=0.122\linewidth]{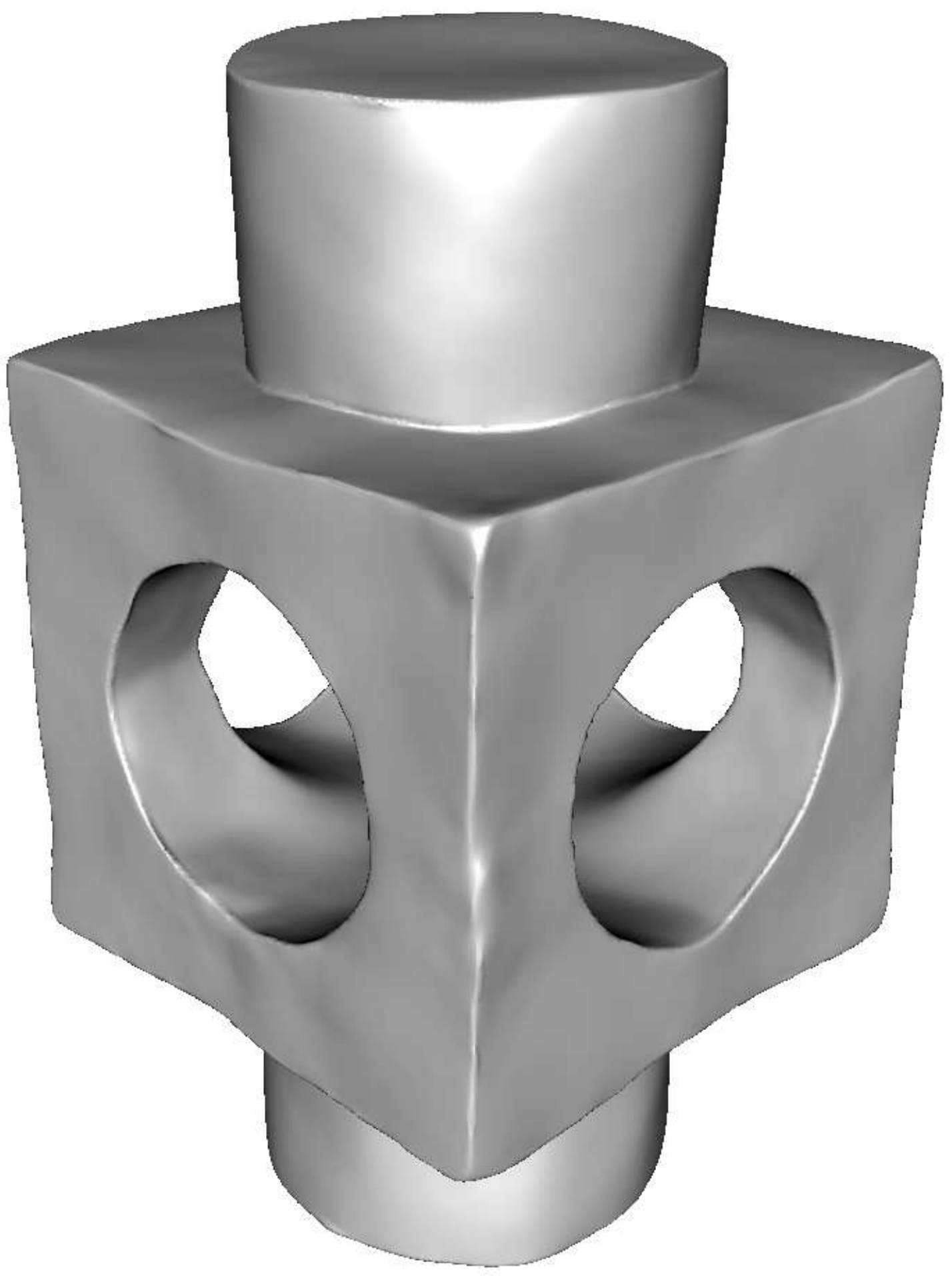}%
	\includegraphics[width=0.122\linewidth]{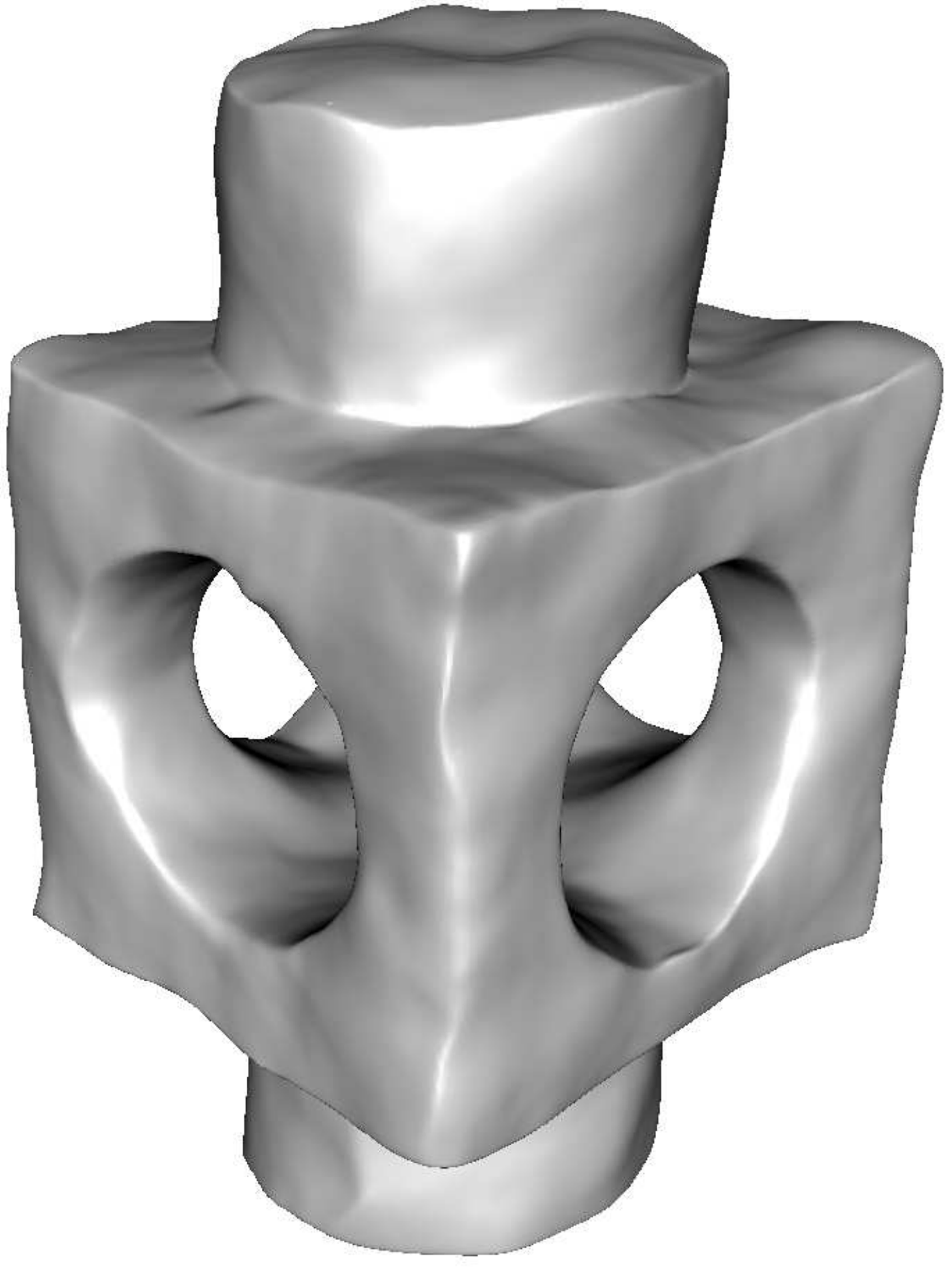}%
	\includegraphics[width=0.122\linewidth]{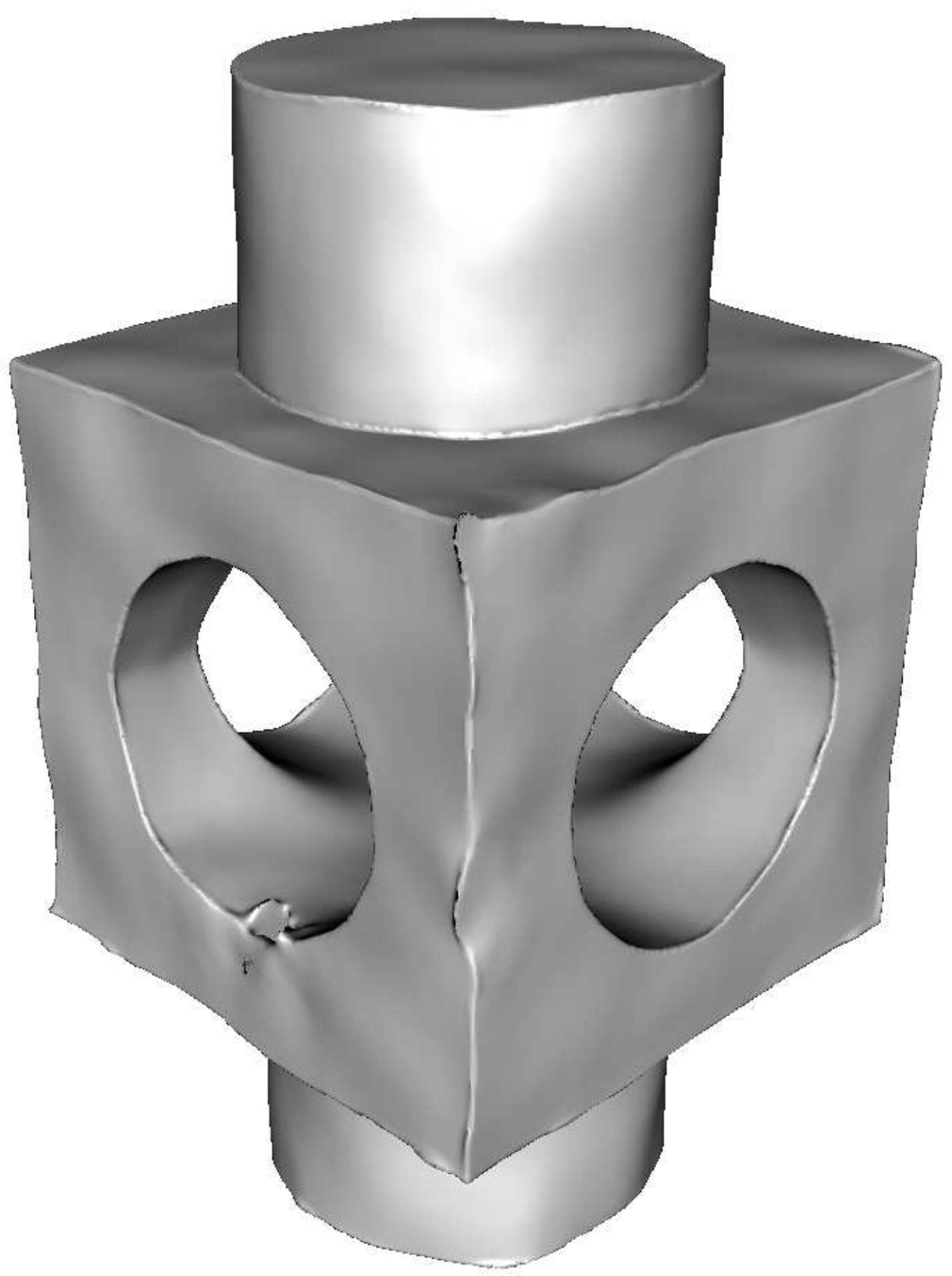}%
	\includegraphics[width=0.122\linewidth]{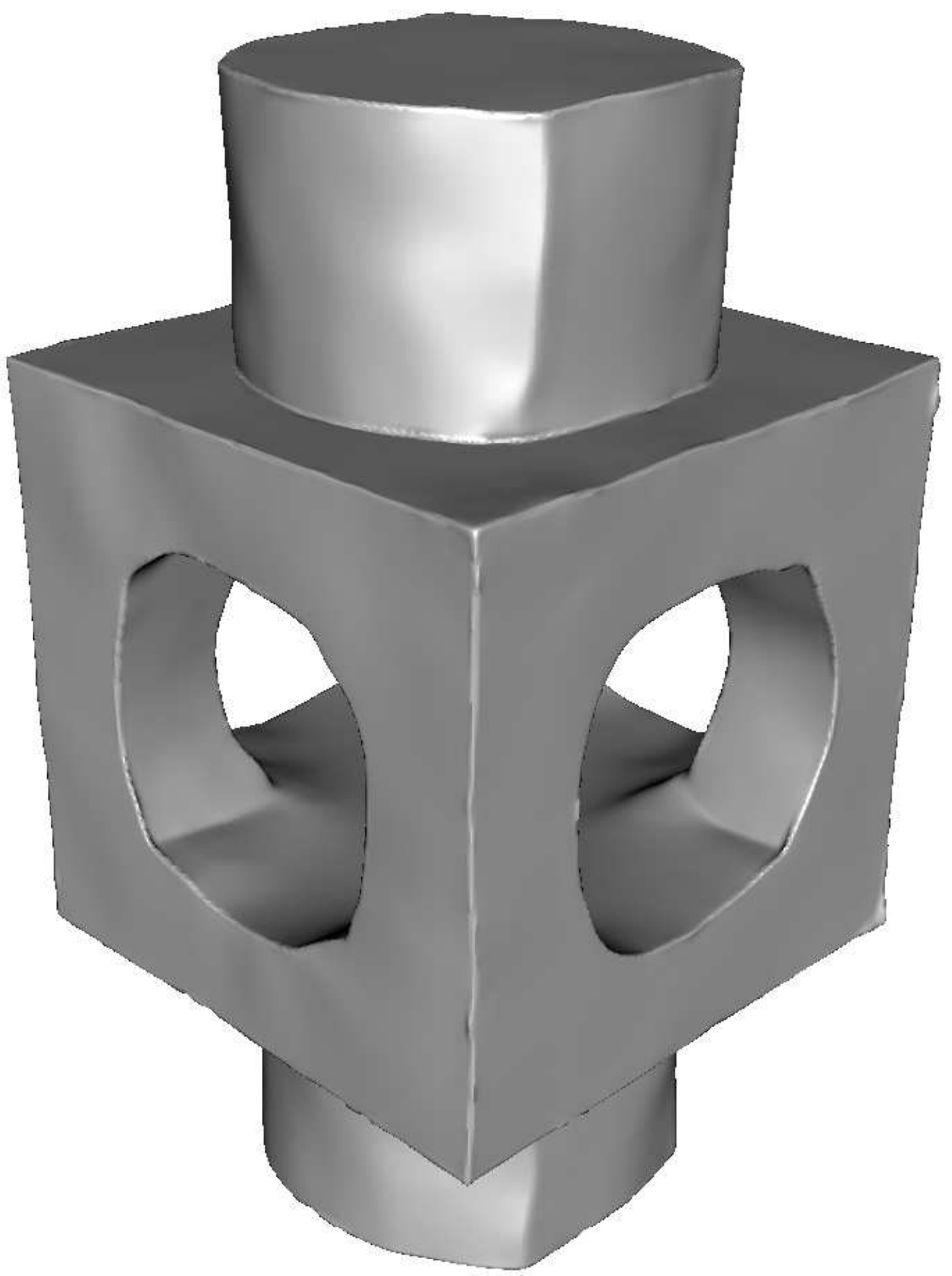}%
	\includegraphics[width=0.122\linewidth]{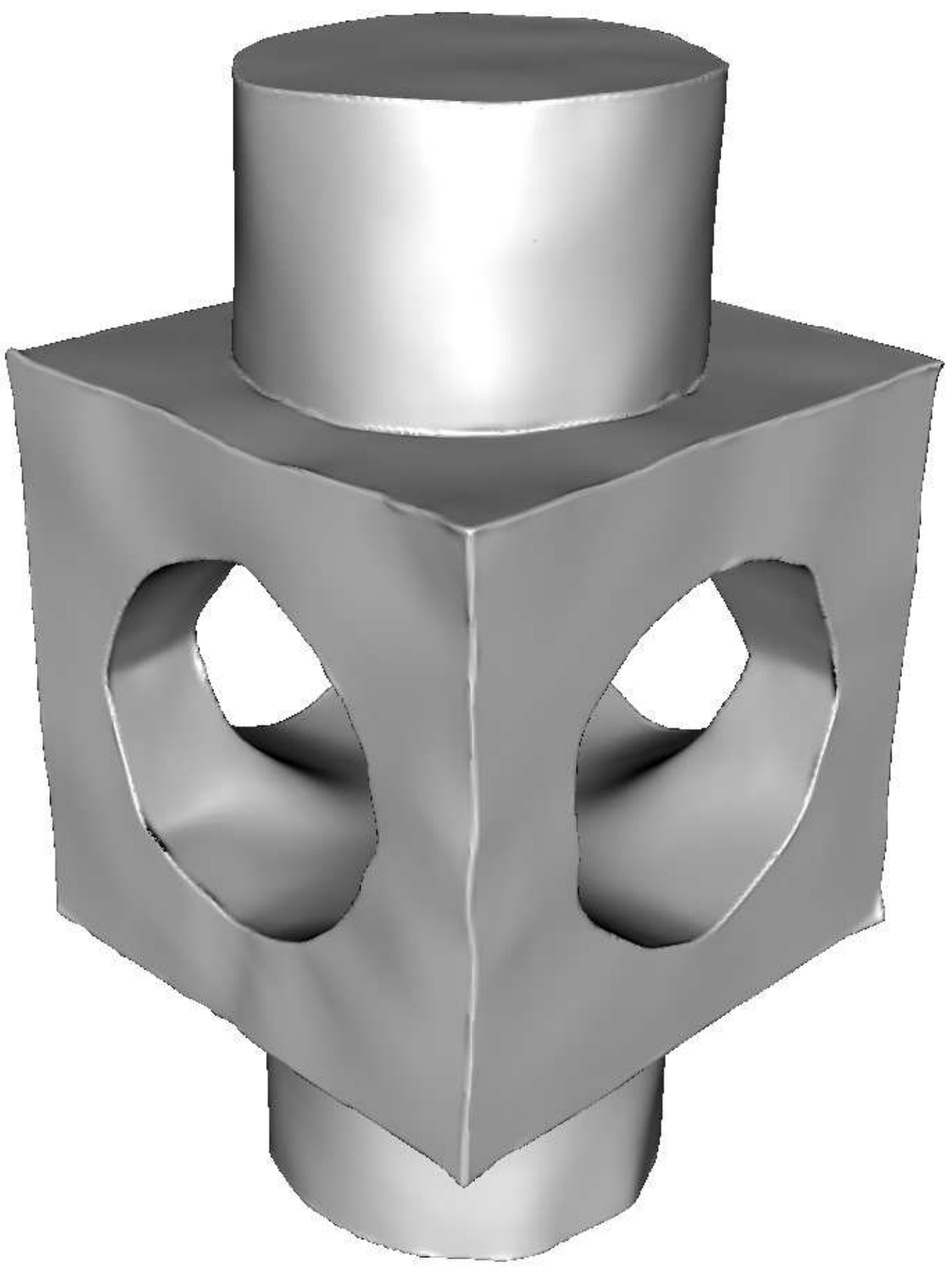}
	
	\includegraphics[width=0.122\linewidth]{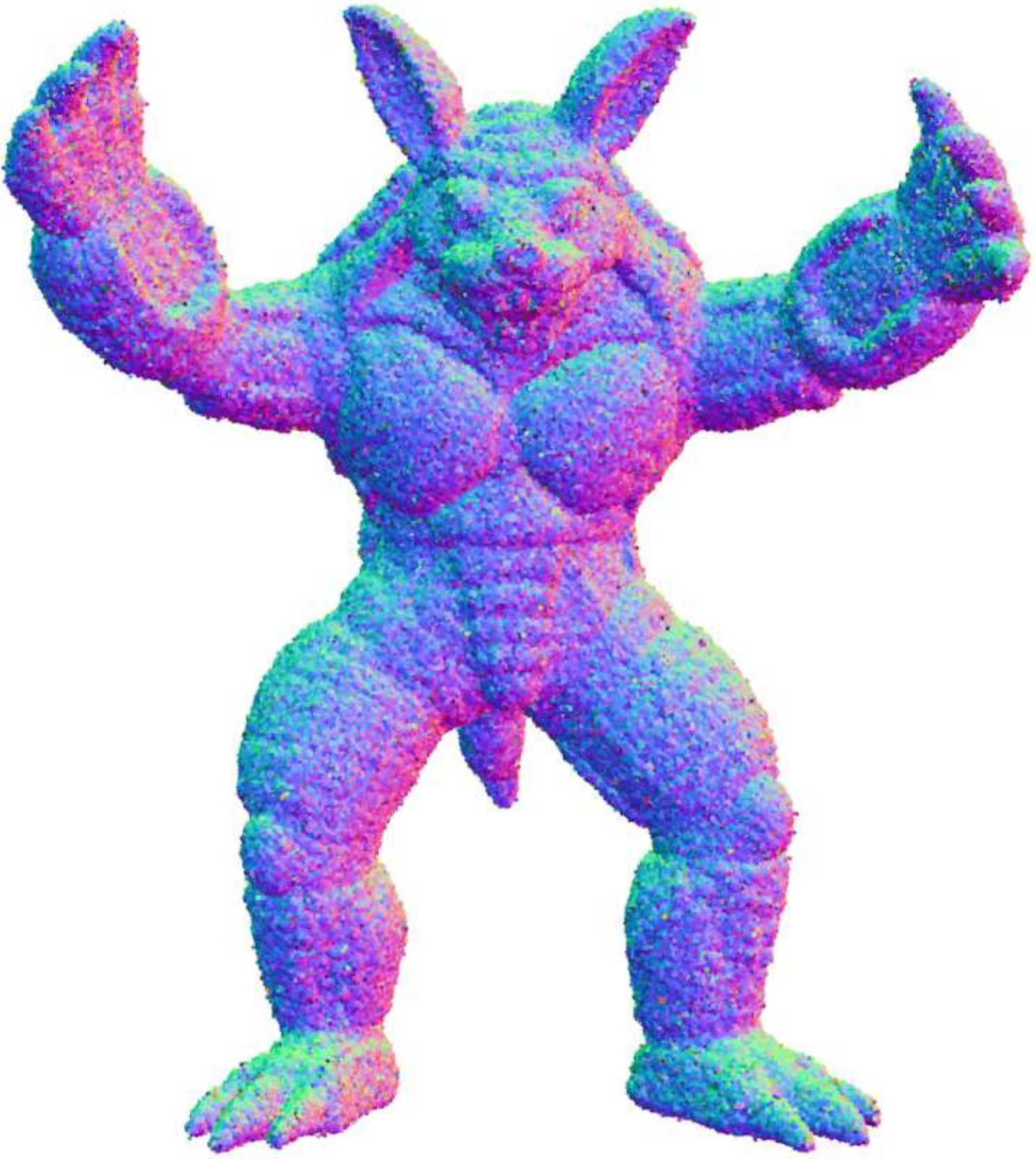}%
	\includegraphics[width=0.122\linewidth]{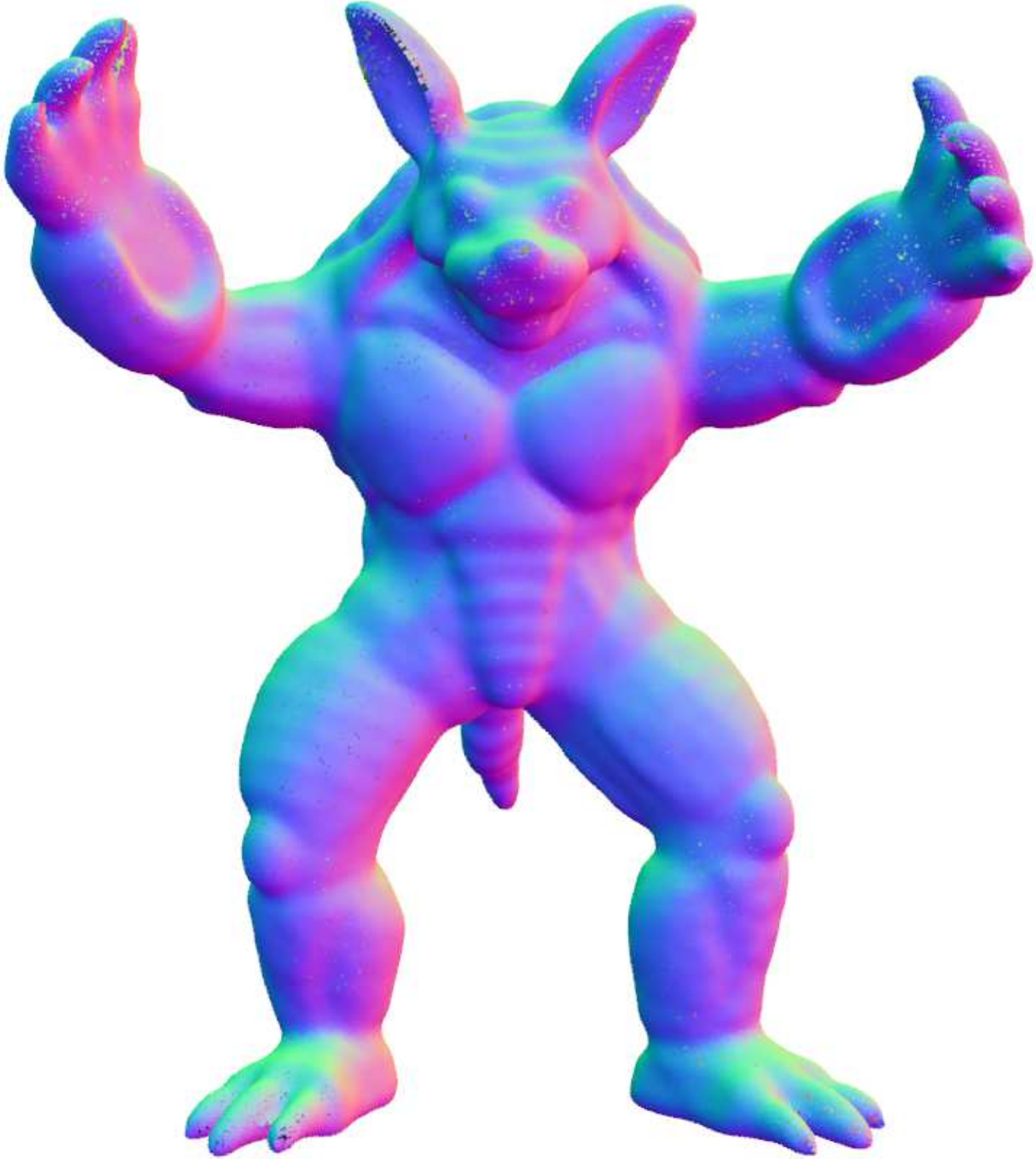}%
	\includegraphics[width=0.122\linewidth]{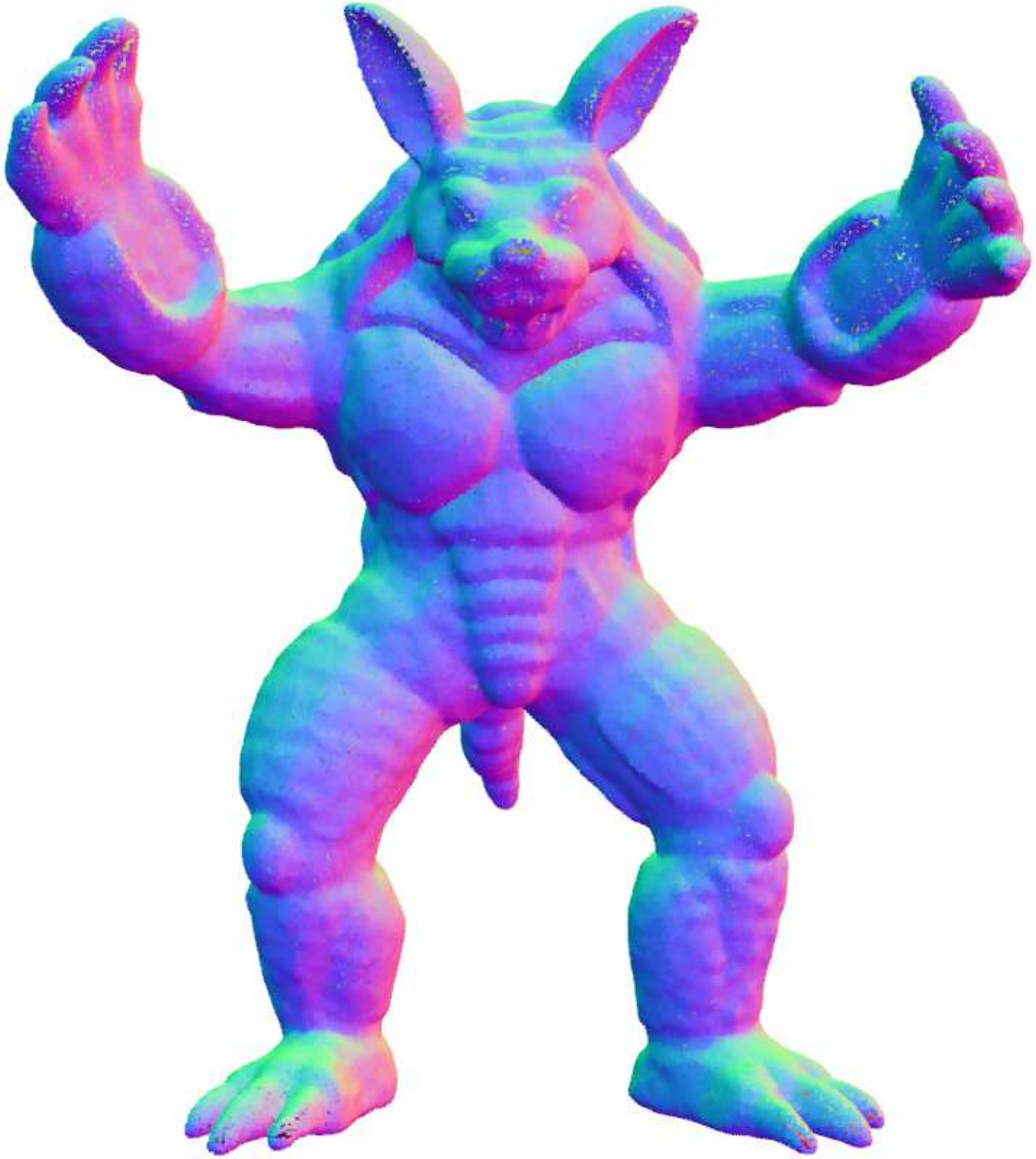}%
	\includegraphics[width=0.122\linewidth]{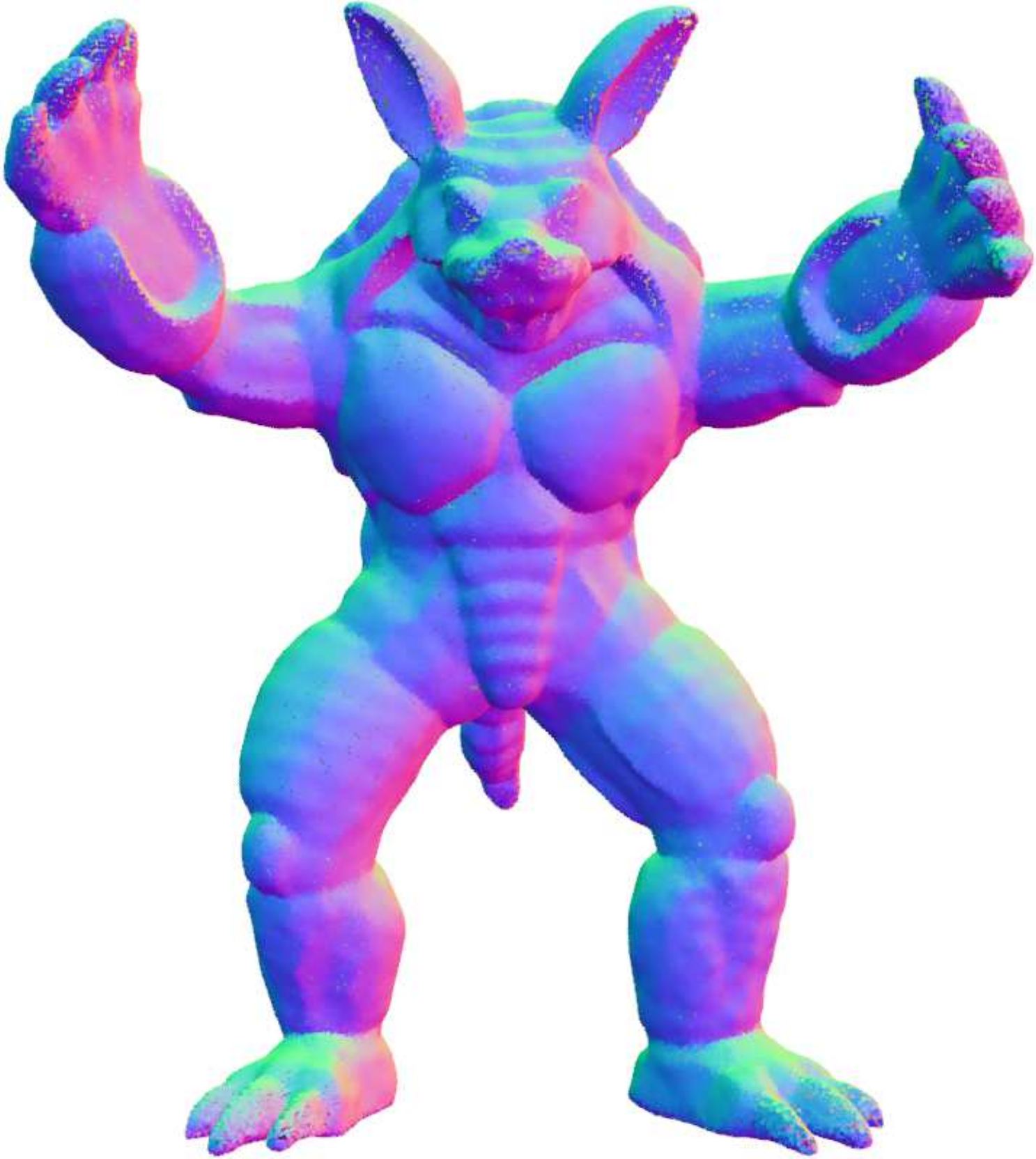}%
	\includegraphics[width=0.122\linewidth]{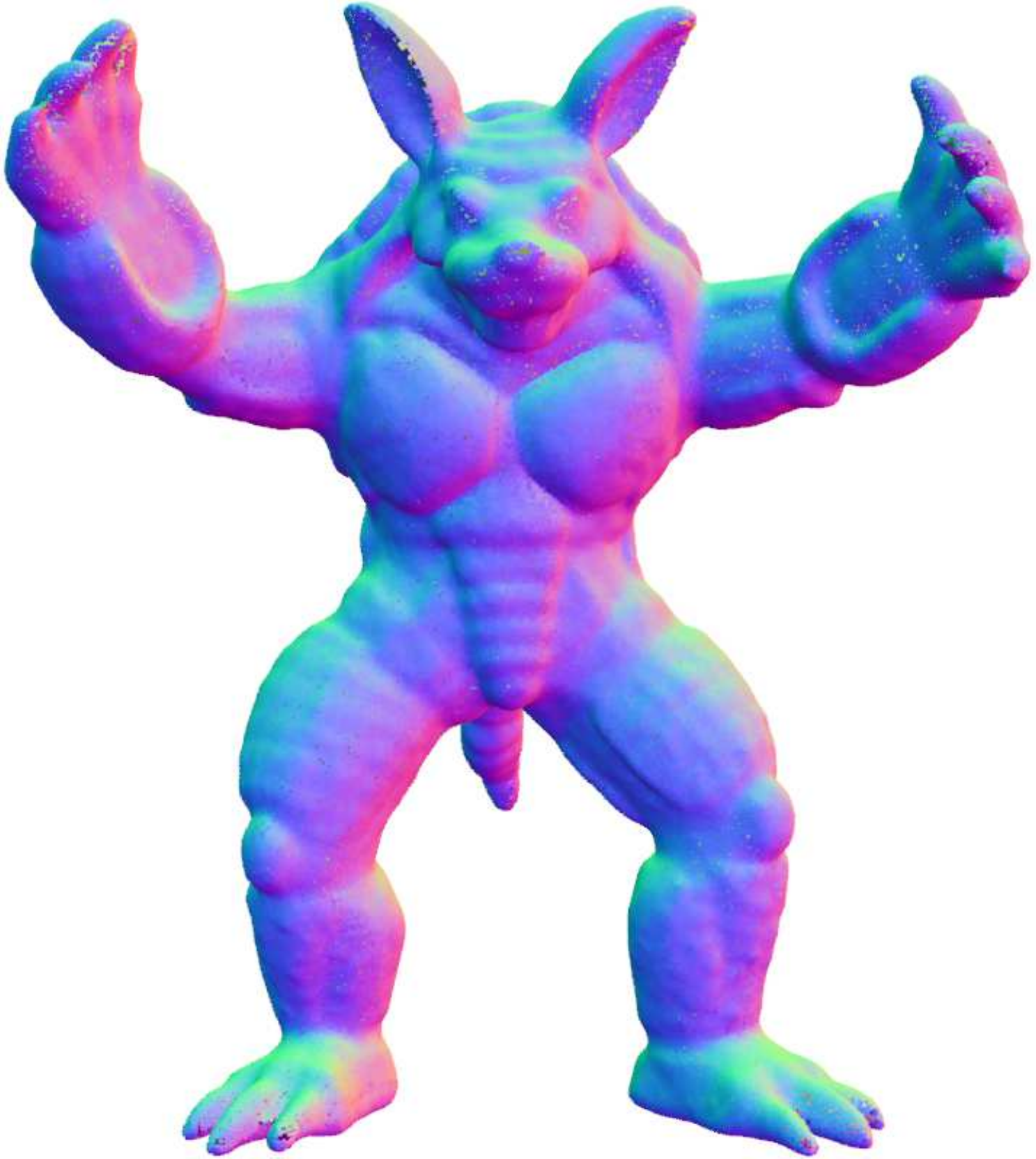}%
	\includegraphics[width=0.122\linewidth]{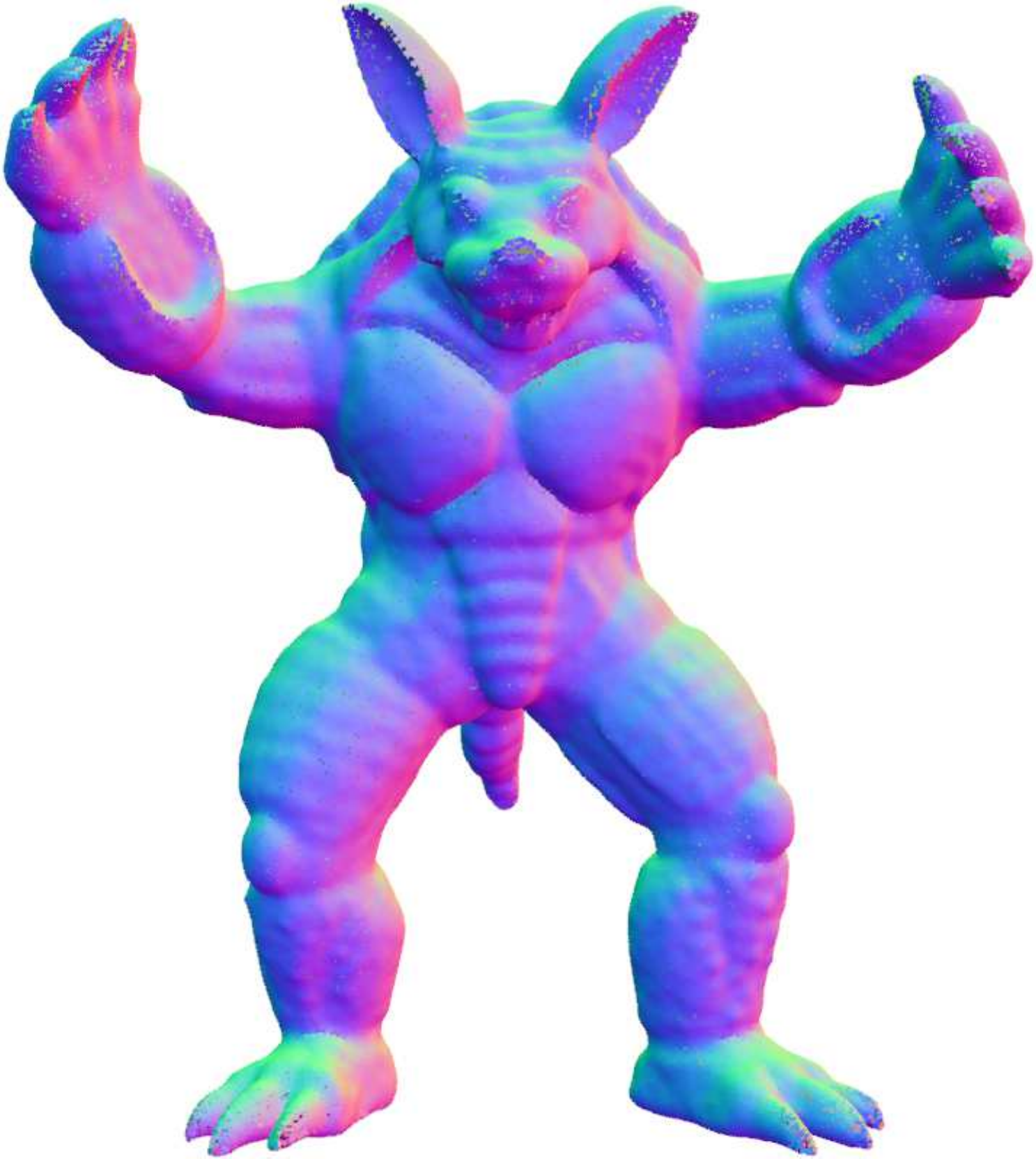}%
	\includegraphics[width=0.122\linewidth]{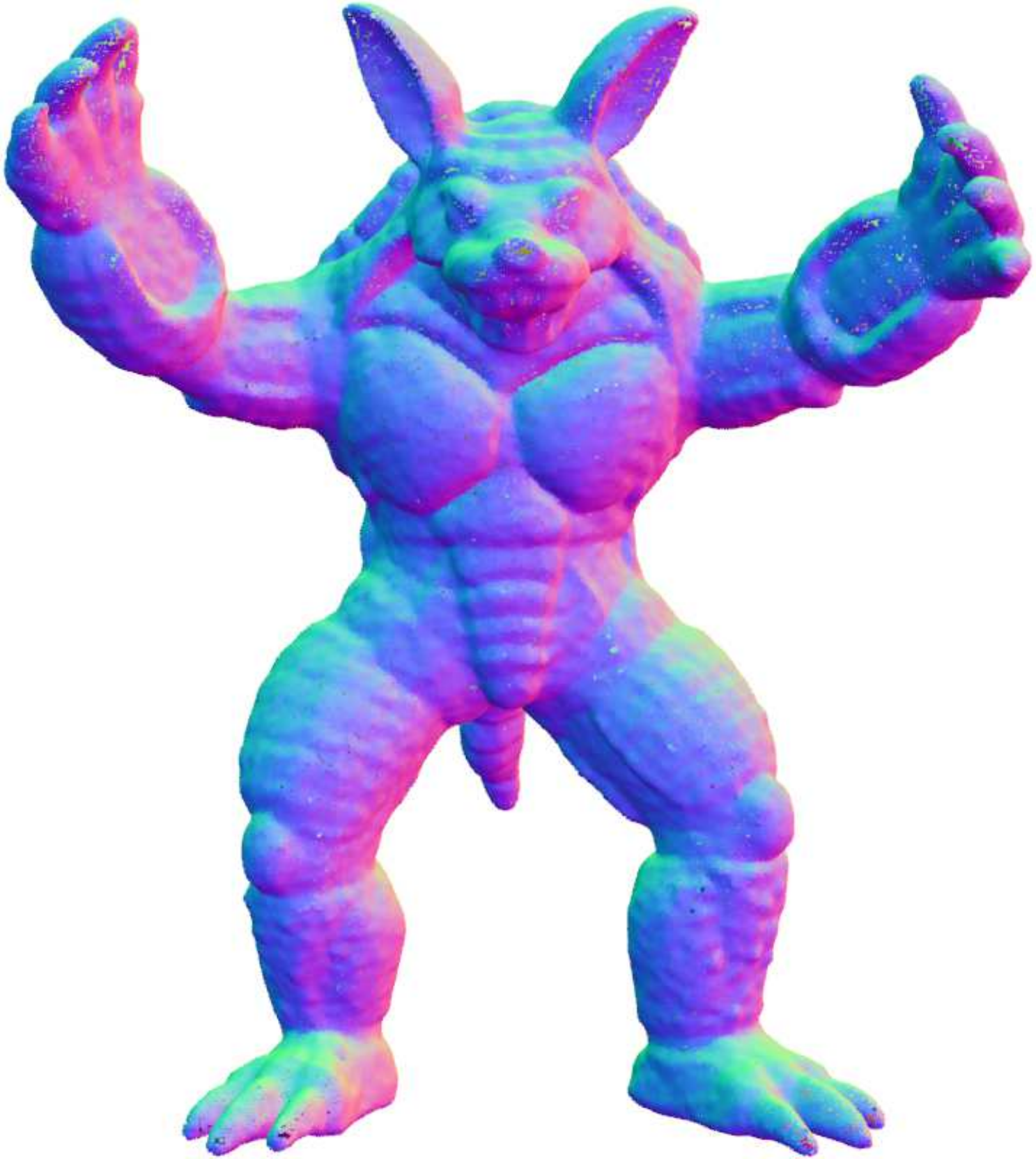}%
	\includegraphics[width=0.122\linewidth]{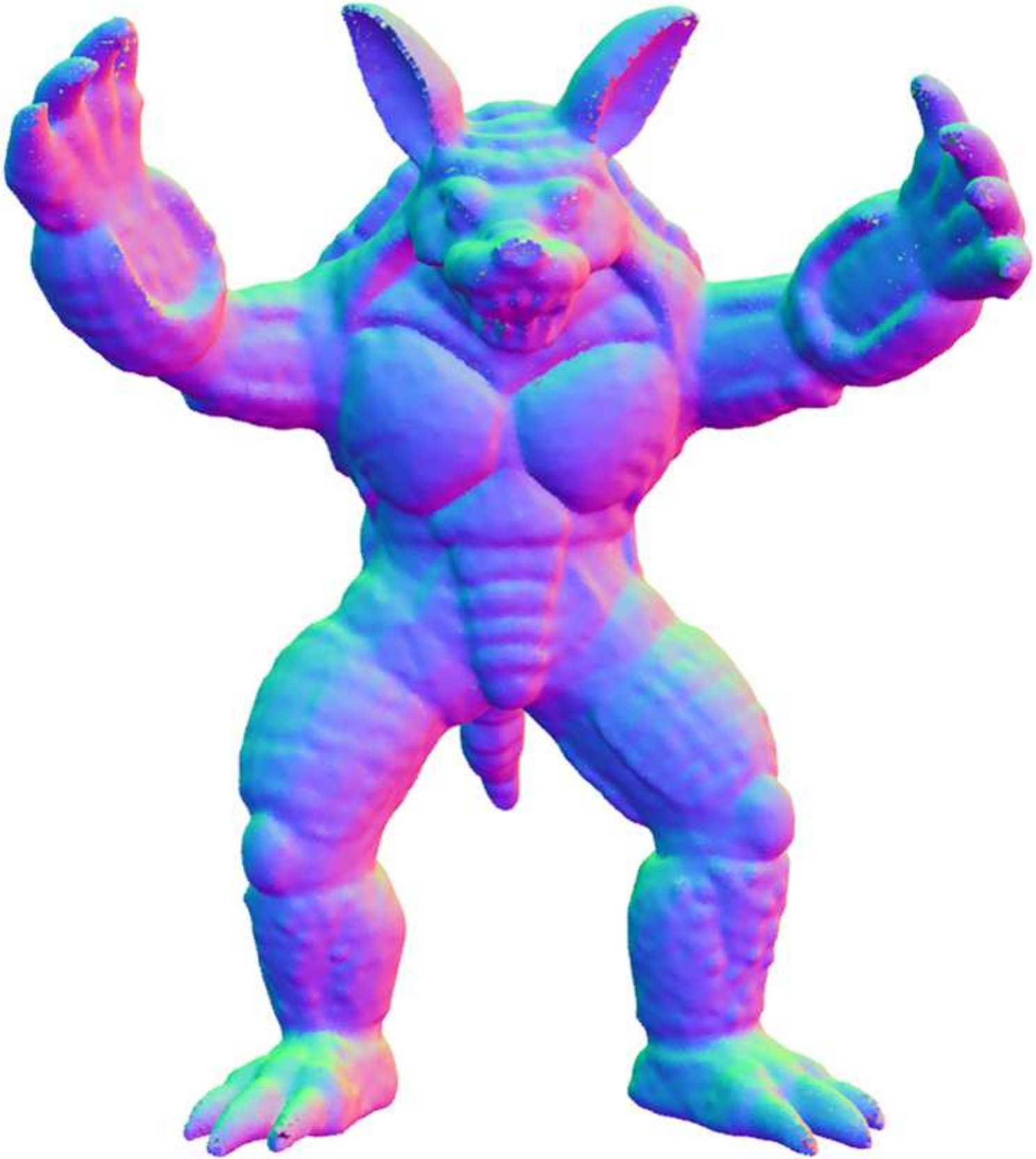}
	
	\vspace{-3pt}
	\subfigure[Noisy input \& GT mesh]{\label{fig:app:armadillo:gt}\includegraphics[width=0.122\linewidth]{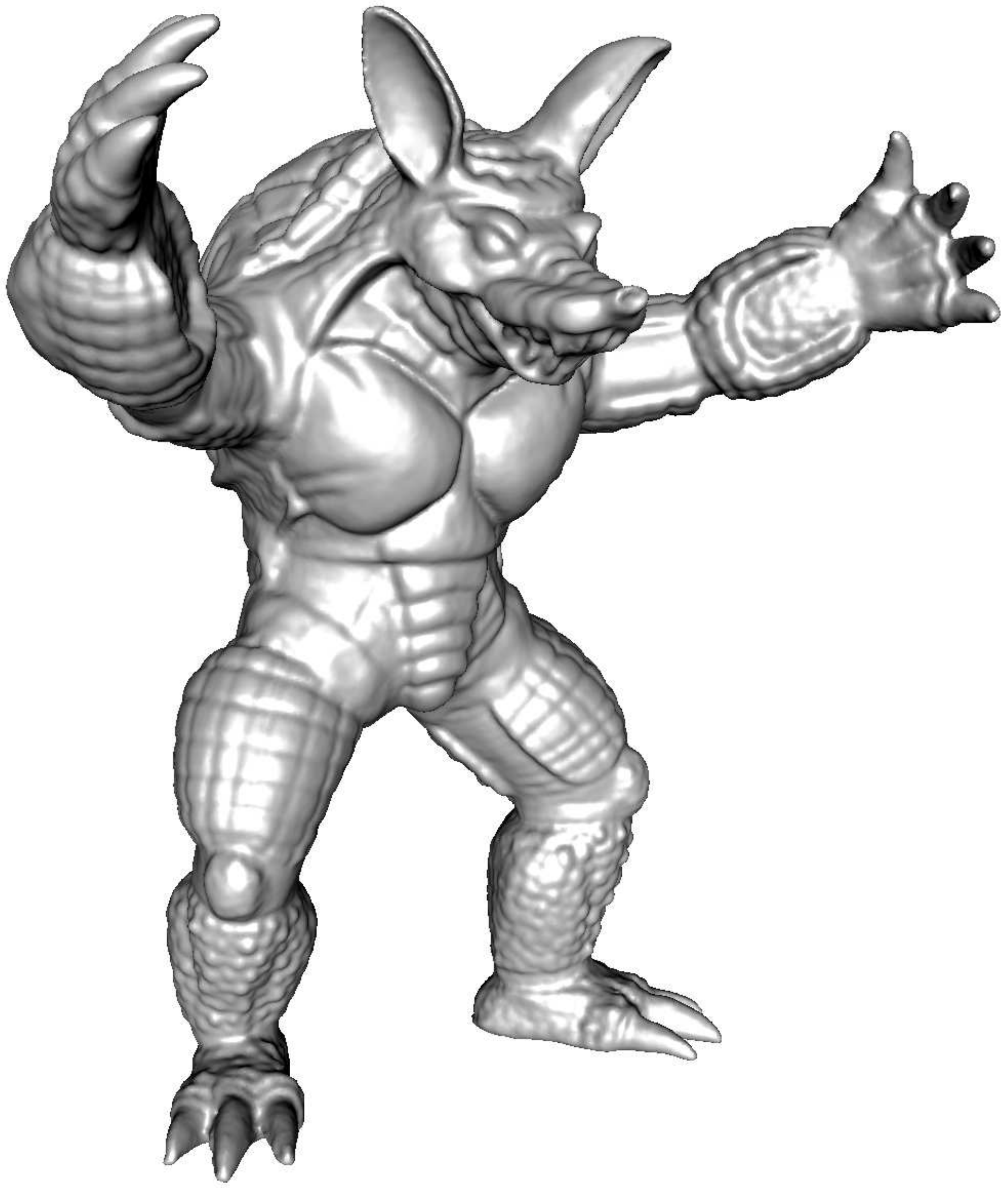}}%
	\subfigure[PCA \cite{hoppe1992surface}]{\label{fig:app:armadillo:pca}\includegraphics[width=0.122\linewidth]{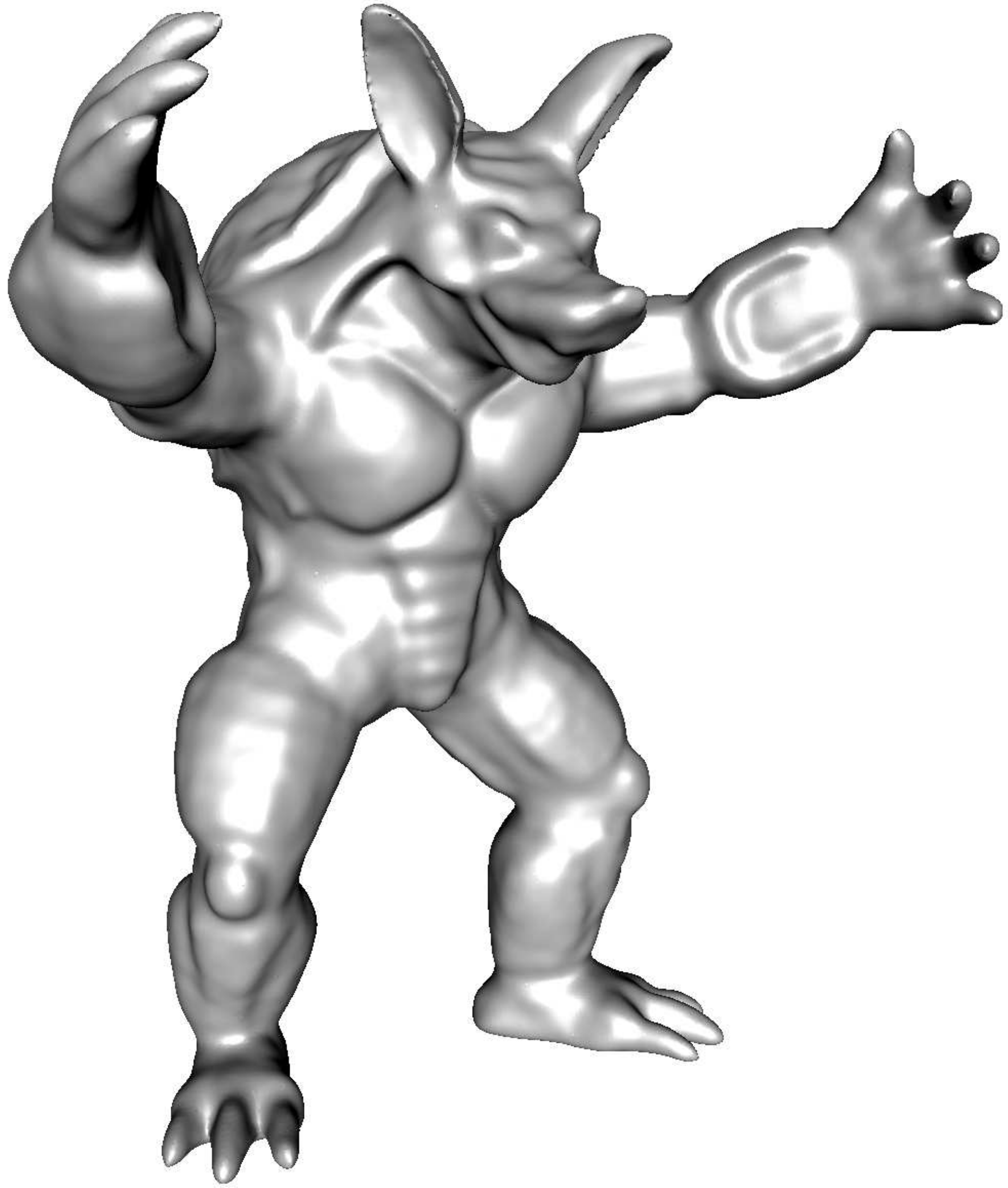}}%
	\subfigure[HF \cite{boulch2012fast}]{\label{fig:app:armadillo:HF}\includegraphics[width=0.122\linewidth]{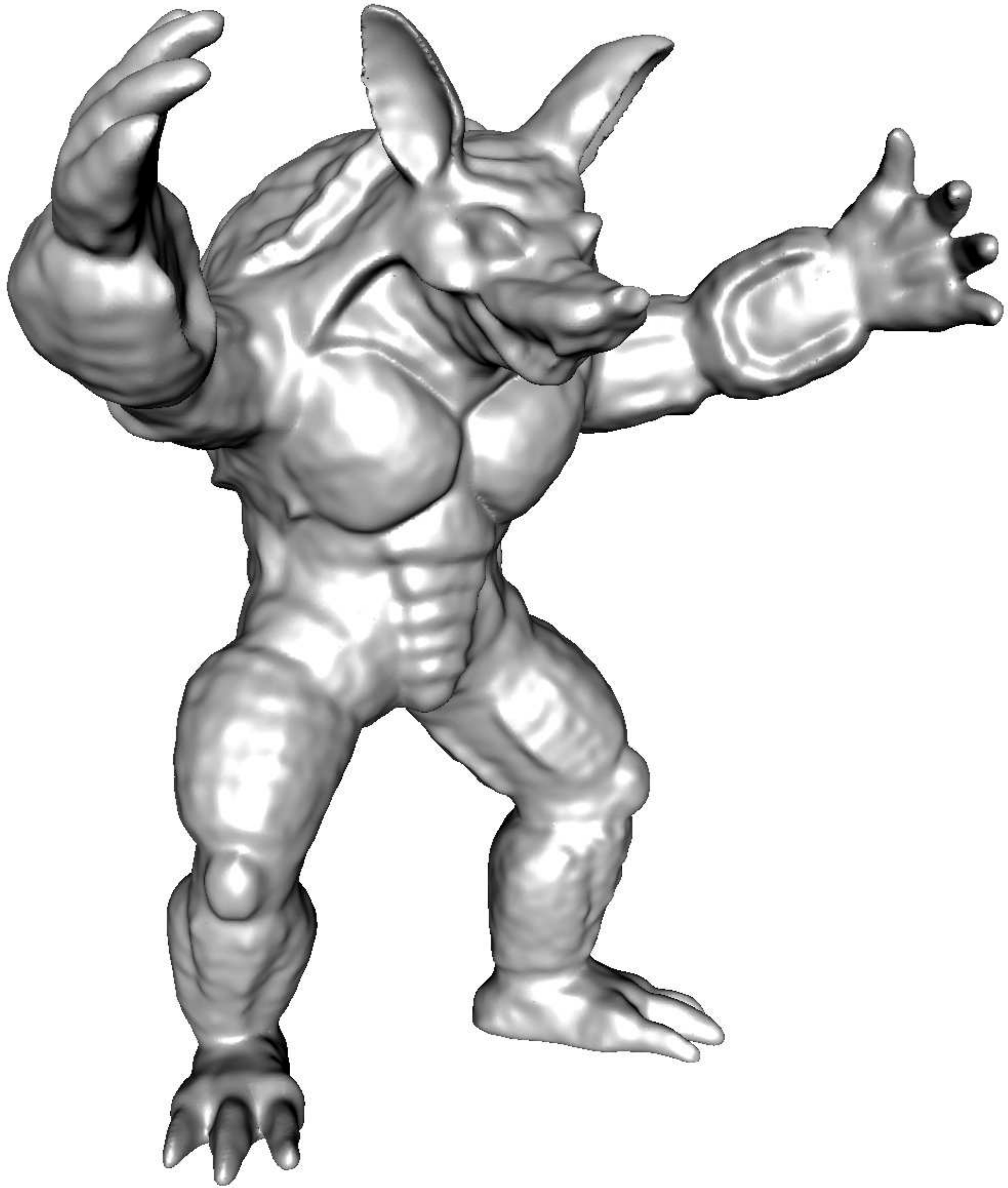}}%
	\subfigure[HoughCNN \cite{boulch2016deep}]{\label{fig:app:armadillo:HoughCNN}\includegraphics[width=0.122\linewidth]{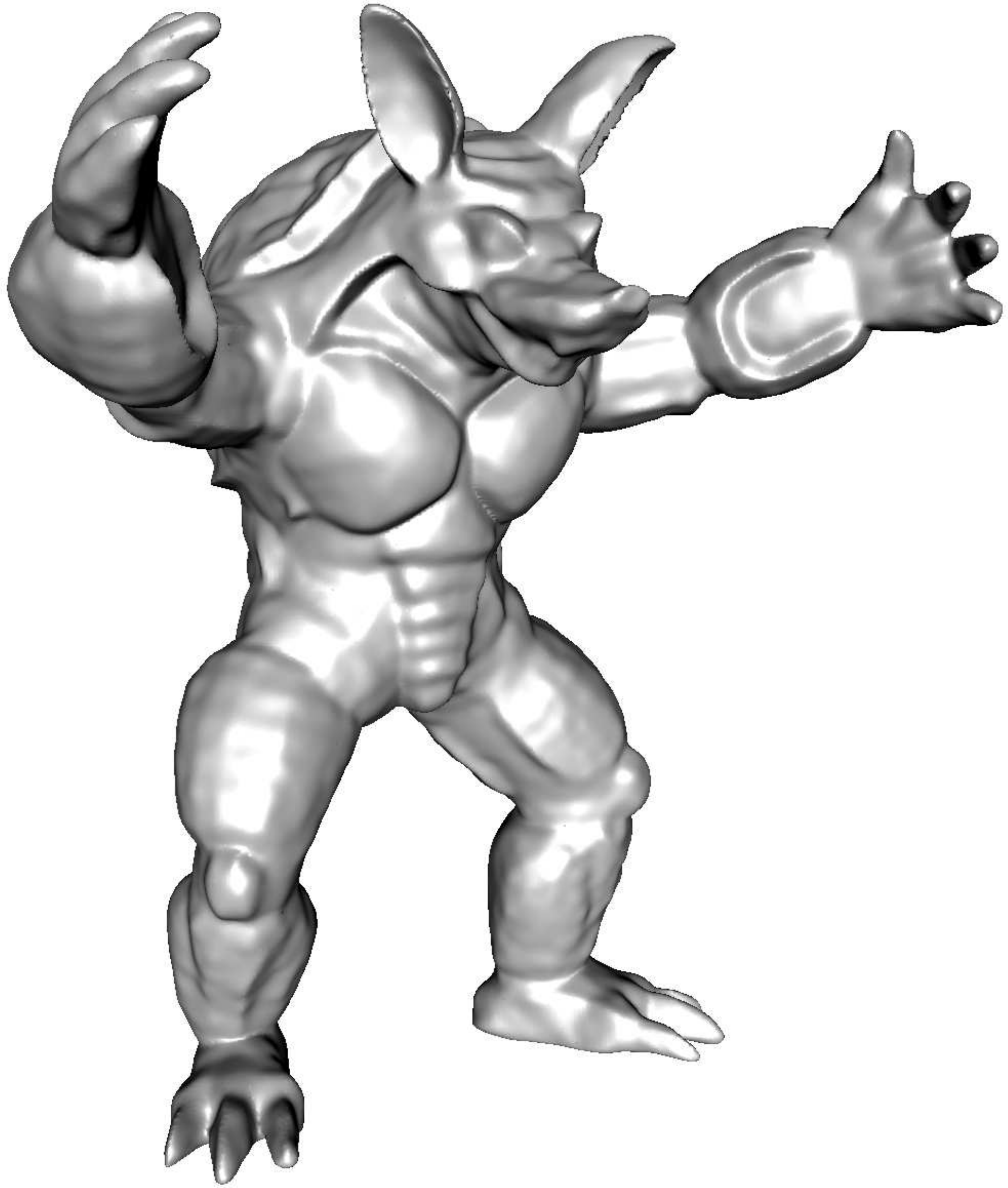}}%
	\subfigure[PCPNet \cite{GuerreroEtAl:PCPNet:EG:2018}]{\label{fig:app:armadillo:pcpnet}\includegraphics[width=0.122\linewidth]{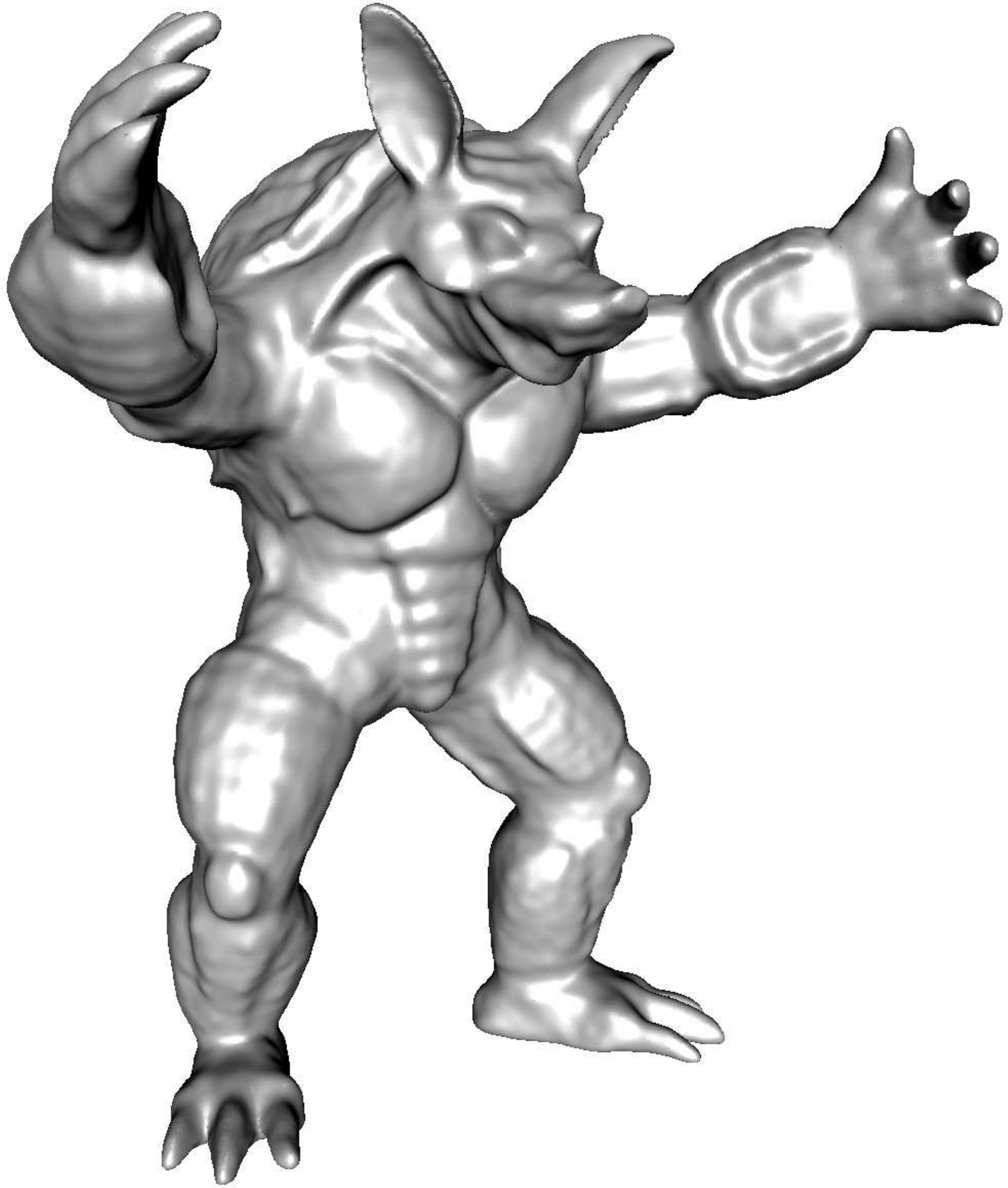}}%
	\subfigure[PCV \cite{zhang2018multi}]{\label{fig:app:armadillo:PCV}\includegraphics[width=0.122\linewidth]{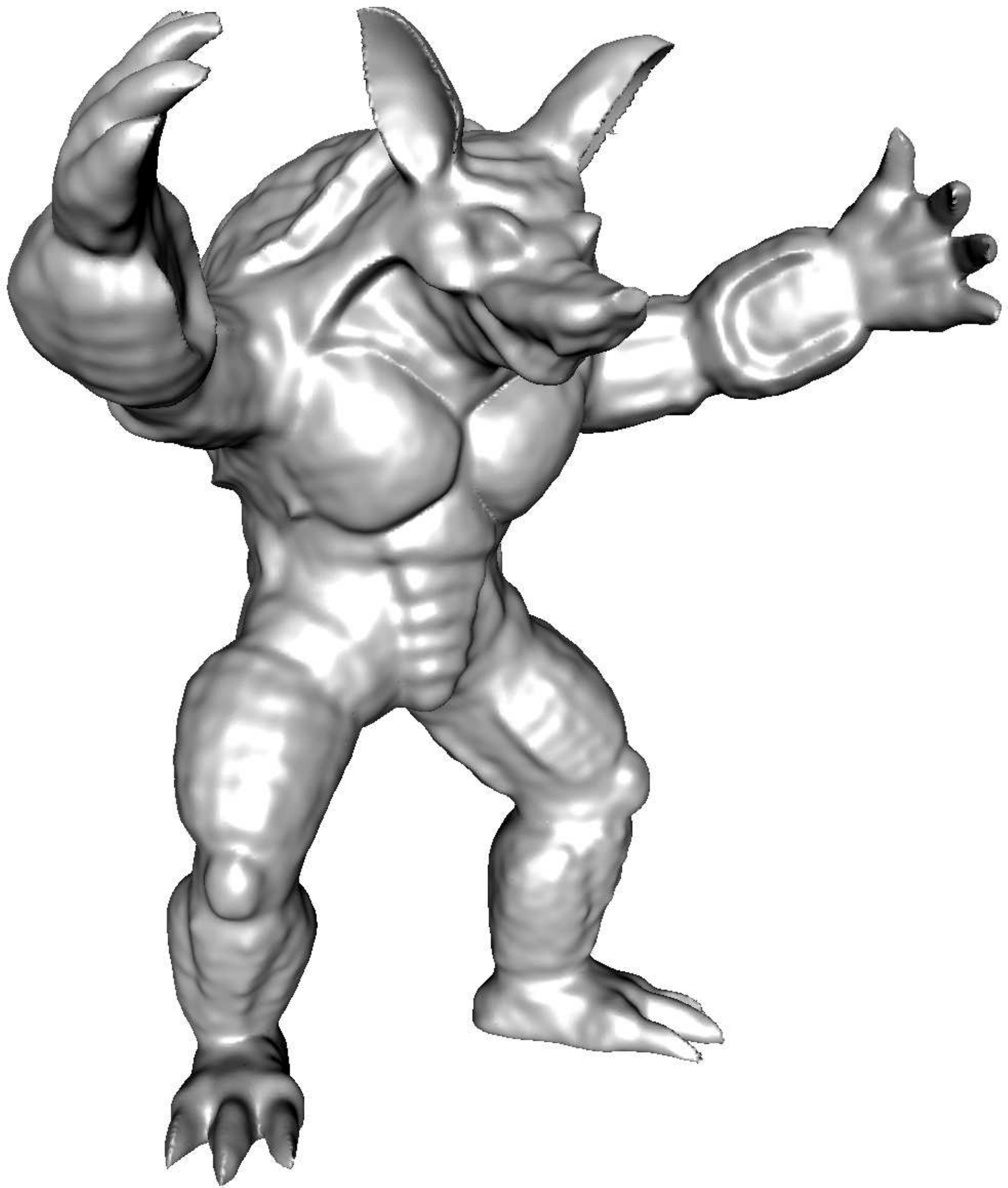}}%
	\subfigure[Nesti-Net \cite{Ben-ShabatLF19}]{\label{fig:app:armadillo:nesti}\includegraphics[width=0.122\linewidth]{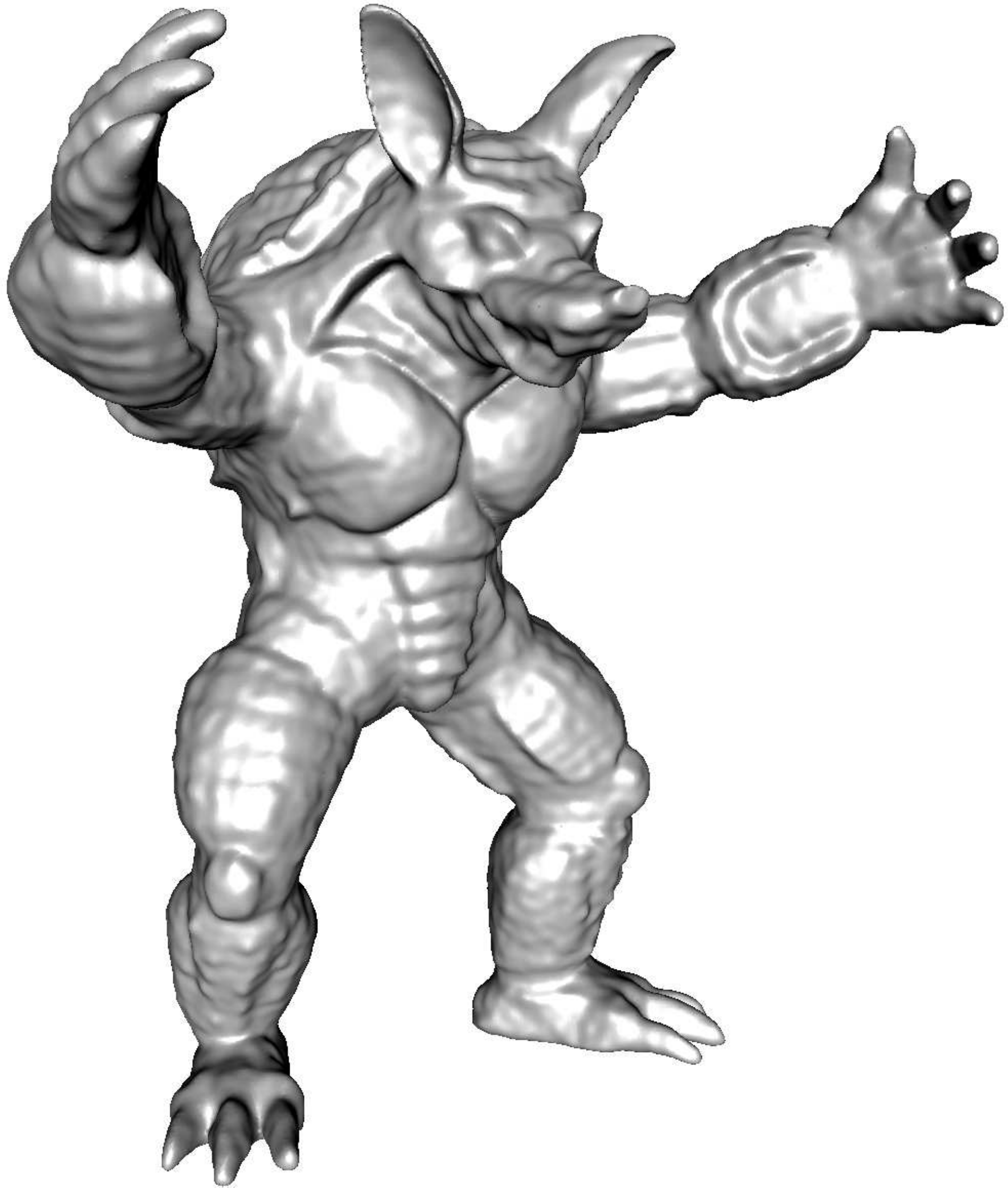}}%
	\subfigure[Ours]{\label{fig:app:armadillo:ours}\includegraphics[width=0.122\linewidth]{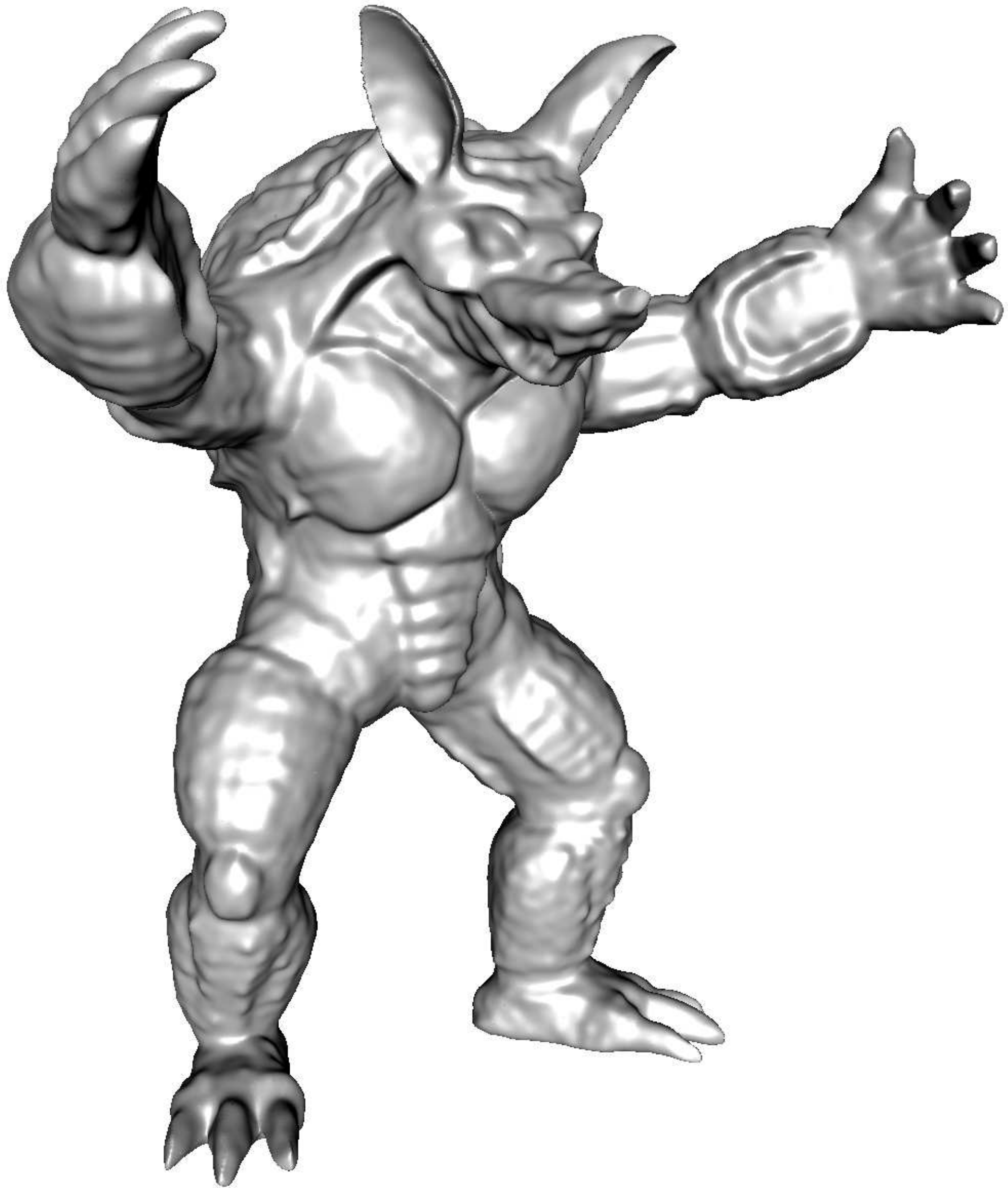}}
	
	\caption{We prove that our method helps produce the best results in several applications. The 1-st and 3-rd rows show the comparison of point cloud denoising results obtained by our update algorithm. The 2-nd and 4-th rows are the corresponding surface reconstruction results. Our method produces the most faithful results on sharp edges as shown in the 1-st and 2-nd rows. And for the model with rich features, our method can better recover geometric details to help produce the reconstruction result.}
	\label{fig:block}
\end{figure*}

\section{Ablation studies}
\label{sec:eval:ablation}
In this section, we explain some of the architecture choices and evaluate the effectiveness of several components in Refine-Net.

\vspace{5pt}
\noindent\textbf{Feature modules.} Compared with similar works \cite{wang2016mesh,GuerreroEtAl:PCPNet:EG:2018}, we introduce two additional feature modules into the Refine-Net framework apart from the main normal feature. Since the local points and HMPs are both generated related to the normal, Refine-Net can utilize the extracted information and improve the initial normal. To demonstrate the advantages of this combination (normals\&points\&HMPs), we design several ablations and evaluate their performances on the synthetic dataset \cite{wang2016mesh} as shown in Tab.~\ref{table:ablation_results}:

$\bullet$ normals - We first consider using normals of each point (from multi-scale bilateral filtering) to recover the ground-truth normal. The filtered normals are concatenated and fed directly to several fully-connected layers (256,128,3). The network setting follows our output module.

$\bullet$ normals\&points - This design solely uses point feature for a refinement. That is, the second refinement is abandoned, and the transformed feature after point module is exported from each branch. The output dimensions are the same as in our Refine-Net.

$\bullet$ normals\&HMPs - Similarly, we test the network which uses normals and HMPs as input. The height-map module outputs a $64 \times 3$ transformation matrix $T$ applied to the normal, which uses the architecture in Tab.~\ref{table:network_architecture} (right). The first refinement (point module) is not included. 


\vspace{5pt}
\noindent\textbf{Connection module choices.} The key idea behind our approach is to learn a transformation term $T$ and to combine the two feature inputs by matrix multiplication. Apart from the adopted choice (weight matrix), we test another two options discussed (rotation matrix and transformation matrix), which are also alternative in other situations. Results are shown in Tab.~\ref{table:ablation_results}. Further, we explore other solutions to combine on two different data inputs:

$\bullet$ concat - The feature module outputs a vector $T$ with the same dimension as $V$. Then, in the connection module of each refinement, the two different feature inputs are concatenated and connected through fully-connected layers (64,64). The output dimensions in each branch are identical to our Refine-Net.

$\bullet$ residual - We replace the matrix $T$ by a single vector with the same size as $V$ ($1 \times 3$ normal), which learns the residual (noise on point normal) between the  input and the ground truth. The intermediate feature $Y$ is obtained by $V$ plus the residual.

Experimental results in Tab.~\ref{table:ablation_results} shows that the multi-feature scheme significantly improves the normal estimation accuracy compared with using single features as input. Note that, our previous work \cite{zhou2020geometry} can be seen as the normals\&HMPs version, and Refine-Net clearly achieves better performance on top of this. On the other hand, the practice that combines two features using matrix multiplication shows its advantages over other learning schemes. The weight matrix choice is better since it captures more information in a high-dimensional vector.

\vspace{5pt}
\noindent\textbf{Initial normals with fitting patch selection.} For the initial normal estimation, we propose a fitting patch selection method to choose one from the multi-scale candidate patches. To quantitatively evaluate this method, we compute the initial normal results without fitting patch selection (simple MFPS), as shown in Tab.~\ref{table:ablation_results}. We can see that the proposed MFPS clearly performs better in all categories, especially on shapes with sharp edges. Without such a selection scheme, it would simply choose the patch with highest score within the neighborhood. However, the most consistent patch may reflect a normal directing the wrong side as illustrated previously in Fig.~\ref{fig:fps}.

\vspace{5pt}
\noindent\textbf{L1 Loss vs L2 loss.} The L1 loss is known to be less sensitive to outliers and better at preserving sharp estimations. In this experiment, we compare the training performances using L1 loss and L2 loss in the task of normal estimation. We use the PCPNet dataset which is categorized according to noise levels and report the rms errors in Tab.~\ref{table:loss}. It shows that the L1 loss performs better on the noise-free point clouds while it is more sensitive to high-level noise. 
Furthermore, in Fig.~\ref{fig:loss2}, we show colormaps reflecting the error distances where a specific color (blue: L1, red: L2) is used to represent points which perform better with L1 or L2 on shapes with different noise levels. The L2 loss performs clearly better on sharp features and tiny details, especially at a lower noise level. Therefore, we choose to use the L2 loss which is more robust to noise and better at recovering normals in challenging regions with suboptimal initial normals.

\begin{table}[t]
	\centering
	\footnotesize
	\setlength{\tabcolsep}{3.5mm}
	\caption{Comparison of the L1 and L2 losses on the PCPNet dataset \cite{GuerreroEtAl:PCPNet:EG:2018}, evaluated as RMS errors.}
	\begin{tabular}{c|cccc} 
		\toprule[1pt]
		& \multicolumn{4}{c}{Noise $\delta$}   \\
		& None & 0.0012 & 0.006 & 0.012 \\ 
		\midrule[0.3pt]
		\midrule[0.3pt]
		L1 	& 6.05 & 9.58 & 17.13 & 23.50  \\
		L2  & 6.27 & 9.18 & 16.59 & 22.57  \\
		\bottomrule[1pt]
	\end{tabular}
	\label{table:loss}
\end{table}

\section{Learning to refine normals}
\label{sec:insight}
To achieve a better understanding of Refine-Net, we provide two insights exploring deeper into the proposed model.

\vspace{5pt}
\noindent\textbf{Learning to refine normals.} Refine-Net is a generic normal refinement framework to ``refine" initial normals. When inputting better initial  estimations, it is expected to also achieve better final normal results. However, we find it is not always true. For example, it shows in Tab.~\ref{table:synthetic_result} and Tab. \ref{table:synthetic_PGP} that combining Refine-Net with MFPS achieves better results than combining Refine-Net with Nesti-Net, though Nesti-Net has better initial normals than MFPS in some of the categories.

In order to explain such phenomena, we explore deeper to see how Refine-Net improves the geometric normals by means of deep learning techniques. In this experiment, we train our full pipeline (MFPS + Refine-Net) and Nesti-Net \cite{Ben-ShabatLF19} on the synthetic dataset from Wang et al. \cite{wang2016mesh}. Then, the Nesti-Net normal results are further refined by our Refine-Net (see Fig.~\ref{fig:insight}(b)). By absorbing prior geometric knowledge, initial normals computed by our MFPS can be more effective for detecting sharp edges and details, while they may also reflect discontinuity of the smooth surface caused by high-level noise (see Fig.~\ref{fig:insight}(c)). On the other hand, Nesti-Net tends to produce piecewisely smooth normals, representing general directions of the underlying surfaces. Refine-Net can polish the initial normals from different methods while the improvements can be better on the MFPS normals (see Fig.~\ref{fig:insight}(e)), by recovering the sharp edges (preserving details) and removing perturbations in the flat areas (denoising). Further, we show another result of refined normals using only point features in Fig.~\ref{fig:insight}(d). We can see that using a single feature module can already produce decent denoised normals on noisy point clouds and Refine-Net with multiple feature inputs has better performances on sharp features.

\vspace{5pt}
\noindent\textbf{Cluster-based scheme.} Inspired by \cite{wang2016mesh}, Refine-Net uses a cluster-based scheme to promote its learning ability, which divides input samples into $K_c$ clusters before training. In all our experiments, including the variants of the proposed network, we set $K_c = 4$ by default. A larger number of clusters slightly improves the performance but heavily expands the network (see Fig.~\ref{fig:cluster} left). This clustering scheme can detect similar features among all training samples especially for points on geometric features that are hard to process (see Fig.~\ref{fig:cluster} right). Thus, the network can focus on a specific kind of input data.

\section{Application}
\label{sec:application}
As known, well-estimated normals can boost the effect of many point cloud processing tasks, like denoising and surface reconstruction. To further verify the advantages of our method, we show denoising results using the same point updating algorithm, yet based on the normal results estimated by different methods.

Denote that $\mathcal N_i$ is the ball neighborhood of $p_i$, we here provide an algorithm to update point positions under the guidance of estimated normals, which is able to remove noise and recover sharp features:
\begin{equation}
p_i' = p_i + \gamma_i \sum_{{p}_{j} \in \mathcal N_i} (p_j-p_i)(w_\sigma(n_i,n_j) n_i^T n_i + \lambda n_j^T n_j),
\end{equation}
where $n_i$ is the normal of $p_i$, $w_\sigma(n_i,n_j) = \exp(-\frac{||n_i-n_j||^2}{\sigma ^2})$ is a weight function, $\lambda = 0.5$ is a tradeoff parameter and $\gamma_i$ is the step size set to $\frac{1}{3|\mathcal N_i|}$ by default. To prevent points from accumulating around edges, we keep the neighboring information unchanged in all iterations \cite{lu2020low}. The iteration number is set to 20 in our experiment.

We demonstrate the visual quality of denoising results and their reconstruction results (see Fig. \ref{fig:block}). Our normal results help produce the most faithful denoising and reconstruction results.

\label{SECevaluation}

\section{Conclusion}

In this work, we propose a normal refinement network (Refine-Net) to estimate normals for noisy point clouds, by introducing multiple important feature representations into the refinement steps of the input initial normals. Our Refine-Net involves several feature modules which can capture different geometric information in order to jointly contribute to the normal recovery. By a novel connection module, our method effectively handles features from distinct domains and incorporates them into the produced  normal feature. In addition to the overall architecture of Refine-Net, we propose a multi-scale fitting patch selection (MFPS) scheme to estimate our own initial normals with better geometric supports. 

Refine-Net is a generic normal refinement framework. We show that Refine-Net is able to repair the suboptimal estimations of any initial normal input by recovering sharp edges (preserving details) and removing perturbations (denoising); thus, it can be applied to improve results from other state-of-the-art networks. More feature modules, representing typical geometric properties of the input point clouds, could be potentially developed to further explore the network ability of our framework. Extensive evaluations demonstrate the clear superiority of Refine-Net over the state-of-the-arts on both synthetic and real-scanned datasets. It also shows the promising prospect of applying our method to several downstream geometric tasks in future researches, such as surface reconstruction, consolidation, and semantic segmentation.


\label{SECconclusion}

\vspace{5pt}
\noindent\textbf{Acknowledgements.}
The authors thank the anonymous reviewers for their careful reading and valuable comments.
This work was supported  in part by the National Natural Science Foundation of China (No. 62032011, No. 62172218, No. 61672273), in part by the Free Exploration of Basic Research Project, Local Science and Technology Development Fund Guided by the Central Government of China (No. 2021Szvup060), in part by the Key Program of Jiangsu Provincial Department of Culture and Tourism (No. 20ZD06), in part by the Research Grants Council of the Hong Kong Special Administrative Region, China (No. 15205919), and in part by the Innovation and Technology Fund - Midstream Research Programme for Universities of Innovation and Technology Commission (No. MRP/022/20X).


%

\ifCLASSOPTIONcompsoc
\else
\fi


\ifCLASSOPTIONcaptionsoff
  \newpage
\fi



\bibliographystyle{IEEEtran}
\bibliography{references}
%

%

\begin{IEEEbiography}[{\includegraphics[width=1in,height=1.25in,clip,keepaspectratio]{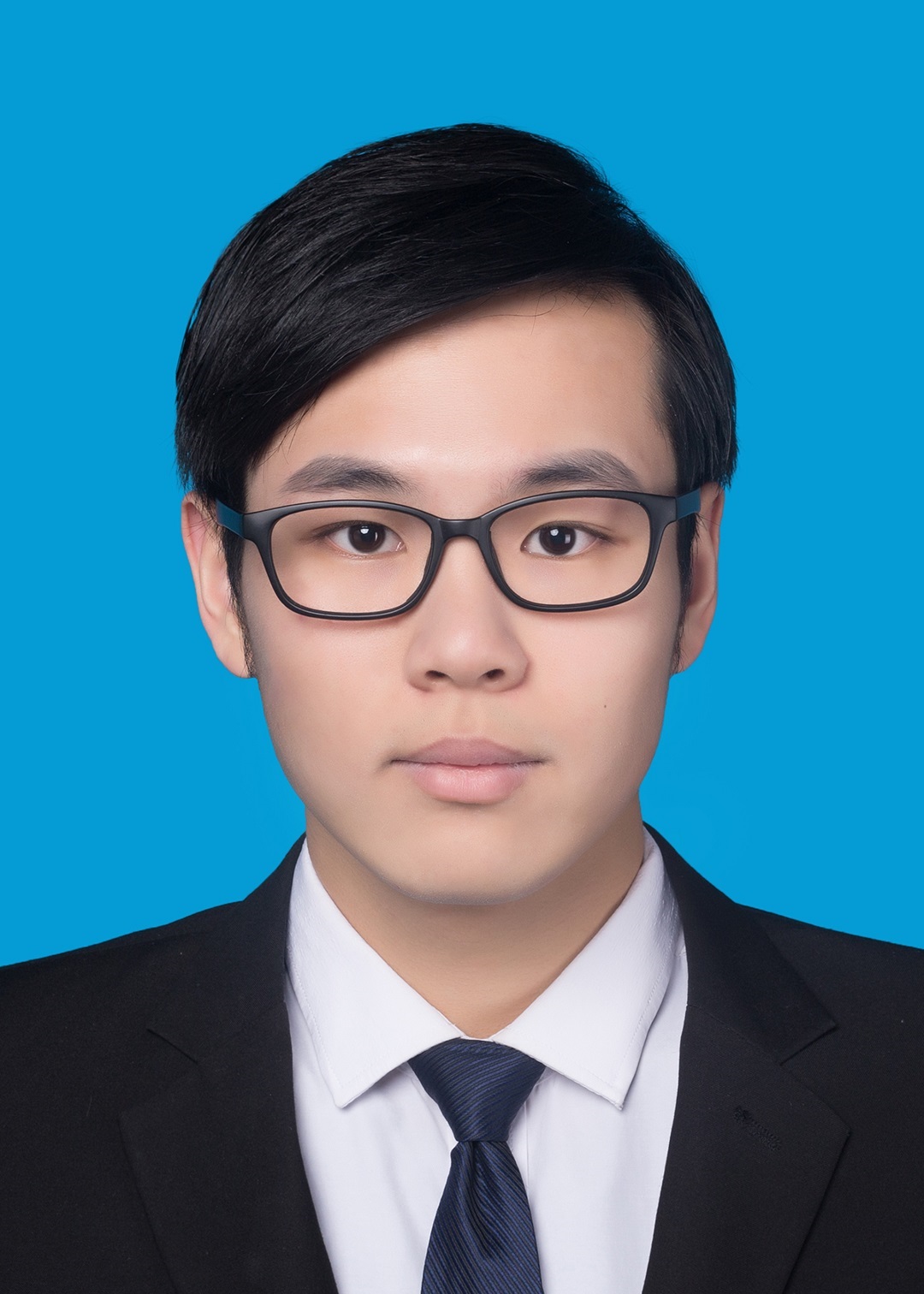}}]{Haoran Zhou}
is a Ph.D. candidate at Nanjing University (NJU). He received his B.Sc. degree from Nanjing University of Aeronautics and Astronautics in 2020. His research interests include computer vision and learning-based geometry processing.
\end{IEEEbiography}

\begin{IEEEbiography}[{\includegraphics[width=1in,height=1.25in,clip,keepaspectratio]{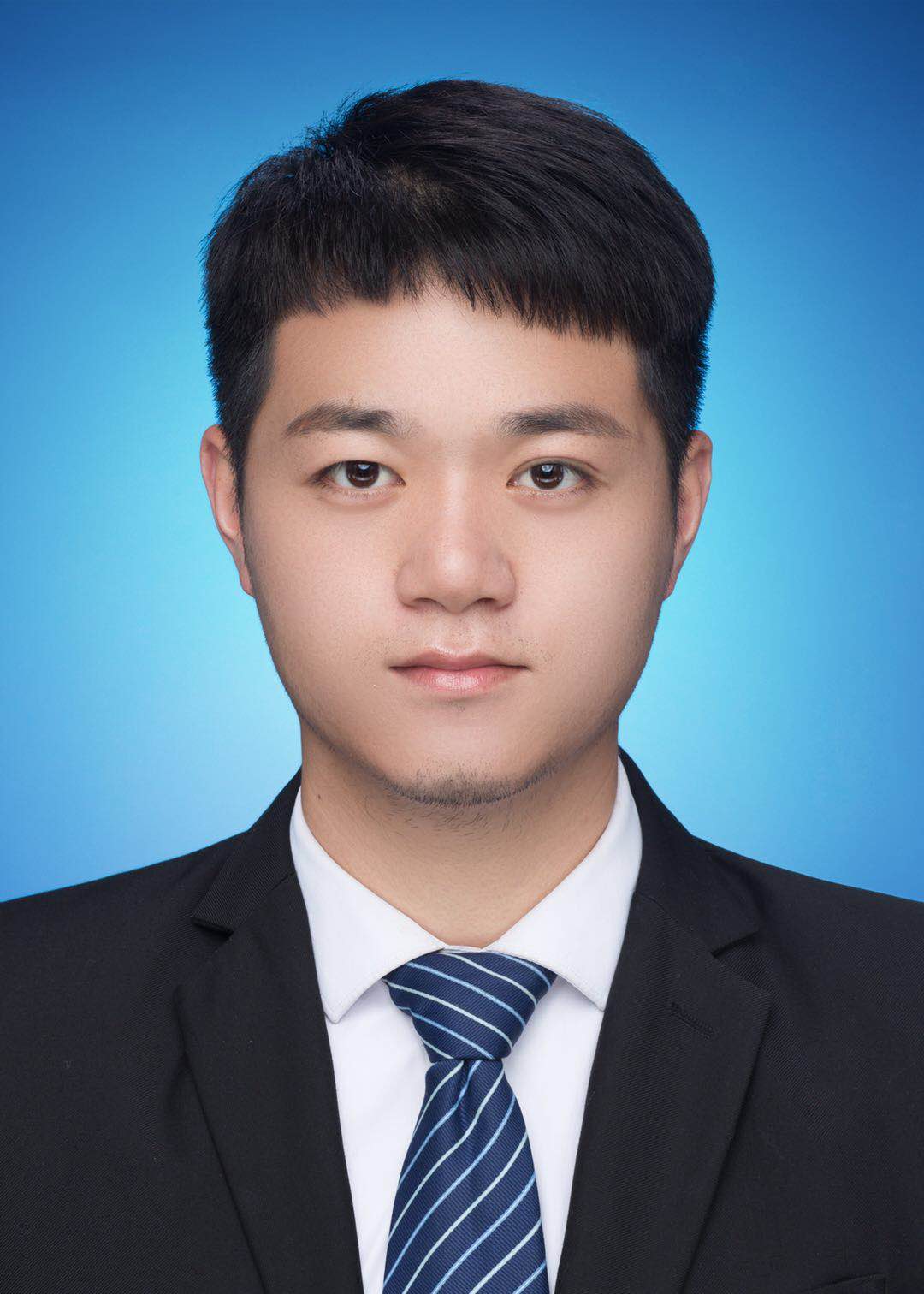}}]{Honghua Chen}
is a Ph.D. candidate at Nanjing
University of Aeronautics and Astronautics (NUAA),
China. He earned his masters degree from
Nanjing Normal University in 2017. His research
interest is learning-based geometry processing.
\end{IEEEbiography}

\begin{IEEEbiography}[{\includegraphics[width=1in,height=1.25in,clip,keepaspectratio]{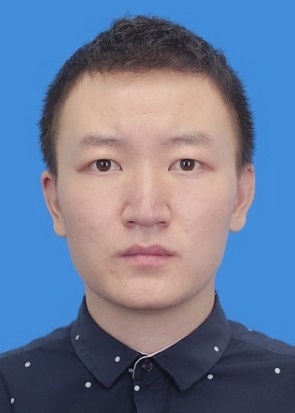}}]{Yingkui Zhang}
is currently a Research Assistant at The Hong Kong Polytechnic University. He received his M.S. degree from Shenzhen Institutes of Advanced Technology, Chinese Academy of Sciences in 2021. His research interests include point cloud processing and 3D vision.
\end{IEEEbiography}

\begin{IEEEbiography}[{\includegraphics[width=1in,height=1.25in,clip,keepaspectratio]{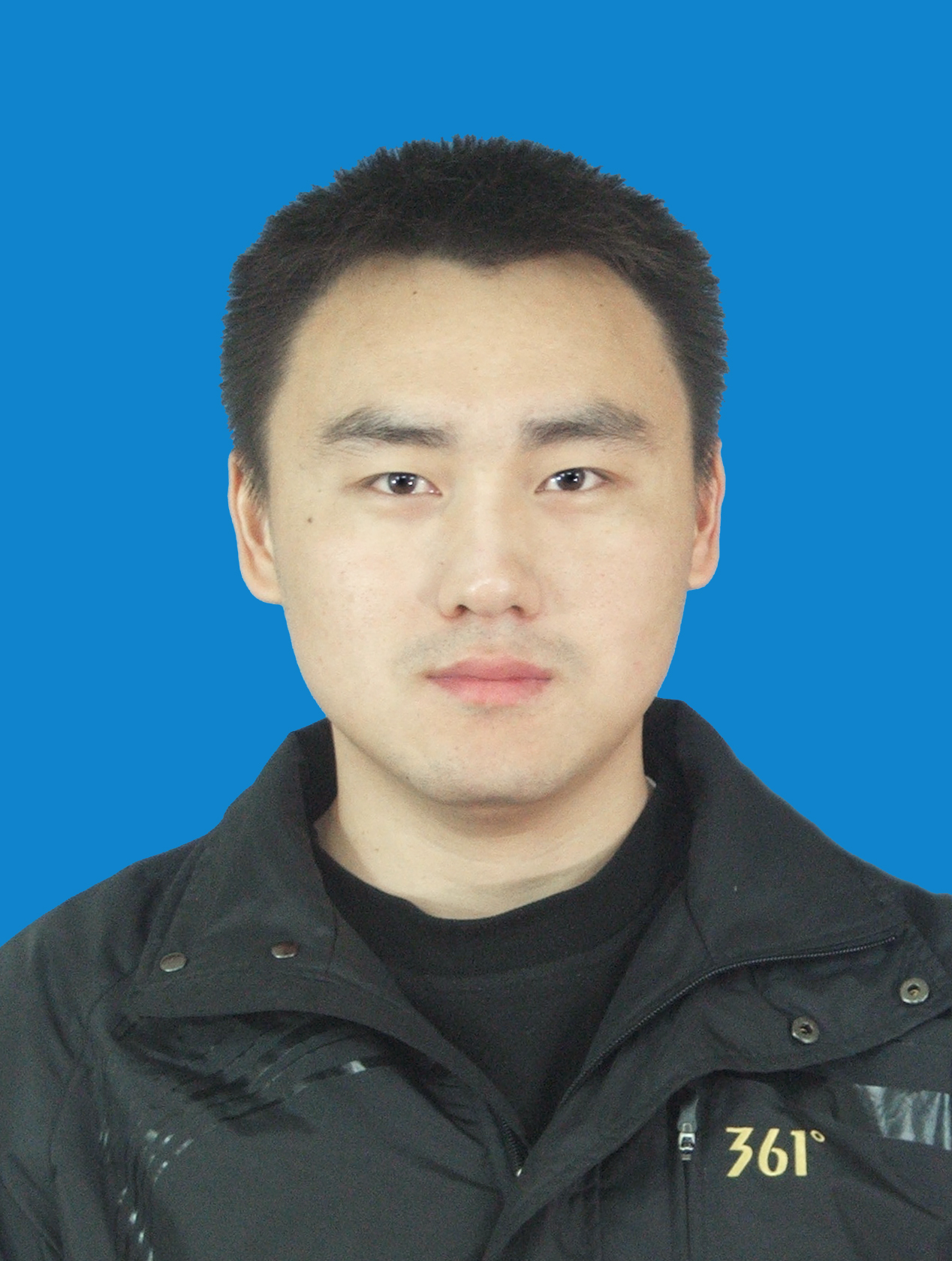}}]{Mingqiang Wei}
(Senior Member, IEEE) received his Ph.D degree (2014) in Computer Science and Engineering from the Chinese University of Hong Kong (CUHK). 
He is a Professor at the School of Computer Science and Technology, Nanjing University of Aeronautics and Astronautics (NUAA). 
Before joining NUAA, he served as an assistant professor at Hefei University of Technology, and a postdoctoral fellow at CUHK. 
He was a recipient of the CUHK Young Scholar Thesis Awards in 2014. 
He is now an Associate Editor for the Visual Computer Journal, Journal of Electronic Imaging, and a Guest Editor for IEEE Transactions on Multimedia. His research interests focus on 3D vision, computer graphics, and deep learning. Prof. Wei is the co-corresponding author.
\end{IEEEbiography}

\begin{IEEEbiography}[{\includegraphics[width=1in,height=1.25in,clip,keepaspectratio]{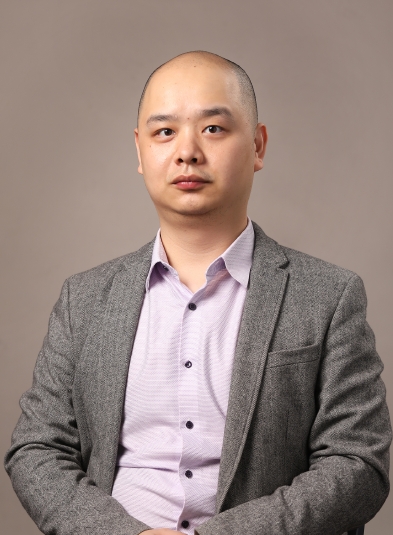}}]{Haoran Xie}
is an Associate Professor at the Department of Computing and Decision Sciences, Lingnan University, Hong Kong. He received his Ph.D. in Computer Science from the City University of Hong Kong. His research interest includes artificial intelligence, big data, and educational technology. He has published 240 research publications, including 114 journal articles. Among all 114 journal articles, there are 89 SCI/SSCI indexed and 12 SCOPUS indexed. He has obtained 14 research awards, including five best paper awards from WI 2020, ICBL 2020, DASFAA 2017, ICBL 2016 and SECOP 2015, the Golden Medal and the special award from International Invention Innovation Competition in Canada and so on. Prof. Xie is the Editor-in-Chief of Computers \& Education: Artificial Intelligence, Associate Editors of Array Journal, Australasian Journal of Educational Technology, Advances in Computational Intelligence, and International Journal of Mobile Learning and Organisation. He has successfully obtained more than 50 research grants; the total amount of these grants is more than HK \$27 million. He is the Senior Member of IEEE and ACM, and the Life Member of AAAI. 
\end{IEEEbiography}


\begin{IEEEbiography}[{\includegraphics[width=1in,height=1.25in,clip,keepaspectratio]{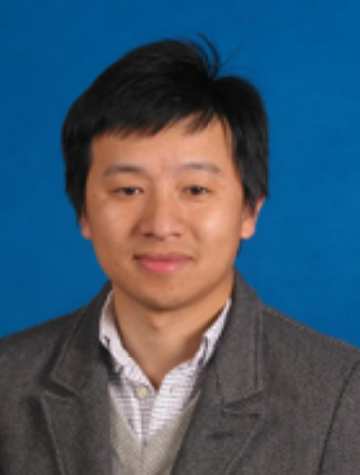}}]{Jun Wang}
is currently a professor at Nanjing University of Aeronautics and Astronautics (NUAA), China. He received his Bachelor and PhD degrees in Computer-Aided Design from NUAA in 2002 and 2007 respectively. From 2008 to 2009, he conducted research as a postdoctoral scholar at the University of California, Davis. Subsequently, he worked as a research associate at the University of Wisconsin, Milwaukee for one year. From 2010 to 2013, he worked as a senior research engineer at Leica Geosystems Inc., USA. In 2013, he paid an academic visit to the Department of Mathematics at Harvard University. His research interests include geometry processing and geometric modeling, especially large-scale LiDAR point data capturing, management, processing and analysis.
\end{IEEEbiography}

\begin{IEEEbiography}[{\includegraphics[width=1in,height=1.25in,clip,keepaspectratio]{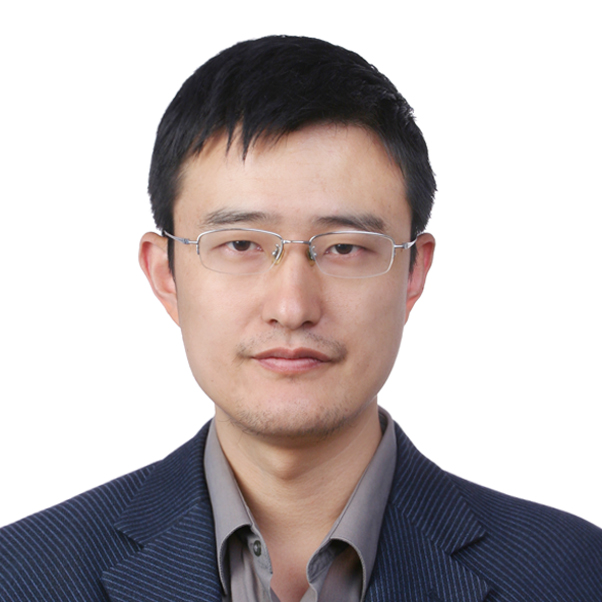}}]{Tong Lu}
received the Ph.D. degree in computer science from Nanjing University in 2005. He received his B.Sc. degree from the same university in 1997. He served as Associate Professor and Assistant Professor in the Department of Computer Science and Technology at Nanjing University from 2007 and 2005. He is now a full Professor at the same university. He also has served as Visiting Scholar at National University of Singapore and Department of Computer Science and Engineering, Hong Kong University of Science and Technology, respectively. He is also a member of the National Key Laboratory of Novel Software Technology in China. He has published over 130 papers and authored 2 books in his area of interest, and issued more than 20 international or Chinese invention patents. His current interests are in the areas of multimedia, computer vision and pattern recognition algorithms/systems. Prof. Lu is the co-corresponding author.
\end{IEEEbiography}

\begin{IEEEbiography}[{\includegraphics[width=1in,height=1.25in,clip,keepaspectratio]{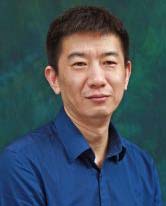}}]{Jing Qin}
is currently an associate professor in School of Nursing, The Hong Kong Polytechnic University and is a key member in the Centre for Smart Health. His research focuses on creatively leveraging advanced virtual reality (VR) and artificial intelligence (AI) techniques in healthcare and medicine applications and his achievements in relevant areas has been well recognized by the academic community. He won the Hong Kong Medical and Health Device Industries Association Student Research Award for his PhD study on VR-based simulation systems for surgical training and planning. He won 3 best paper awards for his research on AI-driven medical image analysis and computer-assisted surgery, including one of the most prestigious awards in this field: MIA-MICCAI best paper award in 2017. He served as a local organization chair for MICCAI 2019, technical program committee (TPC) members for many academic conferences, speakers for many invited talks, and referees for many prestigious journals in relevant fields.
\end{IEEEbiography}

\begin{IEEEbiography}[{\includegraphics[width=1in,height=1.25in,clip,keepaspectratio]{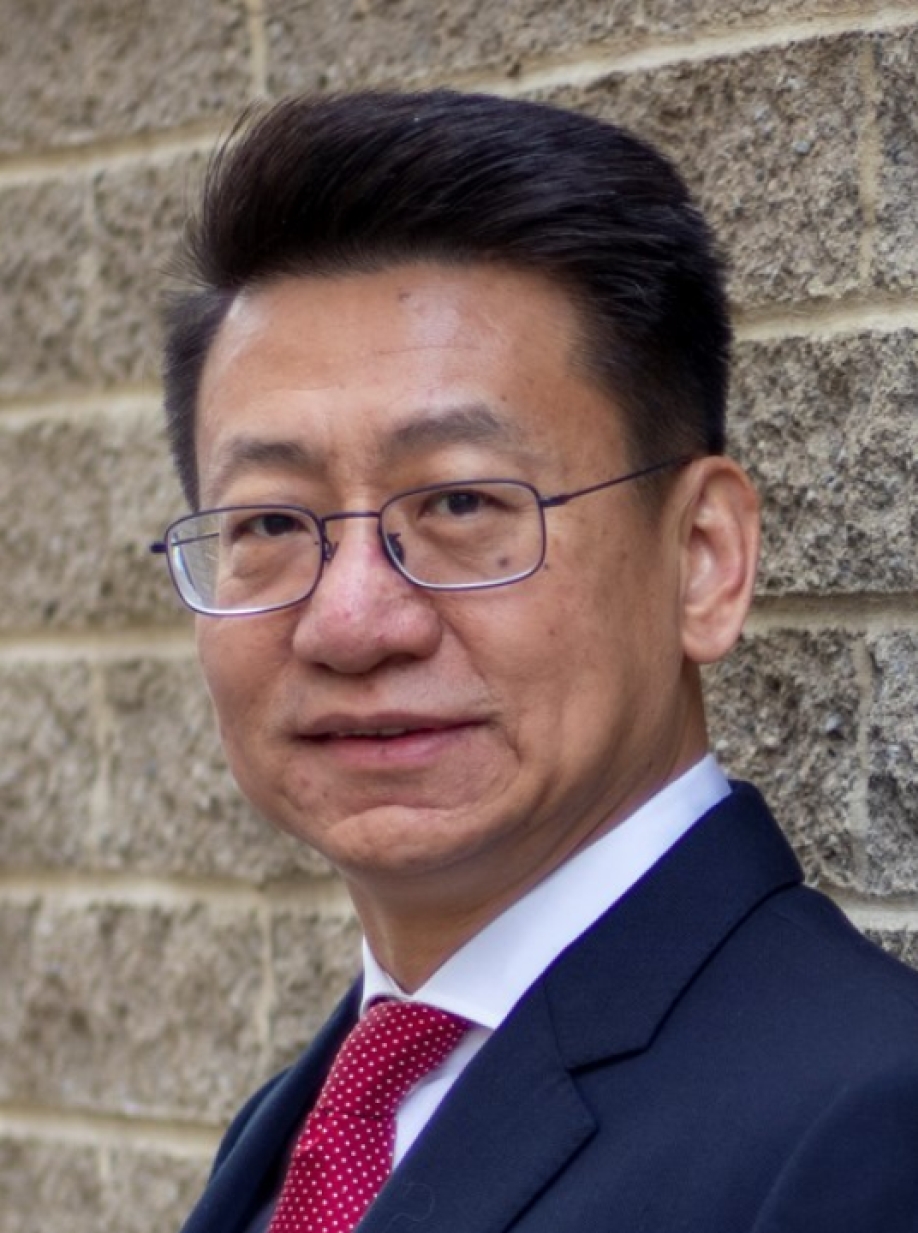}}]{Xiao-Ping Zhang}
received B.S. and Ph.D. degrees from Tsinghua University, in 1992 and 1996, respectively, both in Electronic Engineering. He holds an MBA in Finance, Economics and Entrepreneurship with Honors from the University of Chicago Booth School of Business, Chicago, IL. 

Since Fall 2000, he has been with the Department of Electrical, Computer and Biomedical Engineering, Ryerson University, Toronto, ON, Canada, where he is currently a Professor and the Director of the Communication and Signal Processing Applications Laboratory. He has served as the Program Director of Graduate Studies. He is cross-appointed to the Finance Department at the Ted Rogers School of Management, Ryerson University. He was a Visiting Scientist with the Research Laboratory of Electronics, Massachusetts Institute of Technology, Cambridge, MA, USA, in 2015 and 2017. He is a frequent consultant for biotech companies and investment firms. His research interests include sensor networks and IoT, machine learning, statistical signal processing, image and multimedia content analysis, and applications in big data, finance, and marketing. 

Dr. Zhang is Fellow of the Canadian Academy of Engineering, Fellow of the Engineering Institute of Canada, Fellow of the IEEE, a registered Professional Engineer in Ontario, Canada, and a member of Beta Gamma Sigma Honor Society. He is the general Co-Chair for the IEEE International Conference on Acoustics, Speech, and Signal Processing, 2021. He is the general co-chair for 2017 GlobalSIP Symposium on Signal and Information Processing for Finance and Business, and the general co-chair for 2019 GlobalSIP Symposium on Signal, Information Processing and AI for Finance and Business. He was an elected Member of the ICME steering committee. He is the General Chair for the IEEE International Workshop on Multimedia Signal Processing, 2015. He is the Publicity Chair for the International Conference on Multimedia and Expo 2006, and the Program Chair for International Conference on Intelligent Computing in 2005 and 2010. He served as a Guest Editor for Multimedia Tools and Applications and the International Journal of Semantic Computing. He was a tutorial speaker at the 2011 ACM International Conference on Multimedia, the 2013 IEEE International Symposium on Circuits and Systems, the 2013 IEEE International Conference on Image Processing, the 2014 IEEE International Conference on Acoustics, Speech, and Signal Processing, the 2017 International Joint Conference on Neural Networks and the 2019 IEEE International Symposium on Circuits and Systems. He is Editor-in-Chief for the IEEE JOURNAL OF SELECTED TOPICS IN SIGNAL PROCESSING. He is Senior Area Editor for the IEEE TRANSACTIONS ON IMAGE PROCESSING. He served as Senior Area Editor the IEEE TRANSACTIONS ON SIGNAL PROCESSING and Associate Editor for the IEEE TRANSACTIONS ON IMAGE PROCESSING, the IEEE TRANSACTIONS ON MULTIMEDIA, the IEEE TRANSACTIONS ON CIRCUITS AND SYSTEMS FOR VIDEO TECHNOLOGY, the IEEE TRANSACTIONS ON SIGNAL PROCESSING, and the IEEE SIGNAL PROCESSING LETTERS. He received 2020 Sarwan Sahota Ryerson Distinguished Scholar Award -- the Ryerson University highest honor for scholarly, research and creative achievements. He is selected as IEEE Distinguished Lecturer by the IEEE Signal Processing Society for the term 2020 to 2021, and by the IEEE Circuits and Systems Society for the term 2021 to 2022. 
\end{IEEEbiography}




\end{document}